\documentclass{article}

\usepackage{microtype}
\usepackage{graphicx}

\usepackage{booktabs} % for professional tables

\usepackage{float} 
\usepackage{caption}
\usepackage{subcaption}

\usepackage[dvipsnames]{xcolor}
\usepackage{dblfloatfix}

\usepackage{alphalph}
\usepackage{hyperref}

\usepackage[accepted]{icml2023}

\usepackage{setspace}
\usepackage{amsmath}
\usepackage{amssymb}
\usepackage{mathtools}
\usepackage{amsthm}

\usepackage[capitalize,noabbrev]{cleveref}

\theoremstyle{plain}
\newtheorem{theorem}{Theorem}[section]

\newtheorem{lemma}[theorem]{Lemma}
\newtheorem{corollary}[theorem]{Corollary}
\theoremstyle{definition}

\theoremstyle{remark}

\usepackage[textsize=tiny]{todonotes}

\icmltitlerunning{A Case for Validation Buffer in Pessimistic Actor-Critic}

\begin{document}

\twocolumn[
\icmltitle{A Case for Validation Buffer in Pessimistic Actor-Critic}



\icmlsetsymbol{equal}{*}

\begin{icmlauthorlist}
\icmlauthor{Michal Nauman}{yyy,sch}
\icmlauthor{Mateusz Ostaszewski}{poli}
\icmlauthor{Marek Cygan}{yyy,comp}
\end{icmlauthorlist}

\icmlaffiliation{yyy}{Informatics Institute, University of Warsaw}
\icmlaffiliation{comp}{Nomagic}
\icmlaffiliation{poli}{Warsaw University of Technology}
\icmlaffiliation{sch}{Ideas National Centre for Research and Development}

\icmlcorrespondingauthor{Michal Nauman}{nauman.mic@gmail.com}

\icmlkeywords{Machine Learning, ICML}

\vskip 0.3in
]



\printAffiliationsAndNotice{} % otherwise use the standard text.

\begin{abstract} 
In this paper, we investigate the issue of error accumulation in critic networks updated via pessimistic temporal difference objectives. We show that the critic approximation error can be approximated via a recursive fixed-point model similar to that of the Bellman value. We use such recursive definition to retrieve the conditions under which the pessimistic critic is unbiased. Building on these insights, we propose Validation Pessimism Learning (VPL) algorithm. VPL uses a small validation buffer to adjust the levels of pessimism throughout the agent training, with the pessimism set such that the approximation error of the critic targets is minimized. We investigate the proposed approach on a variety of locomotion and manipulation tasks and report improvements in sample efficiency and performance. 
\end{abstract}

\section{Introduction}

\begin{figure}[h]
\begin{center}
    \begin{subfigure}{0.49\linewidth}
    \includegraphics[width=\textwidth]{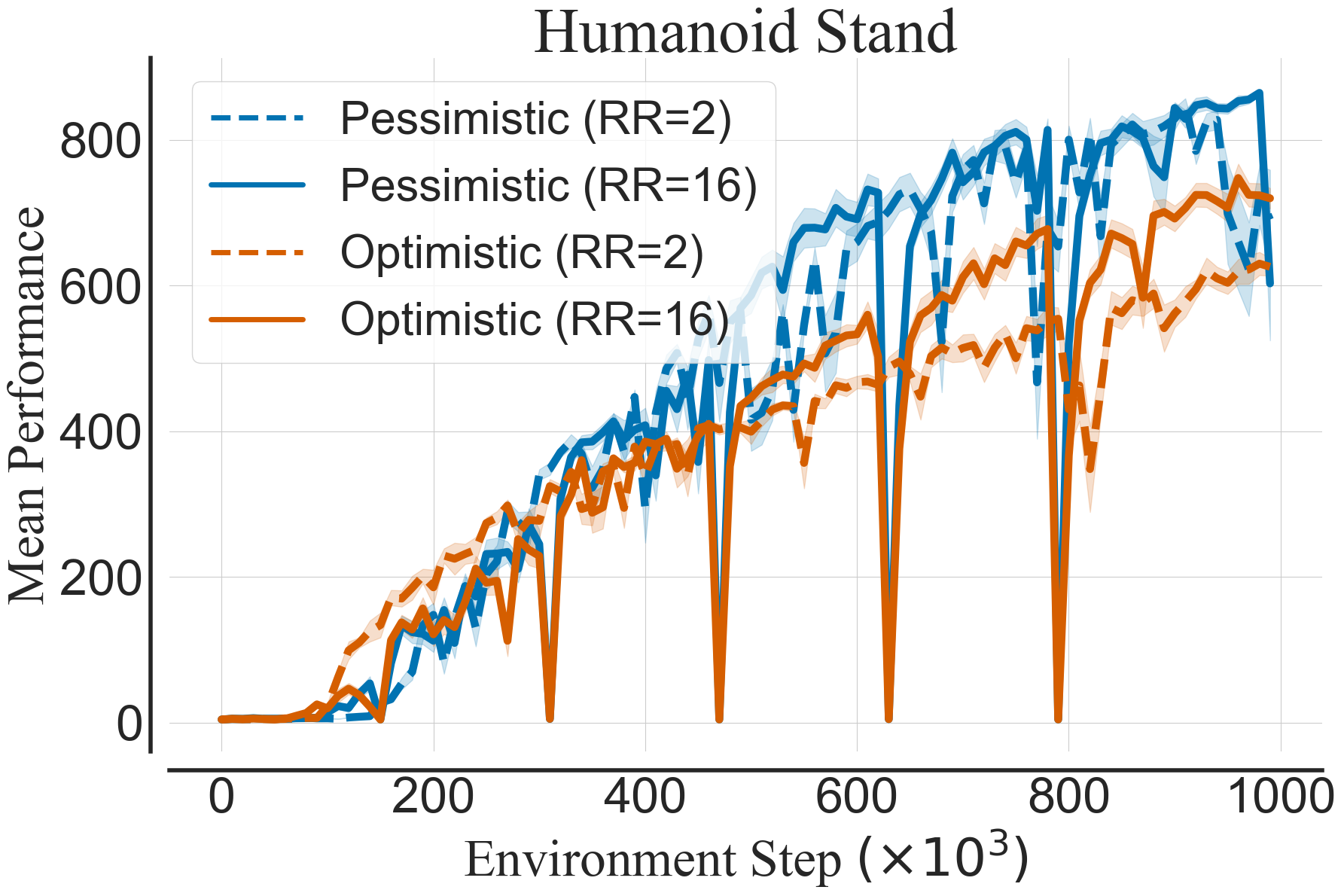}
    \label{fig:pess1}
    \end{subfigure}
    \begin{subfigure}{0.49\linewidth}
    \includegraphics[width=\textwidth]{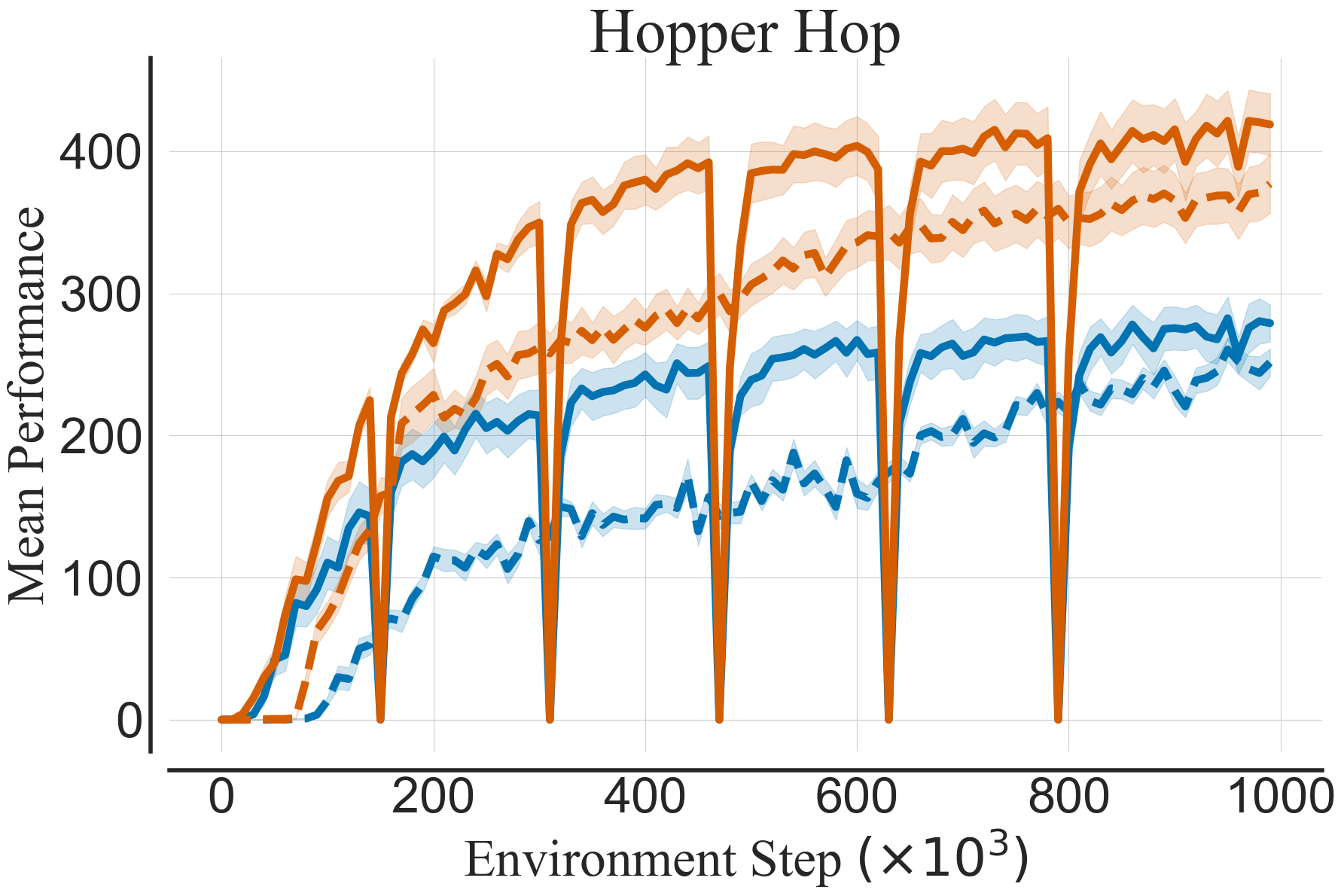}
    \label{fig:pess2}
    \end{subfigure}
\vspace{-0.2in}
\caption{Pessimism adjustment can yield performance benefits exceeding those of increased replay ratio and full-parameter resets. The pessimistic algorithm dominates Humanoid, whereas the optimistic algorithm dominates Hopper. 10 seeds and 95\% CI.}
\label{fig:pessimism}
\end{center}
%\vspace{-0.2in}
\end{figure}

Approximation errors, although ubiquitous in machine learning, are particularly exaggerated in the context of value-based Reinforcement Learning (RL). Such exaggeration stems from Temporal Difference (TD) in which the critic is supervised via value estimate calculated at a different state \citep{silver2014deterministic, mnih2015human, barth2018distributed}. Inaccuracies in this estimate lead to propagated errors in state-action updates, and the use of maximization in value estimation inherently promotes overestimation. Addressing such overestimation has proven to be an effective strategy in discrete and continuous action environments \citep{hasselt2010double, van2016deep, hessel2018rainbow, haarnoja2018soft}. Clipped Double Q-Learning (CDQL), a common solution to overestimation in continuous action actor-critic algorithms aims to mitigate overestimation by balancing errors against a pessimistic lower bound value approximation \citep{fujimoto2018addressing}. However, challenges arise if the lower bound is insufficiently pessimistic, leading to continued overestimation, or overly pessimistic, causing underestimation \citep{cetin2023learning}. The latter, though less recognized, can significantly reduce sample efficiency and degrade actor-critic agents' performance in both low and high replay ratio settings which we show in Figure \ref{fig:pessimism}.

\begin{figure*}[ht!]
\begin{center}
\begin{minipage}[h]{1.0\linewidth}
\centering
    \begin{subfigure}{0.66\linewidth}
    \includegraphics[width=\textwidth]{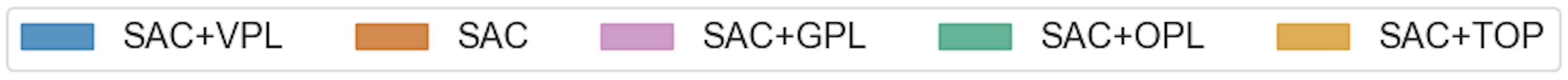}
    \end{subfigure}
\end{minipage}
\medskip
\begin{minipage}[h]{1.0\linewidth}
    \begin{subfigure}{1.0\linewidth}
    \includegraphics[width=0.237\linewidth]{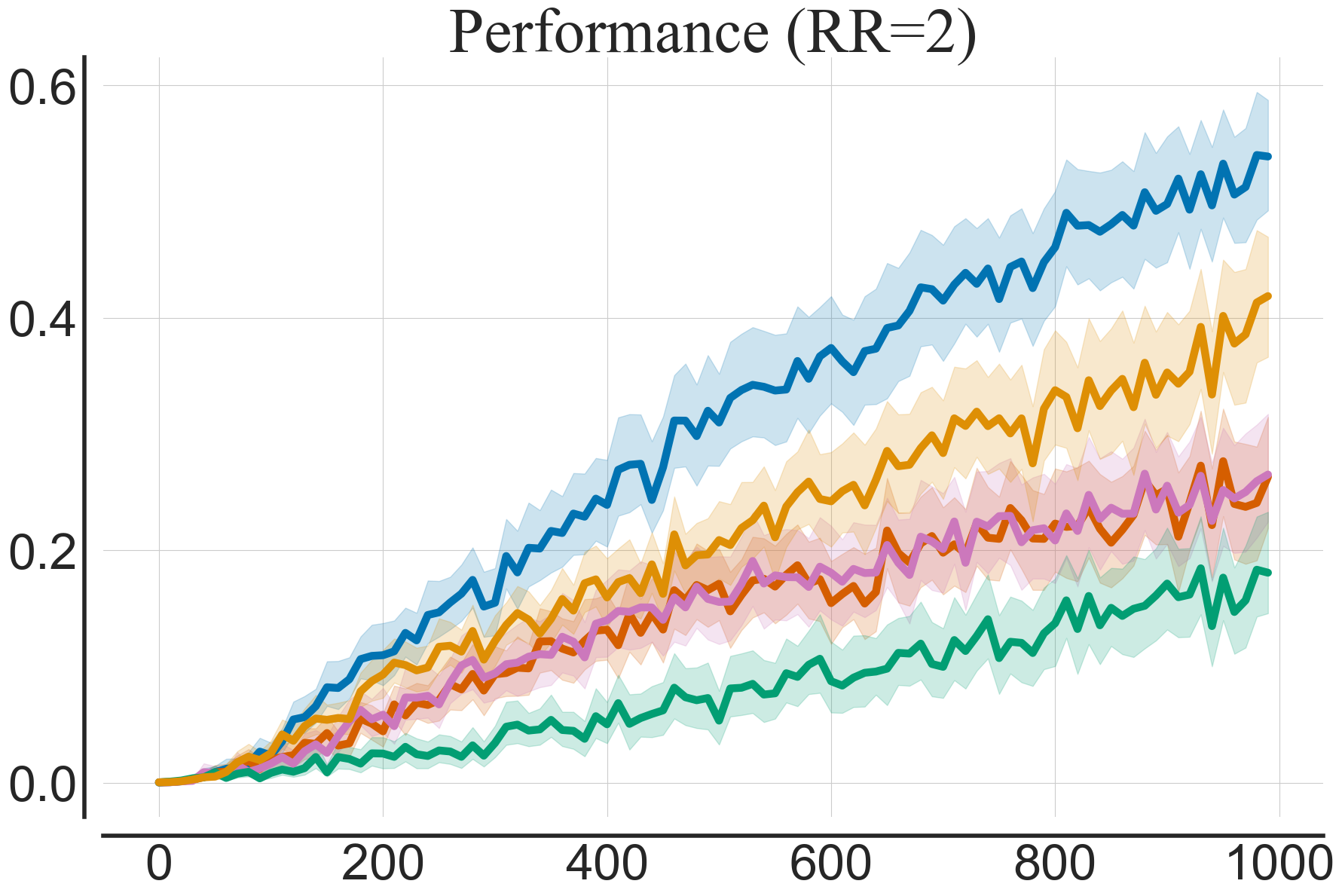}
    \hfill
    \includegraphics[width=0.237\linewidth]{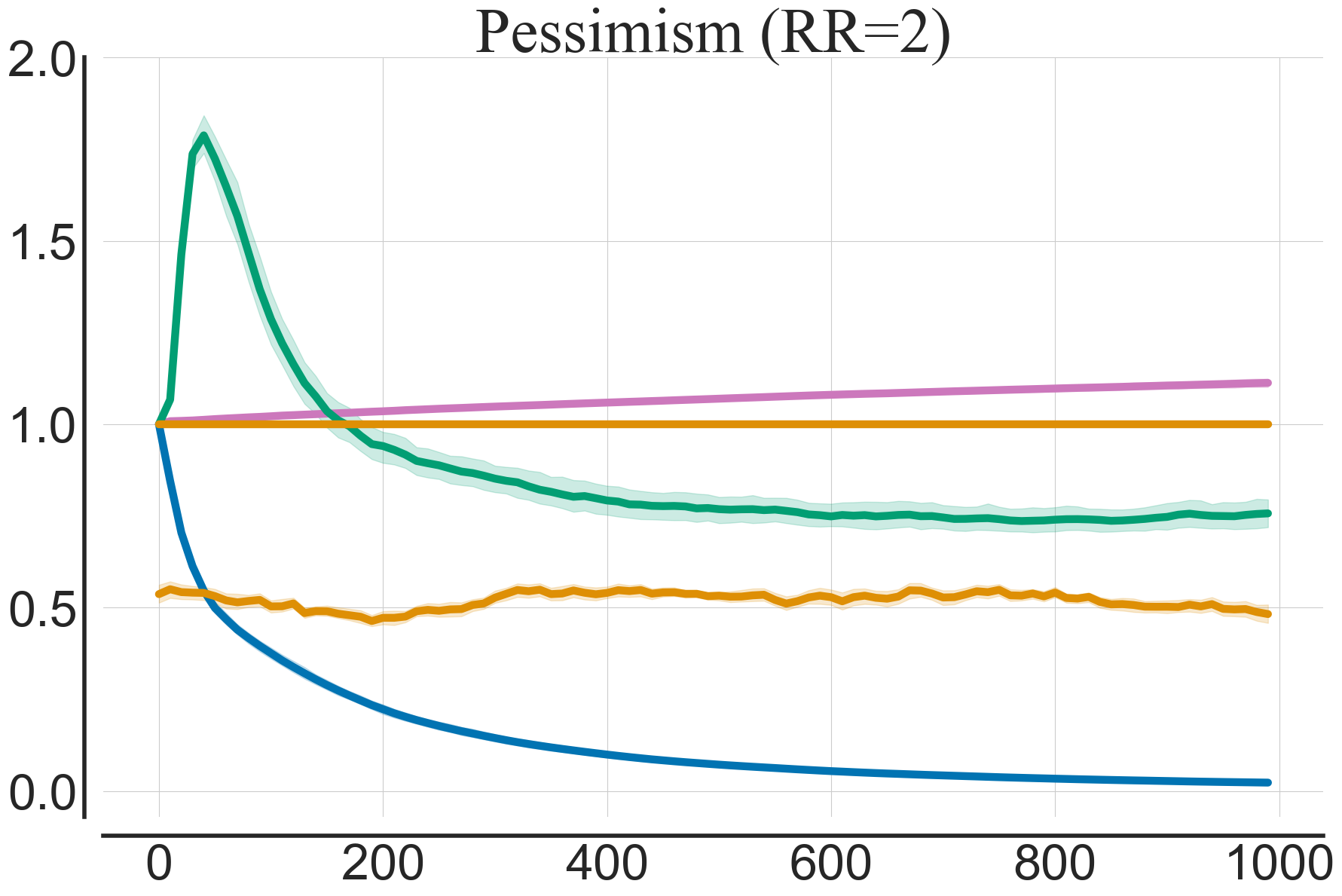}
    \hfill
    \includegraphics[width=0.237\linewidth]{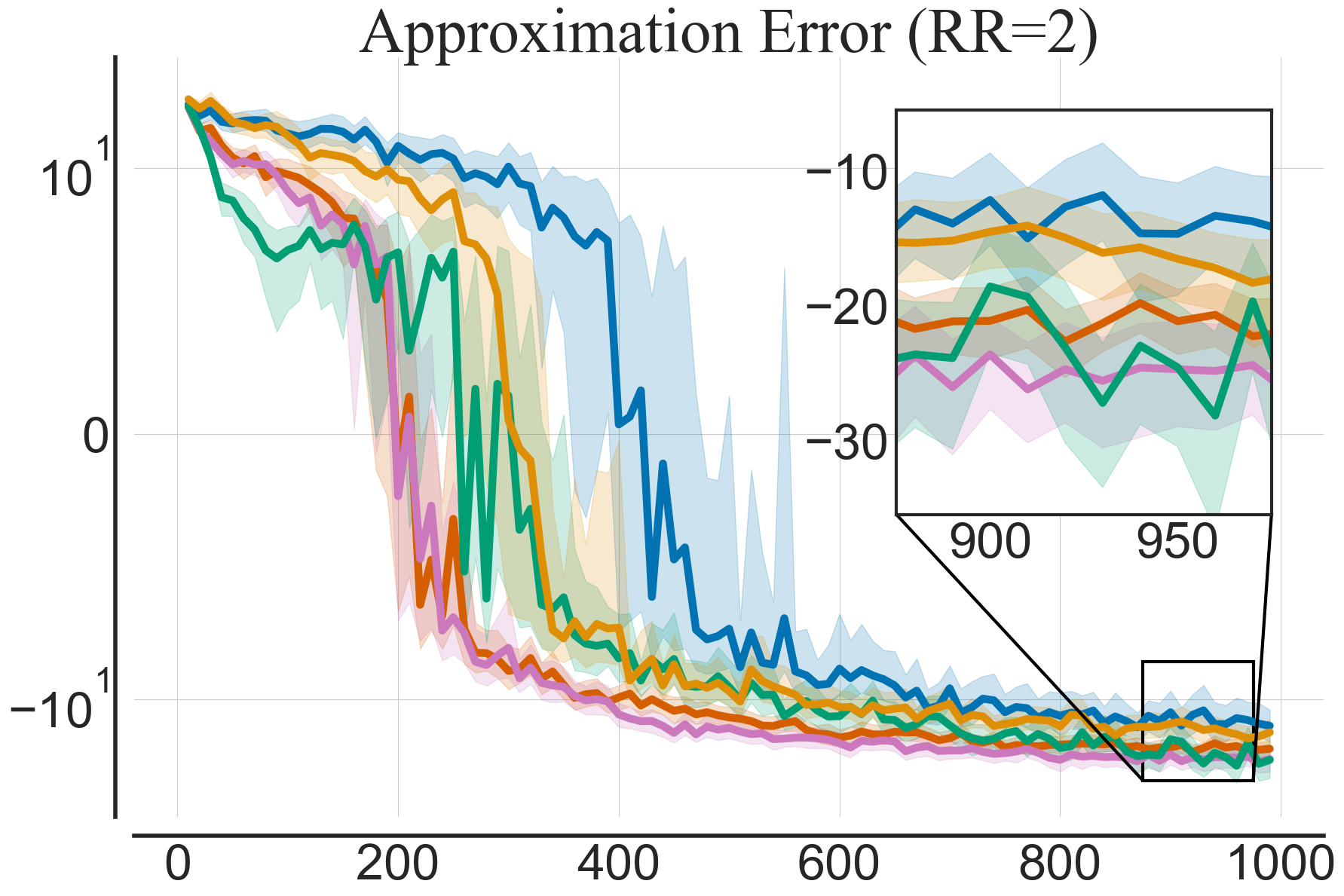}
    \hfill
    \includegraphics[width=0.237\linewidth]{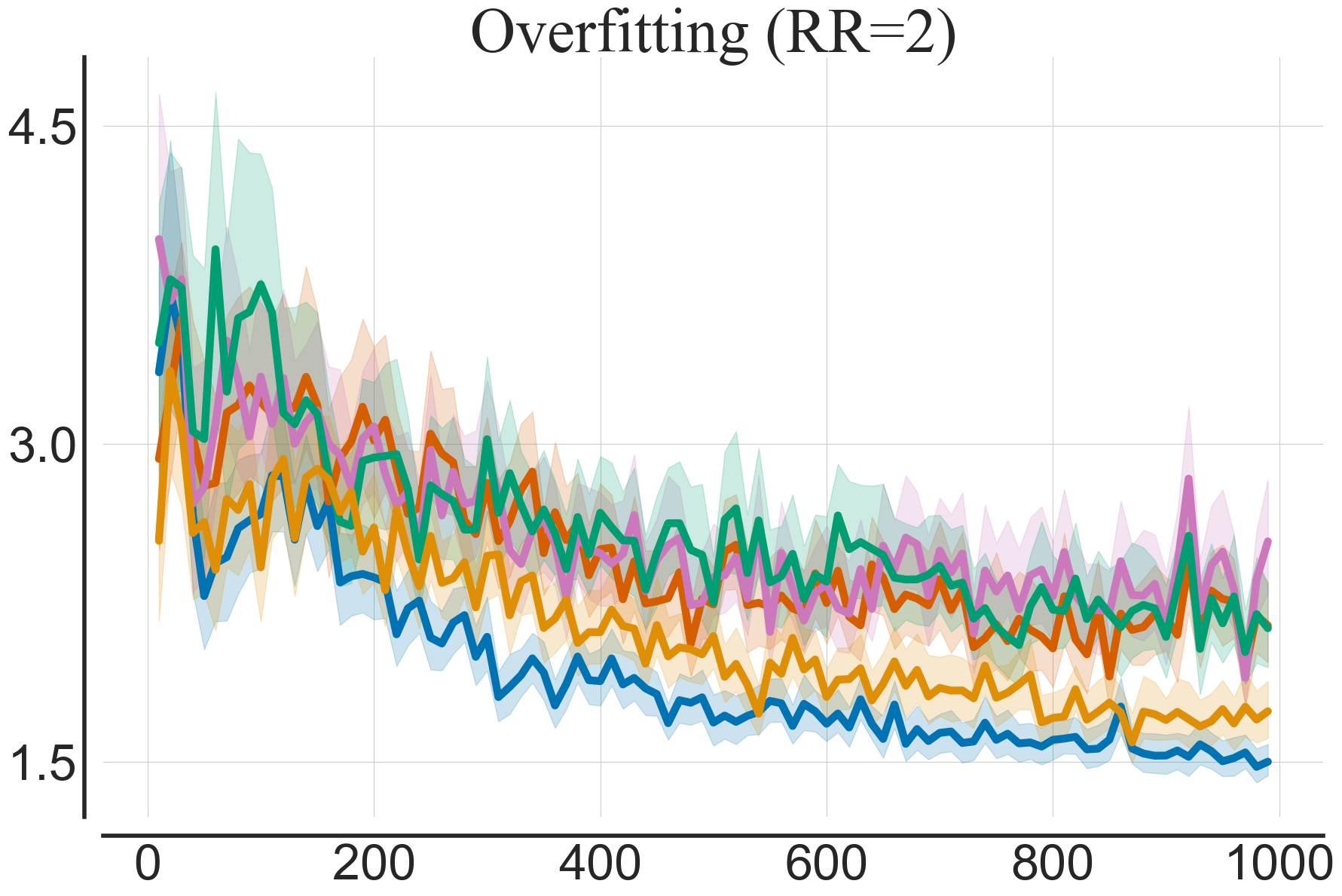}
    %\caption{Replay Ratio = 2}
    \label{fig:intro_perf1}
    \end{subfigure}
\end{minipage}
\begin{minipage}[h]{1.0\linewidth}
    \begin{subfigure}{1.0\linewidth}
    \includegraphics[width=0.237\linewidth]{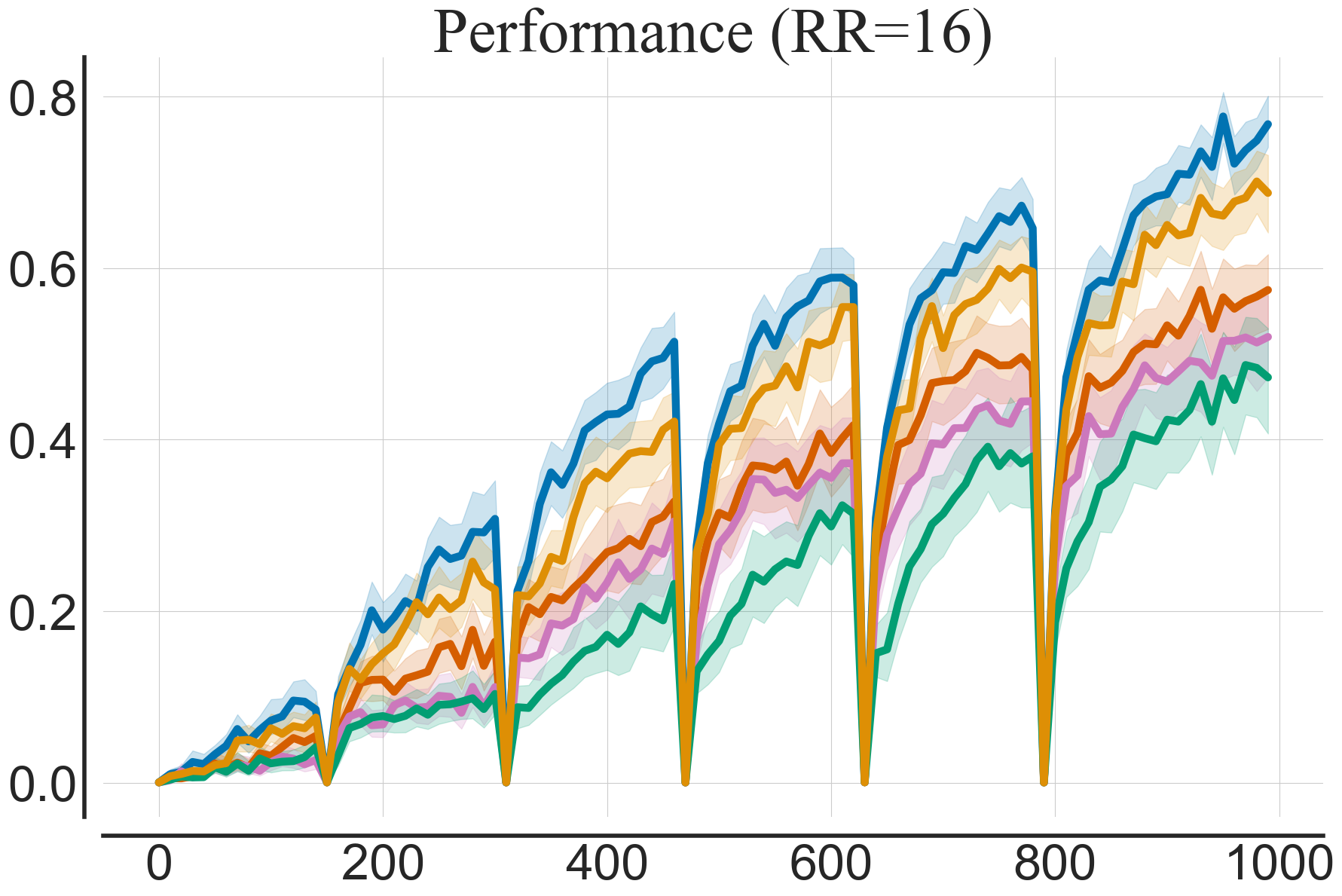}
    \hfill
    \includegraphics[width=0.237\linewidth]{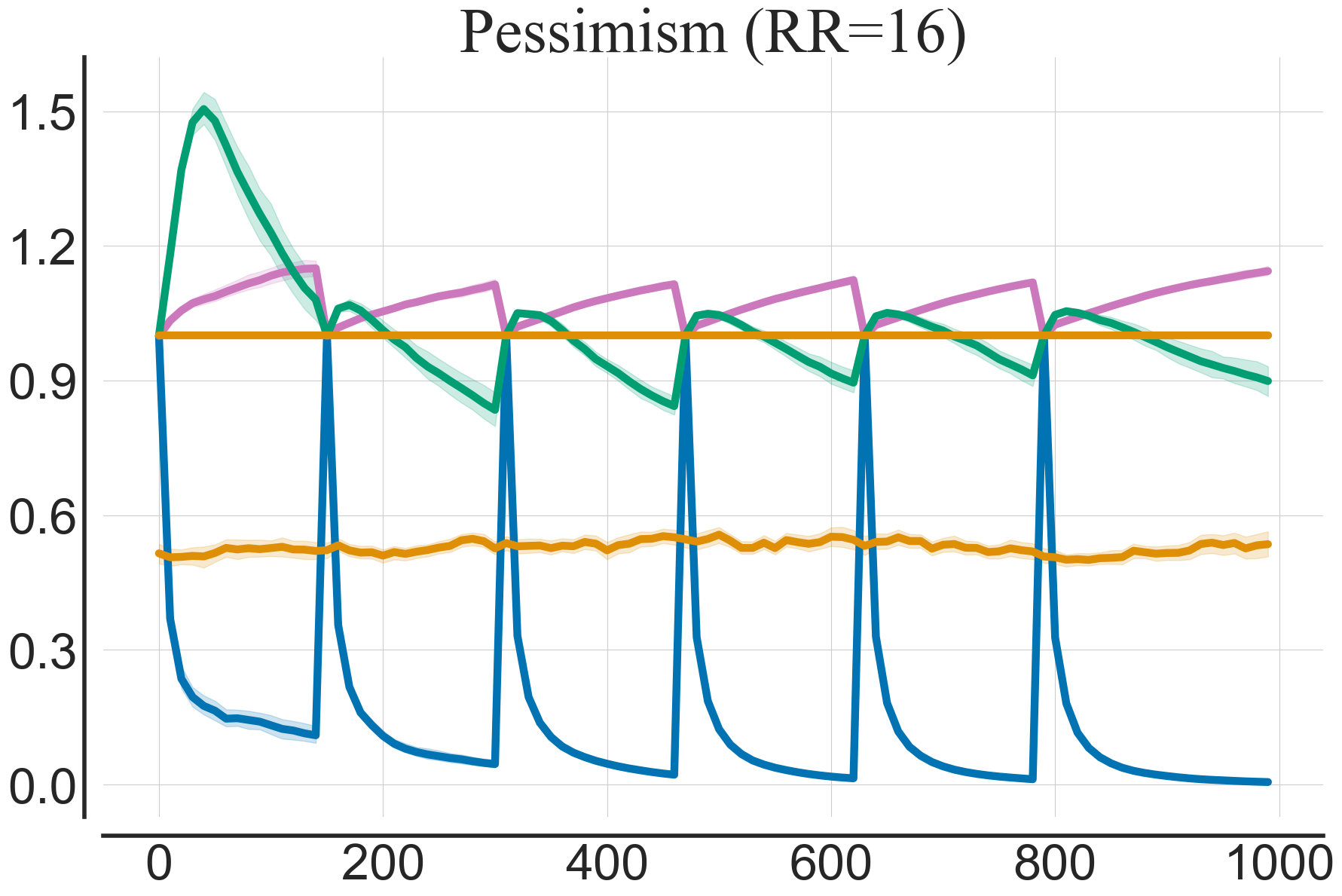}
    \hfill
    \includegraphics[width=0.237\linewidth]{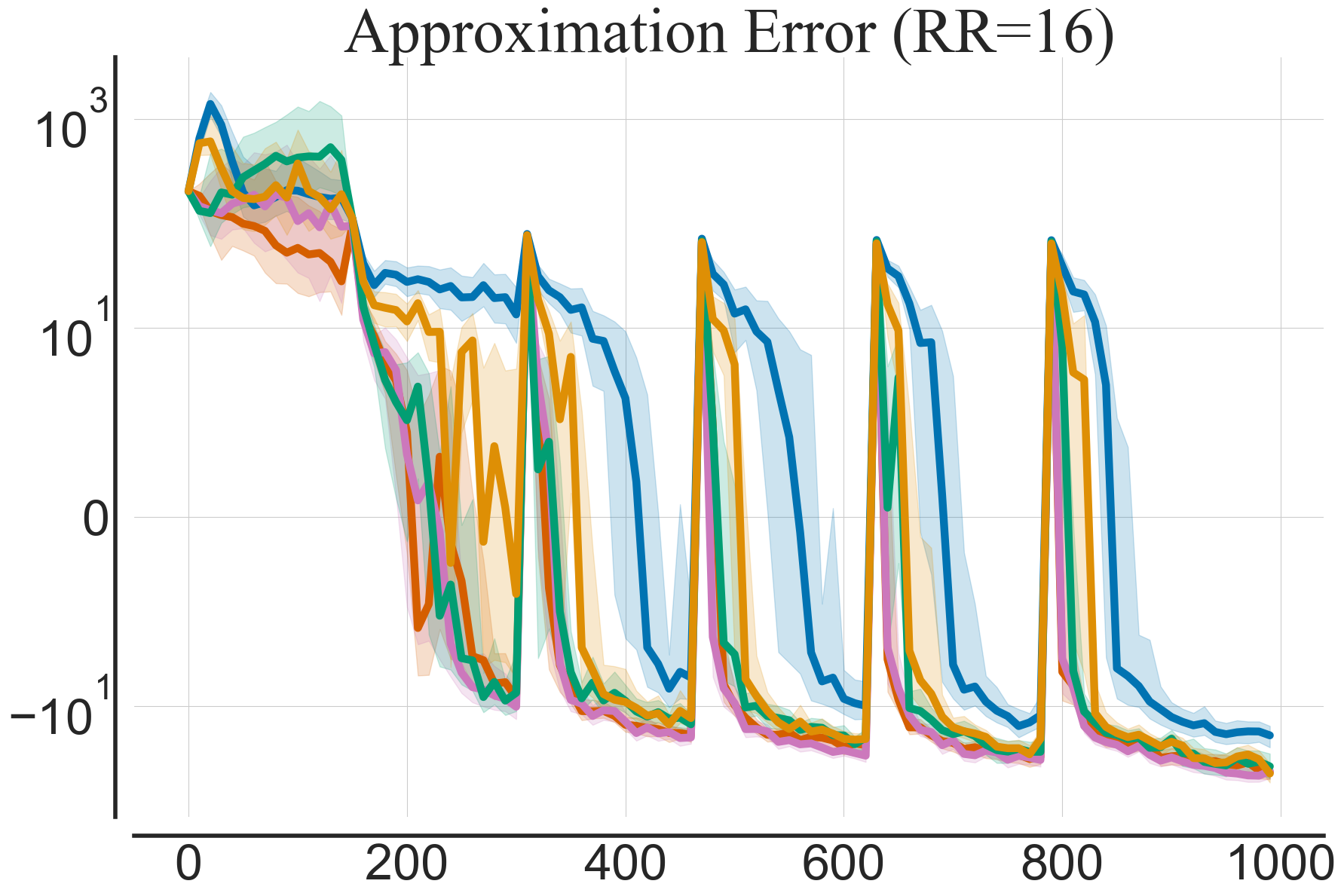}
    \hfill
    \includegraphics[width=0.237\linewidth]{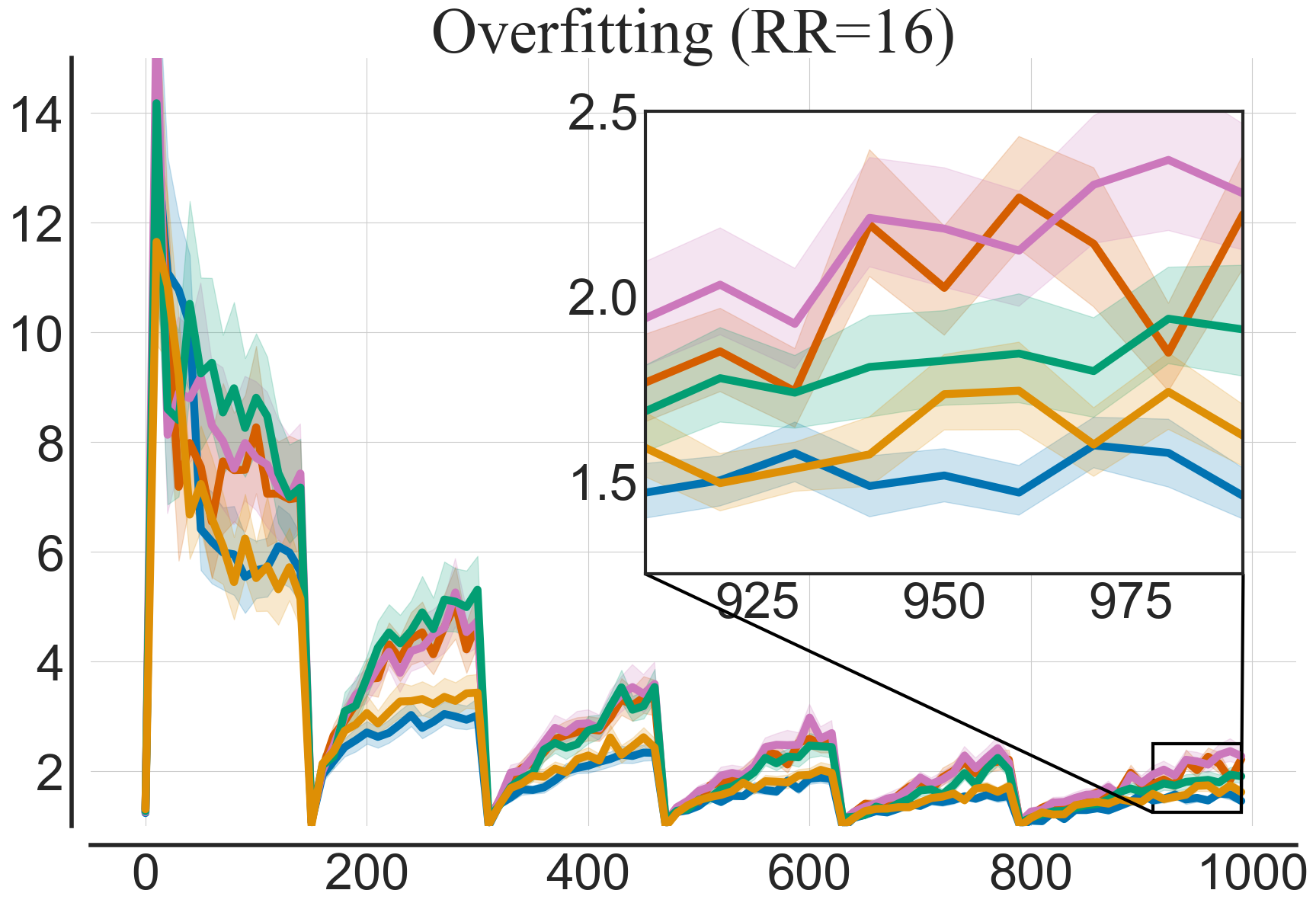}
    %\caption{Replay Ratio = 2}
    \label{fig:intro_perf2}
    \end{subfigure}
\end{minipage}
\caption{We integrate the Soft Actor-Critic (SAC) and the Scaled-By-Resetting SAC (SR-SAC) with various pessimism adjustment algorithms. We evaluate the performance of the integrated algorithms in both low replay (top row) and high replay (bottom row) regimes. All tested algorithms use the same network architectures and hyperparameter settings. As such, the performance difference stems solely from the varying strategy for pessimism adjustment. Notably, despite similar underlying motivations, each algorithm manifests specific levels of pessimism. Our proposed Validation Pessimism Learning (VPL) module demonstrates the lowest approximation error and mitigates value overfitting more effectively than the alternative approaches. These characteristics translate to performance and sample efficiency improvements. The experimental setting is detailed in Sections \ref{sec:experiments} \& \ref{appendix_exp_details}. 20 tasks, 10 seeds per task, IQM and 95\% CI.}
\label{fig:intro_perf}
\end{center}
\end{figure*}

In this paper, we investigate the relationship between pessimism in Q-value approximation and error accumulation in critic networks. We start by characterization of existing strategies for online pessimism adjustment. Furthermore, we analyze the pessimistic critic approximation error and show that such error can be represented recursively forming a fixed-point model, akin to values and Q-values. This recursive representation helps us highlight the bias inherent in pessimistic actor-critic algorithms, examine their convergence dynamics, and identify the conditions under which pessimistic critics can achieve zero error in value approximation. Building on these insights, we propose the Validation Pessimism Learning (VPL) algorithm. VPL employs a small validation replay buffer to adjust the pessimism levels online, aiming to minimize the approximation error of critic targets while preventing overfitting to accumulated experience. We evaluate VPL against existing pessimism adjustment methods on DeepMind control \citep{tassa2018deepmind} and single-task MetaWorld \citep{yu2020meta} platforms. Our findings demonstrate that VPL not only achieves performance improvements but also exhibits less sensitivity to hyperparameter settings compared to the baseline algorithms. We summarize our contributions below:

\begin{itemize}
    \item We show that critic approximation error can be defined recursively through a fixed-point model. We demonstrate that pessimistic TD learning, a method often used in continuous action RL, converges to the true value under strict conditions.
    \item We present an empirical analysis showing that the performance loss associated with not including every transition in the replay buffer diminishes as training progresses. This observation challenges the traditional belief that every transition must be used in value learning for sample-efficient RL and builds a case for employing a validation buffer in an online RL setting.    
    \item We propose VPL, an algorithm that uses a small validation buffer for online adjustment of pessimism associated with lower bound Q-value approximation. We test the effectiveness of VPL and other pessimism adjustment strategies in low and high replay regimes. We show that VPL offers performance improvements across a variety of locomotion and manipulation tasks. 
\end{itemize}

\section{Background}

\subsection{Maximum Entropy Reinforcement Learning}

We analyze an infinite-horizon Markov Decision Process (MDP) \citep{puterman2014markov}, represented by the tuple $(S, A, r, p_0, \gamma)$. In this model, both states $S$ and actions $A$ are continuous. The transition reward is given by $r_{s, a}$, $p_{0}(s)$ defines the initial state distribution, and $\gamma \in (0,1]$ is the discount factor. The policy, $\pi(a|s)$, is a distribution of actions conditioned on states. At any given state, the policy entropy is denoted as $\mathcal{H}(s)$. In an MDP where all states are positive recurrent, a policy-induced discounted stationary distribution $p_{\gamma}(s | \pi)$ also exists. The goal of Maximum Entropy Reinforcement Learning (MaxEnt RL) \citep{ziebart2008maximum, haarnoja2017reinforcement} is to devise a policy that optimizes the expected cumulative sum of discounted returns and entropy.

\begin{equation}
\label{eq:rl_objective}
\begin{split}
    \pi^{*} &= \arg \max_{\pi} \underset{p_0, \pi}{\mathrm{E}} \sum_{t=0}^{\infty} \gamma^{t} \bigl( r_{s_{t}, a_{t}} + \alpha \mathcal{H}(s_t) \bigr),
\end{split}
\end{equation}

where $\alpha$ denotes the temperature parameter which balances the reward and entropy objectives \citep{haarnoja2018soft}. Soft Q-value is defined as the expected discounted return from performing an action at a given state and then following the policy $Q^{\pi}(s,a) = r_{s, a} + \gamma V^{\pi}(s')$. Soft value, denoted as $V^{\pi}(s)$ is calculated as follows:

\begin{equation}
\label{eq:value}
\begin{split}
    V^{\pi}(s) & = \underset{\pi}{\mathrm{E}} ~ \bigl( Q^{\pi}(s,a) - \alpha \log \pi(a|s) \bigr).
\end{split}
\end{equation}

In this context, the term $\log \pi(a|s)$ corresponds to the entropy objective, with $-\mathrm{E}_{\pi}\log \pi(a|s) = \mathcal{H}(s)$. In algorithms like Soft Actor-Critic (SAC), the policy and Q-value functions are modeled via parameterized function approximators, commonly referred to as the actor and critic, respectively \citep{silver2014deterministic}. The parameters of these components are iteratively updated through gradient descent, following objectives derived from the policy iteration algorithm \citep{haarnoja2018soft}. In continuous actor-critic algorithms, the policy parameters $\theta$ are updated such that the policy maximized the value approximate at states $s$, which are sampled from an off-policy replay buffer $\mathcal{D}$:

\begin{equation}
\label{eq:actor_objective}
\begin{split}
    & \theta^{*} = \arg \max_{\theta} \underset{\mathcal{D}}{\mathrm{E}} V_{\phi}^{lb}(s),
\end{split}
\end{equation}
where $V_{\phi}^{lb}(s)$ is the approximate value lower bound calculated via the critic network \citep{haarnoja2018soft} and $s \sim \mathcal{D}$. Similarly, the critic parameters $\phi$ are updated in the policy evaluation step by minimizing temporal-difference variant \citep{ciosek2020expected}:
\begin{equation}
\label{eq:critic_objective}
\begin{split}
    & \phi^{*} = \arg \min_{\phi} \underset{\mathcal{D}}{\mathrm{E}} \bigl(Q_{\phi}(s, a) - r_{s,a} - \vartriangle \gamma V_{\phi}^{lb}(s') \bigr)^2.
\end{split}
\end{equation}

Above, we denote the critic outputs for a given state-action as $Q_{\phi}(s, a)$, $s,a,s' \sim \mathcal{D}$ and use $\vartriangle$ to denote the stop gradient operator. Modern actor-critic algorithms leverage a variety of countermeasures to overestimation of Q-value targets, with bootstrapping using target network \citep{van2016deep} and Clipped Double Q-Learning (CDQL) \citep{fujimoto2018addressing} being most prominent. In CDQL, the algorithm maintains an ensemble of critics to approximate the value lower bound:

\begin{equation}
\label{eq:value_approx}
\begin{split}
    & V^{lb}_{\phi}(s) \approx Q^{lb}_{\phi}(s,a) - \alpha \log \pi_{\theta}(a|s) \quad \text{with } a\sim\pi_{\theta},\\
    & Q^{lb}_{\phi}(s,a) = \min\bigl(Q^{1}_{\phi}(s,a), Q^{2}_{\phi}(s,a) \bigr),
\end{split}
\end{equation}

Where $Q^{lb}_{\phi}(s,a)$ denotes the Q-value lower bound and $Q^{i}_{\phi}(s,a)$ denotes the $i-th$ critic in the critic ensemble. The CDQL was generalized by noticing relation between the minimum operator and ensemble statistics \citep{ciosek2019better, moskovitz2021tactical, cetin2023learning}:

\begin{equation}
\label{eq:cdql_general}
\begin{split}
    & Q^{lb}_{\phi}(s,a) = Q^{\mu}_{\phi}(s,a) - \beta Q^{\sigma}_{\phi}(s,a).
\end{split}
\end{equation}

We denote the critic ensemble mean and standard deviation as $Q^{\mu}_{\phi}$ and $Q^{\sigma}_{\phi}$ respectively. In particular, for $\beta = 1$ the above rule is exactly equal to the CDQL \citep{ciosek2019better, cetin2023learning}. Such lower bound updates the actor-critic parameters in the direction corrected by the critic ensemble disagreement. Such targets are referred to as \textit{pessimistic} with the parameter $\beta$ called \textit{pessimism}.

\subsection{Pessimism Adjustment}
\label{sec:pessimism_adjustment}

The success of pessimistic updates in practice has led to various methods for adjusting pessimism online. These techniques aim to improve the performance and efficiency of the agent by reducing the error in critic approximation. Algorithms such as On-policy Pessimism Learning (OPL) \citep{kuznetsov2021automating} and Generalized Pessimism Learning (GPL) \citep{cetin2023learning} estimate this error and modify pessimism accordingly. Specifically, GPL views the adjustment of pessimism as a dual optimization problem, resulting in the following update rule:

\begin{equation}
\label{eq:cetin_pessimism}
\begin{split}
    & \beta = \arg \min_{\beta} \underset{p_{0}, \pi}{\mathrm{E}} ~ \beta \left( Q^{\pi}(s,a) - r_{s,a} - \vartriangle \gamma V_{\phi}^{lb}(s') \right), \\
    & V_{\phi}^{lb}(s') \approx Q^{\mu}_{\phi}(s,a) - \beta Q^{\sigma}_{\phi}(s,a) - \alpha \log \pi_{\theta}(a|s).
\end{split}
\end{equation}

\begin{figure}[t]
\begin{center}
    \begin{subfigure}{0.8\linewidth}
    \centering
    \includegraphics[width=\textwidth]{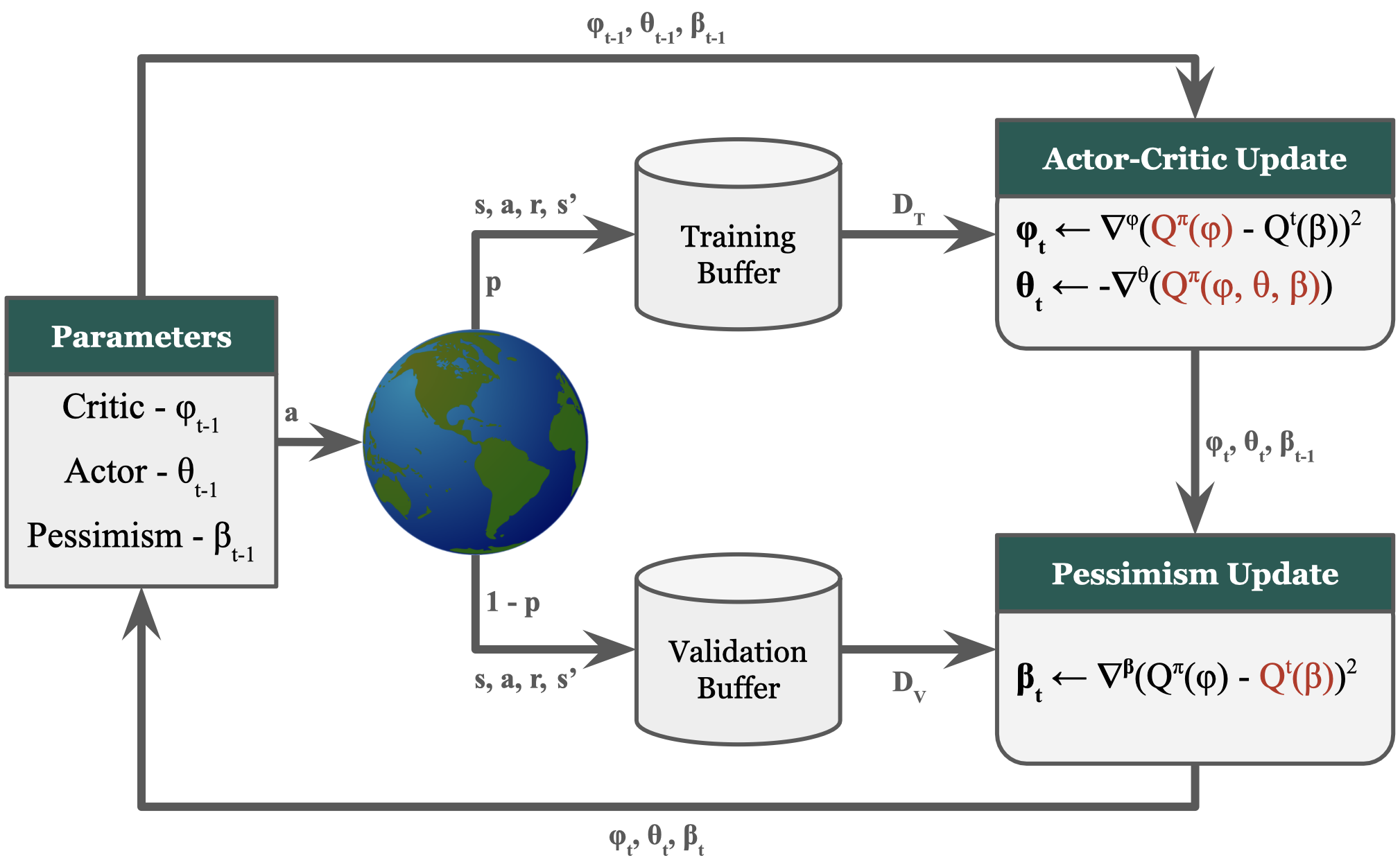}
    \end{subfigure}
\caption{High level overview of the proposed approach. After environment step, the transition is stored in either the training buffer (used for updating actor-critic modules) or the validation buffer (used for updating pessimism module). The pessimism is updated via a "reverse" TD loss, optimisation of which on the training buffer would be prone to overfitting.}
\label{fig:cycle2}
\end{center}
\end{figure}

In this context, $\beta \in (0, \infty)$ is a continuous parameter defining the level of pessimism and the true Q-value is represented by $Q^{\pi}(s,a)$. Since the term is not squared, $\beta$ cannot be trivially optimized by setting it to zero. GPL and OPL focus on aligning pessimism with the error in the pessimistic objective approximation. Since the true Q-values are unknown, they must be estimated. GPL assumes that the critic's output is unbiased for off-policy actions (ie. $Q^{\pi}(s,a) = Q_{\phi}(s,a)$) and calculates the dual optimization pessimism loss using transitions from the replay buffer. However, this approach can lead to overfitting as it relies heavily on the critic output. In contrast, OPL estimates $Q^{\pi}(s,a)$ via $\lambda$-returns calculated using recent transitions, bootstrapped by the critic, which reduces the risk of overfitting. Nevertheless, due to frequent policy updates, even recent transitions may be off-policy. A general limitation of the dual optimization method is that the pessimism adjustment does not correlate with the critic disagreement for specific state-action pairs thus impairing the impact of potential changes to $\beta$. A different strategy, Tactical Optimism and Pessimism (TOP) \citep{moskovitz2021tactical}, adjusts pessimism using an external bandit controller to maximize online episodic rewards. However, this controller is discrete and less effective as possible amount of pessimism values are increased. The further discuss the existing approaches for online pessimism adjustment in Appendix \ref{appendix_related} and summarize key characteristics in Table \ref{table:pessimism_methods}.

\section{Approximation Error and Pessimism}
\label{section_approximationerror}

In this section, we focus on the analysis of critic approximation errors within the framework of pessimistic updates. For simplicity, we consider a fixed policy $\pi_{\theta}$ and use $V(s)$ and $Q(s,a)$ to represent the value and Q-value under this policy. We define the mean and lower bound approximation errors denoted as $U_{\phi}^{\mu}$ and $U_{\phi}^{lb}$ respectively:

\begin{equation}
\label{eq:ensemble_approximation_error}
\begin{split}
    U_{\phi}^{\mu}(s,a) & \triangleq Q(s,a) - Q^{\mu}_{\phi}(s,a), \\
    U_{\phi}^{lb}(s,a) & \triangleq Q(s,a) - Q^{lb}_{\phi}(s,a). \\
\end{split}
\end{equation}

Here, $Q(s,a)$ denotes the true Q-value, the term $Q^{\mu}_{\phi}(s,a)$ represents the mean Q-value estimated by an ensemble of $k$ critics, calculated as $Q^{\mu}_{\phi}(s,a) = \frac{1}{k} \sum^{k} Q^{i}_{\phi}(s,a)$, and $Q^{lb}_{\phi}(s,a)$ is the lower bound Q-value as defined in Equation \ref{eq:cdql_general}. Additionally, we introduce the mean and lower bound temporal critic errors, denoted as $u^{\mu}_{\phi}$ and $u^{lb}_{\phi}$, respectively:

\begin{equation}
\label{eq:ln_approximation_error}
\begin{split}
u^{\mu}_{\phi}(s, a, s') & \triangleq r_{s,a} + \gamma V^{\mu}_{\phi}(s') - Q^{\mu}_{\phi}(s,a), \\
u^{lb}_{\phi}(s, a, s') & \triangleq r_{s,a} + \gamma V^{lb}_{\phi}(s') - Q^{\mu}_{\phi}(s,a). \\
\end{split}
\end{equation}

These temporal critic errors quantify the deviation between the Q-values $Q^{\mu}_{\phi}(s,a)$ and the mean or lower bound Temporal Difference (TD) targets. The value $V_{\phi}^{lb}(s)$ is equal to the expected value of $Q_{\phi}^{lb}(s,a)$ over all state-action pairs under policy $\pi$, such that $V_{\phi}^{lb}(s) = \mathrm{E}_{\pi} Q_{\phi}^{lb}(s,a) - \log \pi_{\theta}(a|s)$. 

\begin{lemma}[Approximation error operator] Given policy $\pi$, $k$ on-policy q-value approximations $Q_{\phi}^{1}, Q_{\phi}^{2}, ..., Q_{\phi}^{k}$, sample mean $Q_{\phi}^{\mu}$ and standard deviation $Q_{\phi}^{\sigma}$, the mean and lower bound approximation errors follow a recursive formula:

\label{lemma:1}
\begin{equation*}
\label{eq:lemma11}
\begin{split}
    & U_{\phi}^{\mu}(s,a) = u_{\phi}^{\mu}(s,a,s') + \gamma \underset{a' \sim \pi}{\mathrm{E}} U_{\phi}^{\mu}(s',a'), \\
    & U_{\phi}^{lb}(s,a) = u_{\phi}^{lb}(s,a,s') + \beta Q_{\phi}^{\sigma}(s,a) + \gamma \underset{a' \sim \pi}{\mathrm{E}} U_{\phi}^{lb}(s',a'), \\
    & U_{\phi}^{lb}(s,a) = U_{\phi}^{\mu}(s,a) + \beta Q_{\phi}^{\sigma}(s,a).
\end{split}
\end{equation*}
\end{lemma}

We expand on Lemma \ref{lemma:1} in Appendix \ref{appendix_derivations}. The lemma reveals that approximation errors exhibit a recurrent pattern analogous to Q-values. Specifically, the temporal errors function as an immediate signal, akin to rewards, while the future approximation errors serve as the bootstrap signal. Furthermore, this observation formalizes the intuitive concept that minimizing the lower-bound approximation error necessitates a precise calibration of the pessimistic correction against the temporal error and the approximation errors of subsequent states. It can be shown that similarly to the Bellman operator, both mean and lower bound error approximation operators are monotonic contractions:

\begin{theorem}[Approximation error contraction] Let $\mathcal{F}$ be the space of functions on domain $S \times A$. We define the mean error and lower bound error operators $\mathcal{U}^{\mu}, \mathcal{U}^{lb}: \mathcal{F} \rightarrow \mathcal{F}$ as: 

\label{lemma:2} 
\begin{equation*}
\label{eq:operator_def}
\begin{split}
    & \mathcal{U}^{\mu} \bigl(f(s,a)\bigr) \triangleq  u_{\phi}^{\mu}(s,a,s') + \gamma \underset{a' \sim \pi}{\mathrm{E}} f(s',a'), \\
    & \mathcal{U}^{lb} \bigl(f(s,a)\bigr) \triangleq  u_{\phi}^{lb}(s,a,s') + \beta Q_{\phi}^{\sigma}(s,a) + \gamma \underset{a' \sim \pi}{\mathrm{E}} f(s',a').
\end{split}
\end{equation*}

Above, $f(s,a): S \times A \rightarrow \mathbb{R}$ represents an estimate of the approximation error. Then it follows that both $\mathcal{U}^{\mu}$ and $\mathcal{U}^{lb}$ are monotonic contractions for any $f_1$ and $f_2$:

\begin{equation*}
\label{eq:contraction}
\begin{split}
    ||\mathcal{U} (f_{1}) - \mathcal{U} (f_2) ||_{\infty} & \leq \gamma ||f_1 - f_2||_{\infty}.
\end{split}
\end{equation*}
\end{theorem}

We provide the relevant derivations in Appendix \ref{appendix_derivations}. As follows from Theorem \ref{lemma:2}, repeated application of the approximation error operator yields a Cauchy sequence, and therefore leads to a fixed point:

\begin{corollary}[Approximation error fixed point] We denote repeated $k$ applications of either approximation error operator to function $f$ as $\mathcal{U}_{k}(f)$. Then, due to Banach fixed point theorem:
\label{theorem:1}

\begin{equation*}
\label{eq:theorem1}
\begin{split}
    \mathcal{U}^{\infty}(f) = f^{*} ~~ \wedge ~~ \mathcal{U}(f^{*}) = f^{*}.
\end{split}
\end{equation*}
\end{corollary}

The corollary shows that the approximation error of values can be effectively modeled using a fixed-point approach, analogous to the treatment of values themselves. The potential ramifications and applications of this concept are further explored in Appendix \ref{appendix_derivations}. Principally, the convergence of a pessimistic value model signifies that the approximation errors converge to zero, implying $U_{\phi}^{\mu} = U_{\phi}^{lb} = 0$. The convergence proof of CDQL indicates that the value model should align with the true on-policy values under the conventional Q-learning convergence assumptions \citep{watkins1992q, fujimoto2018addressing}. Lemma \ref{lemma:1} explicitly shows that for all $s$, $a$ and $s'$, both approximation errors equate to zero iff the following conditions are satisfied:

\begin{equation}
\label{eq:zero_error}
\begin{split}
    Q_{\phi}^{\mu}(s,a) = r + \gamma V_{\phi}^{\mu}(s') ~~ \wedge ~~ \beta Q_{\phi}^{\sigma}(s,a) = 0.
\end{split}
\end{equation}

Consequently, the convergence of a pessimistic model necessitates either the absence of critic ensemble disagreement (i.e., $Q_{\phi}^{\sigma}(s,a) = 0$ for all state-action pairs) or an algorithmic ability to diminish the level of pessimism over time, culminating in $\beta = 0$ asymptotically. Figure \ref{fig:critic_sigma} shows that the critic disagreement does not completely diminish on popular DeepMind Control and MetaWorld benchmarks. Given the improbability of achieving zero critic disagreement in overparameterized deep RL contexts, the adjustment of $\beta$ emerges as a compelling strategy. Additionally, it can be demonstrated that under the scenario of critic underestimation, the lower-bound approximation error exceeds the mean approximation error:

\begin{equation}
    U_{\phi}^{\mu}(s,a) > 0 \implies |U_{\phi}^{\mu}(s,a)| \leq |U_{\phi}^{lb}(s,a)|.
\end{equation}

As follows, pessimistic learning is advantageous only in overestimation, whereas it becomes detrimental in cases of underestimation. To this end, the pessimism levels should be adjusted in tandem with changes in the approximation errors. In practical terms, achieving a zero approximation error for either mean or lower bound is an unrealistic. Given that $U_{\phi}(s,a) \in \mathbb{R}$, one might be interested in optimization of norm of $U_{\phi}^{\mu}(s,a)$ or $U_{\phi}^{lb}(s,a)$. This leads to the possibility of defining an "optimal" level of pessimism, where optimality is considered in relation to minimizing the respective approximation error norm. We note that our analysis yields a different approach to updating pessimism as compared to the method derived from dual optimization \citet{cetin2023learning}, which we discuss in Section \ref{sec:pessimism_adjustment}.

\section{Validation Pessimism Learning Algorithm}
\label{sec:vpl}

Building on the analysis conducted in the previous Section, we propose Validation Pessimism Learning module (VPL). The goal of the VPL module is to adjust the pessimism parameter such that the critic targets (lower bound Q-value approximation) has the least approximation error. As such, VPL can be used as an alternative to CDQL or GPL in conjuction with any off-policy actor-critic algorithm. For our analysis, we utilize the Soft Actor-Critic (SAC) \citep{haarnoja2018soft} as the backbone algorithm. VPL is based on a simple premise of adjusting pessimism via a TD loss. Given that this loss function is concurrently optimized by the critic, such setup is especially prone to overfitting. To mitigate this, the optimization of the pessimism parameter is conducted on a distinct set of \textit{validation} data, which remains unseen by the actor-critic modules. From a theoretical standpoint, VPL can be interpreted as a strategy for pessimism model selection, with the selection process aimed at minimizing the lower bound approximation error delineated in the previous section. A critical aspect of VPL involves conducting the pessimism model selection on validation data. The model selection is achieved through gradient-based optimization of the proposed pessimism loss. The utilization of validation data in this process reduces the probability of overfitting to bootstrapped supervision signals used by TD learning.  We summarize VPL approach in Figure \ref{fig:cycle2} and pseudo-code below, where we colour changes wrt. regular SAC.

\begin{algorithm}[ht!]
\setstretch{1.05}
   \caption{Validation Pessimism Learning Step}
   \label{alg:algo}
\begin{algorithmic}[1]
   \STATE {\bfseries Input:} $\pi_{\theta}$ - actor; $Q_{\phi}$ - critic; $\alpha$ - temperature; $\mathcal{D}_{T}$ - replay buffer; \textcolor{BrickRed}{$\beta$ - pessimism}; \textcolor{BrickRed}{$\mathcal{D}_{V}$ - validation buffer}
   \STATE {\bfseries Hyperparameters:} $B$ - batch size; \textcolor{BrickRed}{$v$ - validation rate}
   \STATE $s', r = \textsc{env.step}(a) \quad with \quad a \sim \pi_{\theta}(a|s)$
   \STATE \textcolor{BrickRed}{$p \sim U(0,1)$}
   \IF{$p > v$:}
    \STATE $\mathcal{D}_{T}\textsc{.add}(s,a,r,s')$
    \ENDIF
   \IF{$p \leq v$:}
    \STATE \textcolor{BrickRed}{$\mathcal{D}_{V}\textsc{.add}(s,a,r,s')$}
    \ENDIF
   \FOR{$i=1$ {\bfseries to} ReplayRatio}
   \STATE $s, a, r, s' \sim \mathcal{D}_{T}\textsc{.sample(B)}$ 
   \STATE \textcolor{BrickRed}{$s_V, a_V, r_V, s_{V}' \sim \mathcal{D}_{V}\textsc{.sample(vB)}$}
   \STATE $\phi \xleftarrow{} \phi - \nabla_{\phi} \bigl(Q^{\pi}_{\phi}(s,a) - r - \vartriangle \gamma V^{lb}_{\phi}(s')\bigr)^{2}$
   \STATE $\theta \xleftarrow{} \theta + \nabla_{\theta}V^{\pi}_{\theta}(s)$
   \STATE $\alpha \xleftarrow{} \alpha - \nabla_{\alpha} \alpha(-\log\pi(a|s) - \mathcal{H}^{*})$ 
   \STATE \textcolor{BrickRed}{$\beta \xleftarrow{} \beta - \nabla_{\beta} \bigl(Q^{\pi}_{\phi}(s_{V},a_{V}) - r_{V} - \gamma V^{lb}_{\phi}(s_{V}')\bigr)^{2}$}
   \ENDFOR
\end{algorithmic}
\end{algorithm}

\subsection{Validation Buffer}
\label{sec:vpl_valid}

\begin{figure*}[ht!]
\begin{center}
    \begin{subfigure}{0.245\linewidth}
    \caption{Regret from validation}
    \includegraphics[width=\textwidth]{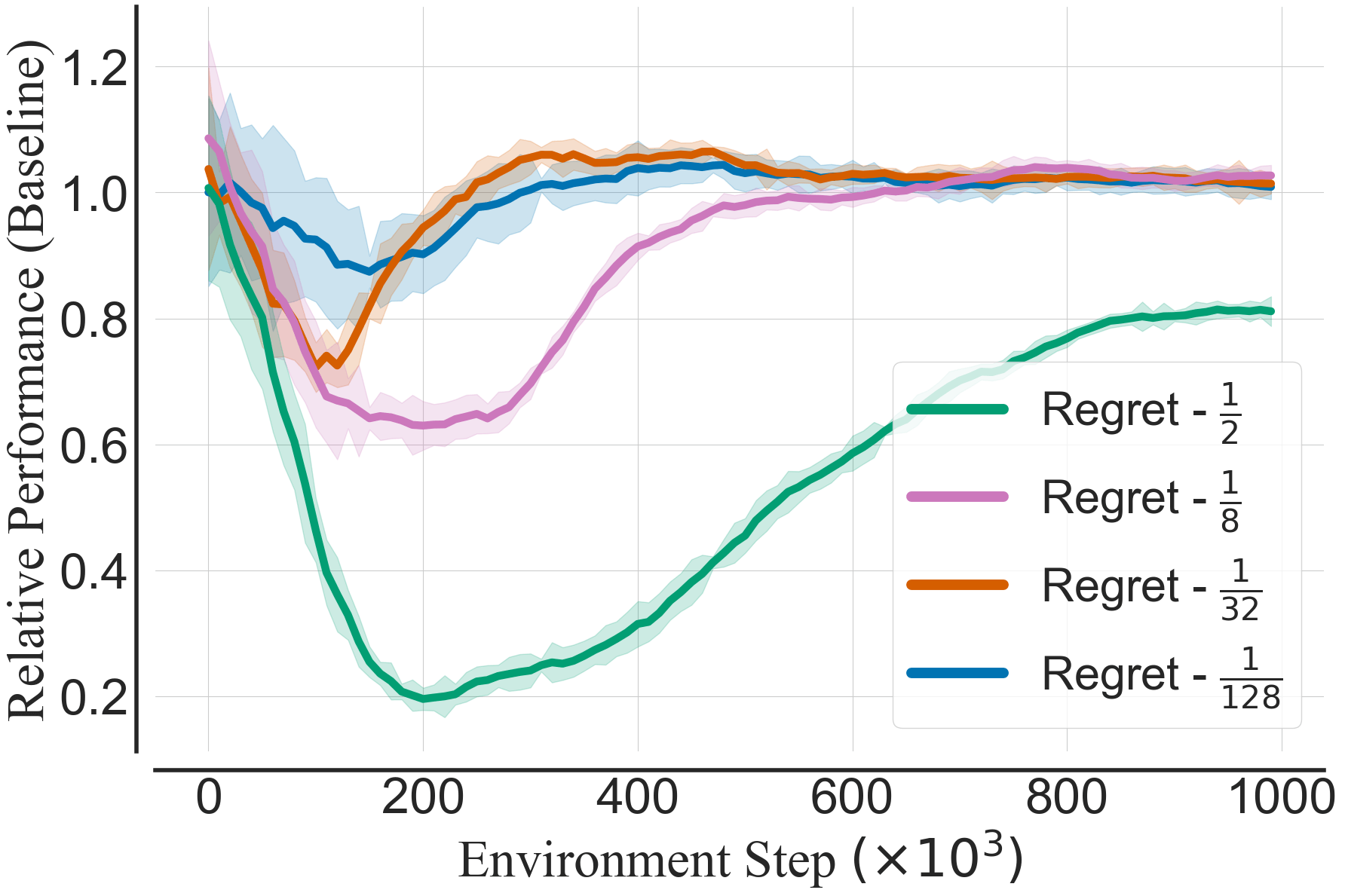}
    \label{fig:buffer1}
    \end{subfigure}
    \hfill
    \begin{subfigure}{0.245\linewidth}
    \caption{Gain from adjustment}
    \includegraphics[width=\textwidth]{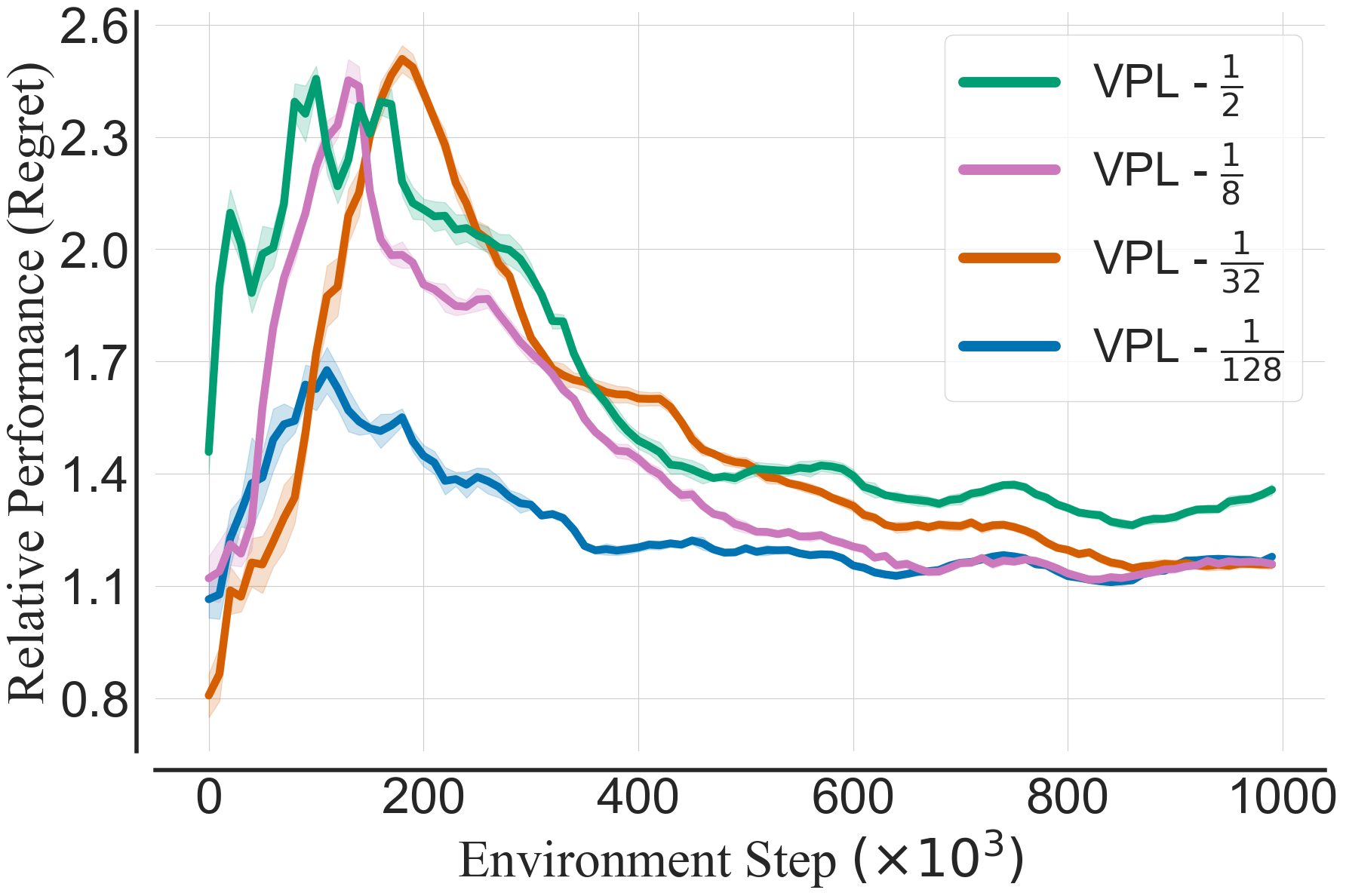}
    \label{fig:buffer2}
    \end{subfigure}
    \hfill
    \begin{subfigure}{0.245\linewidth}
    \caption{Combined effect (normalized)}
    \includegraphics[width=\textwidth]{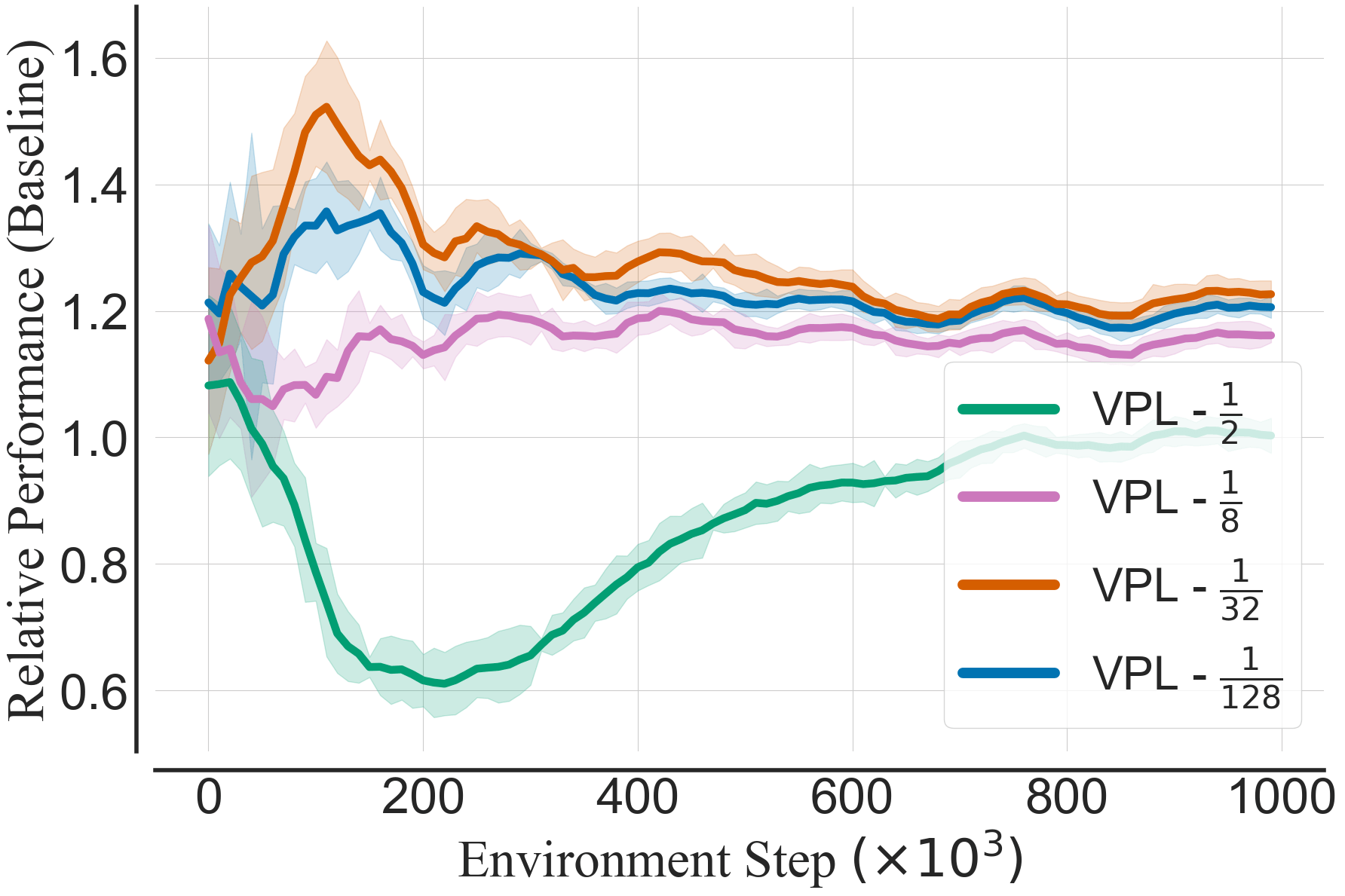}
    \label{fig:buffer3}
    \end{subfigure}
    \hfill
    \begin{subfigure}{0.245\linewidth}
    \caption{Combined effect}
    \includegraphics[width=\textwidth]{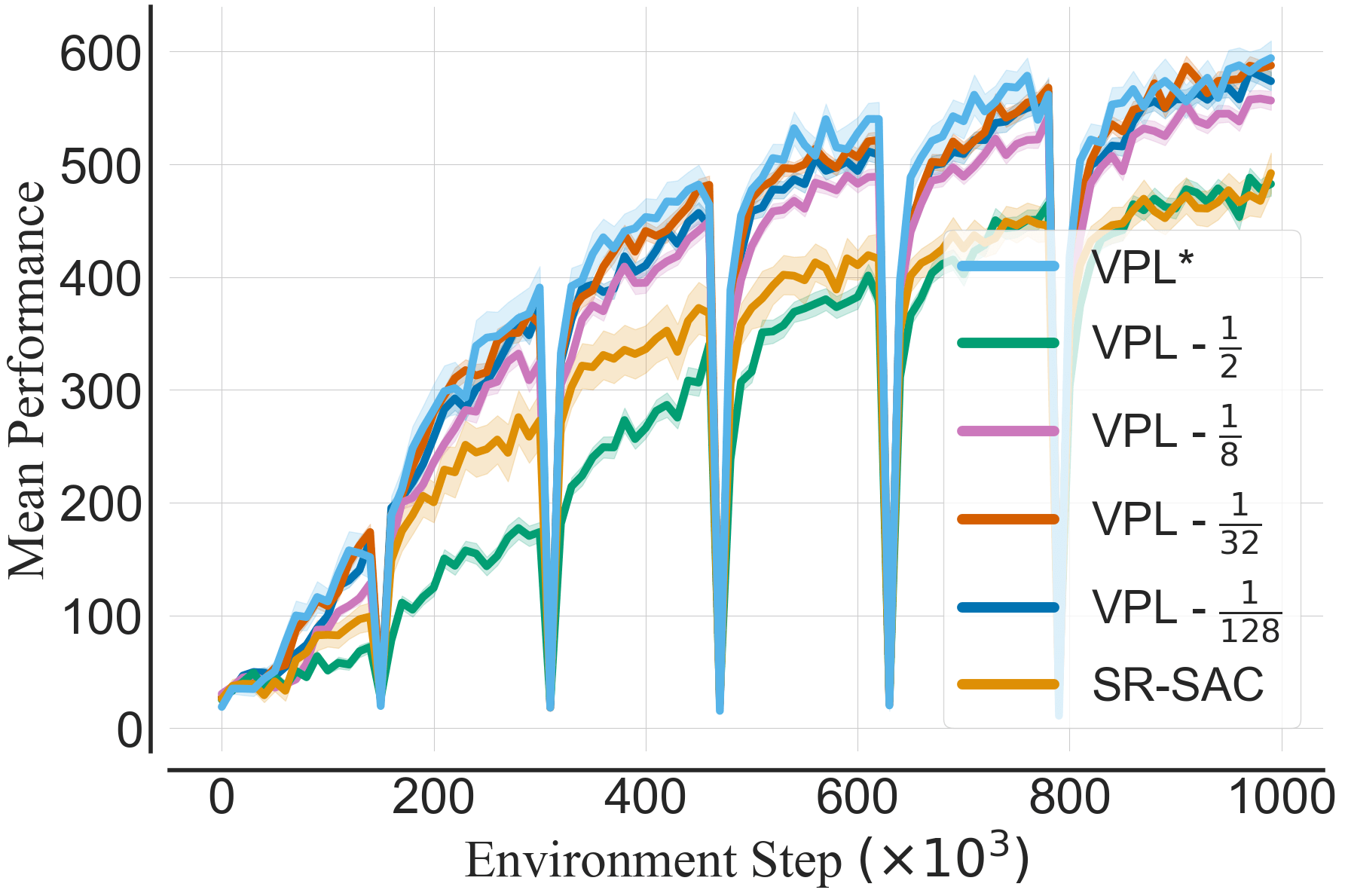}
    \label{fig:buffer4}
    \end{subfigure}
\vspace{-0.2in}
\caption{We measure the impact of maintaining a validation buffer on performance separated from the benefits accrued through pessimism adjustment, across varying proportions of validation samples. Figure \ref{fig:buffer1} shows whether validation agents can match the performance of their validation-free counterparts, despite not utilizing validation samples for pessimism updates. This comparison allows us to quantify the regret associated with allocating a portion of samples to a validation buffer. We find that such regret is completely diminished during training. We observe an exception in the case of 50\% validation samples, where it fails to match the performance of its validation-free agent. Figure \ref{fig:buffer2} measures the performance gains attributable to pessimism adjustment. This is assessed by contrasting the performance of validation agents that do not update pessimism against those that do. We observe that the performance gain stemming from updating pessimism is dependent on the size of the validation dataset, with majority of the possible performance gain already achieved by the 3\% validation ratio. Figures \ref{fig:buffer3} \& \ref{fig:buffer4} picture the cumulative effect of validation pessimism adjustment for different validation ratios, benchmarking it against the baseline performance of SR-SAC and VPL agent with "free" validation (denoted as VPL*). The results show that all configurations except 50\% validation samples outperform the baseline SR-SAC. We detail this experimental setting in Section \ref{sec:experiments_buff}.}
\label{fig:buffer_effects}
\end{center}
\vspace{-0.2in}
\end{figure*}

The employment of validation data is a well-established practice in supervised learning frameworks \citep{bishop2006pattern}. It serves a dual purpose: providing an unbiased assessment of model performance trained on the training dataset, and facilitating regularization techniques such as early stopping \citep{prechelt2002early} or hyperparameter tuning \citep{bergstra2012random}. However, the integration of validation data entails a trade-off, notably the reduction of the training set size. In supervised learning, the regret associated with decreasing the training set can be quantitatively evaluated through the lens of neural scaling laws \citep{rosenfeld2019constructive}. Such regret is, to the best of our knowledge, a relatively understudied area in the context of online RL. In online RL, the notion of a validation buffer is not popular, primarily due to the requisite sacrifice of actor-critic learning on the validation transitions. Given inherent sample inefficiency of RL, this cost is often deemed as overly burdening. Contrary to supervised learning setup, RL is characterized by a high correlation between successive samples, thereby diminishing the marginal utility of processing additional samples from the same trajectory. Consequently, we posit that in online RL, the cost associated with the use of validation data can be counterbalanced, provided the validation data is leveraged to enhance the learning process. In the case of the VPL module, we allocate the validation transitions exclusively for the adjustment of the pessimism parameter. This approach presents a novel utilization of validation data in online RL in a manner that is RL-specific, diverging from traditional supervised methodologies.

\subsection{Pessimism Update Rule}
\label{sec:vpl_pess}

The persistence of critic disagreement throughout training implies that the standard convergence guarantees of the pessimistic temporal difference update towards on-policy values are not upheld when $\beta \neq 0$. Moreover, in cases where minimizing the mean approximation error is not achievable, particularly in scenarios characterized by strong overestimation, the presence of non-zero critic disagreement can be leveraged to decrease the lower bound approximation error by increasing $\beta$. This observation forms the basis for our proposed method of adjusting $\beta$. The aim is to minimize the expected lower bound approximation error $U_{\phi}^{lb}(s,a)$, formulated as follows:

\begin{equation}
\label{eq:pessimism_update_rule}
\begin{split}
    \beta^{*} & = \arg \min_{\beta} \underset{p_0, \pi}{\mathrm{E}} \sum_{t=0}^{\infty} \gamma^{t} U_{\phi}^{lb}(s,a).
\end{split}
\end{equation}

Unfortunately, obtaining $U_{\phi}^{lb}(s,a)$ is challenging as it necessitates an estimate of the true on-policy Q-value. Typically, such estimates are derived through methods like Monte-Carlo (MC) rollouts, TD(n), or TD($\lambda$), with MC being the only unbiased method. However, in the context of off-policy learning or non-terminating environments, employing MC rollouts is impractical. Consequently, we leverage the simple approach proposed by \citet{cetin2023learning} in which it is assumed that the critic output for prerecorded off-policy actions is unbiased. Therefore, we assume that $Q^{\pi}(s,a) = Q_{\phi}^{\mu}(s,a)$ for actions that do not maximize the output of the policy. Additionally, akin to the approach in off-policy actor-critic algorithms, the policy-induced distribution is approximated using an off-policy replay buffer. This approach leads to the formulation of the following:
 
\begin{equation}
\label{eq:pessimism_learning_vpl}
\begin{split}
    & \beta^{*} \approx \arg\min_{\beta} \underset{\mathcal{D}_{v}}{\mathrm{E}} \bigl( Q^{\mu}_{\phi}(s,a) - r_{s,a} - \gamma V^{lb}_{\phi}(s') \bigr)^2.
\end{split}
\end{equation}

\begin{figure*}[ht!]
\begin{center}
    \begin{subfigure}{0.99\linewidth}
    \centering
    \includegraphics[width=0.66\textwidth]{images/legend_1.png}
    \end{subfigure}
    \begin{subfigure}{0.135\linewidth}
    \includegraphics[width=\textwidth]{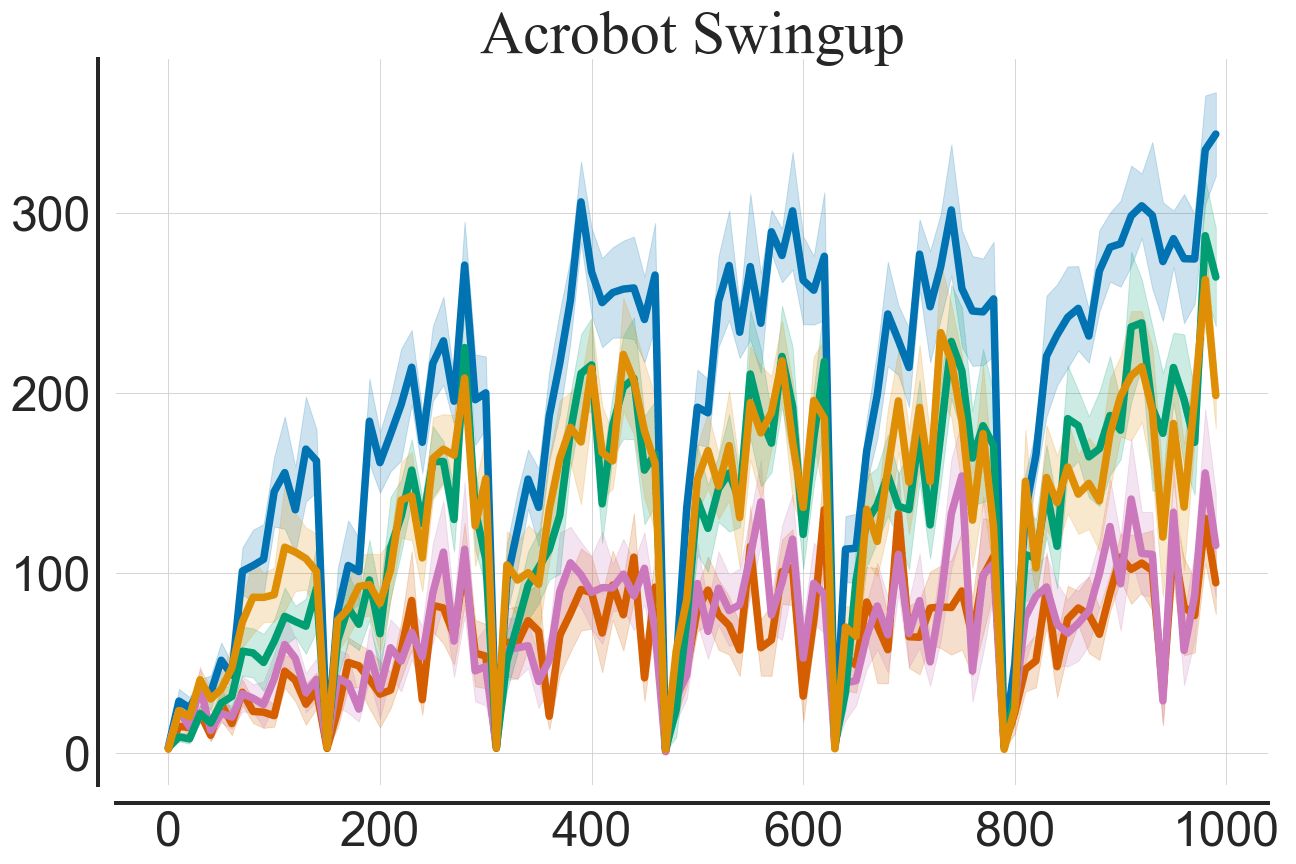}
    \end{subfigure}
    \begin{subfigure}{0.135\linewidth}
    \includegraphics[width=\textwidth]{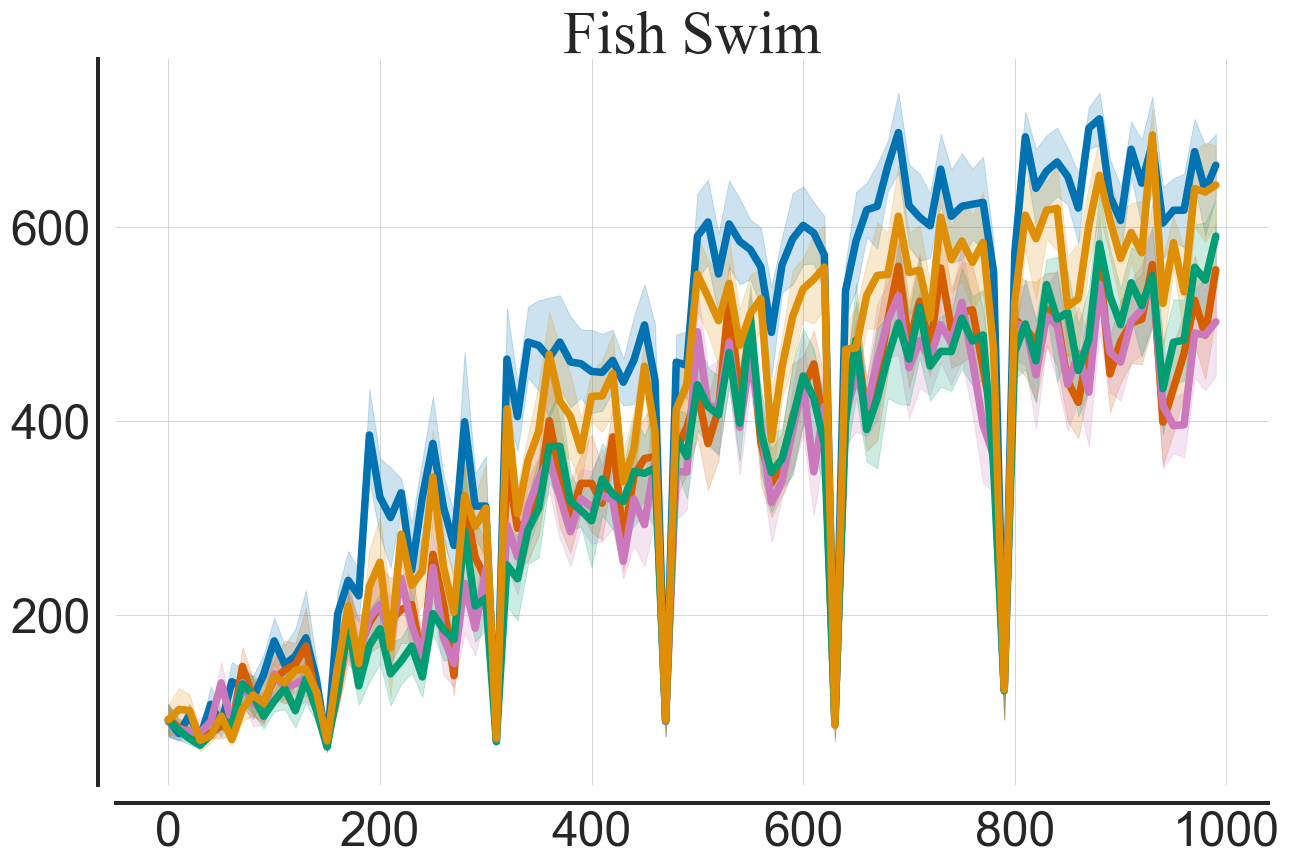}
    \end{subfigure}
    \begin{subfigure}{0.135\linewidth}
    \includegraphics[width=\textwidth]{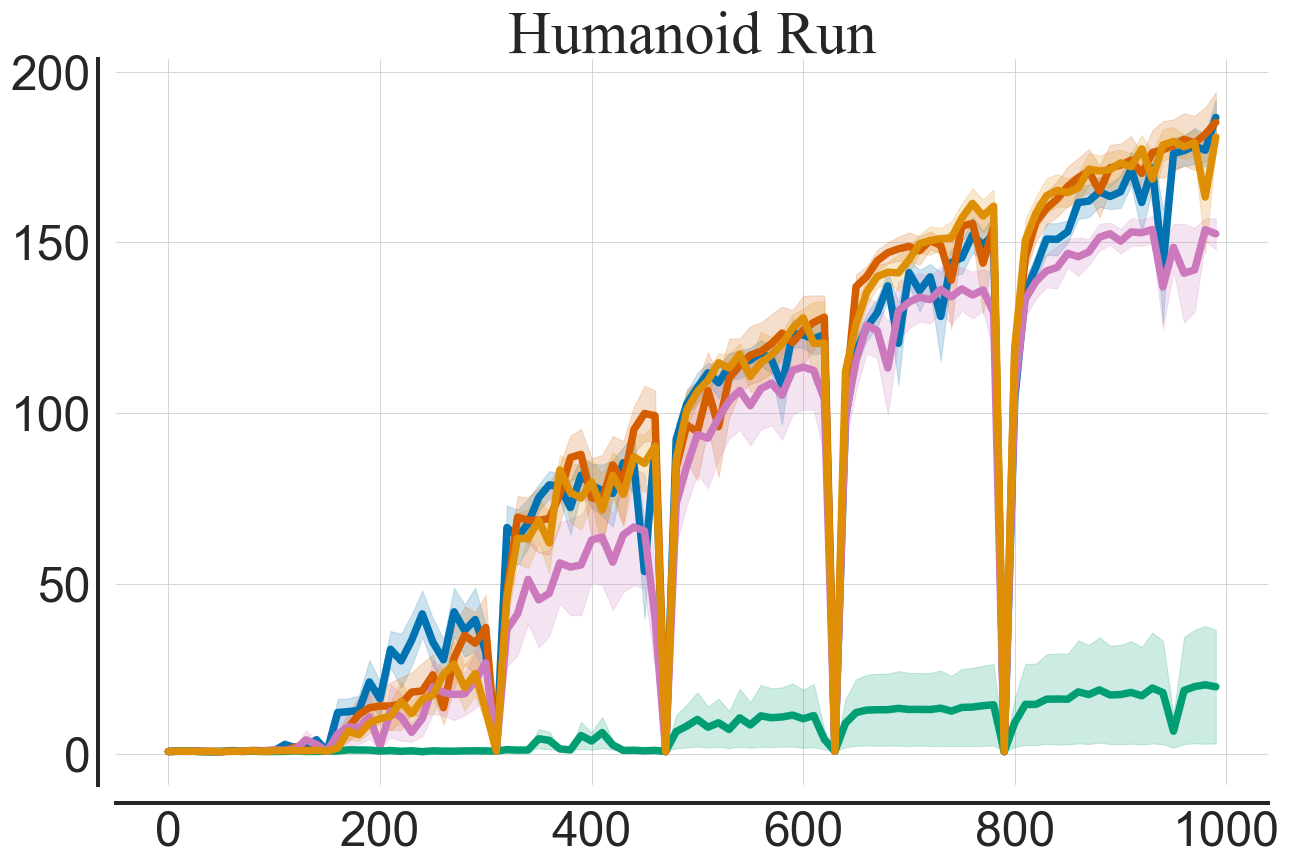}
    \end{subfigure}
    \begin{subfigure}{0.135\linewidth}
    \includegraphics[width=\textwidth]{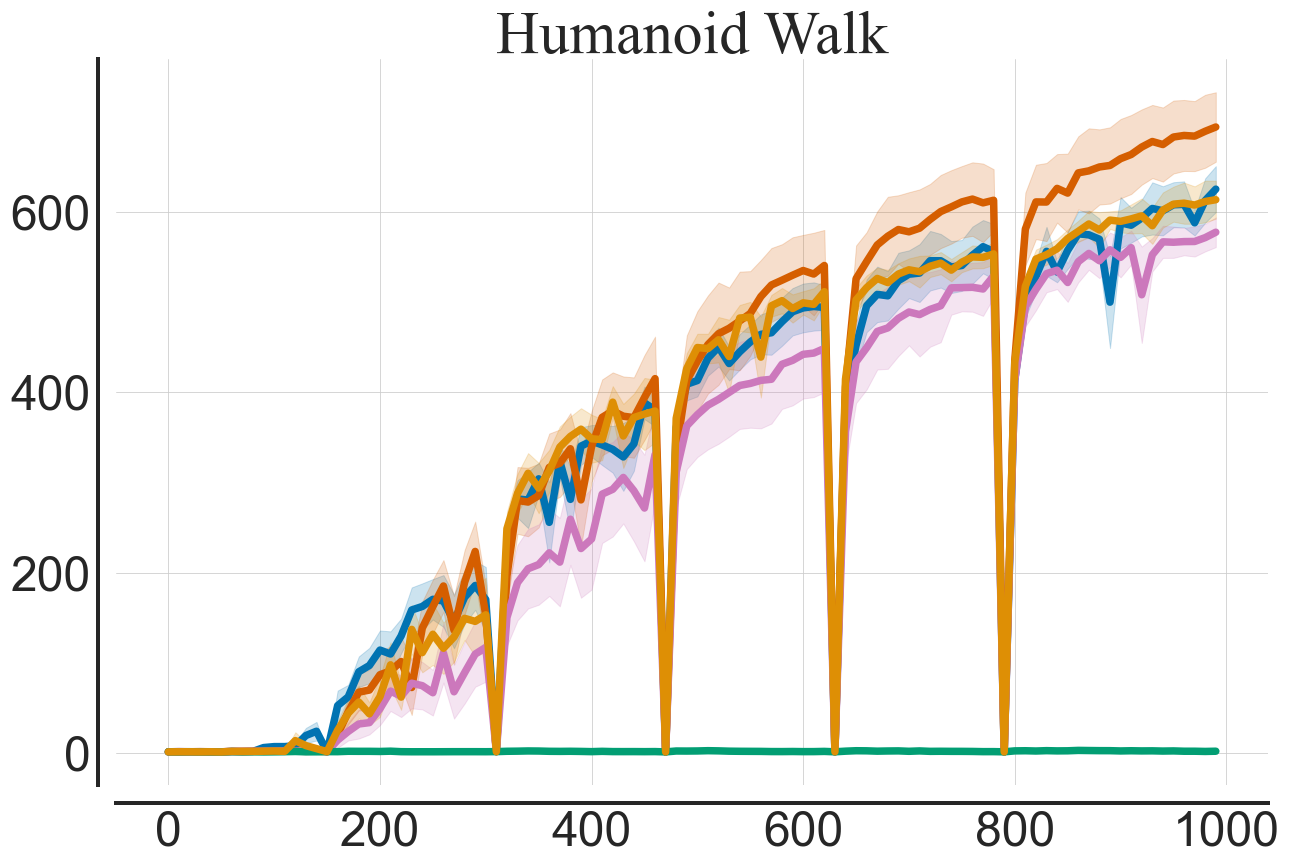}
    \end{subfigure}
    \begin{subfigure}{0.135\linewidth}
    \includegraphics[width=\textwidth]{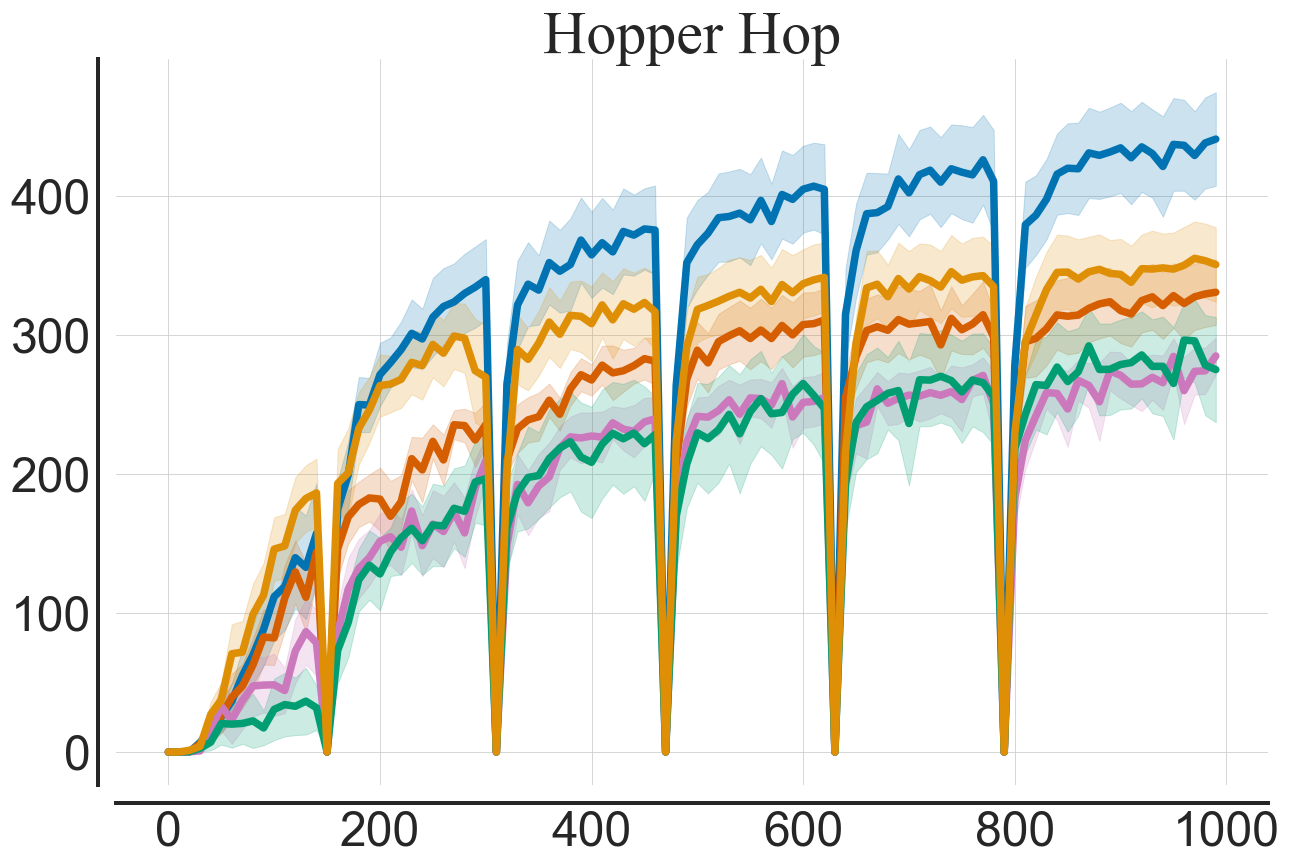}
    \end{subfigure}
    \begin{subfigure}{0.135\linewidth}
    \includegraphics[width=\textwidth]{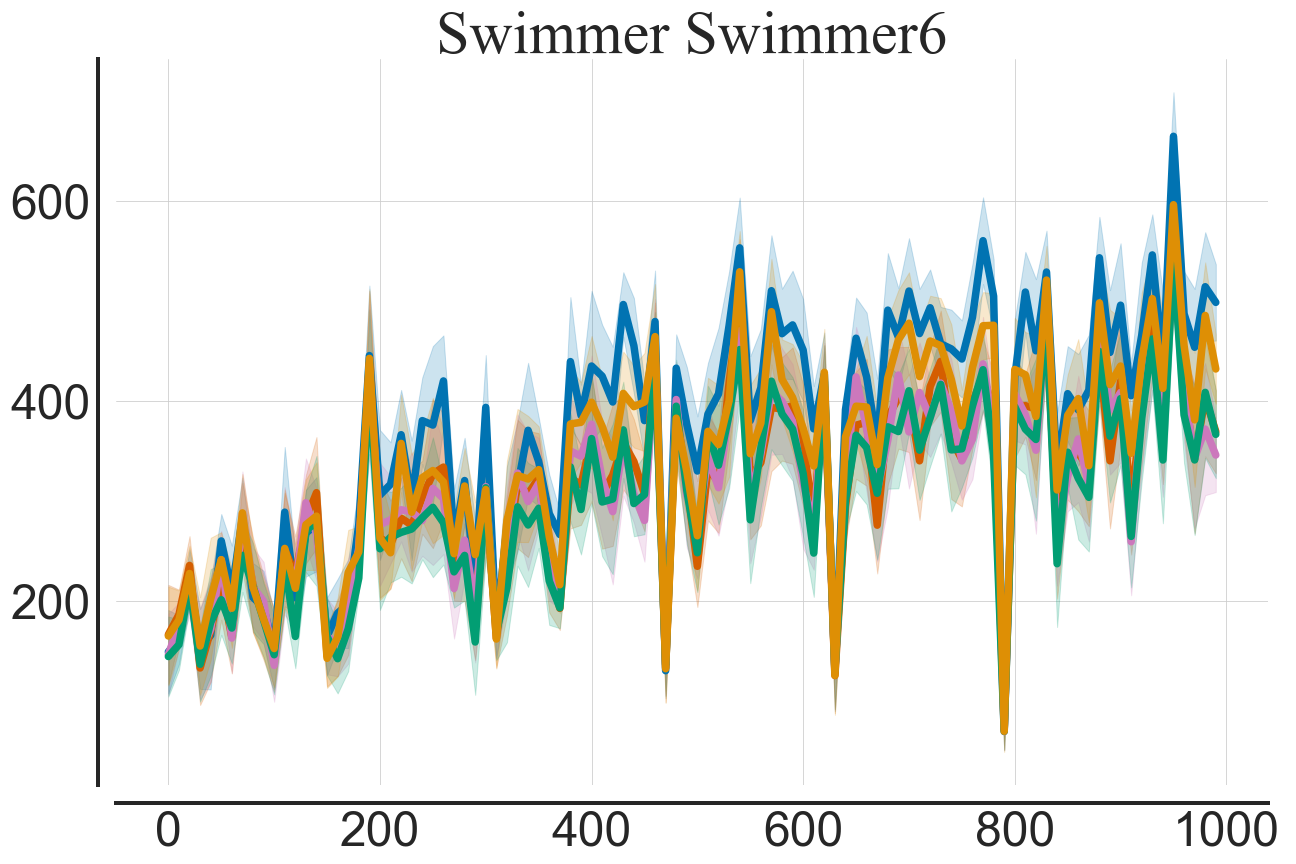}
    \end{subfigure}
    \begin{subfigure}{0.135\linewidth}
    \includegraphics[width=\textwidth]{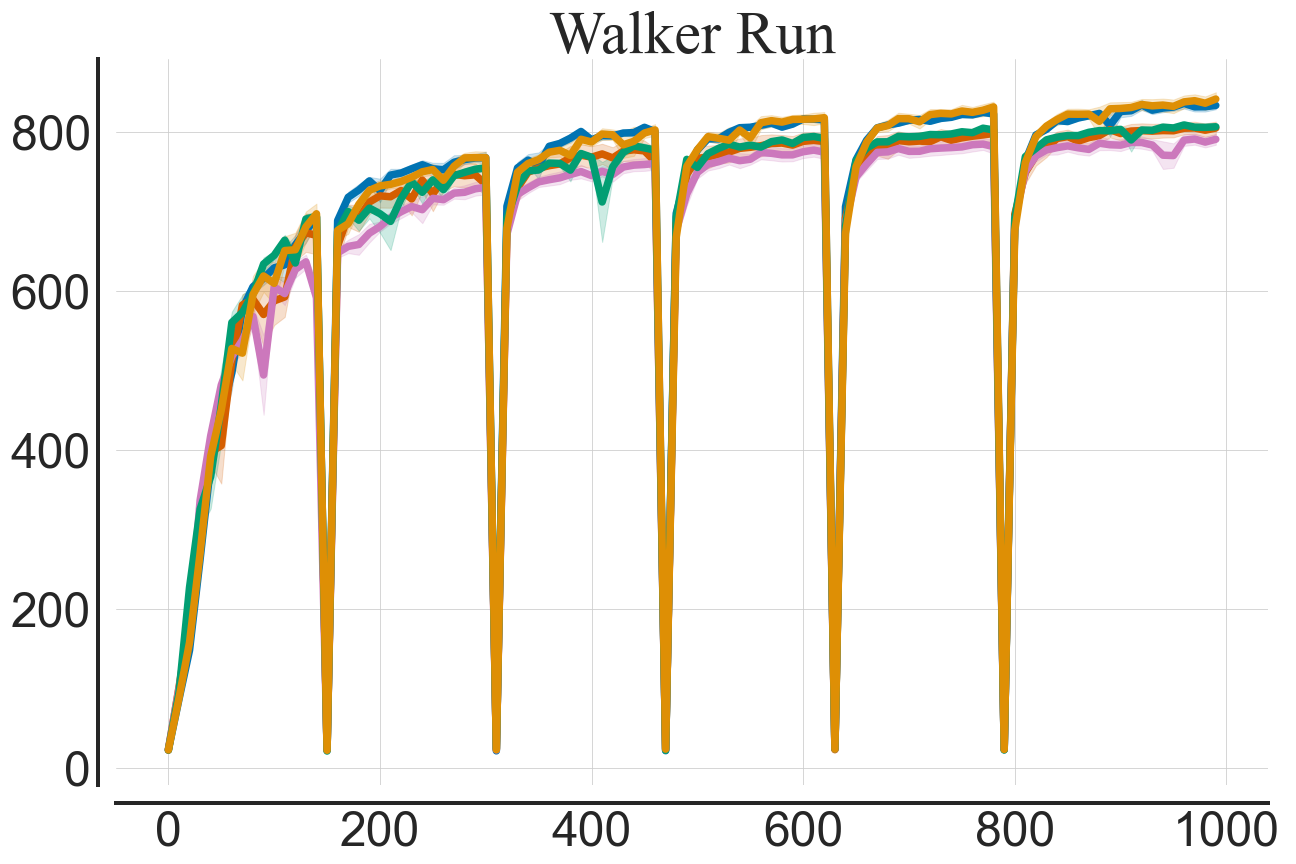}
    \end{subfigure}
    \begin{subfigure}{0.135\linewidth}
    \includegraphics[width=\textwidth]{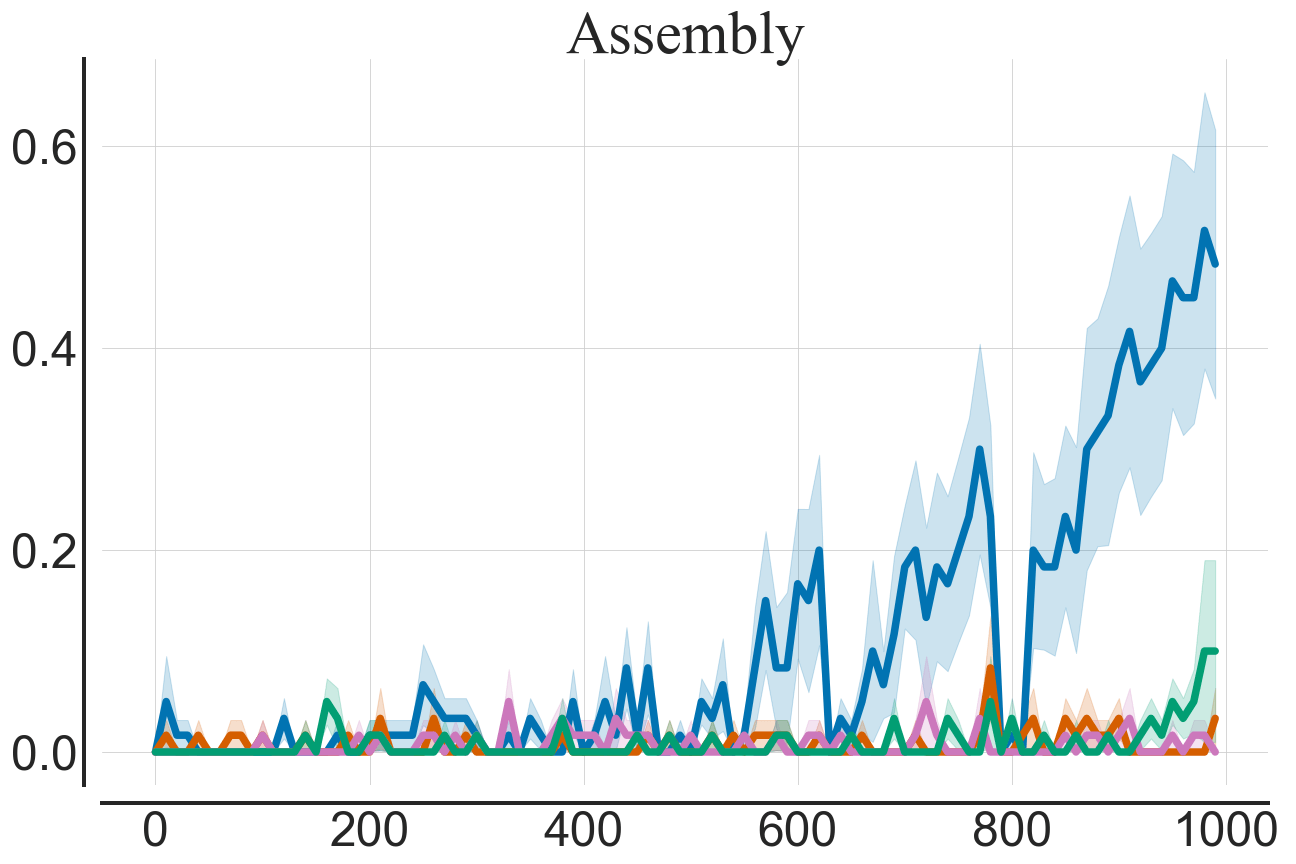}
    \end{subfigure}
    \begin{subfigure}{0.135\linewidth}
    \includegraphics[width=\textwidth]{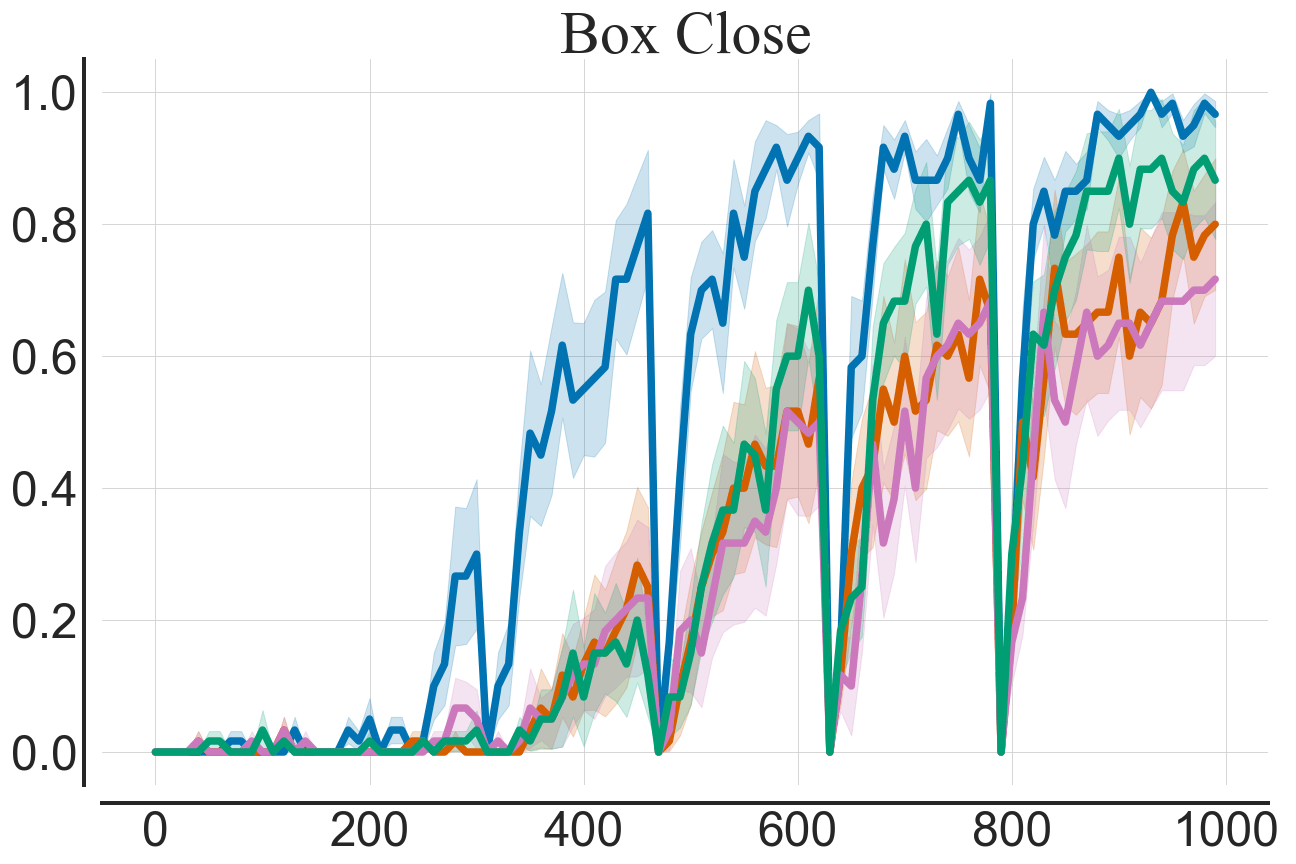}
    \end{subfigure}
    \begin{subfigure}{0.135\linewidth}
    \includegraphics[width=\textwidth]{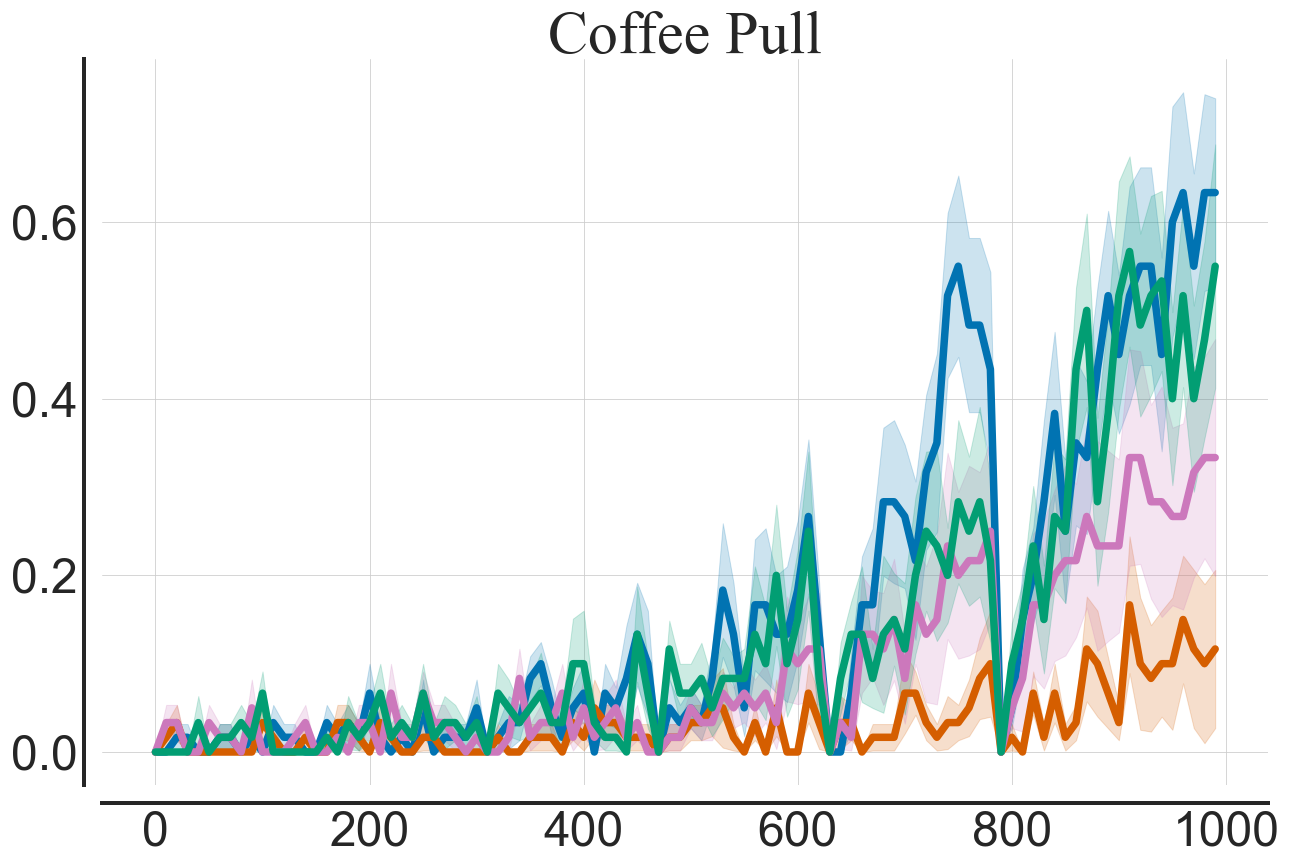}
    \end{subfigure}
    \begin{subfigure}{0.135\linewidth}
    \includegraphics[width=\textwidth]{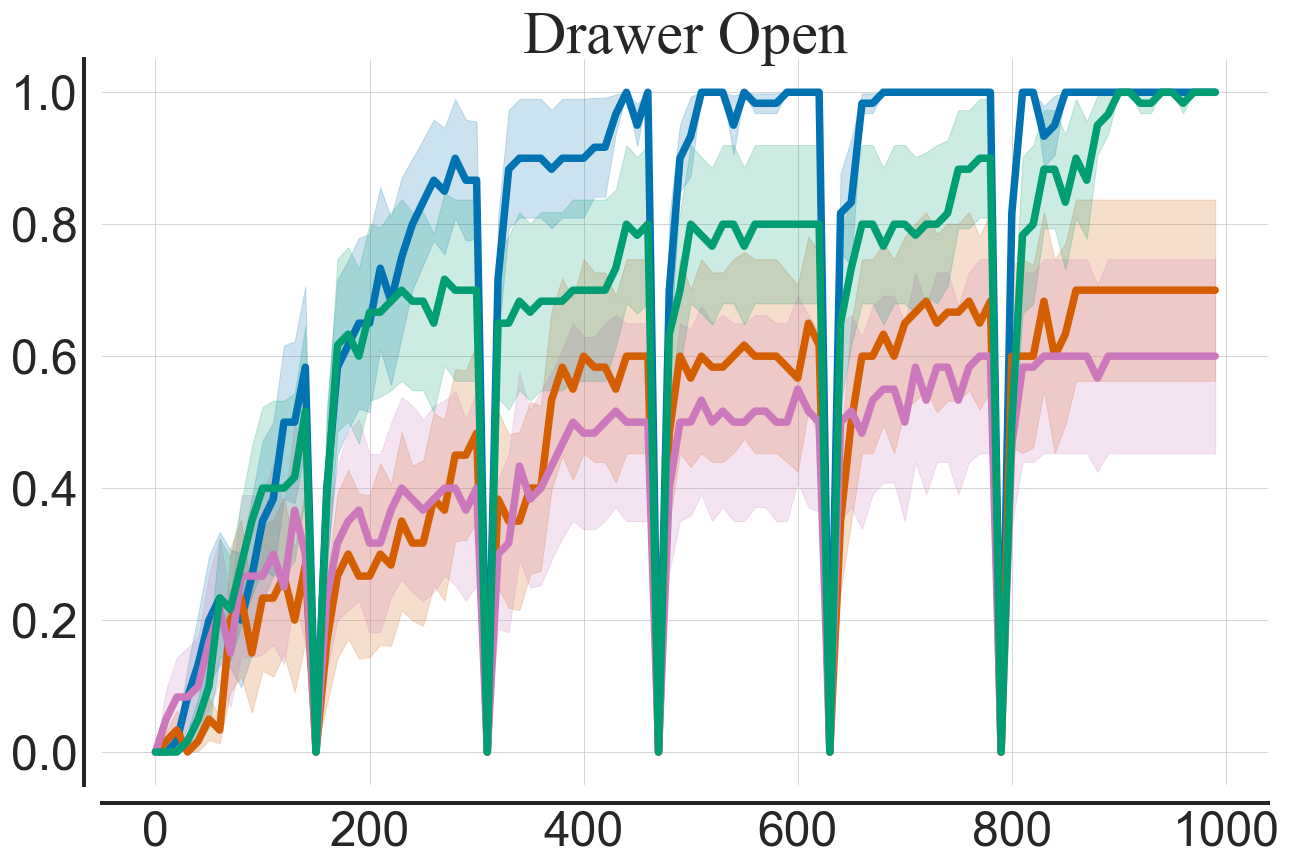}
    \end{subfigure}
    \begin{subfigure}{0.135\linewidth}
    \includegraphics[width=\textwidth]{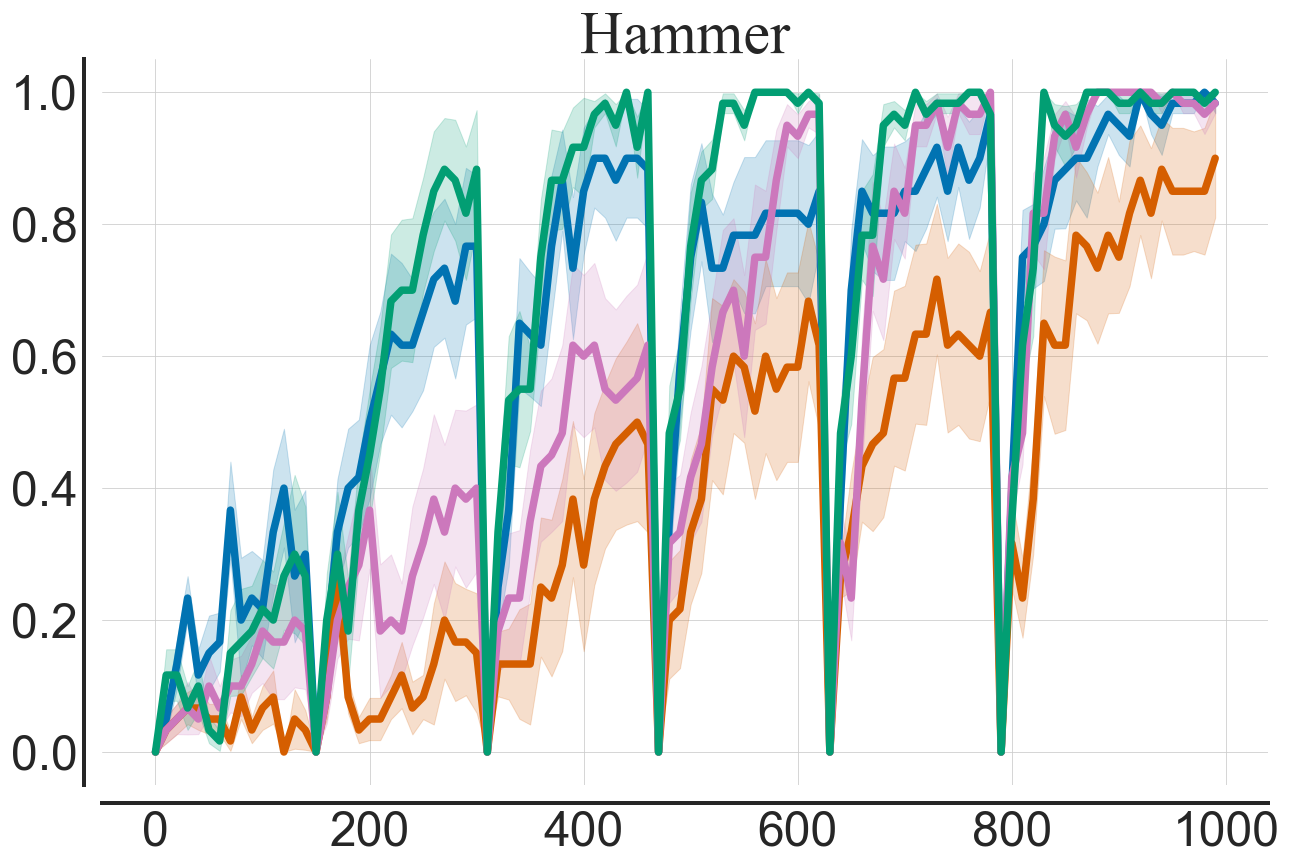}
    \end{subfigure}
    \begin{subfigure}{0.135\linewidth}
    \includegraphics[width=\textwidth]{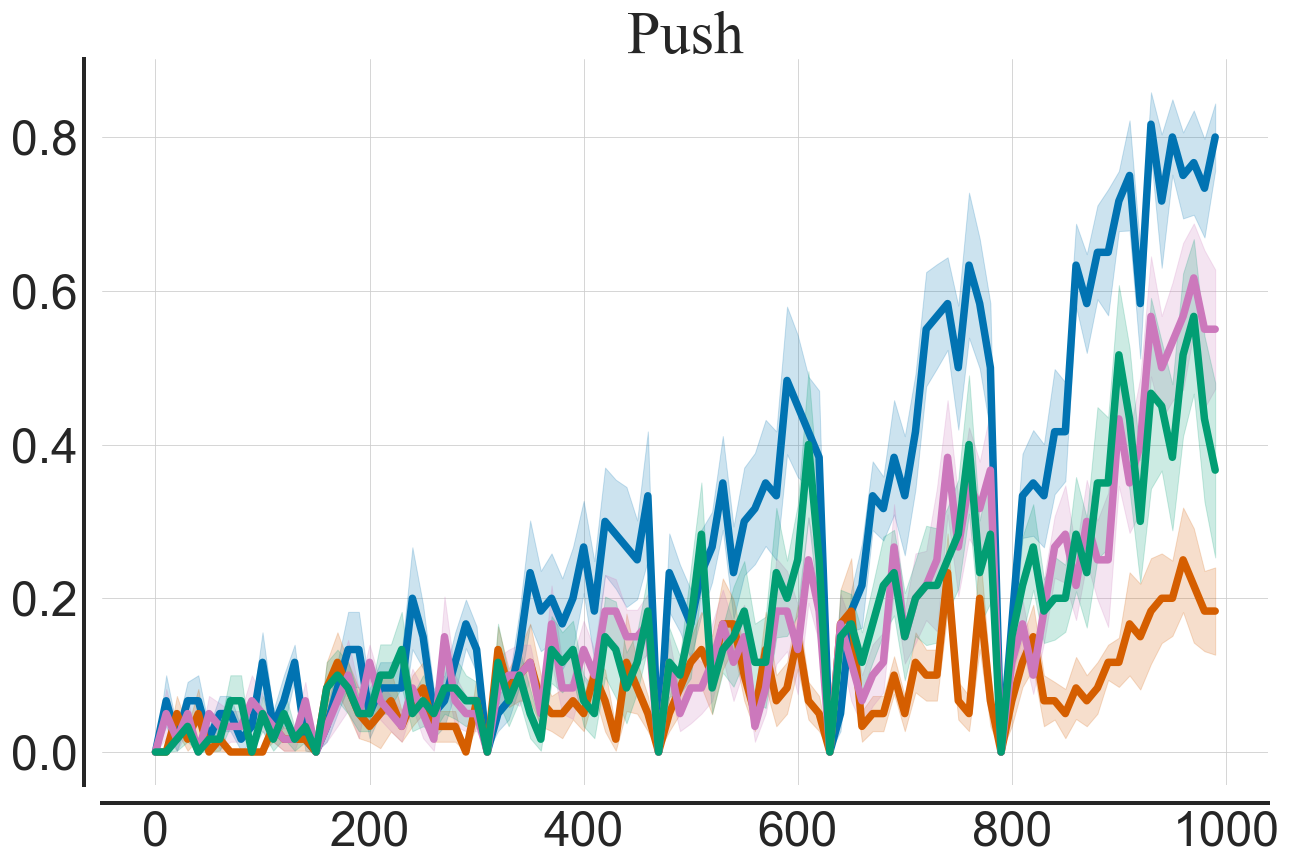}
    \end{subfigure}
    \begin{subfigure}{0.14\linewidth}
    \includegraphics[width=\textwidth]{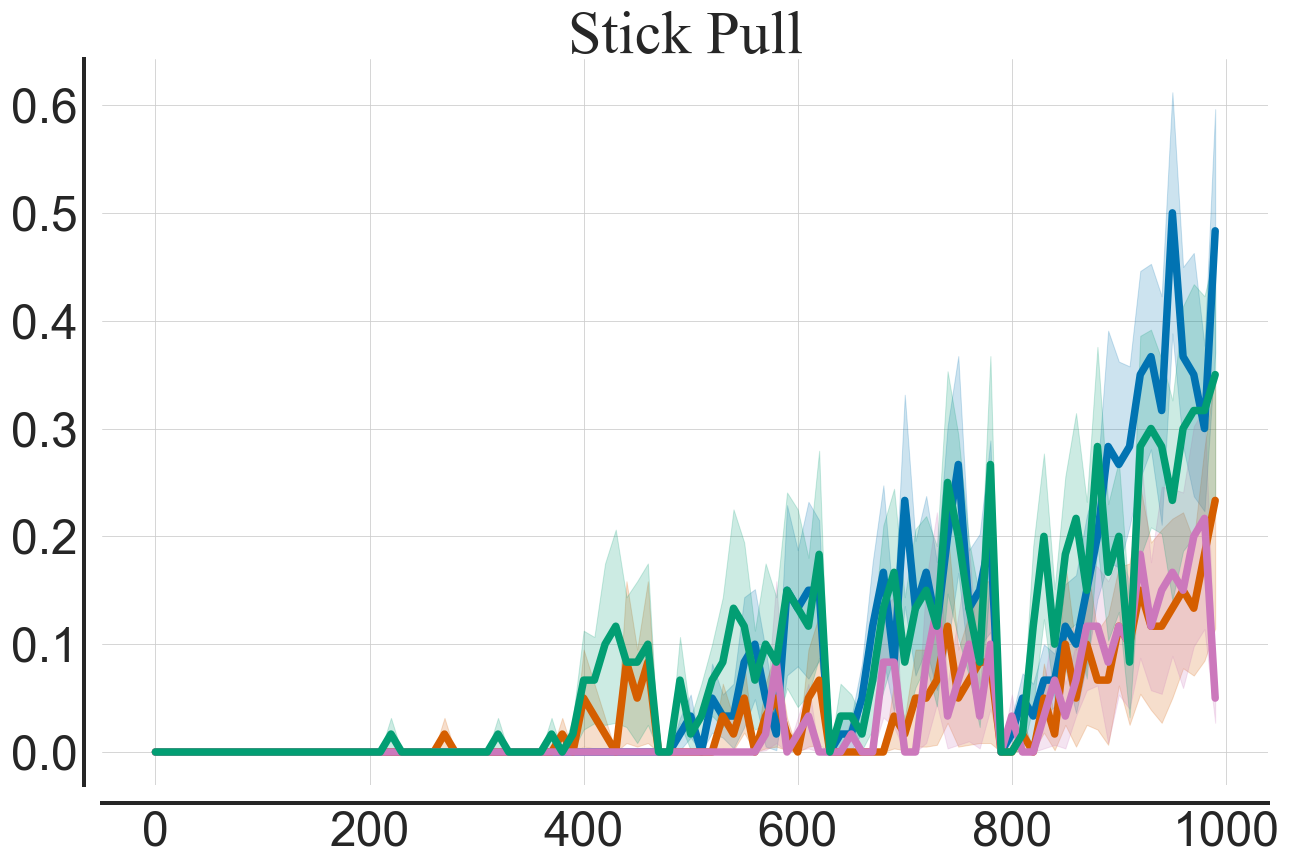}
    \end{subfigure}
\caption{Task-specific performance of high-replay configurations in 14 out of 20 considered tasks. VPL achieves performance improvements, especially in the manipulation tasks. In the case of DMC tasks the y-axis denotes evaluation returns, whereas for MetaWorld tasks it denotes the evaluation success ratio. We detail the experimental setting in Section \ref{sec:experiments_perf}. 10 seeds per task.}
\label{fig:task_perf}
\end{center}
\vspace{-0.2in}
\end{figure*}

In this formulation, $\mathcal{D}_{V}$ represents the validation replay buffer, with $s, a, s'$ denoting transitions sampled from this buffer. In line with other stochastic policy algorithms, we further approximate $V^{lb}_{\phi}(s')$ with the critic output for a single action $a' \sim \pi_{\theta}(a'|s')$. As follows, VPL adjusts the pessimism under assumption that $Q^{\mu}_{\phi}(s,a)$ is a good representation of $Q^{\pi}(s,a)$. Since the actions at which $Q^{\mu}_{\phi}(s,a)$ is evaluated are sampled from the validation buffer and are off-policy, these actions are likely to produce less overestimation than the adversarial actions sampled from a value-maximizing policy. Since $Q^{\mu}_{\phi}(s,a)$ is assumed to be unbiased, VPL thus generally reduces $\beta$ over the training unless the approximated lower bound value evaluated at on-policy actions (ie. $r_{s,a} - \gamma V^{lb}_{\phi}(s')$) is systematically larger than the mean value evaluated at off-policy actions (ie. $Q^{\mu}_{\phi}(s,a)$). This approach contrasts with General Pessimism Learning (GPL) in that it allows for gradient flow through the lower bound approximation, thereby enabling adjustments to $\beta$ that are in proportion to the level of critic disagreement. Moreover, by computing the pessimism loss exclusively on validation samples, which are not utilized by the actor-critic modules, we effectively mitigate the risk of overfitting to the experienced data which we show on Figure \ref{fig:intro_perf}.

\section{Experiments}
\label{sec:experiments}

Our experiments are based on the JaxRL codebase \citep{jaxrl}. Since all considered algorithms use SR-SAC \citep{d2022sample} as their backbonce, we align the common hyperparameters with those recommended for Scaled-By-Resetting SAC (SR-SAC) as per \citet{d2022sample}. This includes using the same network architectures and a two-critic ensemble, in accordance with established practices \citep{fujimoto2018addressing, haarnoja2018soft, ciosek2019better, moskovitz2021tactical, cetin2023learning}. We conduct our experiments in two environments: the DeepMind Control (DMC) suite \citep{tassa2018deepmind} and the single-task MetaWorld \citep{yu2020meta}. Our study encompasses two replay regimes: a compute-efficient setup with 2 gradient steps per environment step without resets, and a sample-efficient setup with 16 gradient steps per environment step, including full-parameter resets every 160k steps, as suggested by \citet{d2022sample}. We provide robust analysis using the RLiable package \citep{agarwal2021deep} and detail the experimental setting in Appendix \ref{appendix_exp_details}.

\subsection{Performance and Sample Efficiency}
\label{sec:experiments_perf}

Firstly, we test the performance and sample efficiency of the proposed approach. To this end, we compare SR-SAC \citep{d2022sample} (DMC state of the art) to four algorithms that extend SR-SAC with online pessimism adjustment: GPL \citep{cetin2023learning}; OPL \citep{kuznetsov2021automating}; TOP \citep{moskovitz2021tactical}; and VPL (the proposed approach). We run the tested algorithms in both replay regimes for 1mln environment steps on 20 medium to hard tasks (10 from DMC and 10 from MetaWorld). We discuss the chosen baselines in Sections \ref{sec:pessimism_adjustment} \& \ref{appendix_related}. We discuss hyperparameter selection in Appendix \ref{appendix_exp_hyperparams} and the tested tasks in \ref{appendix_exp_environ}. We report the results of this experiment in Figures \ref{fig:intro_perf}, \ref{fig:task_perf} \& \ref{fig:aggregates}. We find that the proposed approach surpasses baseline algorithms, demonstrating 48\% and 27\% higher performance than the baseline SR-SAC in low and high replay regimes, respectively. As depicted in Figure \ref{fig:task_perf}, VPL exhibits particular effectiveness in MetaWorld manipulation tasks, developing robust policies in environments where other approaches fail, such as the assembly task.

\begin{table}[h]
\caption{We measure runtimes for $2000$ runs of each algorithm and find that the pessimism adjustment methods have trivial wall-clock overhead as compared to SAC/SR-SAC.}
\label{table:compute_cost}
\begin{center}
{\renewcommand{\arraystretch}{1.2}%
\begin{tabular}{||l|| c | c | c | c||}
\hline
\textsc{Method} & \textsc{GPL} & \textsc{OPL} & \textsc{TOP} & \textsc{VPL} \\
\hline\hline
\textsc{RR}$=2$ & $0.3\%$ & $6.3\%$ & $0.3\%$ & $3.5\%$ \\
\textsc{RR}$=16$ & $0.5\%$ & $1.1\%$ & $0.1\%$ & $3.8\%$ \\
\hline
\end{tabular}}
\end{center}
\vskip -0.1in
\end{table}

\subsection{Validation Buffer Regret}
\label{sec:experiments_buff}

\begin{figure*}[ht!]
\begin{center}
    \begin{subfigure}{0.495\linewidth}
    \centering
    \caption{Replay Ratio = 2}
    \includegraphics[width=0.995\textwidth]{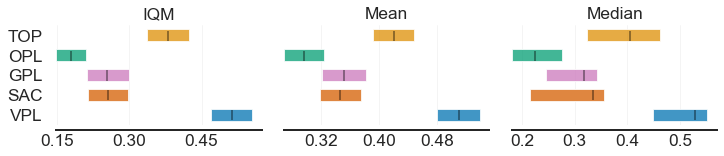}
    \label{fig:aggro1}
    \end{subfigure}
    \hfill
    \begin{subfigure}{0.495\linewidth}
    \centering
    \caption{Replay Ratio = 16}
    \includegraphics[width=0.995\textwidth]{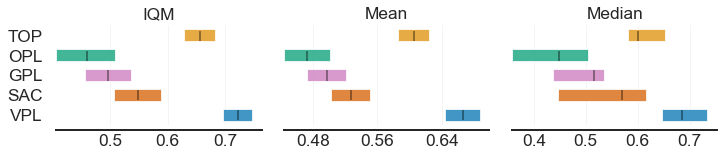}
    \label{fig:aggro2}
    \end{subfigure}
\vspace{-0.1in}
\caption{RLiable final performance metrics for the main experiment detailed in Section \ref{sec:experiments_perf}. VPL outperforms baseline algorithms in both replay regimes. The performance metrics are calculated on 20 tasks listed in Table \ref{table:tasks_20} with 10 random seeds per task.}
\label{fig:aggregates}
\end{center}
\vspace{-0.2in}
\end{figure*}

To understand the impact of a validation buffer on online RL training, we analyze three distinct agent setups: \textit{baseline} SR-SAC, which operates without a validation buffer, thus updating actor-critic modules with all experienced transitions; \textit{regret} SR-SAC, which maintains a validation buffer but does not employ validation transitions for pessimism adjustment; and SR-SAC-VPL, which not only maintains a validation buffer but also utilizes validation transitions for pessimism adjustment. This comparative analysis aims to isolate the performance loss attributable to the presence of a validation buffer and the efficiency gains derived from employing VPL for updating pessimism. We evaluate these agents in high-replay regime on 4 tasks (listed in Table \ref{table:tasks_4}) over 1mln environment steps, using varying ratios of validation to training samples, specifically at proportions of $\frac{1}{128}$, $\frac{1}{32}$, $\frac{1}{8}$, and $\frac{1}{2}$. The results for this experiment are presented in Figure \ref{fig:buffer_effects}. We observe that the regret associated with maintaining a validation buffer, and thus not utilizing it for actor-critic updates, diminishes over the course of training. Specifically, the \textit{regret} SR-SAC reaches parity with the SR-SAC in performance for all validation proportions except at $\frac{1}{2}$. We note that the rate of regret reduction correlates with the size of the validation proportion, with smaller proportions converging to baseline performance more rapidly. When examining the effectiveness of pessimism adjustment, we observe its most pronounced impact during the early stages of training. This trend aligns with the expectation of reducing critic disagreement over time. Additionally, the extent of performance gain appears to be influenced by the size of the validation buffer, where larger proportions yield greater improvements. This effect is likely due to the increased diversity of environment transitions available for pessimism adjustment in larger buffers. When considering the combined effects on performance, our findings indicate that, except for the $\frac{1}{2}$ proportion, all validation proportions successfully compensate for the performance loss due to validation buffer maintenance. This result is in line with the broader experimental results presented in Figures \ref{fig:intro_perf} \& \ref{fig:aggregates}.

\subsection{Hyperparameter Sensitivity \& Other Experiments}
\label{sec:experiments_other}

\begin{figure}[ht!]
\begin{center}
    \begin{subfigure}{0.7\linewidth}
    \includegraphics[width=\textwidth]{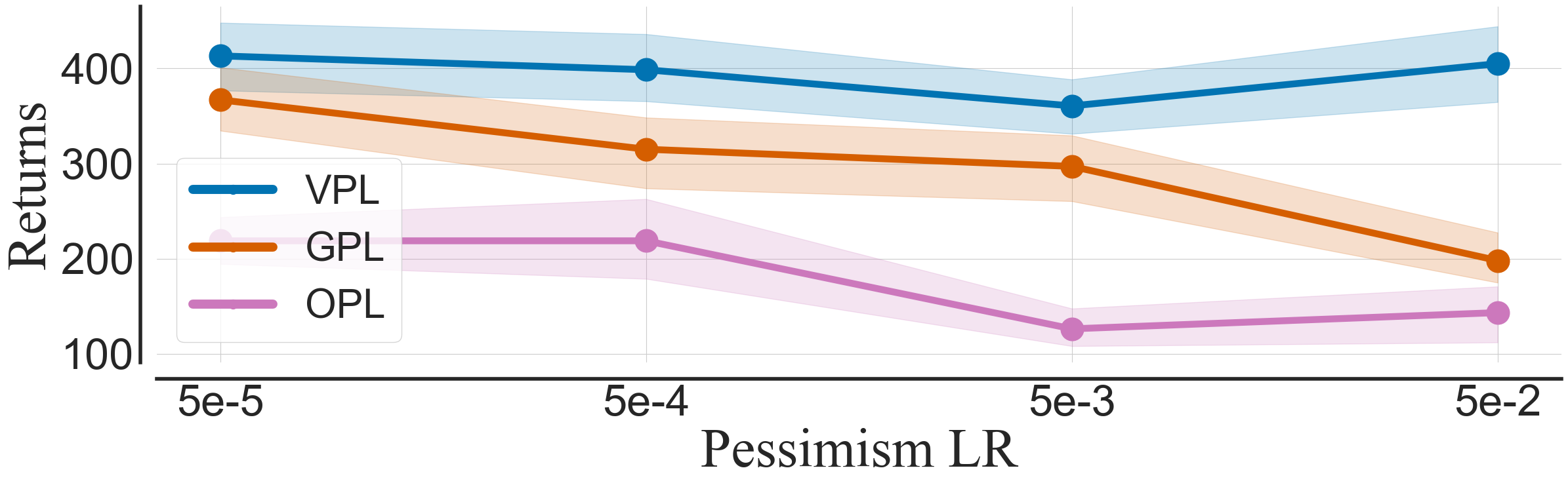}
    \end{subfigure}
\vspace{-0.1in}
\caption{VPL exhibits substantially less sensitivity to the learning rate of the pessimism module. 4 tasks, 10 seeds per task.}
\label{fig:lr_sensitivity}
\end{center}
\end{figure}

We investigate the sensitivity of VPL to varying pessimism learning rates as compared other pessimism adjustment algorithms. Given the dependency of such learning rate on reward scales and environmental dynamics, determining an optimal rate a priori is challenging, which is a significant restriction for practical applications. To address this, we test the performance of VPL, GPL, and OPL across four environments detailed in Table \ref{table:tasks_4} in the high-replay regime. We evaluate agents after 500k environments steps for learning rates of $[5e-5, 5e-4, 5e-3, 5e-2]$. The results, presented in Figure \ref{fig:lr_sensitivity}, indicate that VPL exhibits less sensitivity to changes in the pessimism learning rate than the other considered algorithms. Furthermore, we investigate the importance of the two proposed design elements: the use of a validation buffer and the VPL pessimism loss as formulated in Equation \ref{eq:pessimism_learning_vpl}. To this end, we compare the performance of six agents, each employing different combinations of pessimism loss – either the dual optimization pessimism loss or the VPL pessimism loss – along with varying sources for pessimism updates. These sources include samples from the replay buffer, the validation buffer, and the most recent transitions. The results of this analysis are presented in Figure \ref{fig:ablation_perf_res}. In our final analysis, we focus on validating the premise of VPL: its effectiveness in reducing approximation error and mitigating overfitting compared to baseline algorithms. Our methodology for quantifying approximation error and overfitting are described in Appendix \ref{appendix_exp_details}. We conducted these measurements across both low and high replay regimes, using a selection of 20 tasks from the DMC and MetaWorld as listed in Table \ref{table:tasks_20}. The findings, depicted in Figure \ref{fig:intro_perf} and Appendix \ref{appendix_exp_results}, confirm that VPL achieves the lowest levels of critic overfitting and approximation error in both replay scenarios.

\section{Limitations}

The primary challenge of VPL lies in estimating the lower-bound approximation error necessary for the pessimism adjustment mechanism. This estimation currently relies on a simplistic assumption from inherited from GPL and discussed in Section \ref{sec:vpl_pess}. Exploring alternative estimation methods is a promising avenue for future research. Surprisingly, our experiments (see Figure \ref{fig:buffer_effects}) reveal that using a validation buffer does not detrimentally impact agent performance in high-replay scenarios, except in extremely sample-scarce environments (eg. fewer than 250k environment steps).

\section{Conclusions}

In this paper, we examined the approximation error in critic networks optimized via temporal difference variants. We introduced a fixed-point model for estimating mean and lower bound errors and used this model to analyze the convergence of pessimistic actor-critic algorithms. We proposed the VPL algorithm, which dynamically adjusts pessimism levels to minimize approximation errors of critic supervision in validation samples. We tested VPL against baseline algorithms in various locomotion and manipulation tasks, showing improvements in performance and sample efficiency. We explored the impact of VPL components and their sensitivity to hyperparameter selection. Our results confirm VPLs effectiveness in complex continuous action tasks. We share our code under this \href{https://anonymous.4open.science/r/Valdation-Pessimism-Learning-6D4F/}{link}.

\subsubsection*{Impact Statement}

This paper focuses on the issue of approximation error in temporal-difference learning agents within Reinforcement Learning (RL). While the successful application of RL has the potential to influence society in many ways, our work, focused primarily on a technical advancement in RL algorithms, does not introduce novel ethical considerations beyond those already inherent in the broader field of RL.

\bibliography{bib}
\bibliographystyle{icml2023}

%%%%%%%%%%%%%%%%%%%%%%%%%%%%%%%%%%%%%%%%%%%%%%%%%%%%%%%%%%%%%%%%%%%%%%%%%%%%%%%
%%%%%%%%%%%%%%%%%%%%%%%%%%%%%%%%%%%%%%%%%%%%%%%%%%%%%%%%%%%%%%%%%%%%%%%%%%%%%%%
% APPENDIX
%%%%%%%%%%%%%%%%%%%%%%%%%%%%%%%%%%%%%%%%%%%%%%%%%%%%%%%%%%%%%%%%%%%%%%%%%%%%%%%
%%%%%%%%%%%%%%%%%%%%%%%%%%%%%%%%%%%%%%%%%%%%%%%%%%%%%%%%%%%%%%%%%%%%%%%%%%%%%%%
\newpage
\appendix
\onecolumn

%\section*{Acknowledgments}
%Marek Cygan is co-financed by the National Centre for Research and Development as a part of the EU-supported Smart Growth Operational Programme 2014-2020 (POIR.01.01.01-00-0392/17-00). The experiments were performed using the Entropy cluster funded by NVIDIA, Intel, the Polish National Science Center grant UMO-2017/26/E/ST6/00622, and the ERC Starting Grant TOTAL.

\section*{Appendix Contents}

We divide the Appendix into the following sections:

\begin{enumerate}
    \item Derivations (Appendix \ref{appendix_derivations}) - we present the derivations associated with statements presented in Section \ref{section_approximationerror}. Furthermore, we discuss the further implications of our propositions. 
    \item Related Work (Appendix \ref{appendix_related}) - we discuss the works related to the proposed method. In particular, we discuss pessimistic actor-critic algorithms, approaches for online pessimism adjustment and theoretical work on approximation error in TD learning.
    \item Future Work (Appendix \ref{appendix_future_work}) - we discuss avenues for potential further research related to the proposed method. 
    \item Experimental Details (Appendix \ref{appendix_exp_details}) - we detail all experiments presented throughout the manuscript. 
    \item Tested Environments (Appendix \ref{appendix_exp_environ}) - we list all tested environments from the DeepMind Control and MetaWorld environments. 
    \item Additional Experimental Results (Appendix \ref{appendix_exp_results}) - we present additional experimental results.
    \item Hyperparameters (Appendix \ref{appendix_exp_hyperparams}) - we discuss the procedure for hyperparamenter selection for all algorithms and list all used hyperparameters. 
    \item Learning Curves (Appendix \ref{appendix:learning_curves}) - we present learning curves for the experiments.
\end{enumerate}

\section{Derivations}
\label{appendix_derivations}

In this section, we derive statements presented in Section \ref{section_approximationerror}. For simplicity, we consider a fixed policy $\pi_{\theta}$ and use $V(s)$ and $Q(s,a)$ to represent the value and Q-value under this policy. We define the mean and lower bound approximation errors denoted as $U_{\phi}^{\mu}$ and $U_{\phi}^{lb}$ respectively:

\begin{equation}
\label{eq:ensemble_approximation_error_a}
\begin{split}
    U_{\phi}^{\mu}(s,a) & \triangleq Q(s,a) - Q^{\mu}_{\phi}(s,a) \\
    U_{\phi}^{lb}(s,a) & \triangleq Q(s,a) - Q^{lb}_{\phi}(s,a) \\
\end{split}
\end{equation}

$Q(s,a)$ denotes the true Q-value, the term $Q^{\mu}_{\phi}(s,a)$ represents the mean Q-value estimated by an ensemble of $k$ critics, calculated as $Q^{\mu}{\phi}(s,a) = \frac{1}{k} \sum^{k} Q^{i}{\phi}(s,a)$, and $Q^{lb}{\phi}(s,a)$ is the lower bound Q-value as defined as follows:

\begin{equation}
\label{eq:ln_approximation_error_a1}
\begin{split}
Q^{lb}_{\phi}(s,a) & = Q^{\mu}_{\phi}(s,a) - \beta Q^{\sigma}_{\phi}(s,a)
\end{split}
\end{equation}

Similarly, we define lower bound value:

\begin{equation}
\label{eq:value_error_a}
\begin{split}
V^{lb}_{\phi}(s) & = \underset{a \sim \pi_{\theta}}{\mathrm{E}} \bigl(Q^{\mu}_{\phi}(s,a) - \beta Q^{\sigma}_{\phi}(s,a) - \alpha \log \pi_{\theta}(a|s) \bigr)
\end{split}
\end{equation}

We also introduce the mean and lower bound temporal critic errors, denoted as $u^{\mu}_{\phi}$ and $u^{lb}_{\phi}$, respectively:

\begin{equation}
\label{eq:ln_approximation_error_a}
\begin{split}
u^{\mu}_{\phi}(s, a, s') & \triangleq r_{s,a} + \gamma V^{\mu}_{\phi}(s') - Q^{\mu}_{\phi}(s,a) \\
u^{lb}_{\phi}(s, a, s') & \triangleq r_{s,a} + \gamma V^{lb}_{\phi}(s') - Q^{\mu}_{\phi}(s,a) \\
\end{split}
\end{equation}

These temporal critic errors quantify the deviation between the Q-values $Q^{\mu}_{\phi}(s,a)$ and the mean or lower bound Temporal Difference (TD) targets. 

\subsection{Approximation Error Operator}

Firstly, we note that for the true Q-value the following always holds:

\begin{equation}
\label{eq:qval_bell_a}
Q(s,a) = r_{s,a} + \gamma V(s') = r_{s,a} + \gamma \underset{a' \sim \pi_{\theta}}{\mathrm{E}} \bigl( Q(s',a') - \alpha \log \pi_{\theta}(a'|s') \bigr)
\end{equation}

Then, using Equations \ref{eq:ensemble_approximation_error_a}, \ref{eq:ln_approximation_error_a} \& \ref{eq:qval_bell_a} we write:

\begin{equation}
\label{eq:error1_a}
\begin{split}
    U_{\phi}^{\mu}(s,a) & = Q(s,a) - Q^{\mu}_{\phi}(s,a) \\
    & = r_{s,a} + \gamma V(s') - r_{s,a} - \gamma V_{\phi}^{\mu}(s') + u^{\mu}_{\phi}(s, a, s') \\
    & = u^{\mu}_{\phi}(s, a, s') + \gamma \bigl(V(s') - V_{\phi}^{\mu}(s')\bigr) \\
    & = u^{\mu}_{\phi}(s, a, s') + \gamma \underset{a' \sim \pi_{\theta}}{\mathrm{E}} \bigl( Q(s',a') - \alpha \log \pi_{\theta}(a'|s') - Q_{\phi}^{\mu}(s',a') +  \alpha \log \pi_{\theta}(a'|s') \bigr) \\
    & = u^{\mu}_{\phi}(s, a, s') + \gamma \underset{a' \sim \pi_{\theta}}{\mathrm{E}} U_{\phi}^{\mu}(s',a') 
\end{split}
\end{equation}

Similarly, we calculate $U_{\phi}^{lb}(s,a)$:

\begin{equation}
\label{eq:error2_a}
\begin{split}
    U_{\phi}^{lb}(s,a) & = Q(s,a) - Q^{\mu}_{\phi}(s,a) + \beta Q^{\sigma}_{\phi}(s,a) \\
    & = u^{\mu}_{\phi}(s, a, s') + \beta Q^{\sigma}_{\phi}(s,a) + \gamma \underset{a' \sim \pi_{\theta}}{\mathrm{E}} U_{\phi}^{\mu}(s',a') \\
    & = u^{lb}_{\phi}(s, a, s') + \beta Q^{\sigma}_{\phi}(s,a) + \gamma \underset{a' \sim \pi_{\theta}}{\mathrm{E}} U_{\phi}^{lb}(s',a') 
\end{split}
\end{equation}

As such, both $U_{\phi}^{\mu}(s,a)$ and $U_{\phi}^{lb}(s,a)$ can be expressed as a function of combination of $U_{\phi}^{\mu}(s',a')$ or $U_{\phi}^{lb}(s',a')$ and $u^{\mu}_{\phi}(s, a, s')$ or $u^{lb}_{\phi}(s, a, s')$.

\subsection{Approximation Error Contraction}

We show that both approximation error operators are contractions wrt. infinity norm with similar argument to Bellman values \citep{puterman2014markov}. 

\begin{equation}
\label{eq:contract}
\begin{split}
    ||\mathcal{U} (f_{1}) - \mathcal{U} (f_2) ||_{\infty} 
    & = \sup_{s,a}|u^{\mu}_{\phi}(s, a, s') + \gamma\underset{a' \sim \pi_{\theta}}{\mathrm{E}}f_1(s',a')-u^{\mu}_{\phi}(s, a, s') - \gamma\underset{a' \sim \pi_{\theta}}{\mathrm{E}}f_2(s',a')| \\
    & = \gamma| \underset{a' \sim \pi_{\theta}}{\mathrm{E}}f_1(s',a') - \underset{a' \sim \pi_{\theta}}{\mathrm{E}}f_2(s',a')|\\
    & \leq  \gamma \underset{a' \sim \pi_{\theta}}{\mathrm{E}}|f_1(s',a') -f_2(s',a')|\\
    & \leq \gamma ||f_1 - f_2||_{\infty}.
\end{split}
\end{equation}

\section{Related Work}
\label{appendix_related}
 
\subsection{Pessimistic Actor-Critic}

Recent model-free, off-policy algorithms address the overestimation bias in critic's TD-targets through diverse methods \citep{thrun2014issues, hasselt2010double}. These include leveraging multiple function approximators to conservatively estimate expected returns \citep{fujimoto2018addressing, haarnoja2018soft, ciosek2019better, lee2021sunrise, andrychowicz2021matters}. Notably, Clipped Double Q-learning (CDQL) employs a pessimistic approach by calculating the critic’s TD-targets as the minimum of two action-value model outputs \citep{fujimoto2018addressing}. Weighted Double Q-Learning (WDQL) introduces a weighted sum of mean and minimum targets for TD calculations \citep{zhang2017weighted}. Furthermore, \citet{kuznetsov2020controlling} suggest using a quantile distributional critic with interquantile statistics for TD target computations. An alternative method proposes reducing approximation errors in TD loss by varying batch sample weights as to counteract the negative interaction of approximation bias and the data-collecting distribution \citep{kumar2020discor}. Pessimism was also studied in the context of model-based RL \citep{ha2018recurrent, asadi2018lipschitz, janner2019trust, ball2020ready, seyde2022learning, wang2022sample}. A popular approach is to avoid or reduce the impact of simulated trajectories which the dynamics model deems uncertain \citep{buckman2018sample, yu2020mopo, yao2021sample, mendonca2021discovering}. In this context, similarly to value, model ensemble disagreement is a very popular approach to uncertainty quantification \citep{janner2019trust, yu2020mopo, pan2020trust, yao2021sample}.

\subsection{Pessimism Adjustment}

Adjusting pessimism levels has become more dynamic with the development of methods that represent the minimum target as a function of the mean and critic ensemble disagreement \citep{kuznetsov2021automating, moskovitz2021tactical, cetin2023learning}. GPL, for instance, modifies pessimism using a dual optimization objective, calculating the loss on replay samples \citep{cetin2023learning}. OPL uses an online approach to adjust pessimism levels by comparing critic outputs with on-policy return estimators \citep{kuznetsov2021automating}. TOP uses an auxiliary bandit to select optimal pessimism levels for maximizing online returns \citep{moskovitz2021tactical}. These approaches are detailed in Table \ref{table:pessimism_methods}.

\begin{table*}[ht!]
\caption{Considered algorithms differ by the pessimism domain, strategy for critic error estimation, as well as the pessimism update rule.}
\label{table:pessimism_methods}
\begin{center}
{\renewcommand{\arraystretch}{1.05}%
\begin{tabular}{||c||c|c|c||}
\hline
 & $\beta$ \textsc{Domain} & \textsc{Critic Error Estimated Via} & \textsc{Pessimism Update Rule}  \\
\hline\hline
\textsc{TOP} & $0, 1$ & \textsc{NA} & Auxiliary bandit maximizing episodic returns \\
\textsc{GPL} & $\left[0, \infty \right]$ & Critic output on the replay transitions & Dual optimization update \\
\textsc{OPL} & $\left[0, \infty \right]$ & Bootstrapped $\lambda$-returns on recent transitions & Dual optimization update \\
\textsc{VPL} & $\left[0, \infty \right]$ & Critic output on the validation transitions & Minimization of approximation error\\
\hline
\end{tabular}}
\end{center}
\vskip -0.1in
\end{table*}

\subsection{Approximation Error in RL}

The regret caused by errors in critic approximation has been explored in approximate value iteration algorithms \citep{de2000existence, van2006performance, munos2005error, munos2007performance, munos2008finite, farahmand2010error}. When a policy is greedy with respect to the critic estimates,  value approximation errors can greatly influence the policy and the resulting returns. Therefore, there's been significant work to understand how these approximation errors affect performance \citep{munos2005error, munos2007performance, munos2008finite, farahmand2010error}. These ideas have also been revisited in the area of deep reinforcement learning \citep{kumar2019stabilizing, kumar2020discor}. In particular, \citet{kumar2020discor} examines the detailed patterns in non-pessimistic value approximation errors. Those results remain relevant for off-policy actor-critic algorithms such as SAC, as it can be described as an approximate policy iteration algorithm \citep{haarnoja2018soft}.

\section{Future Work}
\label{appendix_future_work}

While our implementation is based on the vanilla SR-SAC algorithm, recent studies have demonstrated that simple regularization methods applied to the critic can significantly enhance performance \citep{hiraoka2021dropout, li2022efficient, ball2023efficient}. Consequently, integrating VPL with network regularization appears to be a promising approach. Specifically, layer normalization and spectral normalization have been effective in continuous action off-policy agents \citep{ba2016layer, gogianu2021spectral}. Similarly, it has been observed that deep RL agents experience a reduced ability to learn over time, a phenomenon known as 'plasticity loss'. Addressing this diminishing capacity has been shown to be empirically beneficial \citep{janner2019trust, nikishin2022primacy, d2022sample, lyle2023understanding}. Although our approach involves full-parameter resets in the high replay regime, employing multiple techniques to address plasticity loss has proven advantageous \citet{lee2023plastic}. Therefore, combining VPL with strategies like CReLU \citet{shang2016understanding} or Sharpness Aware Minimization (SAM) \citet{foret2020sharpness} could potentially lead to further performance improvements. Given that VPL employs a more controlled use of the critic ensemble compared to standard SAC/TD3 methods, increasing the critic ensemble size in VPL may create synergies, potentially surpassing the benefits seen in conventional ensemble AC approaches \citep{chen2020randomized, lee2021sunrise, januszewski2021continuous, ball2023efficient}. Additionally, the integration of a distributional critic setup into the pessimism adjustment framework \citep{moskovitz2021tactical}, which has been shown to enhance RL learning \citep{bellemare2017distributional, rowland2019statistics, rowland2023analysis}, suggests that incorporating distributional critics into VPL could yield notable performance gains.

\section{Experimental Details}
\label{appendix_exp_details}

\subsection{Performance and Sample Efficiency}

We run the tested algorithms for 1mln environment steps on 20 DMC/MetaWorld tasks listed in Table \ref{table:tasks_20} using hyperparameters described in Section \ref{appendix_exp_hyperparams}. All algorithms are evaluated via greedy policies every 10k environment steps. We calculate the final performance presented in Figure \ref{fig:aggregates} by averaging over last 10 policy evaluations (ie. the last 100k environment steps). The results for this setup are presented in Figures \ref{fig:intro_perf}, \ref{fig:task_perf} and \ref{fig:aggregates}, as well as in the final Appendix section. 

\subsection{Approximation Error and Overfitting}

Throughout the manuscripts we present estimates of critic approximation error and overfitting. Here, we discuss our methodology for calculating both.

\subsubsection{Approximation Error}

We calculate the critic approximation via:

\begin{equation}
    U_{\phi}^{\mu}(s,a) = - (Q(s,a) - Q^{\mu}_{\phi}(s,a))
\end{equation}

Where we add the negative sign such that the positive approximation error represents overestimation, and negative approximation error represents underestimation. Above, $Q^{\mu}_{\phi}(s,a)$ represents the output of the critic. Estimating the reference Q-value $Q(s,a)$ in our context demands solving two problems. Firstly, as all algorithms use SAC as their backbone, the Q-value is soft, ie. it represents the sum of returns and entropies. This contrasts with regular DDPG-style critic which approximates the returns alone. Secondly, both MetaWorld and DMC are infinite-horizon MDPs. As such, obtaining an unbiased Monte-Carlo rollout value is non-trivial. To this end, we calculate the reference Q-value via:

\begin{equation}
    Q(s,a) = \frac{1}{1-\gamma} \bigl(\hat{R}_{s,a} - \alpha \log \hat{\pi}(a|s)\bigr)
\end{equation}

Where $\hat{R}_{s,a}$ denotes the average reward gathered when performing action $a$ at state $s$ and following the policy afterwards, and $\log \hat{\pi}$ denotes the average policy log-probability when following the policy from a given state-action. The term $\frac{1}{1-\gamma} $ stems from the sum of a geometric series, reflecting the infinite-horizon that the critic models. We estimate $\hat{R}$ with using a Monte-Carlo rollout, and use the entropy target to calculate $\log \hat{\pi}$. For each approximation error measurement, we average the error over 5 different starting states. 

\subsubsection{Overfitting}

We calculate the critic overfitting (denoted as $\mathcal{O}(\phi)$) with the following equation:

\begin{equation}
   \mathcal{O}(\phi) = \frac{\underset{\mathcal{D}_V}{\mathrm{E}} u^{lb}_{\phi}(s_V, a_V, s_{V}^{'})}{\underset{\mathcal{D}_T}{\mathrm{E}} u^{lb}_{\phi}(s, a, s')}
\end{equation}

Where $\mathcal{D}_T$ and $\mathcal{D}_V$ denote the training and validation replay buffers respectively. Furthermore, $s_V, a_V, s_{V}^{'}$ denote the transitions sampled from the validation buffer, and $s, a, s'$ denote the training replay buffer transitions. As such, we calculate our overfitting metric by comparing the temporal difference for unseen validation transitions and the training transitions which were the critic is trained on. For each algorithm, we gather such validation samples during the evaluation rollouts. As such, the validation buffer gathers 5,000 new transitions every 10,000 training transitions. We estimate the expectation on batches of 256 transitions sampled from each buffer. Such definition of overfitting is easily interpretable - when $\mathcal{O}(\phi)$ is close to $1$ then both validation and training TD errors are comparable and therefore there is little to none overfitting. When $\mathcal{O}(\phi)$ is greater than $1$ then it means that the validation TD errors are relatively larger than the ones in the training, indicating overfitting. 

\subsection{Validation Buffer Regret}

Our study examines three agent configurations: (1) baseline SR-SAC, which updates actor-critic modules with all transitions and lacks a validation buffer; (2) regret SR-SAC, featuring a validation buffer but not using it for pessimism adjustment; and (3) SR-SAC-VPL, which includes a validation buffer and employs validation transitions for pessimism adjustment. The analysis focuses on identifying performance differences caused by the validation buffer and the benefits of using VPL for pessimism updates. We tested these agents across four tasks (see Table \ref{table:tasks_4}) for 1 million environment steps, exploring various validation to training sample ratios, namely $\frac{1}{128}$, $\frac{1}{32}$, $\frac{1}{8}$, and $\frac{1}{2}$. The experiments are performed in high replay setting with full parameter resets every 160k environment steps \citep{d2022sample}. Experimental results are detailed in Figure \ref{fig:buffer_effects}.

\subsection{Learning Rate Sensitivity}

We run GPL, OPL and VPL for 500k environment steps in high replay setup on 4 DMC tasks listed in Table \ref{table:tasks_4}. We test each algorithms performance on four pessimism learning rate values: $[5e-5, 5e-4, 5e-3, 5e-2]$. We present the results in Figure \ref{fig:lr_sensitivity}. 

\subsection{Design Choices}

Finally, we evaluate the impact of each of VPL contributions, namely the VPL update rule and performing updates on the validation buffer. We evaluate six agents, each utilizing different forms of pessimism loss - either dual optimization or VPL pessimism loss - in combination with various sources for pessimism updates. These sources encompass samples from the replay buffer (ie. GPL), the validation buffer (ie. VPL), and the most recent online transitions (ie. OPL). We run the agents for 1mln environment steps on 5 dmc tasks shown in Figure \ref{fig:ablation_perf_res}.

\section{Tested Environments}
\label{appendix_exp_environ}

Tables below list tasks from DeepMind Control and MetaWorld considered in our experiments.

\begin{table*}[ht!]
\parbox{.45\linewidth}{
\caption{20 DMC and MetaWorld tasks used for the main evaluation.}
\vspace{0.1in}
\label{table:tasks_20}
\begin{center}
{\renewcommand{\arraystretch}{1.05}%
 \begin{tabular}{||c|c||} 
 \hline
 \textsc{DeepMind Control} & \textsc{MetaWorld} \\
 \hline \hline
 \textsc{acrobot-swingup} & \textsc{assembly} \\
 \textsc{fish-swim} & \textsc{box-close} \\
 \textsc{hopper-hop} & \textsc{button-press} \\
 \textsc{hopper-stand} & \textsc{coffee-pull} \\
 \textsc{humanoid-run} & \textsc{coffee-push} \\
 \textsc{humanoid-stand} & \textsc{drawer-open} \\
 \textsc{humanoid-walk} & \textsc{hammer} \\
 \textsc{quadruped-run} & \textsc{push} \\
 \textsc{swimmer-swimmer6} & \textsc{stick-pull} \\
 \textsc{walker-run} & \textsc{sweep} \\
 \hline \hline
\end{tabular}}
\end{center}
\vskip -0.1in} 
\hfill 
\parbox{.45\linewidth}{
\caption{4 DMC tasks used in additional experiments.}
\vspace{0.1in}
\label{table:tasks_4}
\begin{center}
{\renewcommand{\arraystretch}{1.05}%
 \begin{tabular}{||c||} 
 \hline
 \textsc{DeepMind Control} \\
 \hline \hline
 \textsc{acrobot-swingup} \\
 \textsc{hopper-hop} \\
 \textsc{humanoid-walk} \\
 \textsc{quadruped-run} \\
 \hline \hline
\end{tabular}}
\end{center}
\vskip -0.1in}
\end{table*}

\section{Hyperparameters}
\label{appendix_exp_hyperparams}

All considered algorithms use SR-SAC \citep{d2022sample} as their backbonce, we align the common hyperparameters with those recommended for SR-SAC \citep{d2022sample}. This includes using the same network architectures, a two-critic ensemble \citep{fujimoto2018addressing, haarnoja2018soft, ciosek2019better, moskovitz2021tactical, cetin2023learning} and ADAM optimizer \citep{kingma2014adam}. We choose VPL, GPL, and OPL pessimism adjustment learning rate by performing search over the same domain for all algorithms which we present in Figure \ref{fig:lr_sensitivity}. We choose the validation ratio for VPL in experiments presented in Figure \ref{fig:buffer_effects}. We choose TOP bandit setting following the best performing configurations presented \citet{moskovitz2021tactical}. We use consistent hyperparamenters between all environments and both replay regimes. All experiments are run without any action repeat wrappers (ie. we use an action repeat of 1). The hyperparameters are summarized in Table \ref{table:3}.

\begin{table}[ht!]
\centering
\caption{Hyperparameter values used in the experiments.}
\label{table:3}
%\vspace{5pt}
{\renewcommand{\arraystretch}{1.03}%
 \begin{tabular}{||c |c | c ||} 
 \hline
 \textsc{Hyperparameter} & \textsc{Notation} & \textsc{Value} \\
 \hline \hline
  \multicolumn{3}{||c||}{\textsc{Joint}} \\
 \hline \hline
 \textsc{Network Size} & \textsc{na} & $(256, 256)$ \\
 \textsc{Optimizer} & \textsc{na} & \textsc{Adam} \\
 \textsc{Learning Rate} & \textsc{na} & $3e-4$ \\
 \textsc{Batch Size} & $B$ & $256$ \\
 \textsc{Discount} & $\gamma$ & $0.99$ \\
 \textsc{Initial Temperature} & $\alpha_{0}$ & $1.0$ \\
 \textsc{Initial Steps} & $\textsc{na}$ & $10000$ \\
 \textsc{Target Entropy} & $\mathcal{H}^{*}$ & $|\mathcal{A}|/2$ \\
 \textsc{Polyak Weight} & $\tau$ & $0.005$ \\
 
 \hline \hline
 \multicolumn{3}{||c||}{\textsc{TOP}} \\
 \hline \hline
 \textsc{Pessimism Values} & $\beta$ & $0,1$ \\
 \textsc{Bandit Learning Rate} & \textsc{na} & $0.1$ \\
 \hline \hline
 \multicolumn{3}{||c||}{\textsc{GPL}} \\
 \hline \hline
 \textsc{Pessimism Learning Rate} & \textsc{na} & $5e-5$ \\
 \textsc{Initial Pessimism} & $\beta$ & $1.0$ \\
 \hline \hline
 \multicolumn{3}{||c||}{\textsc{OPL}} \\
 \hline \hline
 \textsc{Pessimism Learning Rate} & \textsc{na} & $5e-5$ \\
 \textsc{On-policy Trajectory Length} & \textsc{na} & $8$ \\
 \textsc{TD}-$\lambda$ & \textsc{na} & $0.95$ \\
 \textsc{Initial Pessimism} & $\beta$ & $1.0$ \\
 \hline \hline
 \multicolumn{3}{||c||}{\textsc{VPL}} \\
 \hline \hline
 \textsc{Pessimism Learning Rate} & \textsc{na} & $5e-5$ \\
 \textsc{Validation Proportion} & \textsc{v} & $1/32$ \\
 \textsc{Initial Pessimism} & $\beta$ & $1.0$ \\
 \hline \hline
\end{tabular}}
\end{table}

\section{Additional Experimental Results}
\label{appendix_exp_results}

\begin{figure*}[h!]
\begin{center}
\begin{minipage}[h]{1.0\linewidth}
\centering
    \begin{subfigure}{0.88\linewidth}
    \includegraphics[width=\textwidth]{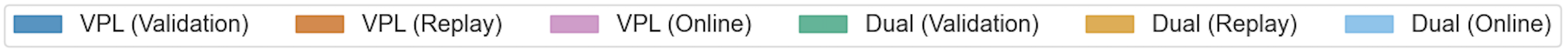}
    \end{subfigure}
\end{minipage}
\medskip
\begin{minipage}[h]{1.0\linewidth}
    \begin{subfigure}{1.0\linewidth}
    \includegraphics[width=0.245\linewidth]{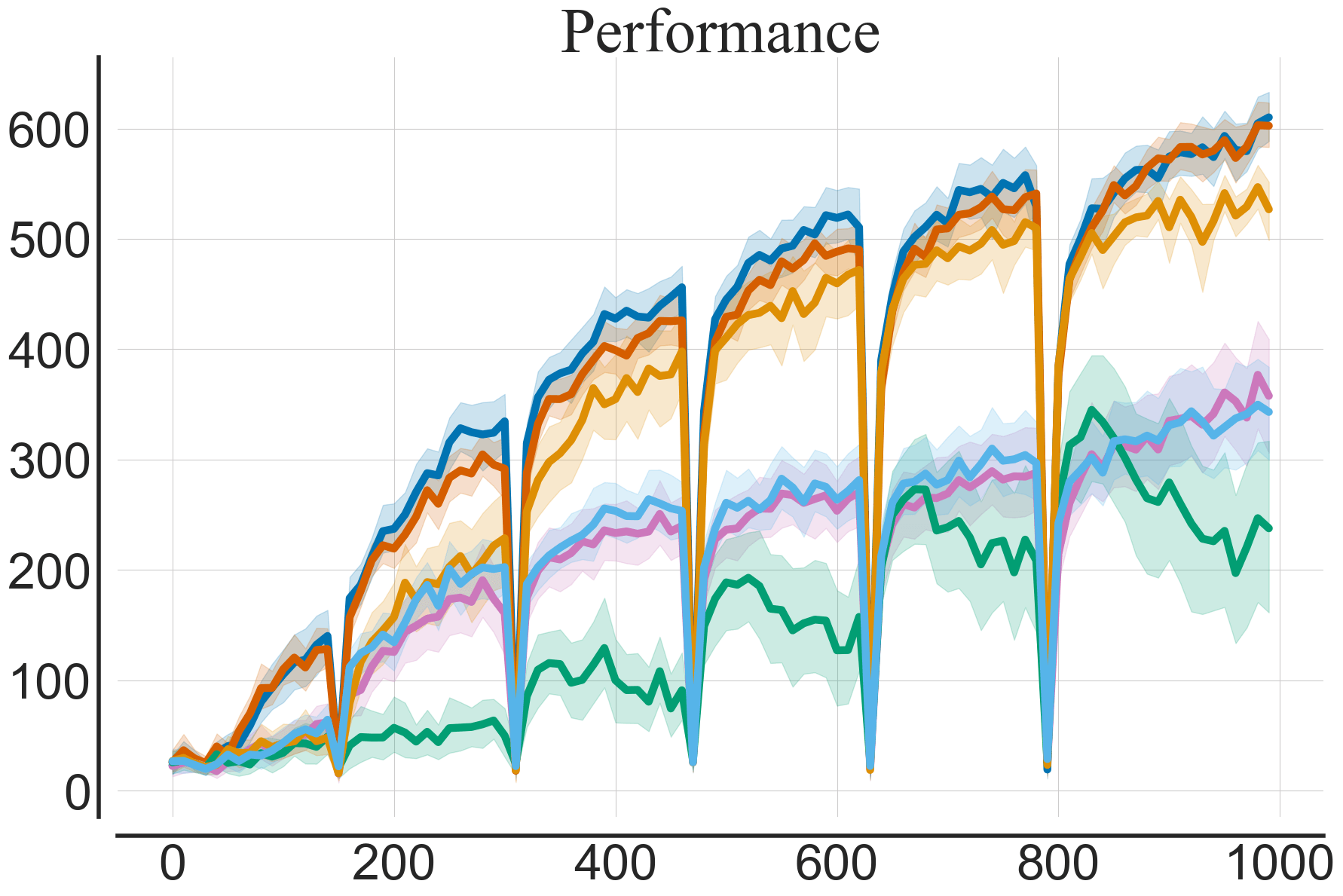}
    \hfill
    \includegraphics[width=0.245\linewidth]{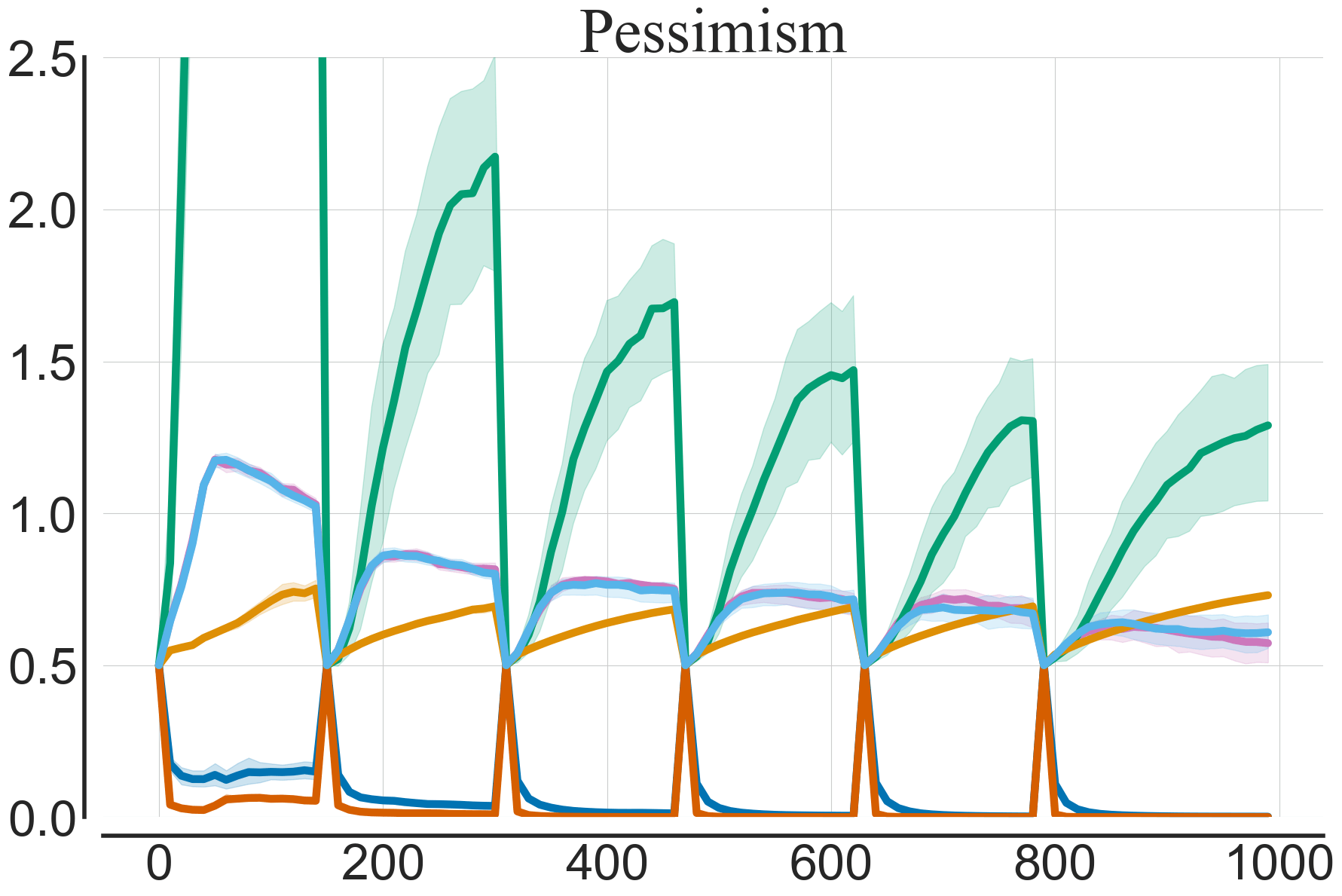}
    \hfill
    \includegraphics[width=0.245\linewidth]{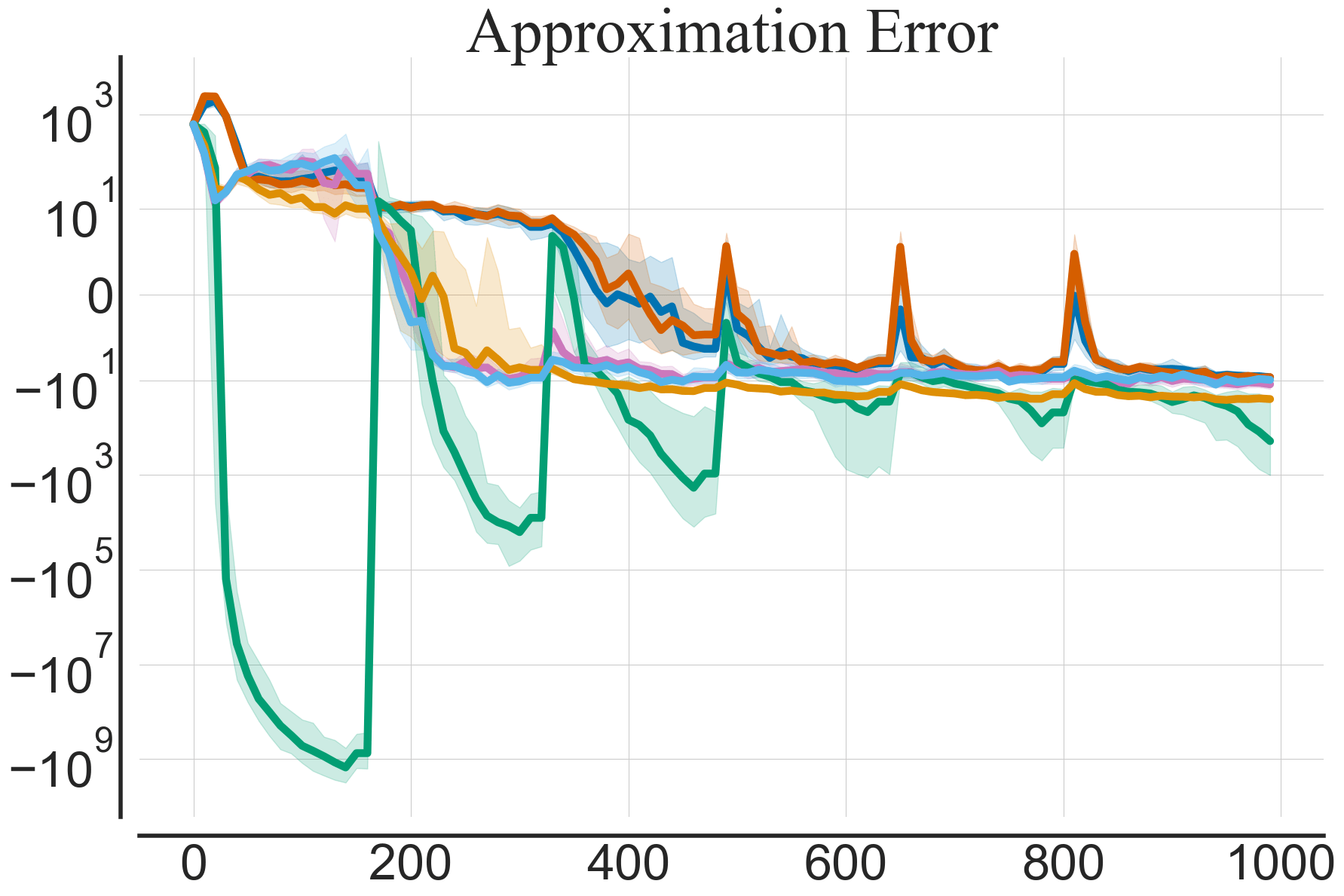}
    \hfill
    \includegraphics[width=0.245\linewidth]{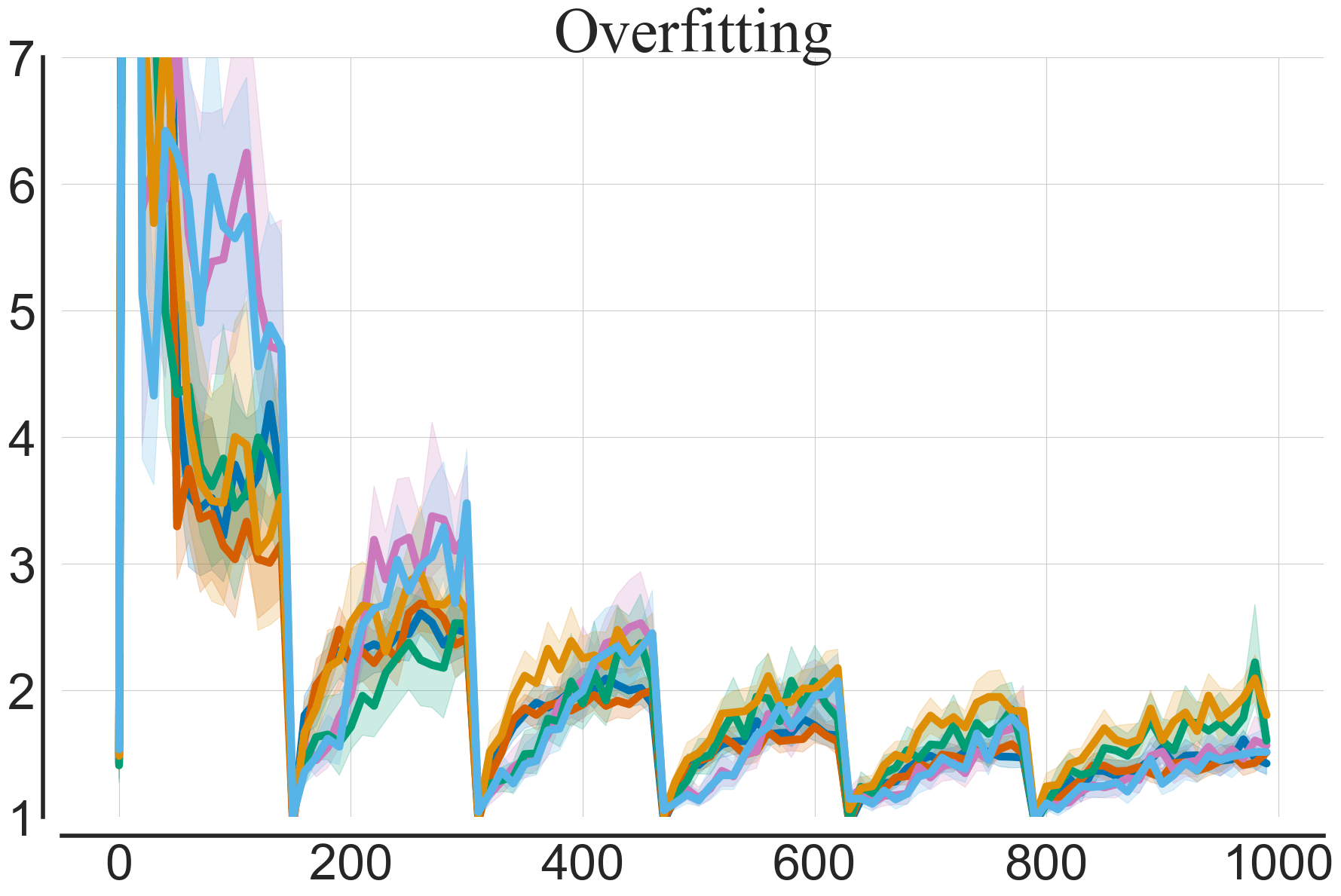}
    \end{subfigure}
\end{minipage}
\caption{Ablation on design of the pessimism module. 4 tasks, 10 seeds per task.}
\label{fig:ablation_perf_res_aggregate}
\end{center}
\end{figure*}

\begin{figure*}[h!]
\begin{center}
\begin{minipage}[h]{1.0\linewidth}
\centering
    \begin{subfigure}{0.88\linewidth}
    \includegraphics[width=\textwidth]{images/legend_ablation.png}
    \end{subfigure}
\end{minipage}
\medskip
\begin{minipage}[h]{1.0\linewidth}
    \begin{subfigure}{1.0\linewidth}
    \includegraphics[width=0.19\linewidth]{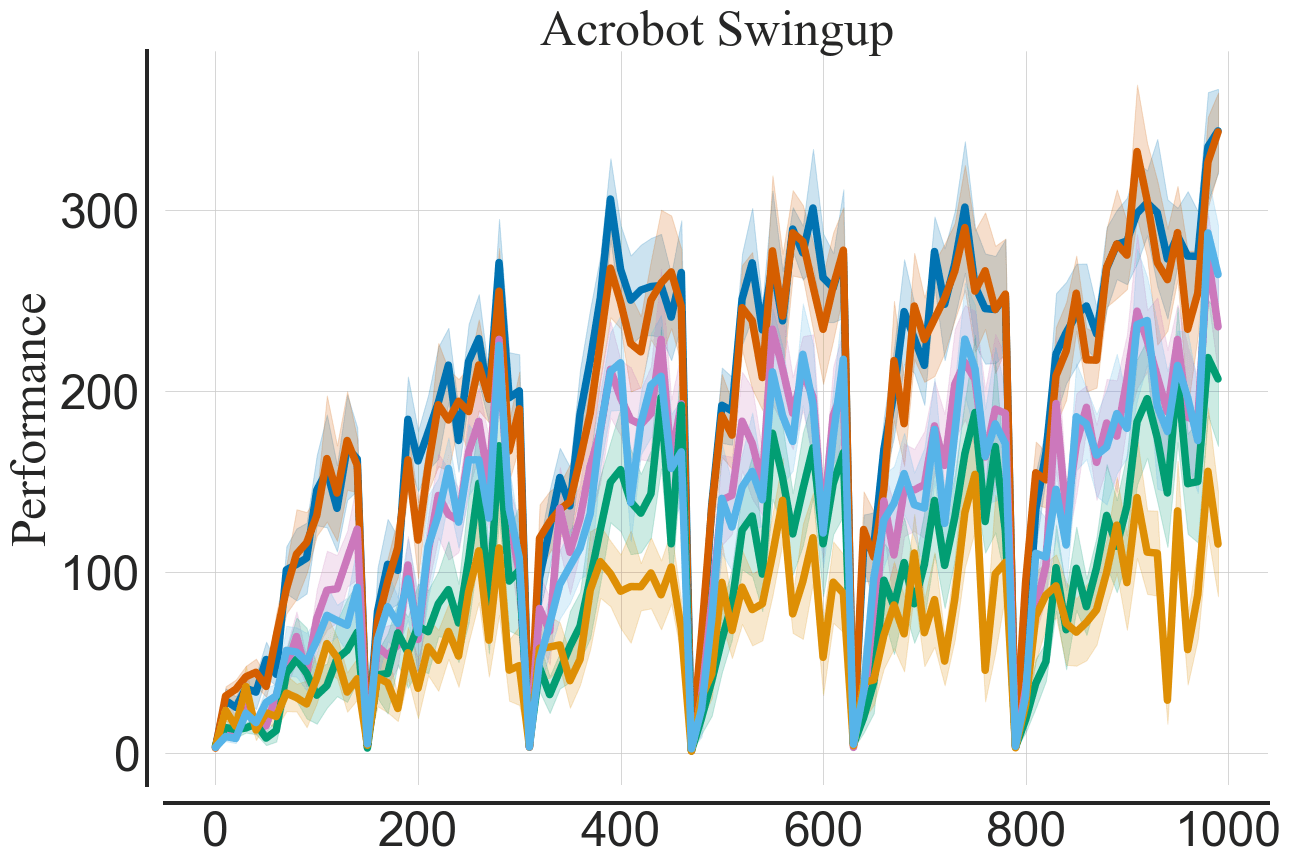}
    \hfill
    \includegraphics[width=0.19\linewidth]{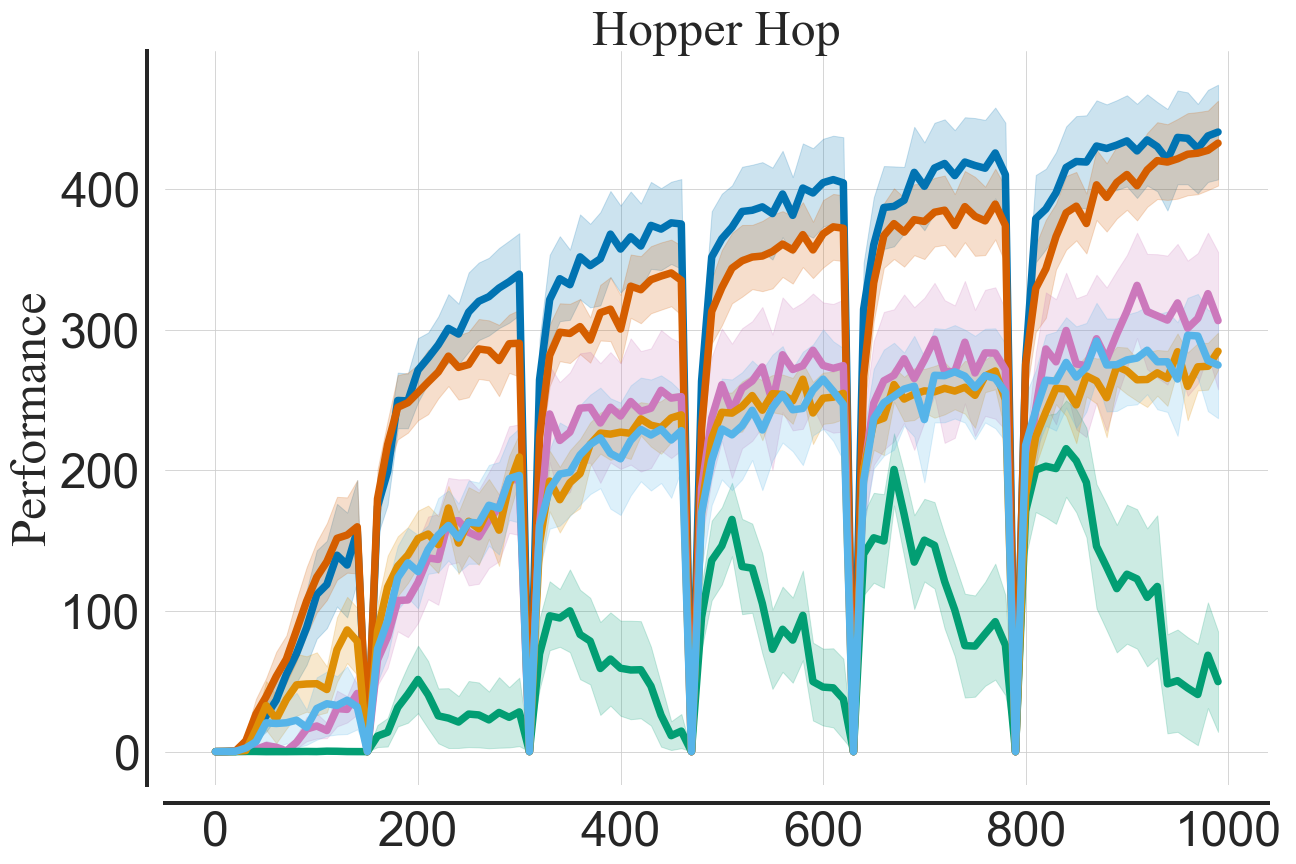}
    \hfill
    \includegraphics[width=0.19\linewidth]{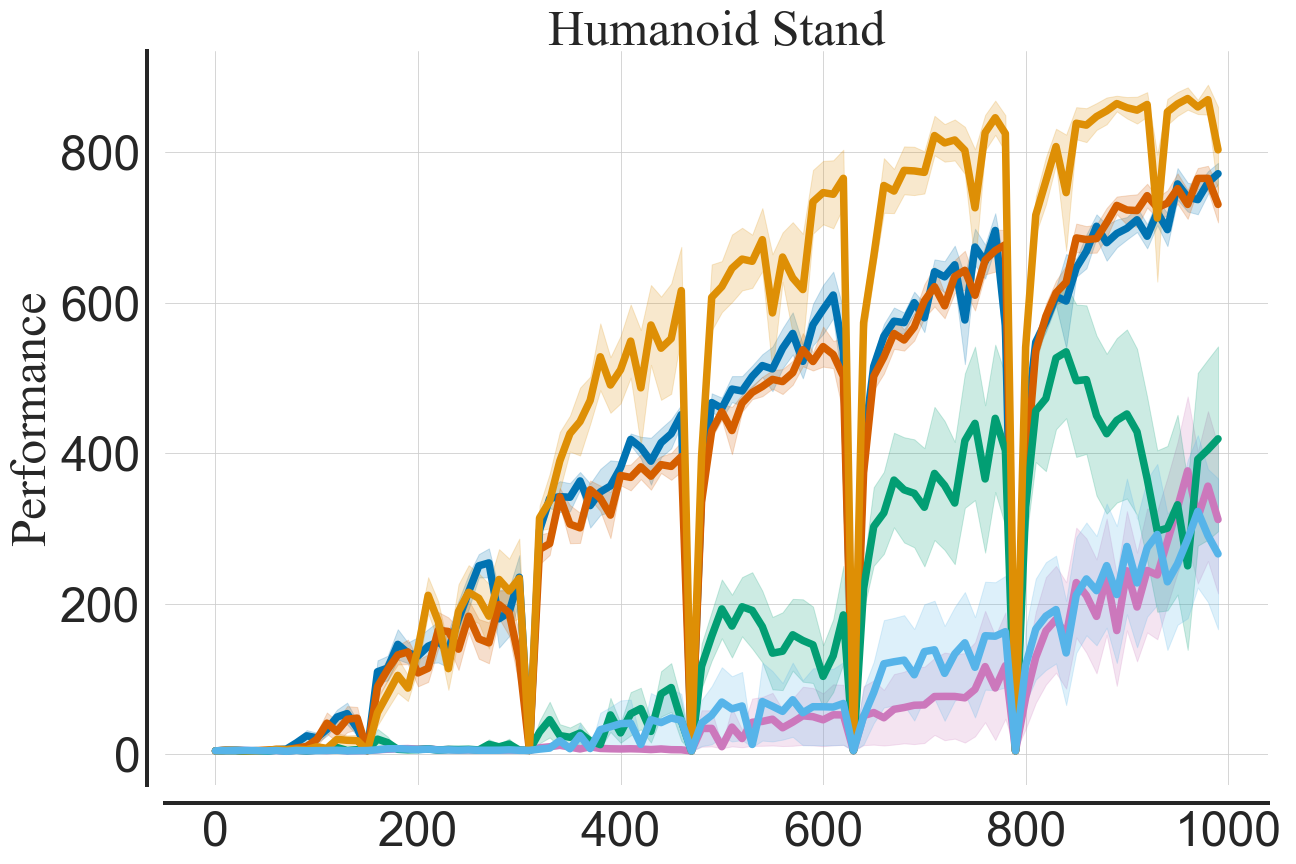}
    \hfill
    \includegraphics[width=0.19\linewidth]{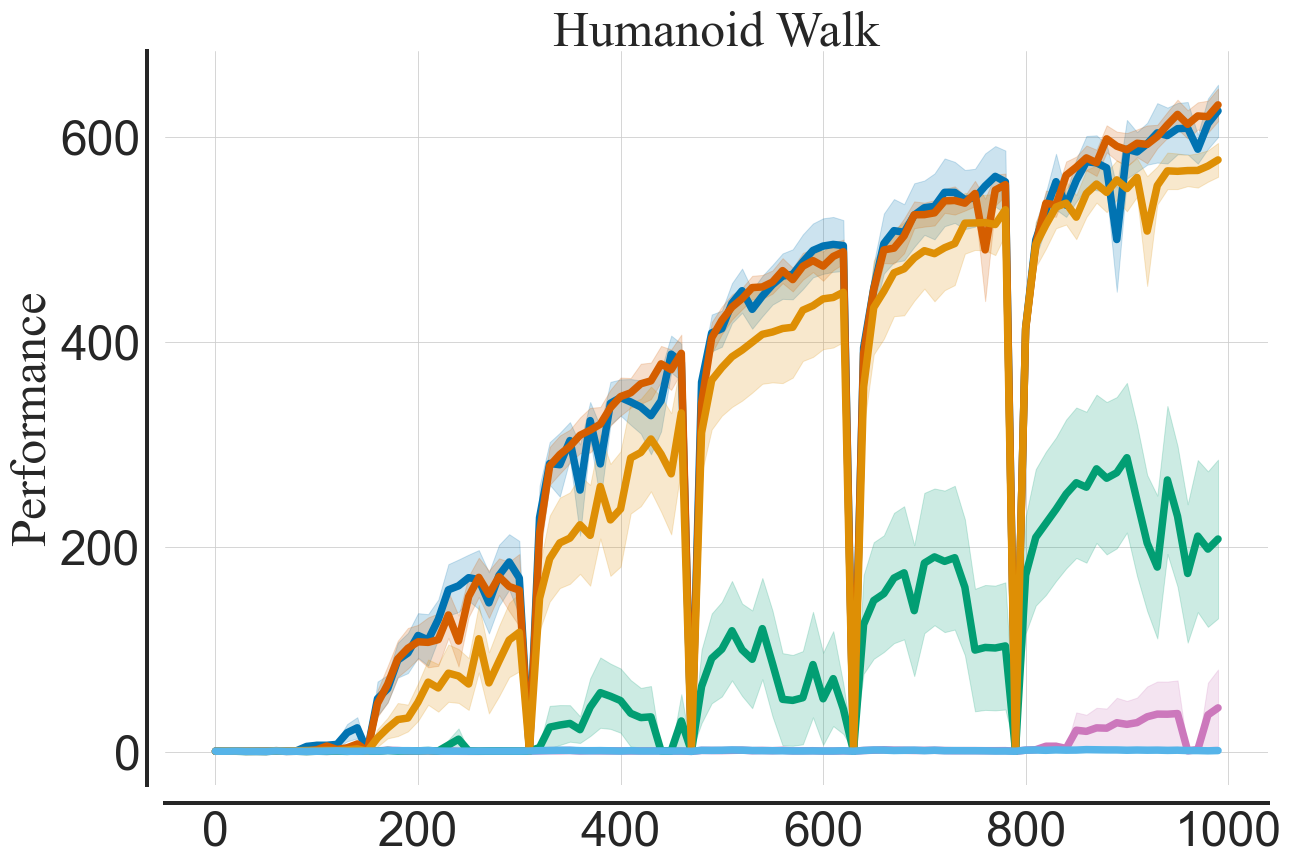}
    \hfill
    \includegraphics[width=0.19\linewidth]{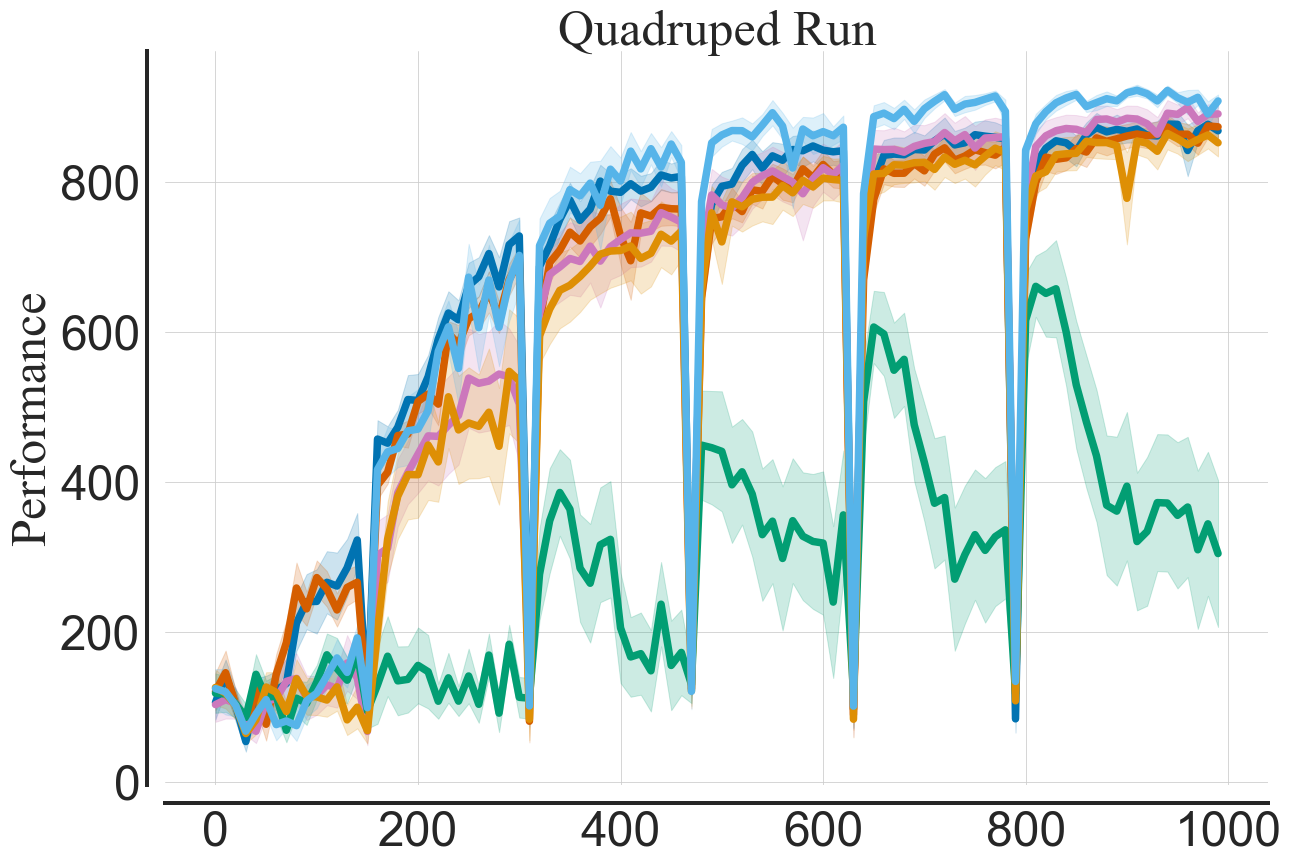}
    \label{fig:ablation_perf}
    \end{subfigure}
\end{minipage}
\begin{minipage}[h]{1.0\linewidth}
    \begin{subfigure}{1.0\linewidth}
    \includegraphics[width=0.19\linewidth]{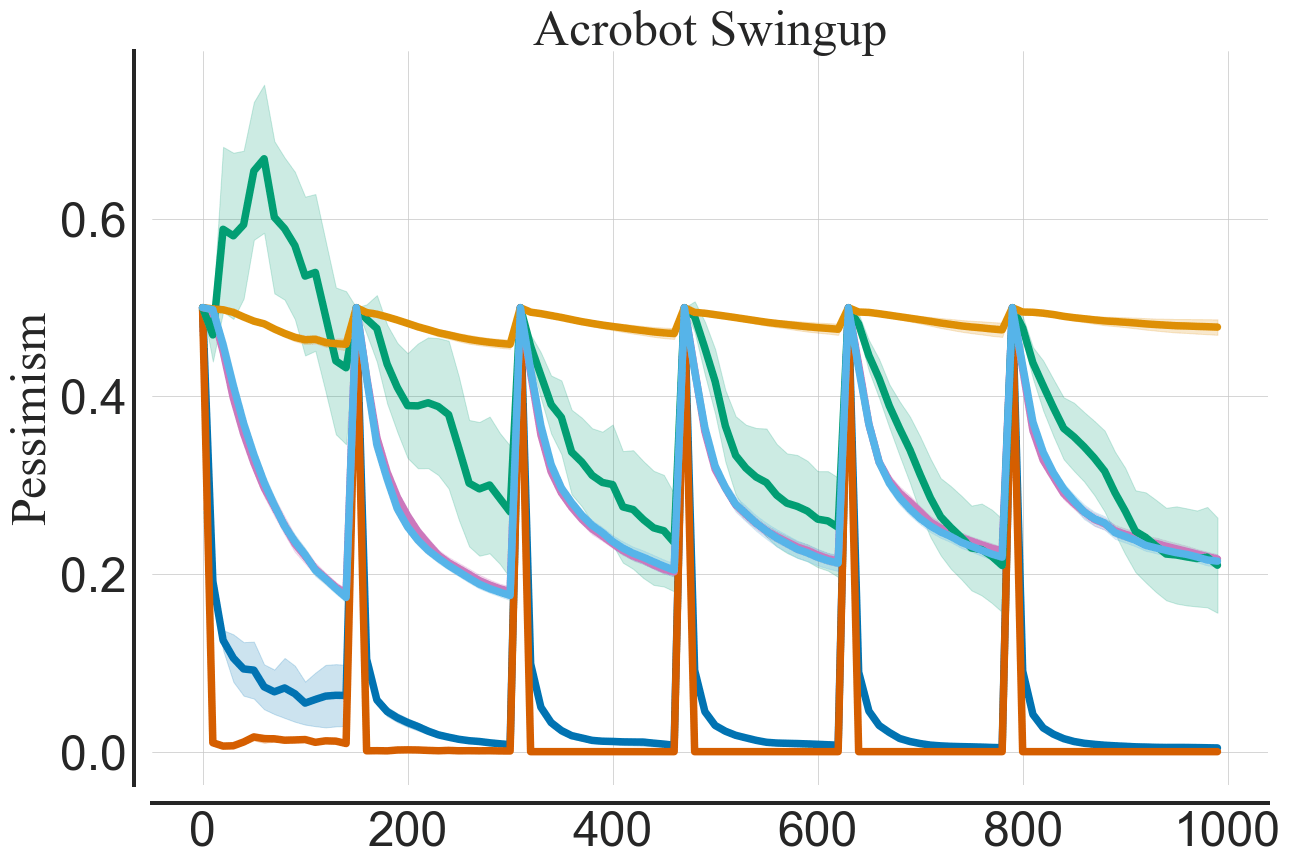}
    \hfill
    \includegraphics[width=0.19\linewidth]{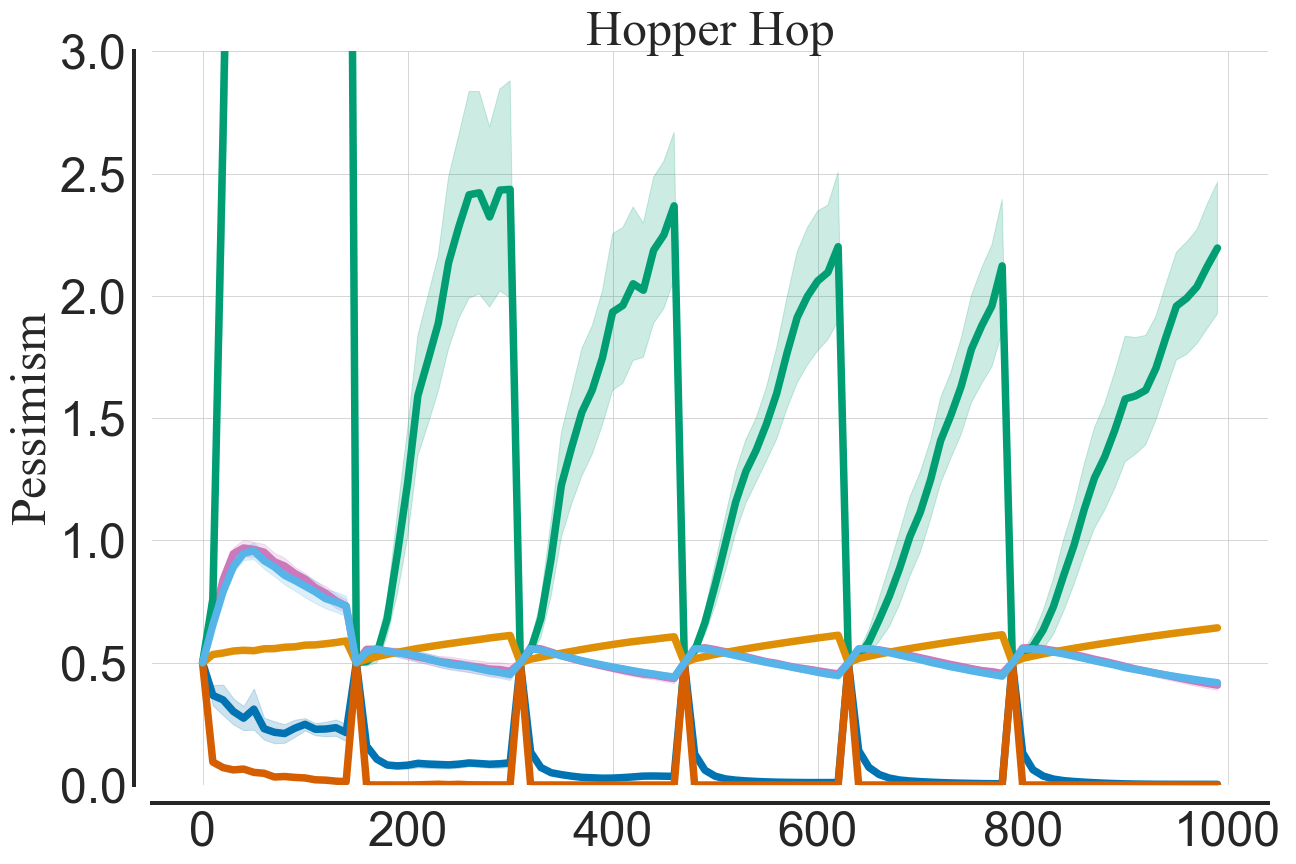}
    \hfill
    \includegraphics[width=0.19\linewidth]{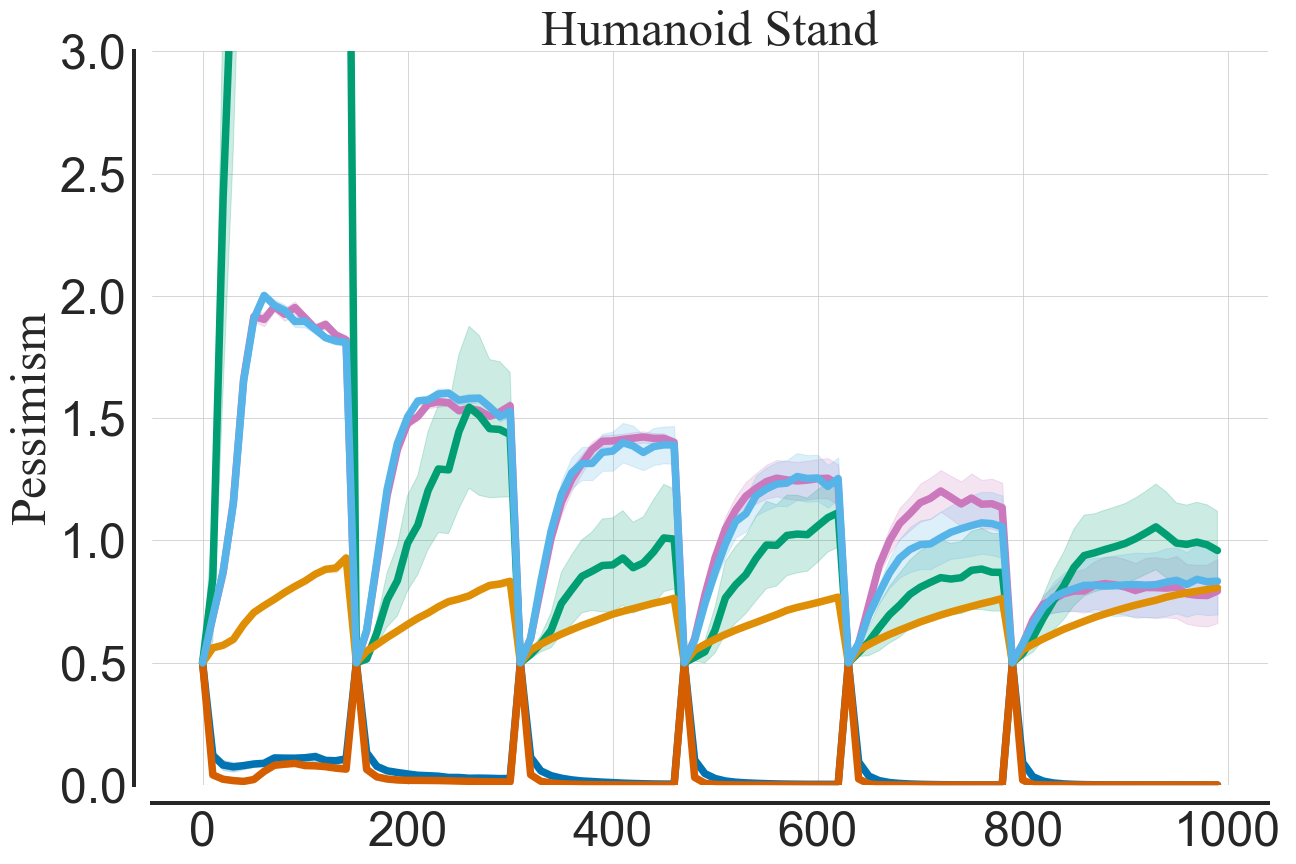}
    \hfill
    \includegraphics[width=0.19\linewidth]{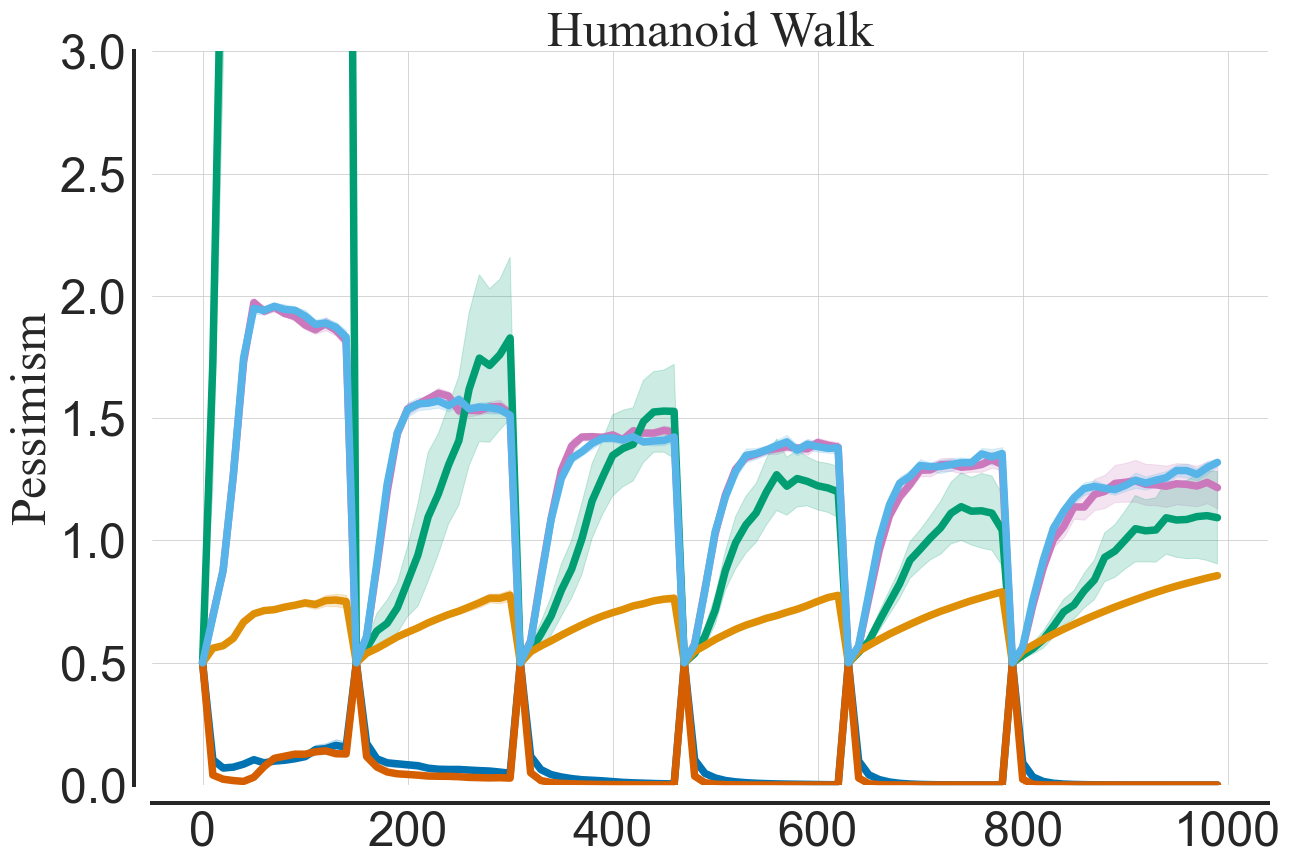}
    \hfill
    \includegraphics[width=0.19\linewidth]{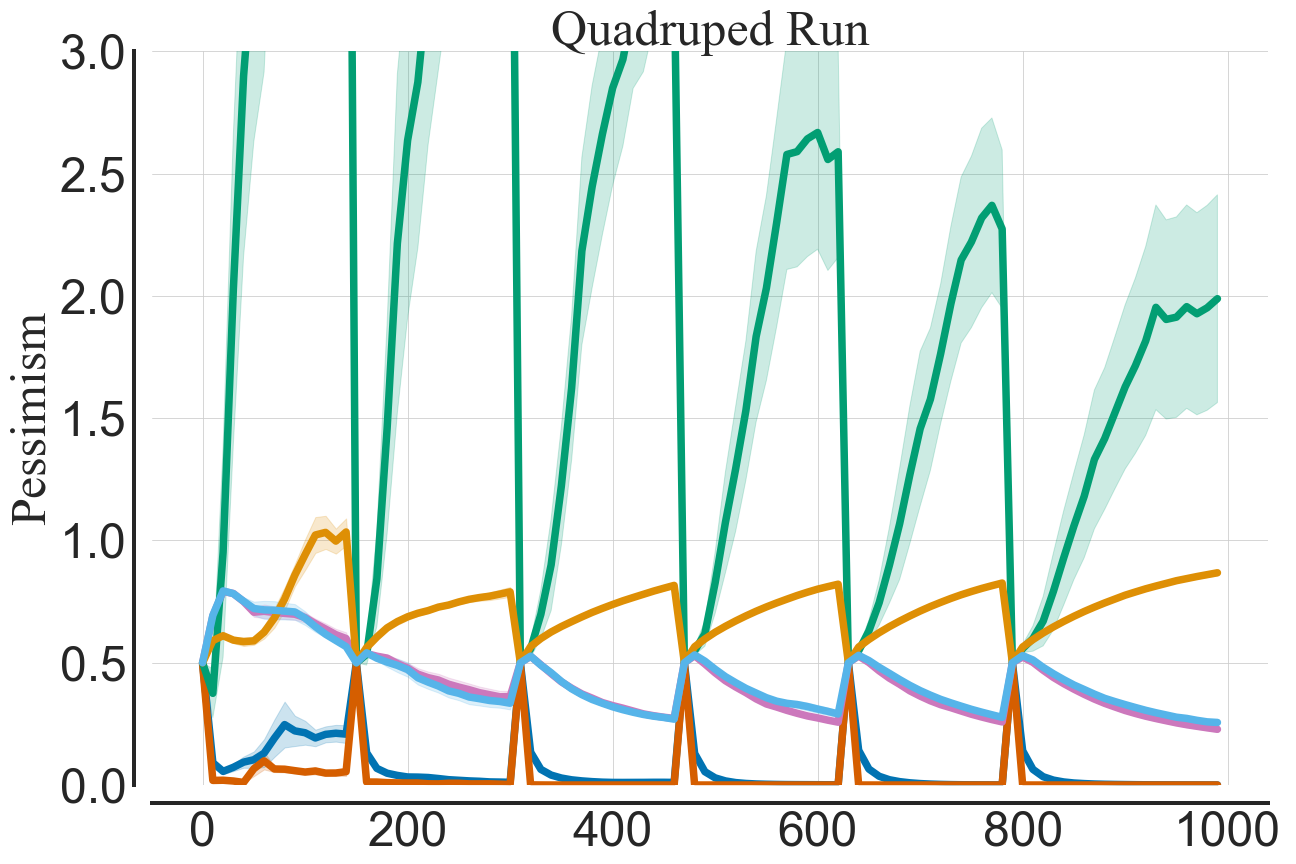}
    \label{fig:ablation_pess}
    \end{subfigure}
\end{minipage}
\caption{Ablation on design of the pessimism module. 10 seeds per task.}
\label{fig:ablation_perf_res}
\end{center}
\end{figure*}

We evaluate the impact of each of VPL contributions, namely the VPL update rule and performing updates on the validation buffer. We evaluate six agents, each utilizing different forms of pessimism loss - either dual optimization (denoted as "Dual") or VPL pessimism loss (denoted as "VPL") - in combination with one of three sources of data for pessimism updates. These sources encompass samples from the validation buffer (denoted as "Validation"), the replay buffer (denoted as "Replay" and used originally in GPL \citep{cetin2023learning}), and the most recent online transitions (denoted as "Online" and used originally in OPL \citep{kuznetsov2021automating}). The performance, pessimism, approximation error and overfitting of these agents is presented in Figures \ref{fig:ablation_perf_res_aggregate} and \ref{fig:ablation_perf_res}.

We find that the VPL pessimism adjustment loss accounts for the majority of the performance improvements of VPL algorithm. As such, we find that updating the pessimism on validation buffer accounts for minor performance improvements over updates performed on the replay buffer. We still deem this result as significant, since the validation agent skips $\frac{1}{16}$ of training transitions. Furthermore, we find that performing VPL pessimism updated on recent transitions leads to pessimism increases and subpar performance. Interestingly, we find that performing dual optimization pessimism updates on the validation buffer leads to the worse performing agent. In terms of pessimism patterns, we observe that performing pessimism updates on the online transitions leads to more pessimistic agents than then other data sources for pessimism updates. Finally, we find that the VPL pessimism updates performed on the replay data leads to sharper decreases of pessimism that the regular validation VPL.

Finally, we investigate whether the critic disagreement diminishes to zero in a popular training regime of 1mln environment steps. As noted in the paper, the convergence of pessimistic actor-critic depends on the critic disagreement being equal to zero. We find that the critic disagreement indeed does not completely diminish. Interestingly, we observe that the same environments yield the most disagreement in both low and high replay regimes.

\begin{figure*}[h!]
\begin{center}
\begin{minipage}[h]{1.0\linewidth}
    \begin{subfigure}{1.0\linewidth}
    \hfill
    \includegraphics[width=0.24\linewidth]{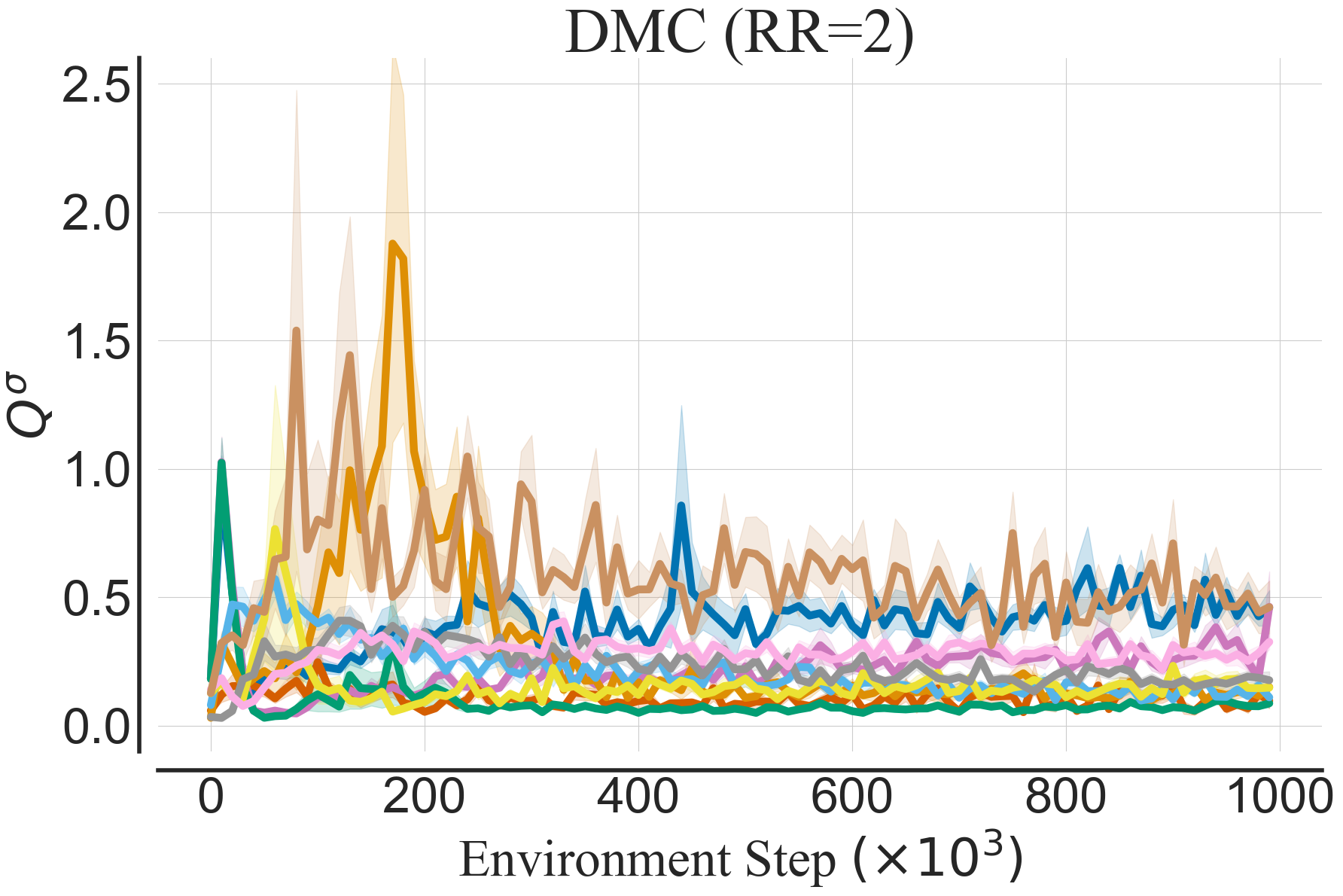}
    \hfill
    \includegraphics[width=0.24\linewidth]{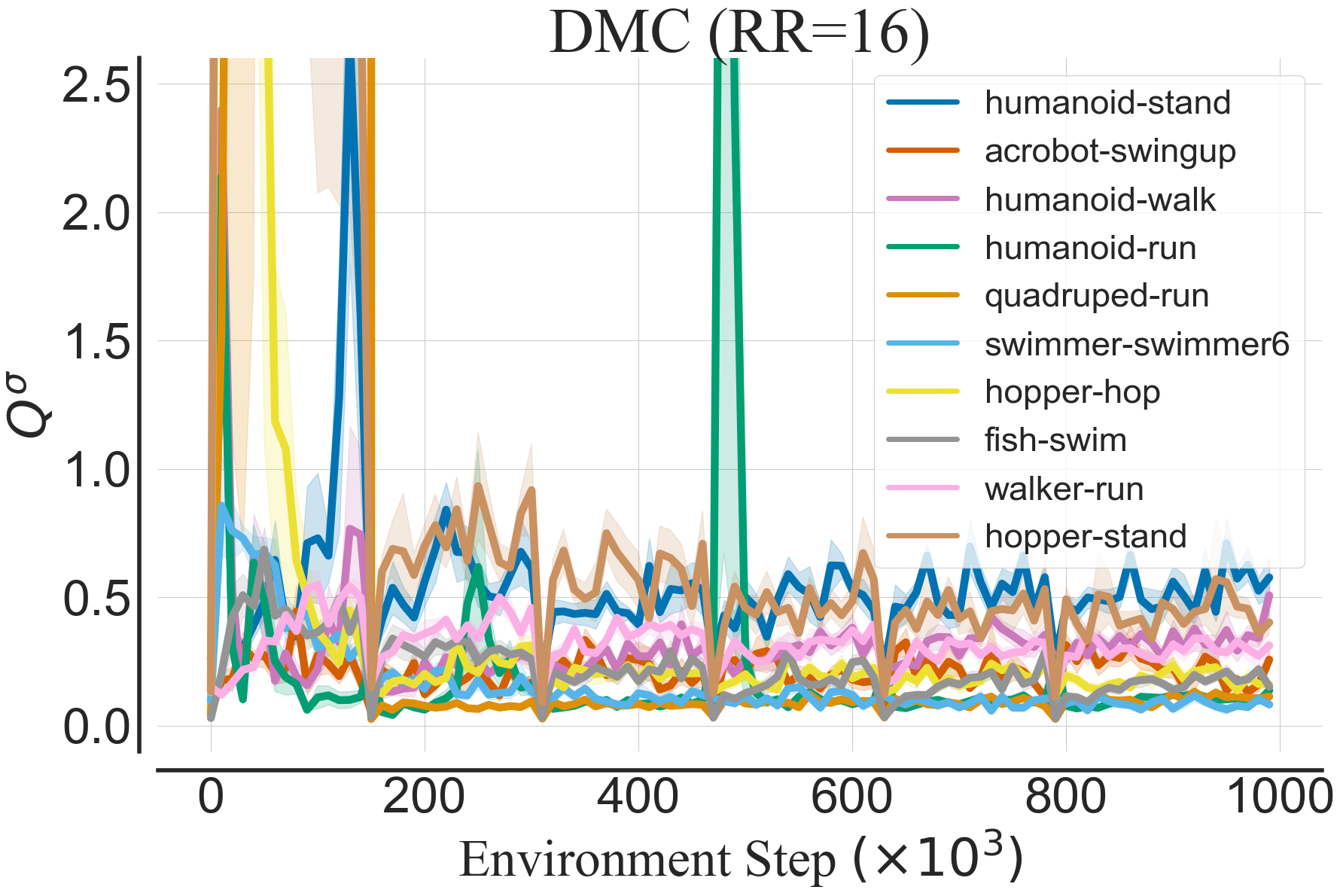}
    \hfill
    \includegraphics[width=0.24\linewidth]{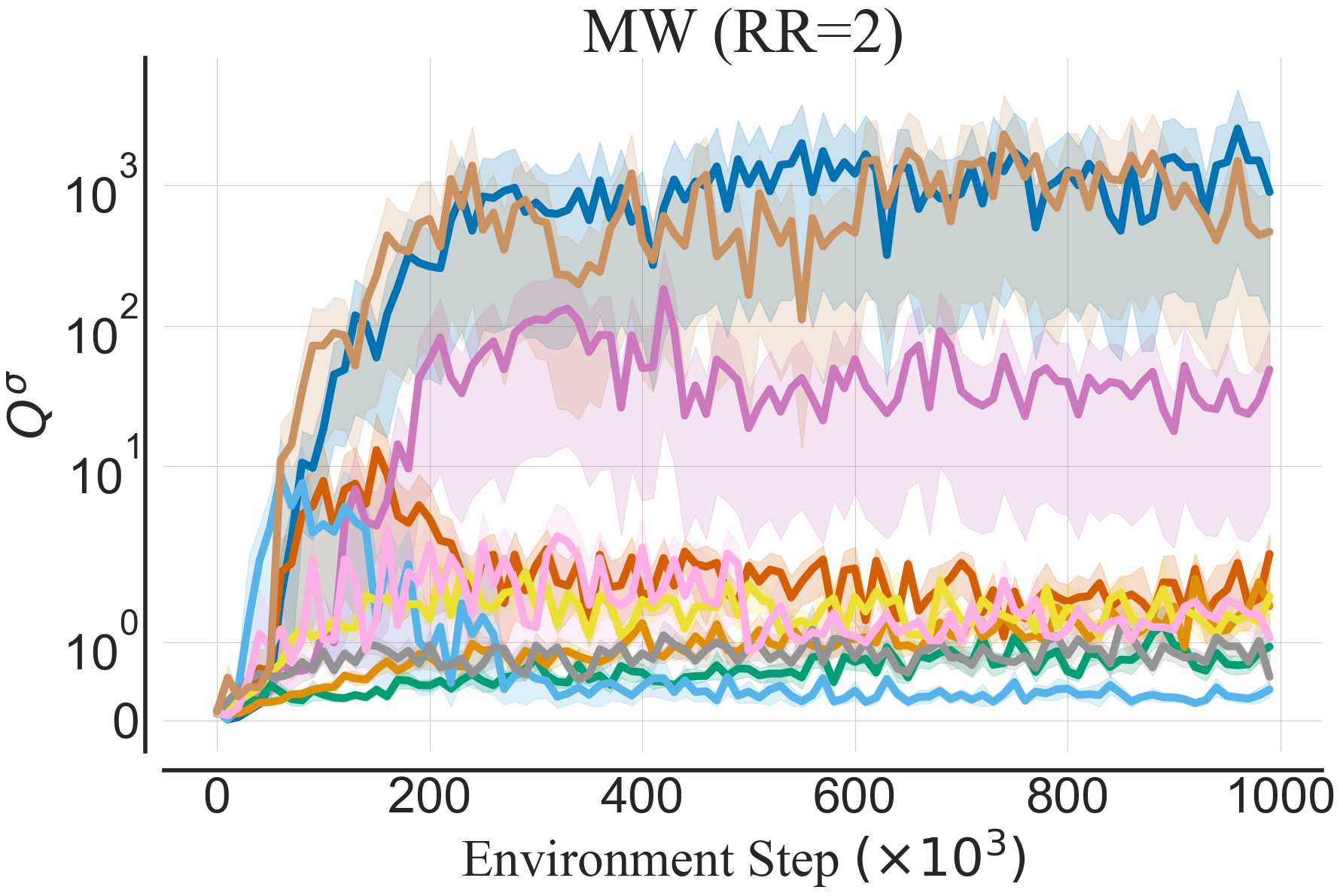}
    \hfill
    \includegraphics[width=0.24\linewidth]{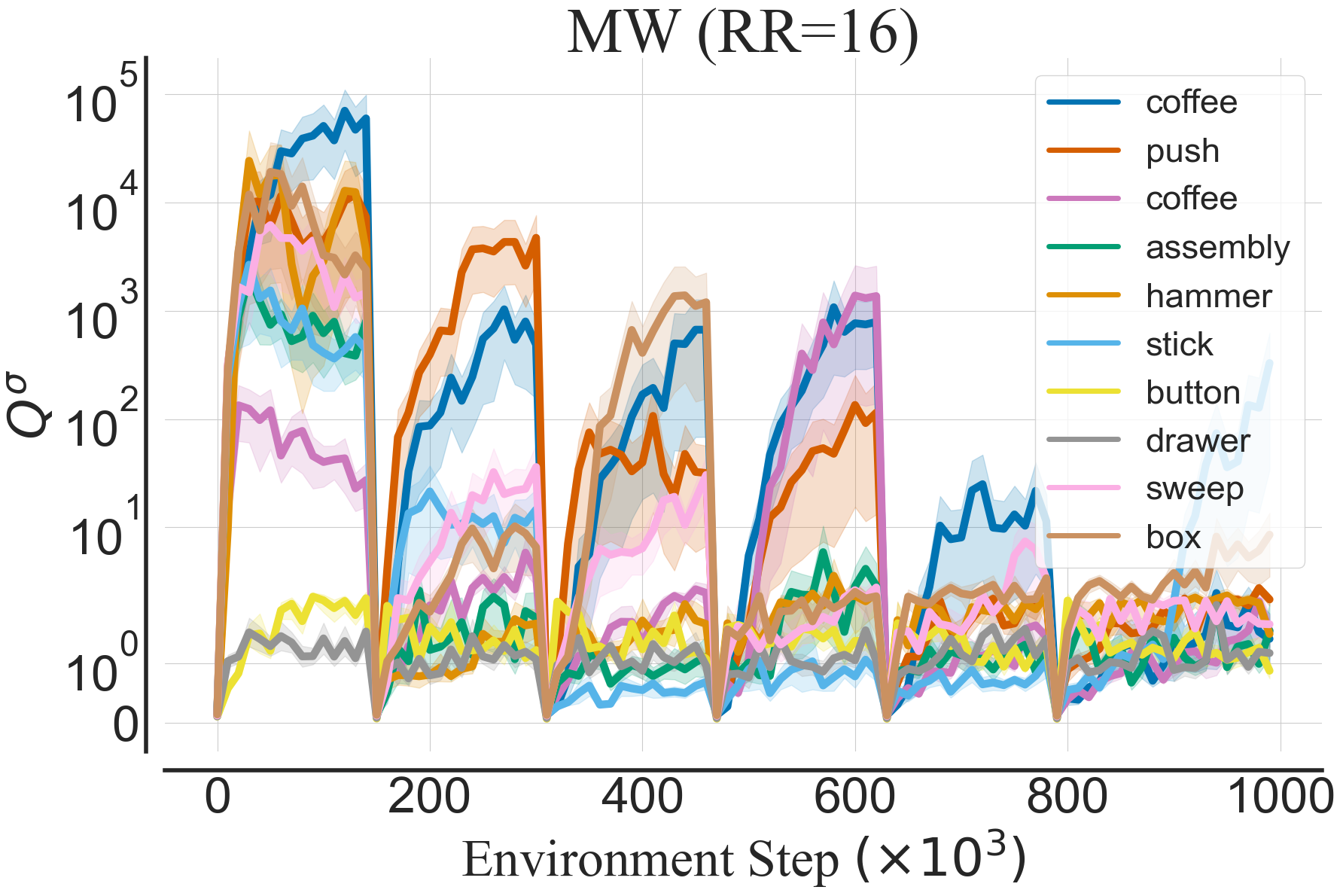}
    \hfill
    \end{subfigure}
\end{minipage}
\caption{The critic disagreement of SAC algorithm does not completely diminish in the considered training regime of 1 million environment steps, making adjustments to the pessimism a viable strategy. 10 seeds per task.}
\label{fig:critic_sigma}
\end{center}
\end{figure*}

\section{Learning Curves}
\label{appendix:learning_curves}

\begin{figure*}[ht!]
\begin{center}
\begin{minipage}[h]{1.0\linewidth}
\centering
    \begin{subfigure}{0.88\linewidth}
    \includegraphics[width=\textwidth]{images/legend_1.png}
    \end{subfigure}
\end{minipage}
\bigskip
\begin{minipage}[h]{1.0\linewidth}
    \begin{subfigure}{1.0\linewidth}
    \hfill
    \includegraphics[width=0.23\linewidth]{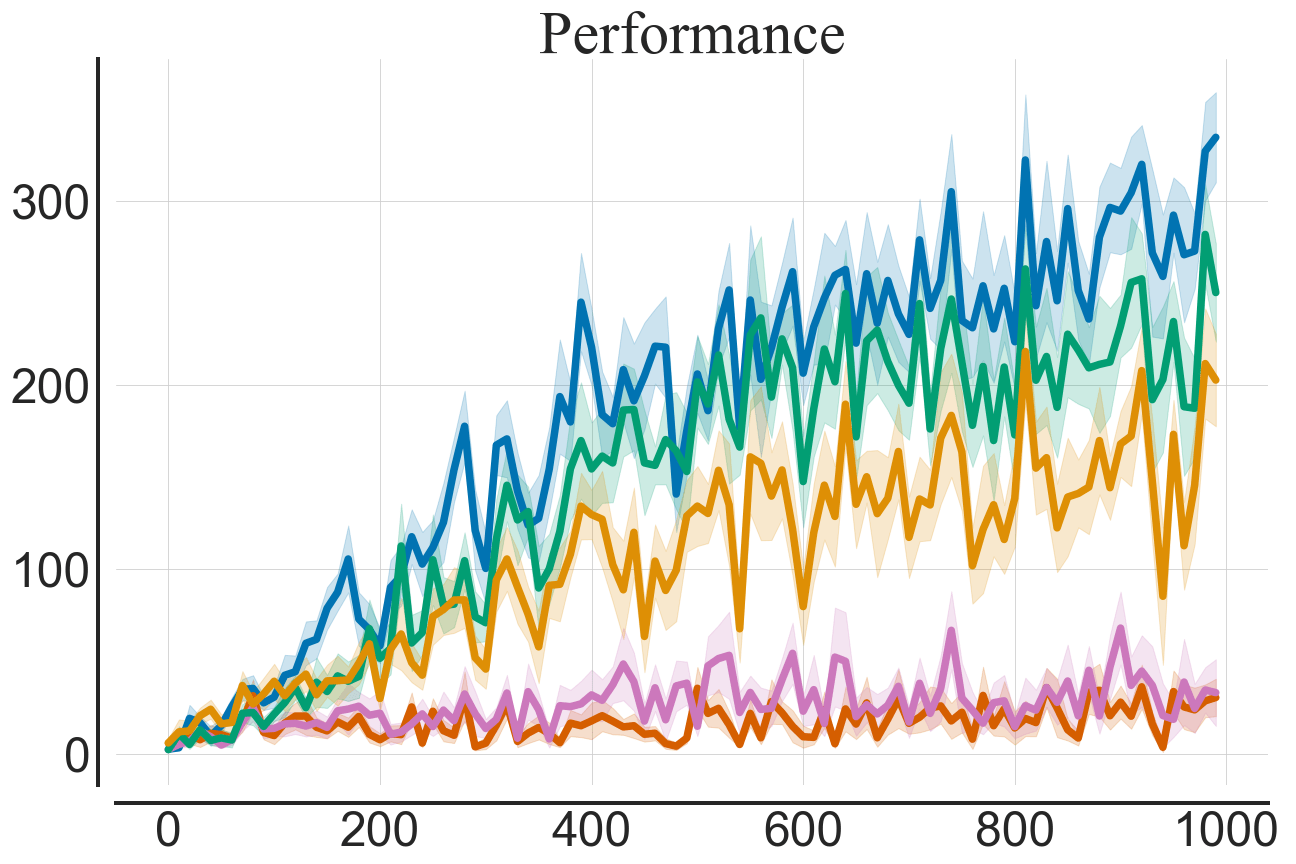}
    \hfill
    \includegraphics[width=0.23\linewidth]{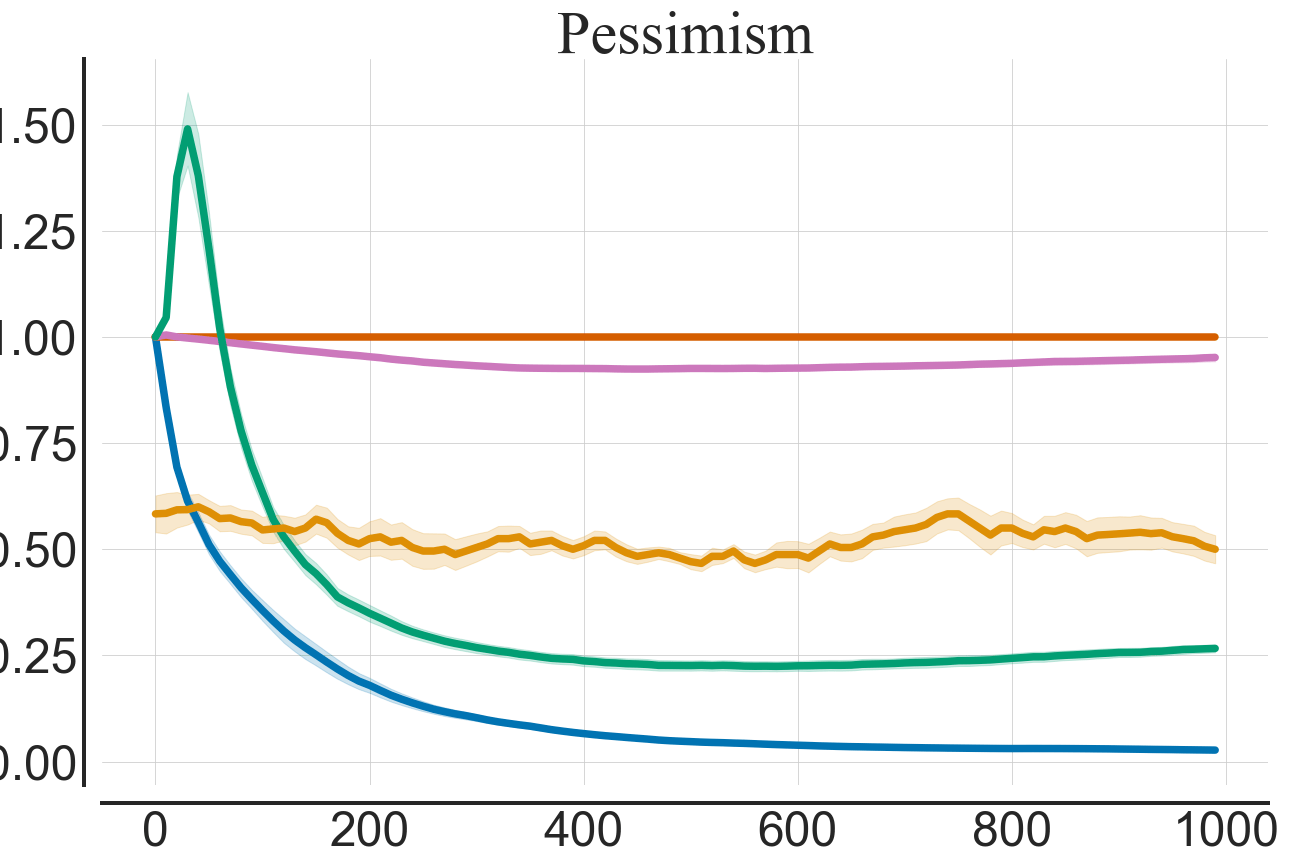}
    \hfill
    \includegraphics[width=0.23\linewidth]{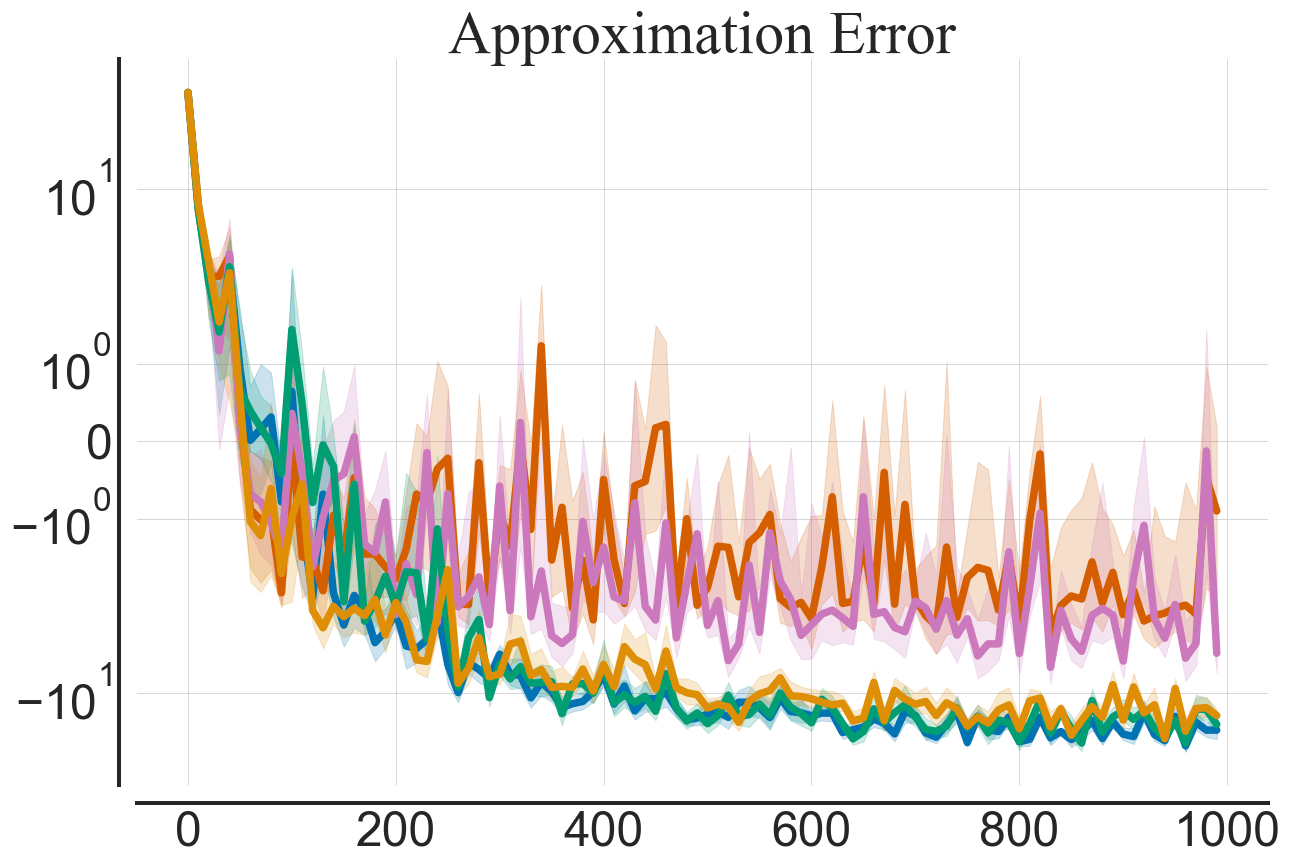}
    \hfill
    \includegraphics[width=0.23\linewidth]{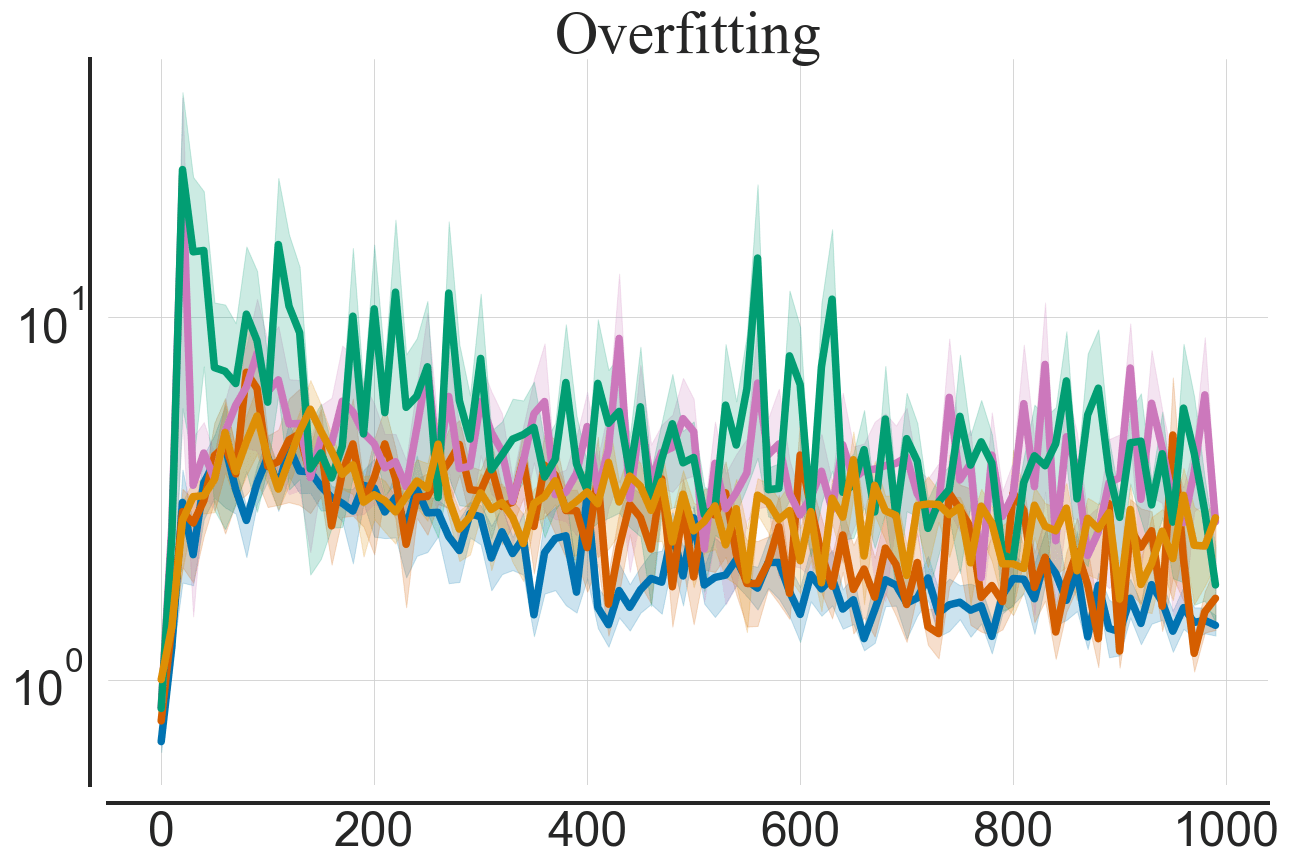}
    \hfill
    \end{subfigure}
    \subcaption{Acrobot Swingup}
\end{minipage}
\bigskip
\begin{minipage}[h]{1.0\linewidth}
    \begin{subfigure}{1.0\linewidth}
    \hfill
    \includegraphics[width=0.23\linewidth]{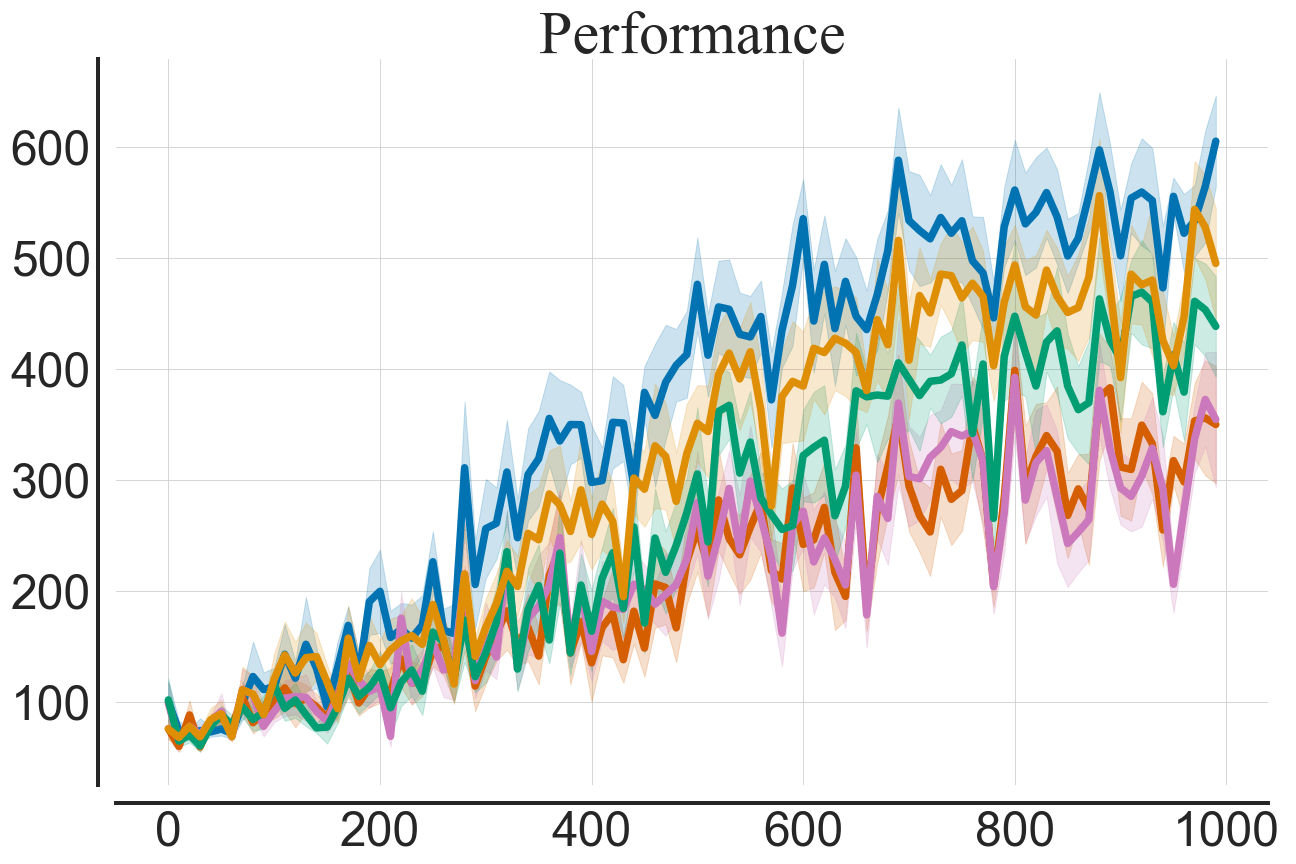}
    \hfill
    \includegraphics[width=0.23\linewidth]{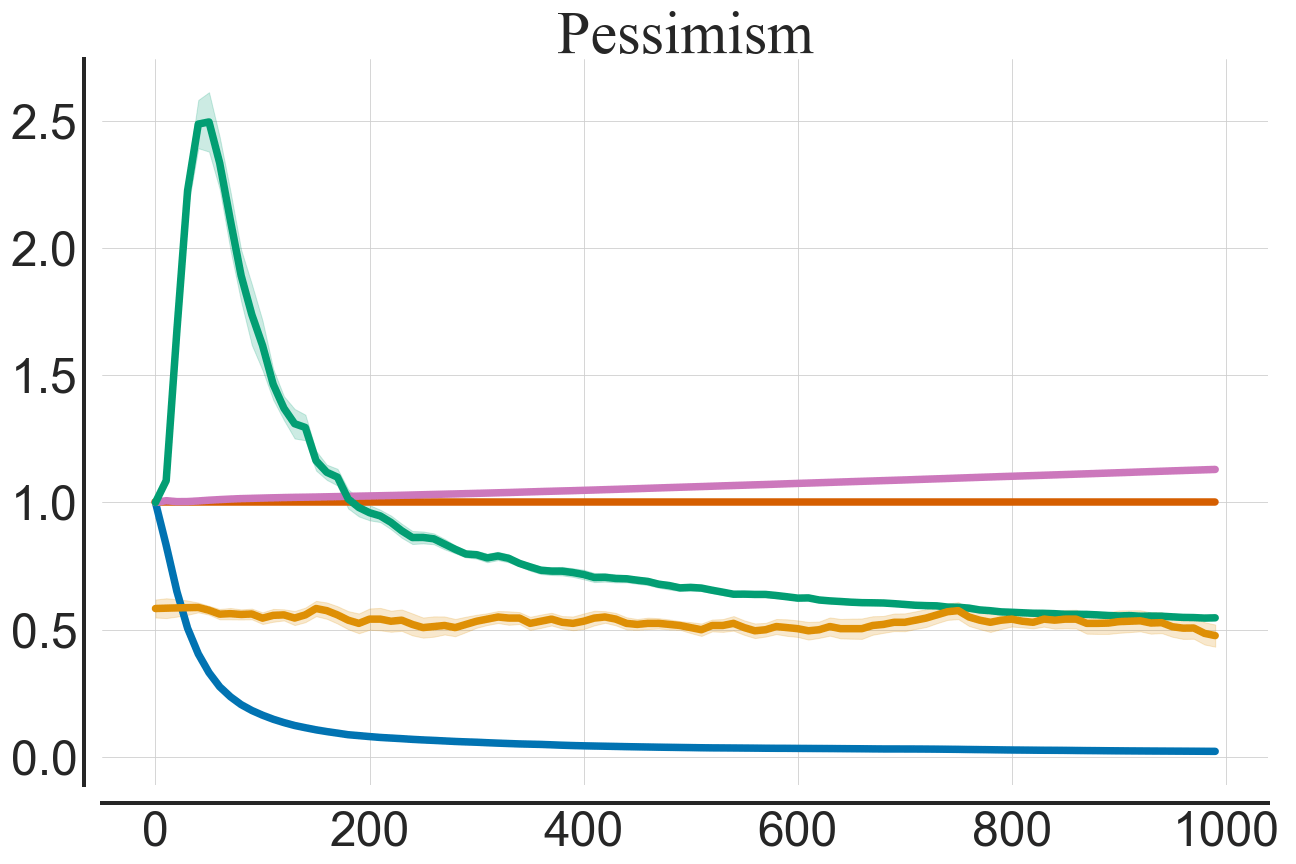}
    \hfill
    \includegraphics[width=0.23\linewidth]{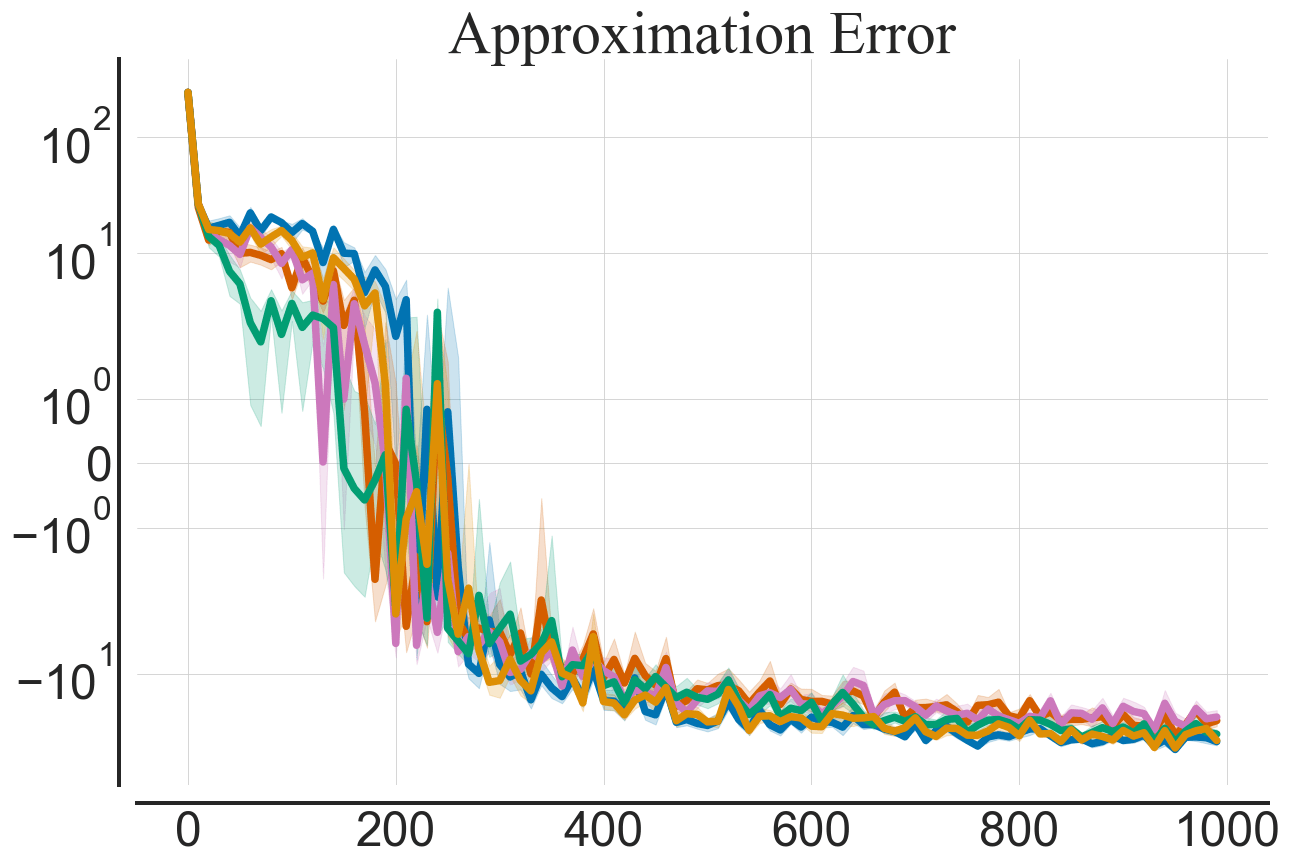}
    \hfill
    \includegraphics[width=0.23\linewidth]{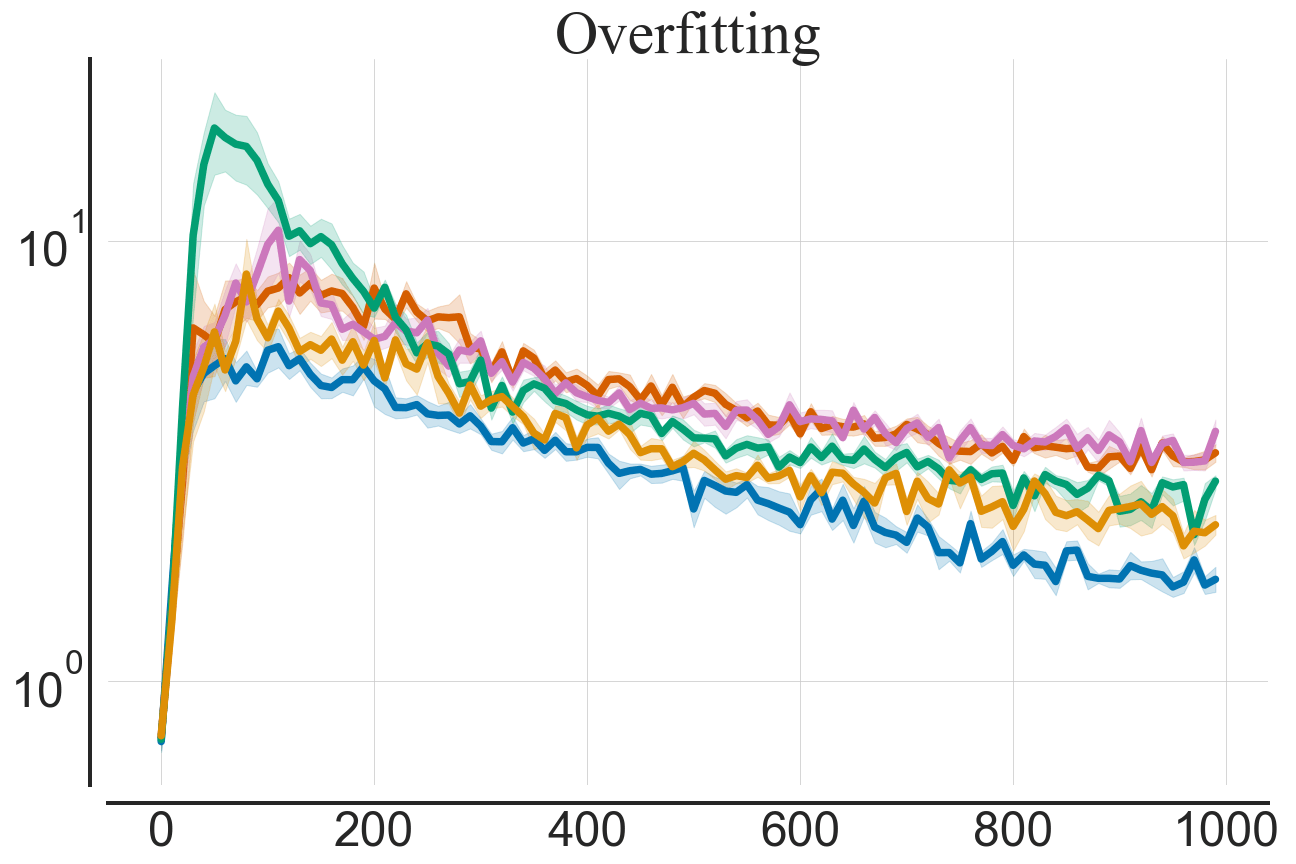}
    \hfill
    \end{subfigure}
    \subcaption{Fish Swim}
\end{minipage}
\bigskip
\begin{minipage}[h]{1.0\linewidth}
    \begin{subfigure}{1.0\linewidth}
    \hfill
    \includegraphics[width=0.23\linewidth]{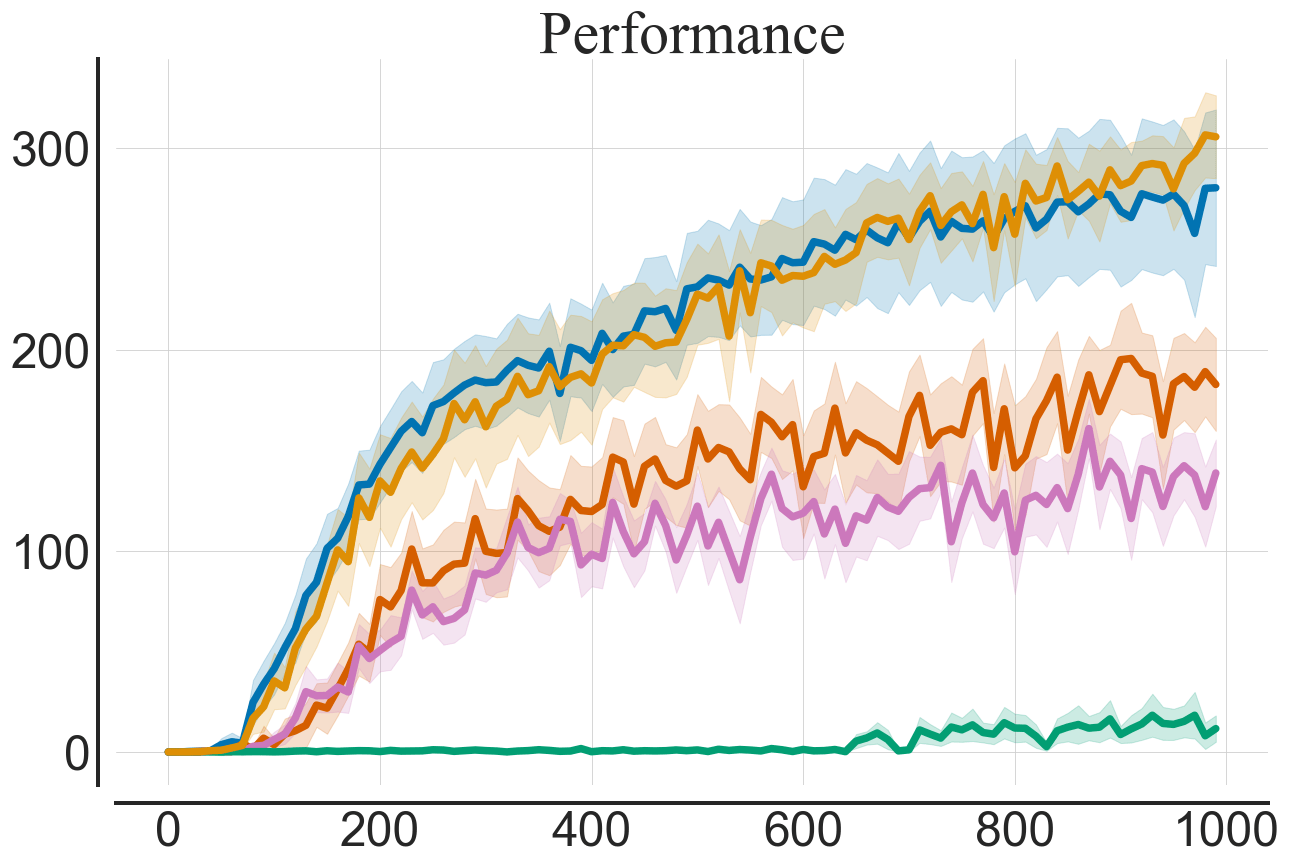}
    \hfill
    \includegraphics[width=0.23\linewidth]{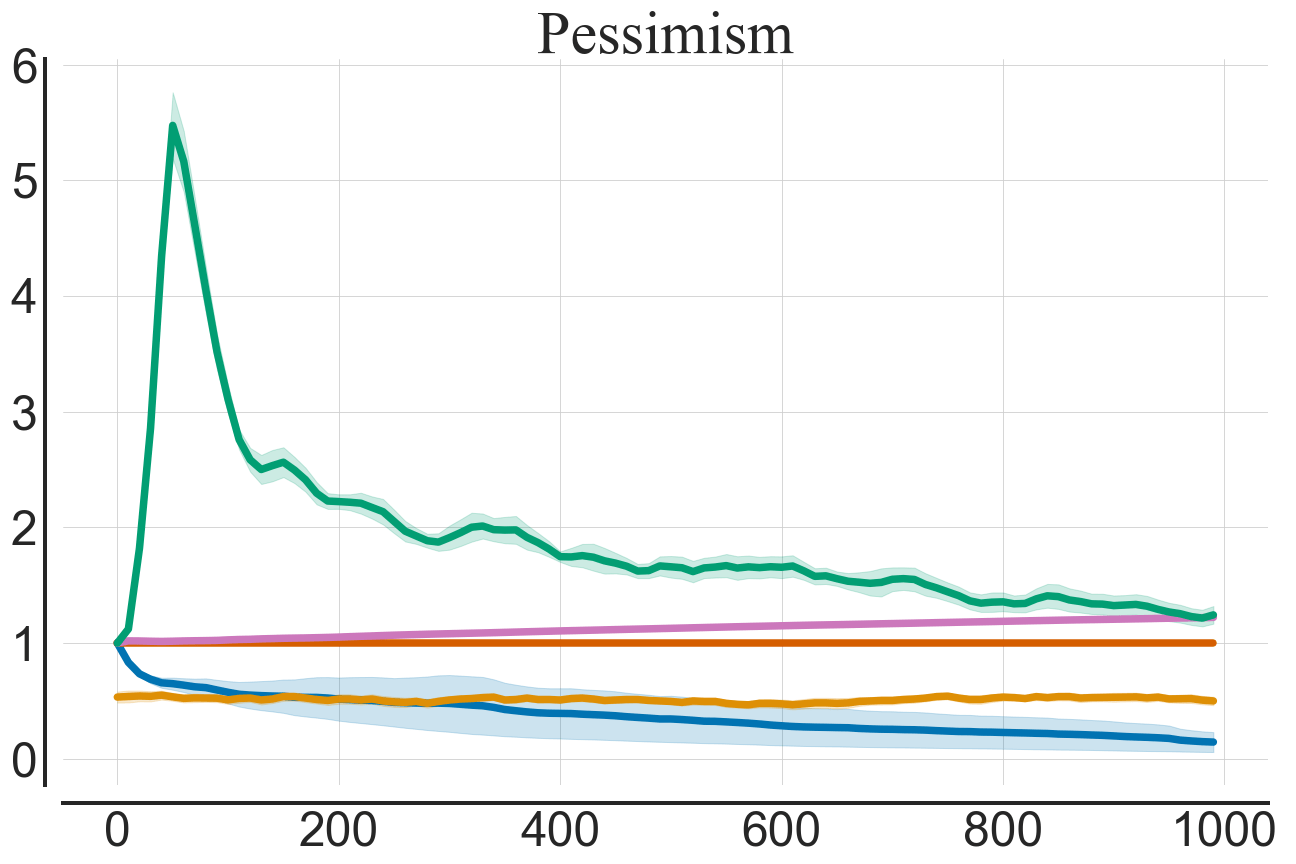}
    \hfill
    \includegraphics[width=0.23\linewidth]{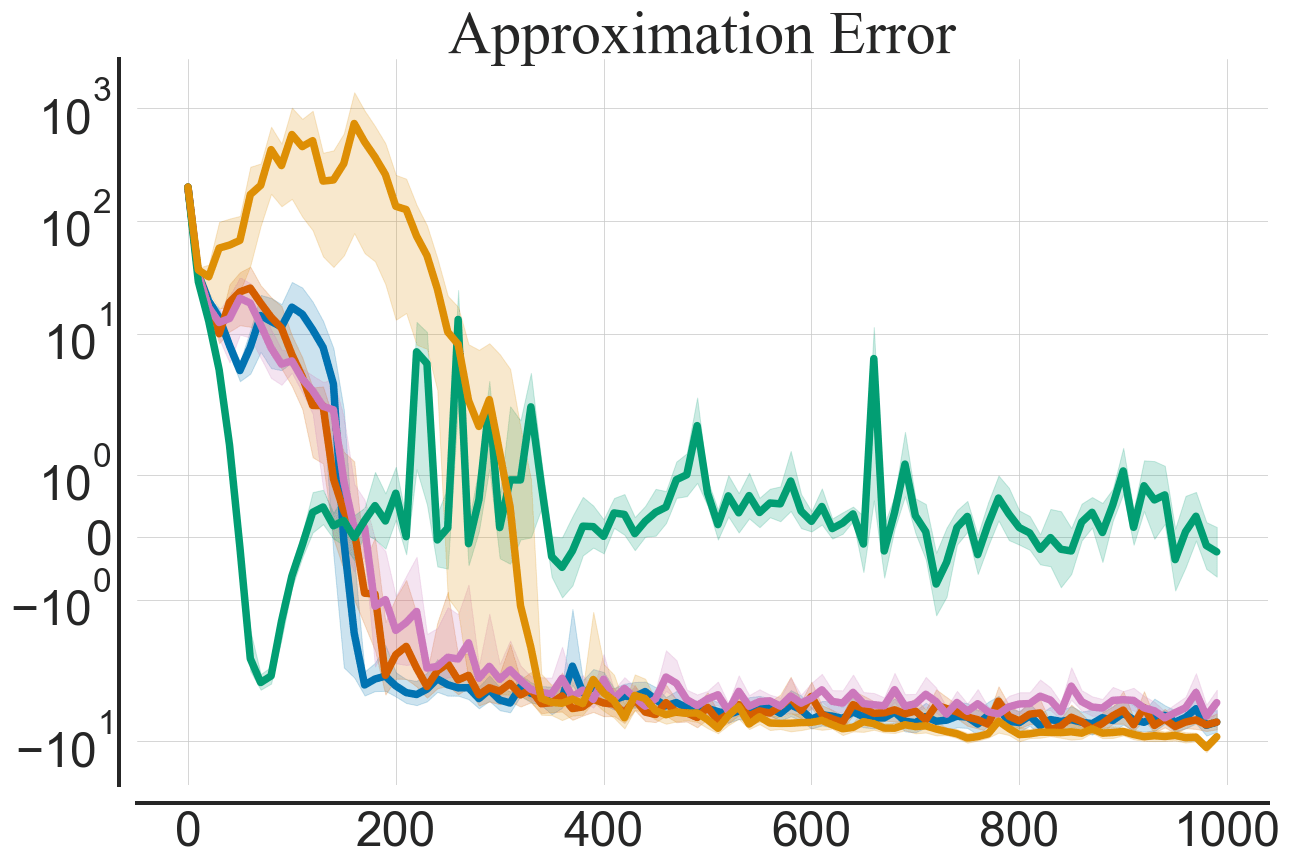}
    \hfill
    \includegraphics[width=0.23\linewidth]{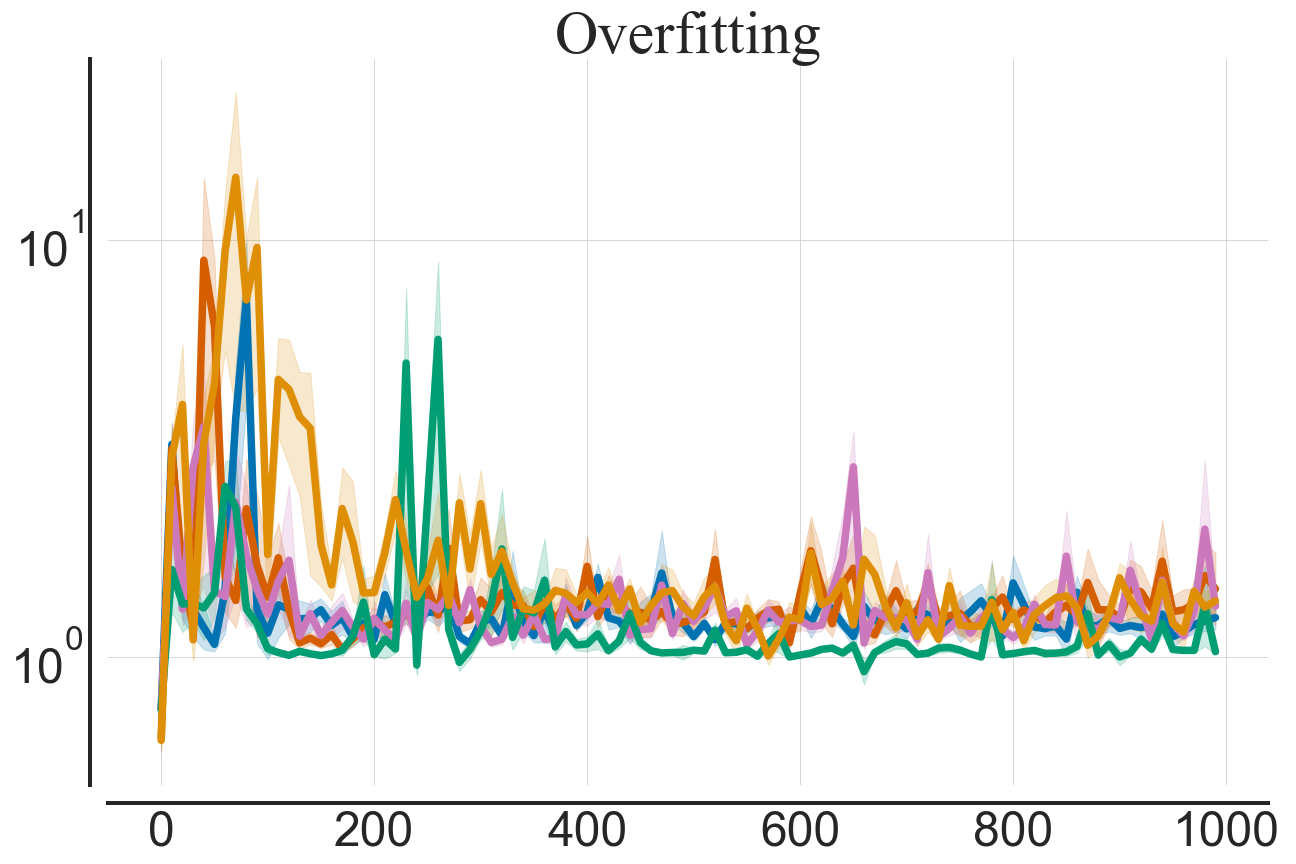}
    \hfill
    \end{subfigure}
    \subcaption{Hopper Hop}
\end{minipage}
\bigskip
\begin{minipage}[h]{1.0\linewidth}
    \begin{subfigure}{1.0\linewidth}
    \hfill
    \includegraphics[width=0.23\linewidth]{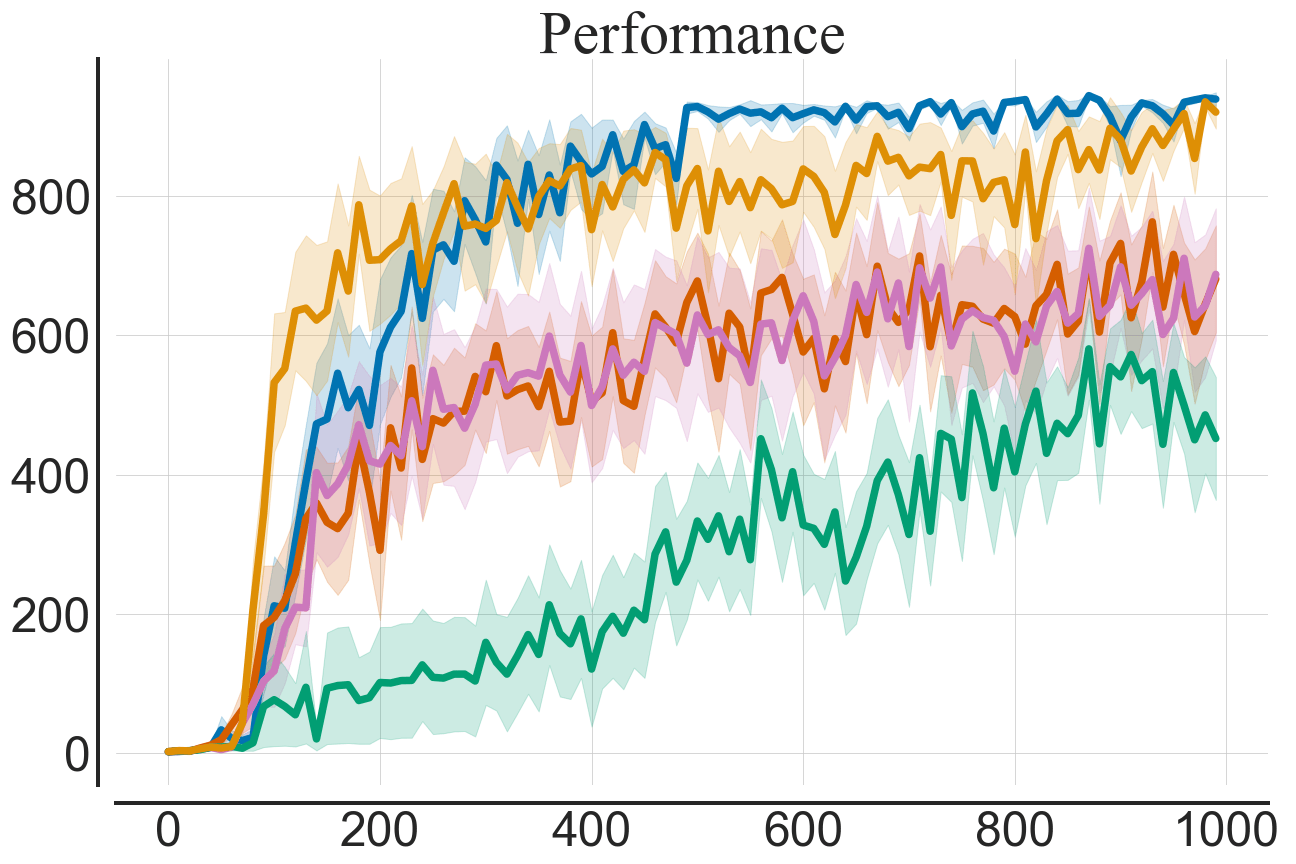}
    \hfill
    \includegraphics[width=0.23\linewidth]{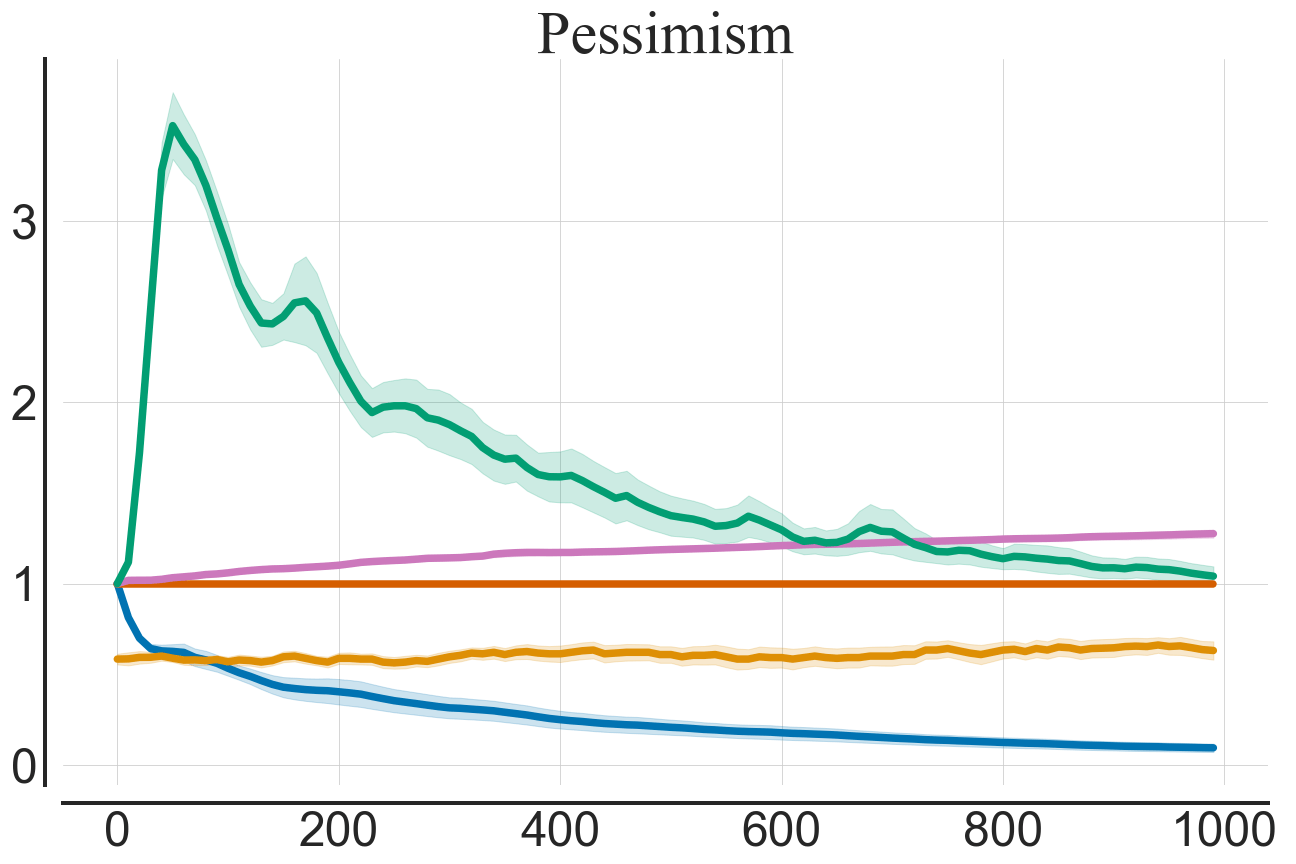}
    \hfill
    \includegraphics[width=0.23\linewidth]{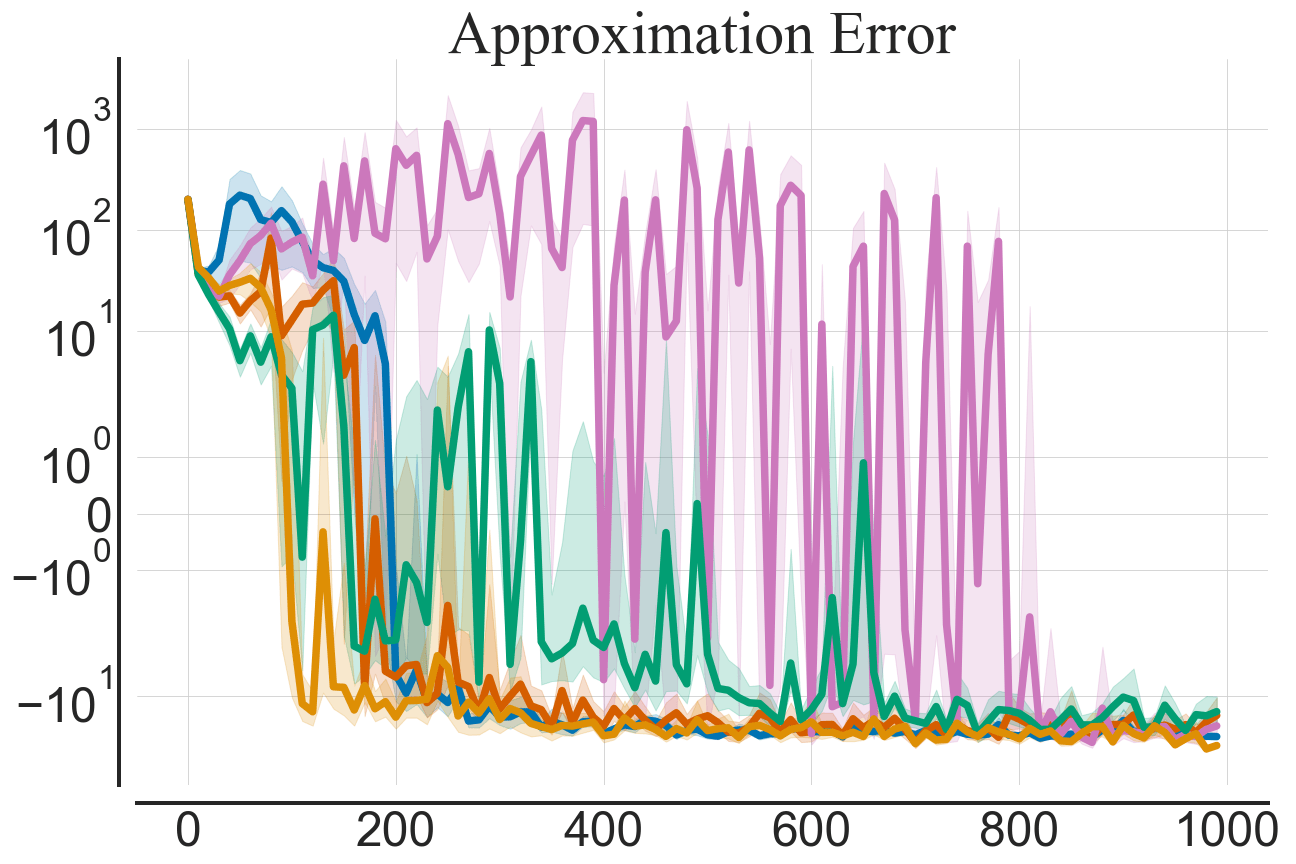}
    \hfill
    \includegraphics[width=0.23\linewidth]{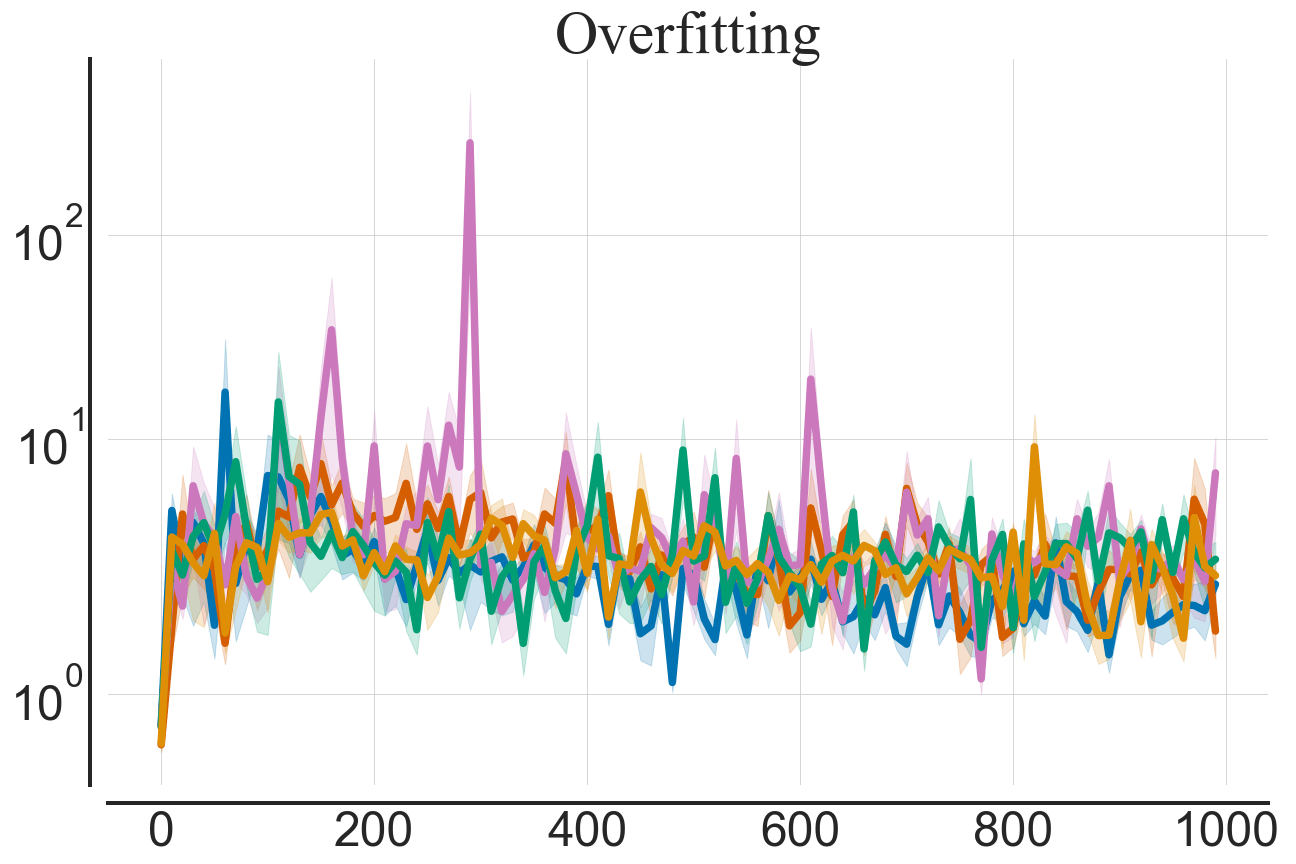}
    \hfill
    \end{subfigure}
    \subcaption{Hopper Stand}
\end{minipage}
\bigskip
\begin{minipage}[h]{1.0\linewidth}
    \begin{subfigure}{1.0\linewidth}
    \hfill
    \includegraphics[width=0.23\linewidth]{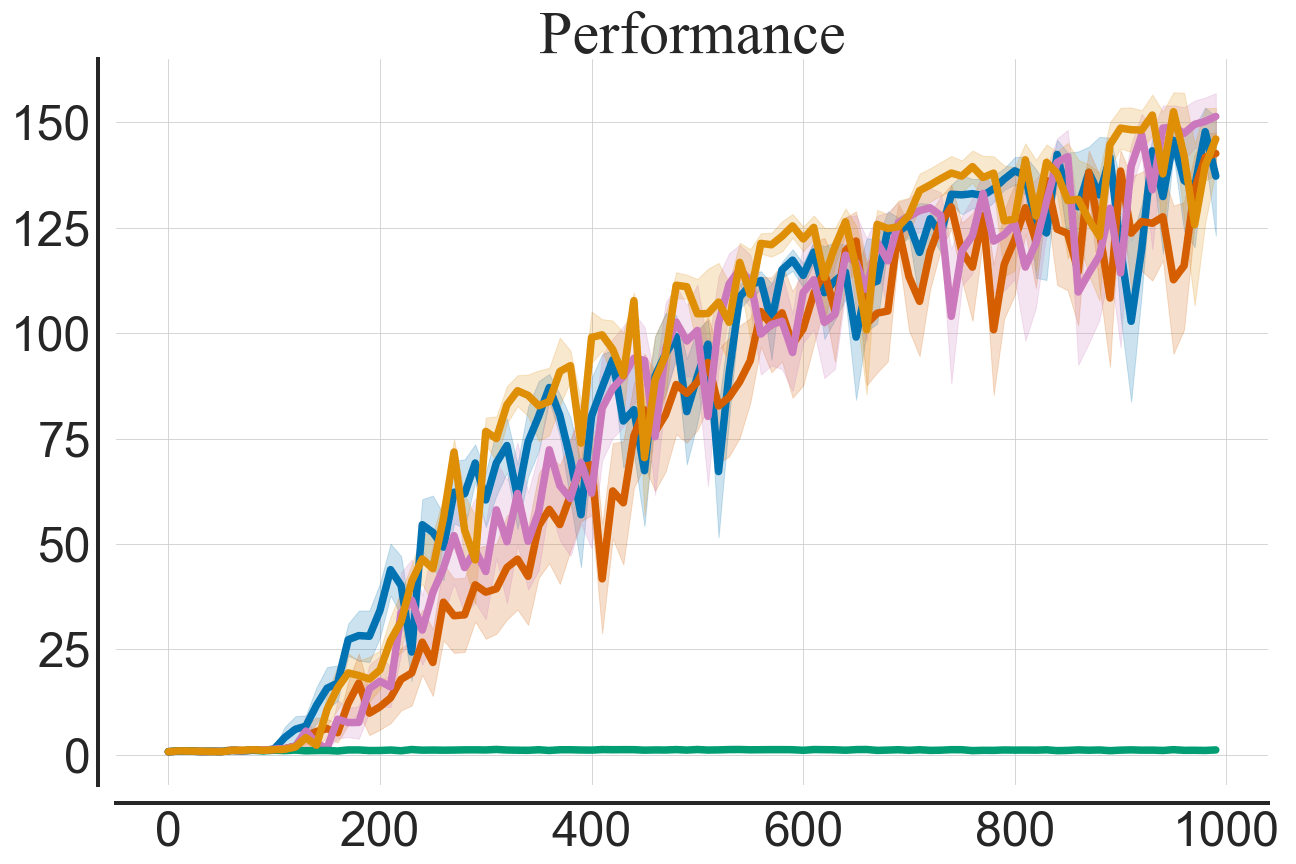}
    \hfill
    \includegraphics[width=0.23\linewidth]{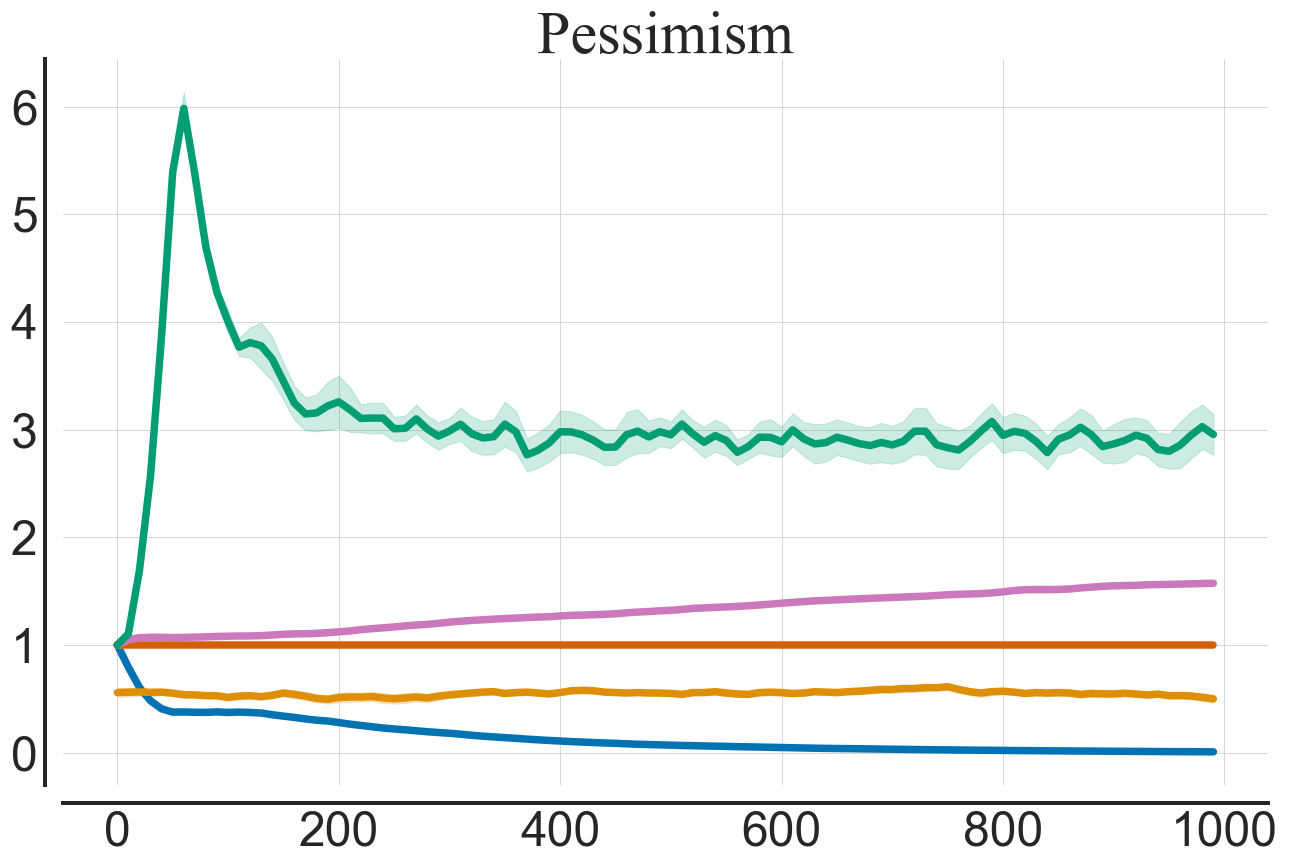}
    \hfill
    \includegraphics[width=0.23\linewidth]{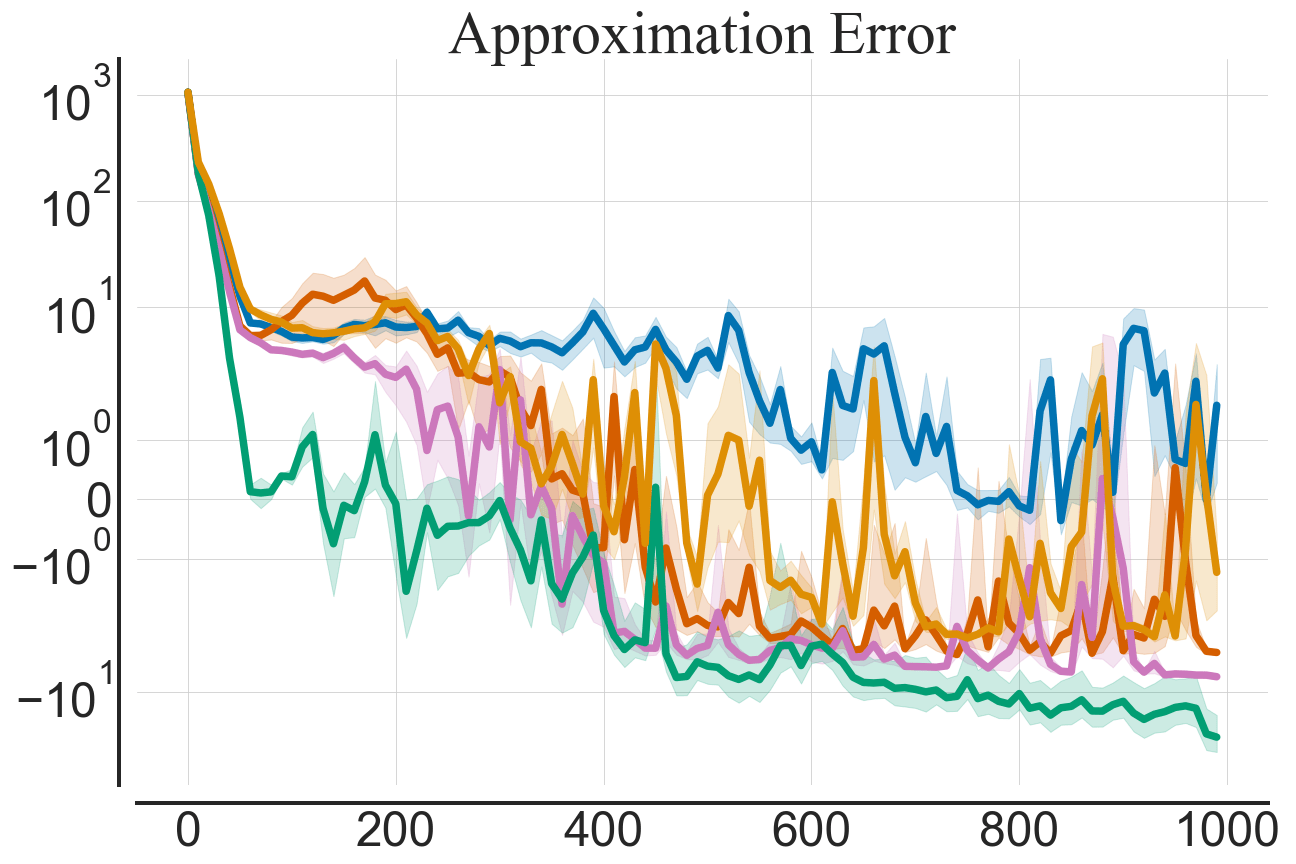}
    \hfill
    \includegraphics[width=0.23\linewidth]{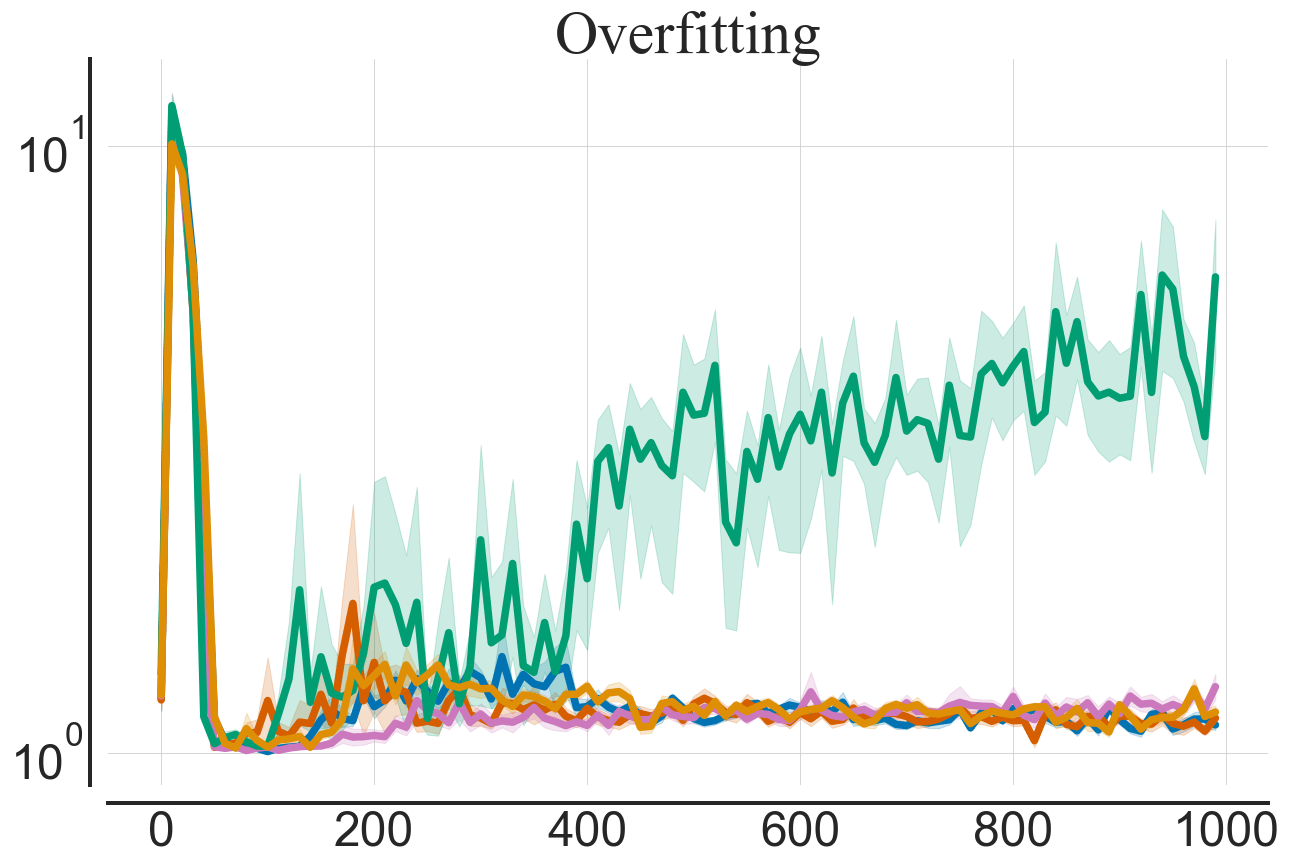}
    \hfill
    \end{subfigure}
    \subcaption{Humanoid Run}
\end{minipage}
\caption{Low replay regime results for each considered task (1/4). 10 seeds per task, mean and 3 standard deviations.}
\label{fig:learning_curves1}
\end{center}
\end{figure*}

\begin{figure*}[ht!]
\begin{center}
\begin{minipage}[h]{1.0\linewidth}
\centering
    \begin{subfigure}{0.88\linewidth}
    \includegraphics[width=\textwidth]{images/legend_1.png}
    \end{subfigure}
\end{minipage}
\bigskip
\begin{minipage}[h]{1.0\linewidth}
    \begin{subfigure}{1.0\linewidth}
    \hfill
    \includegraphics[width=0.23\linewidth]{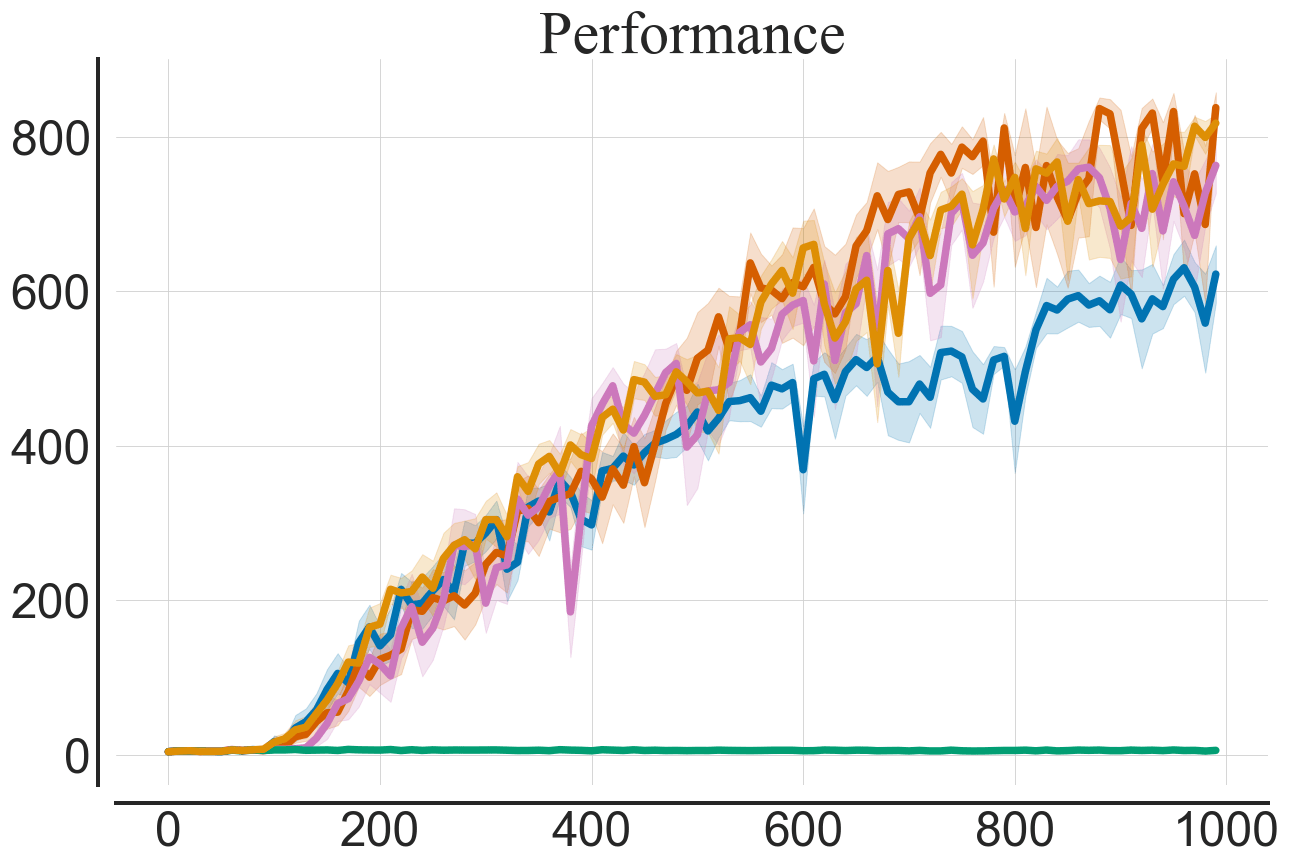}
    \hfill
    \includegraphics[width=0.23\linewidth]{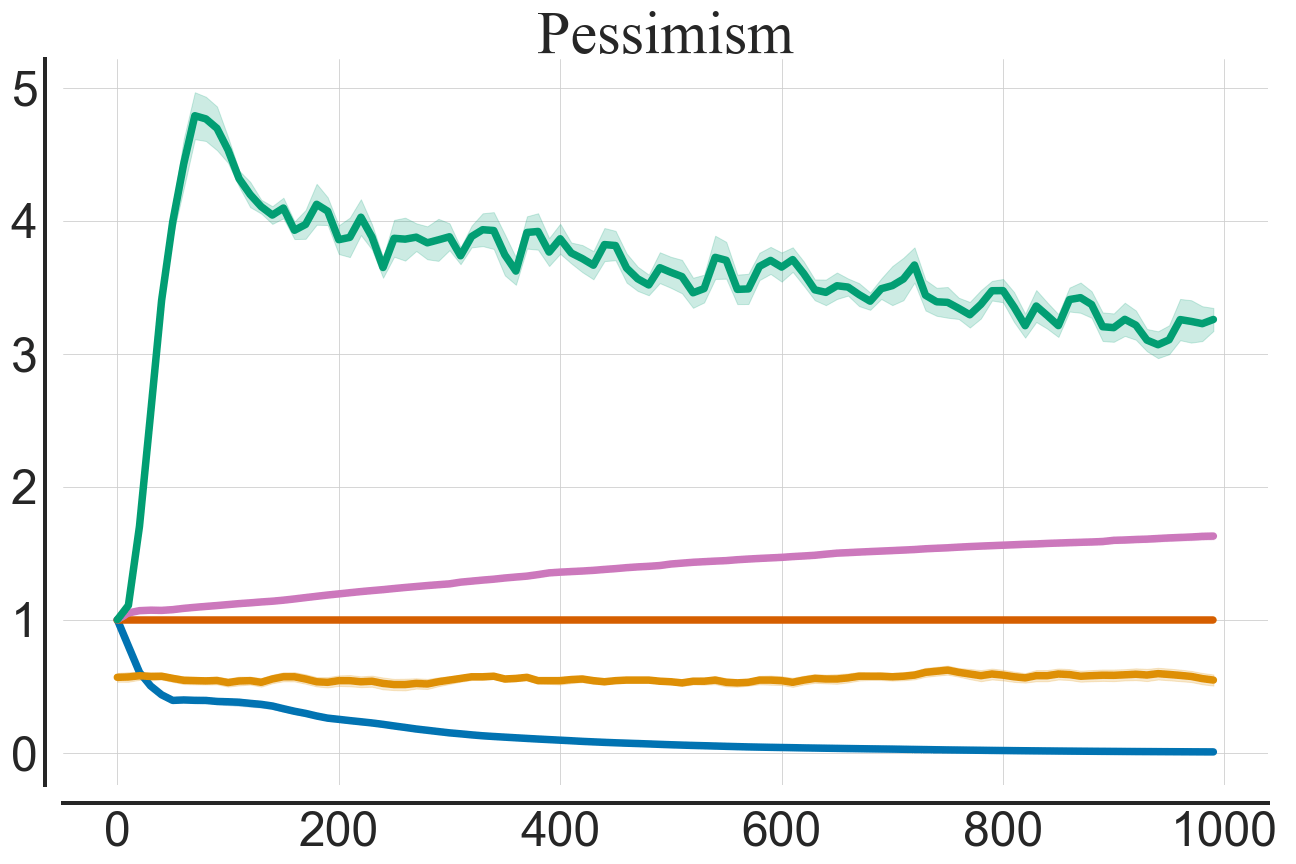}
    \hfill
    \includegraphics[width=0.23\linewidth]{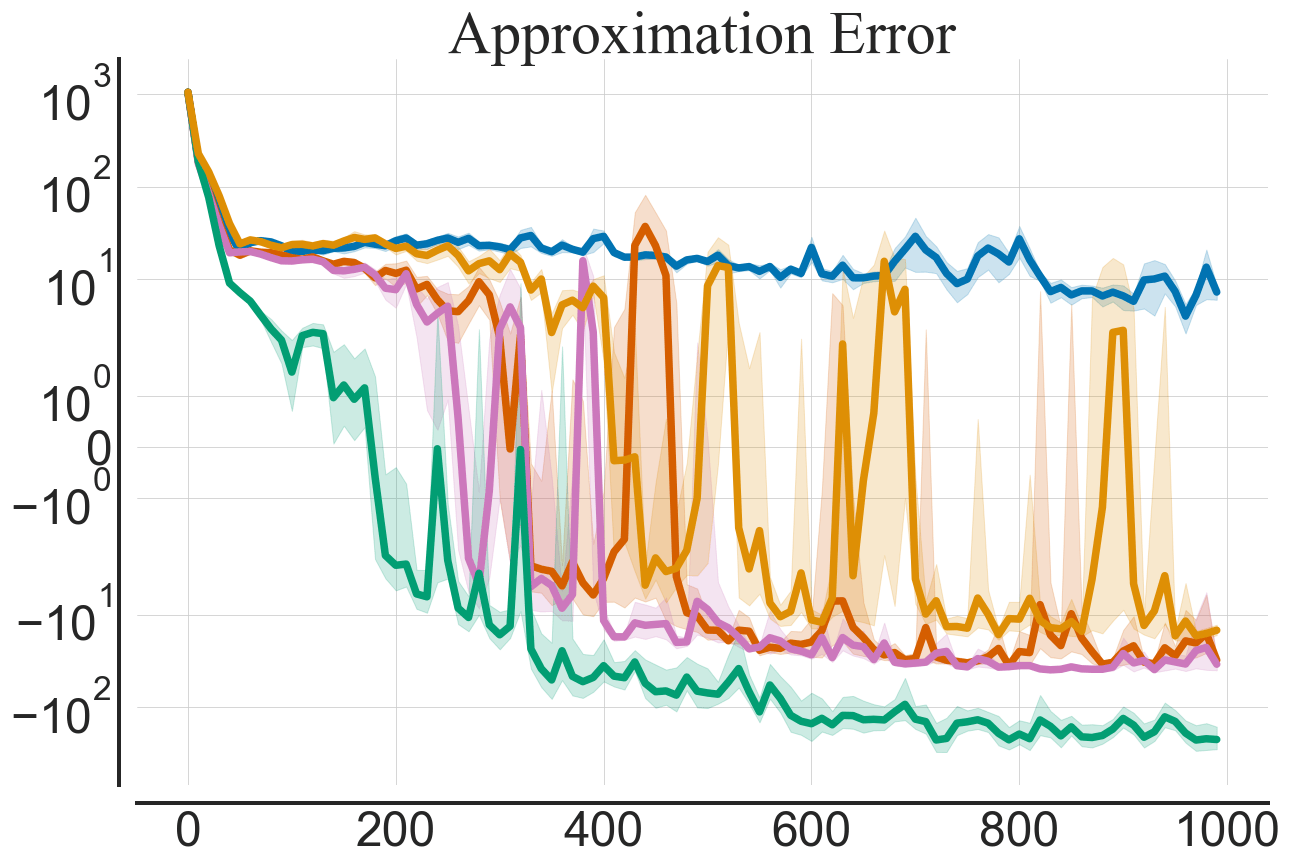}
    \hfill
    \includegraphics[width=0.23\linewidth]{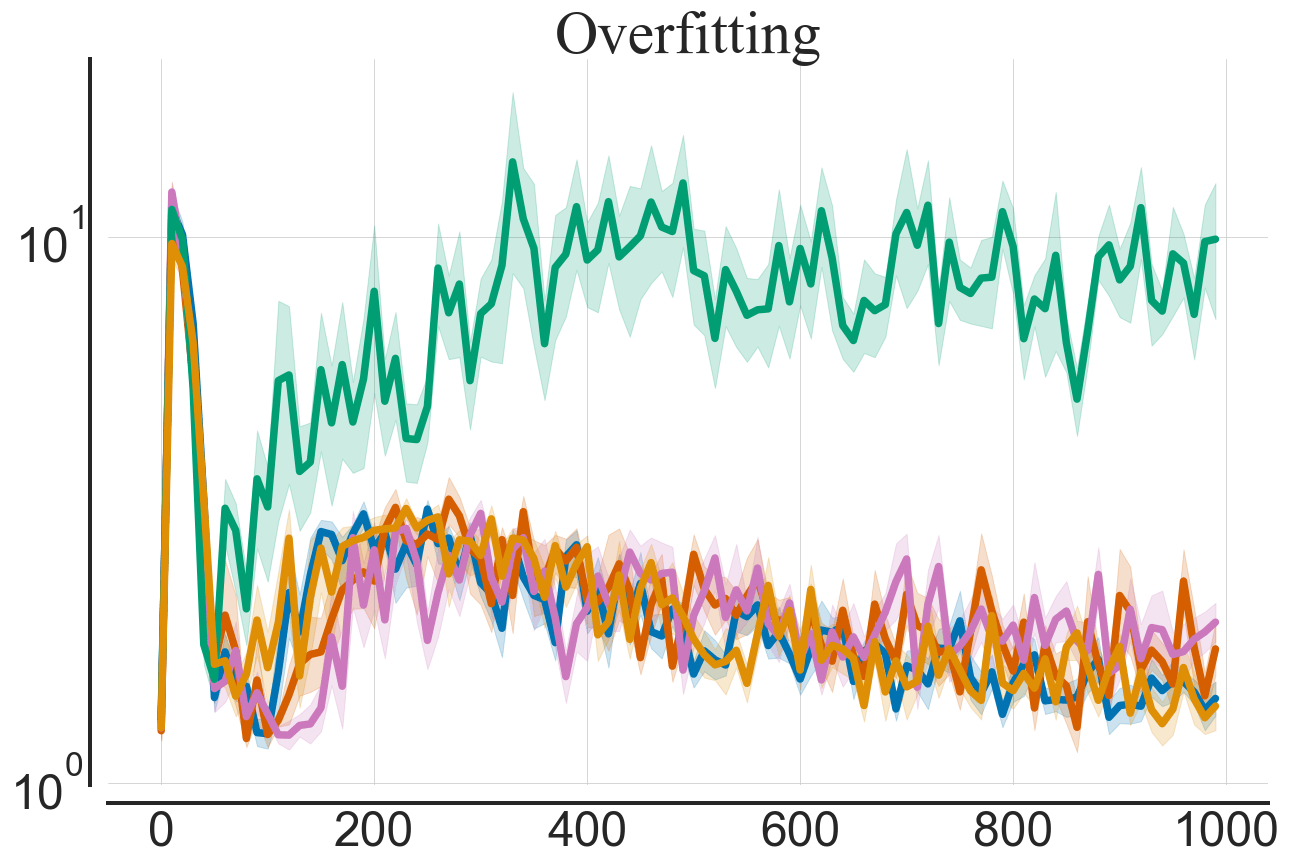}
    \hfill
    \end{subfigure}
    \subcaption{Humanoid Stand}
\end{minipage}
\bigskip
\begin{minipage}[h]{1.0\linewidth}
    \begin{subfigure}{1.0\linewidth}
    \hfill
    \includegraphics[width=0.23\linewidth]{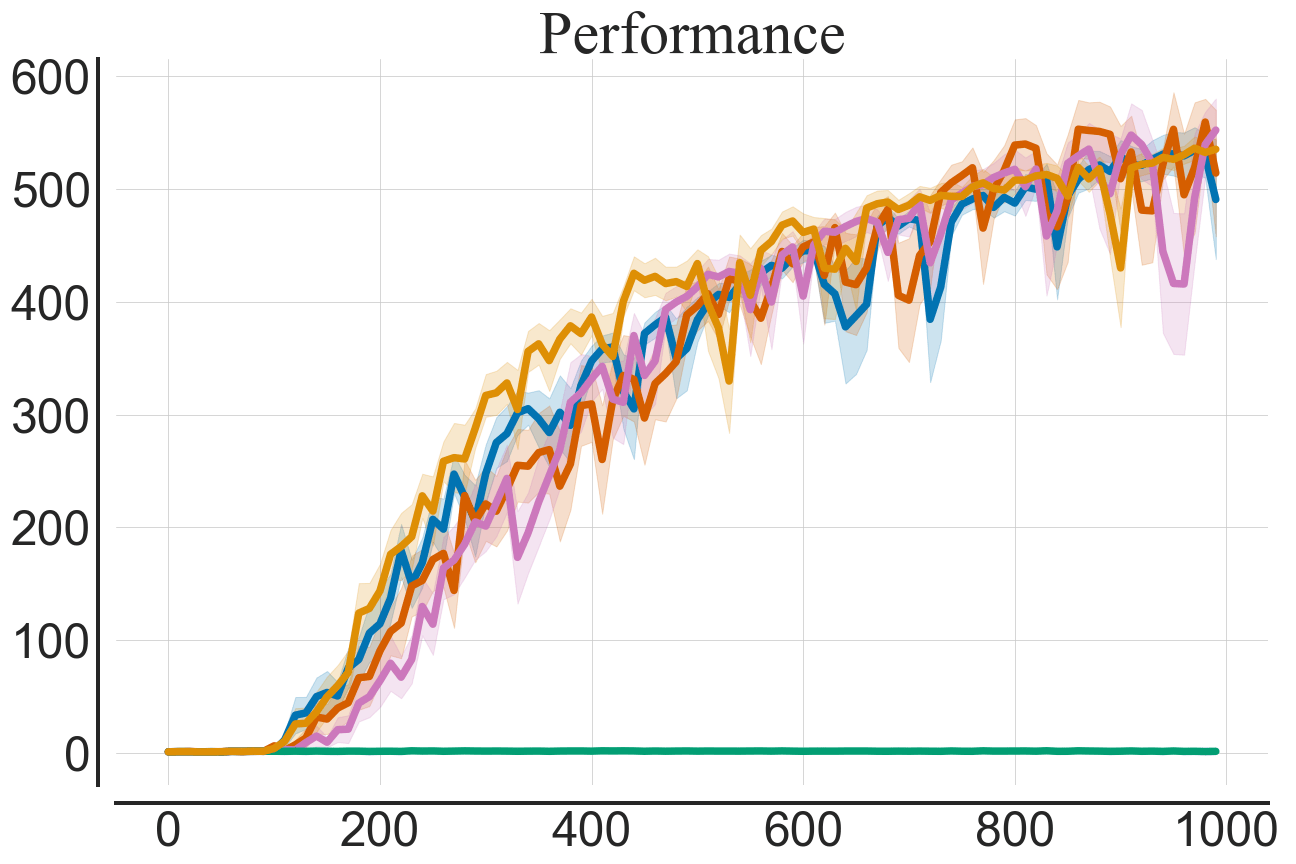}
    \hfill
    \includegraphics[width=0.23\linewidth]{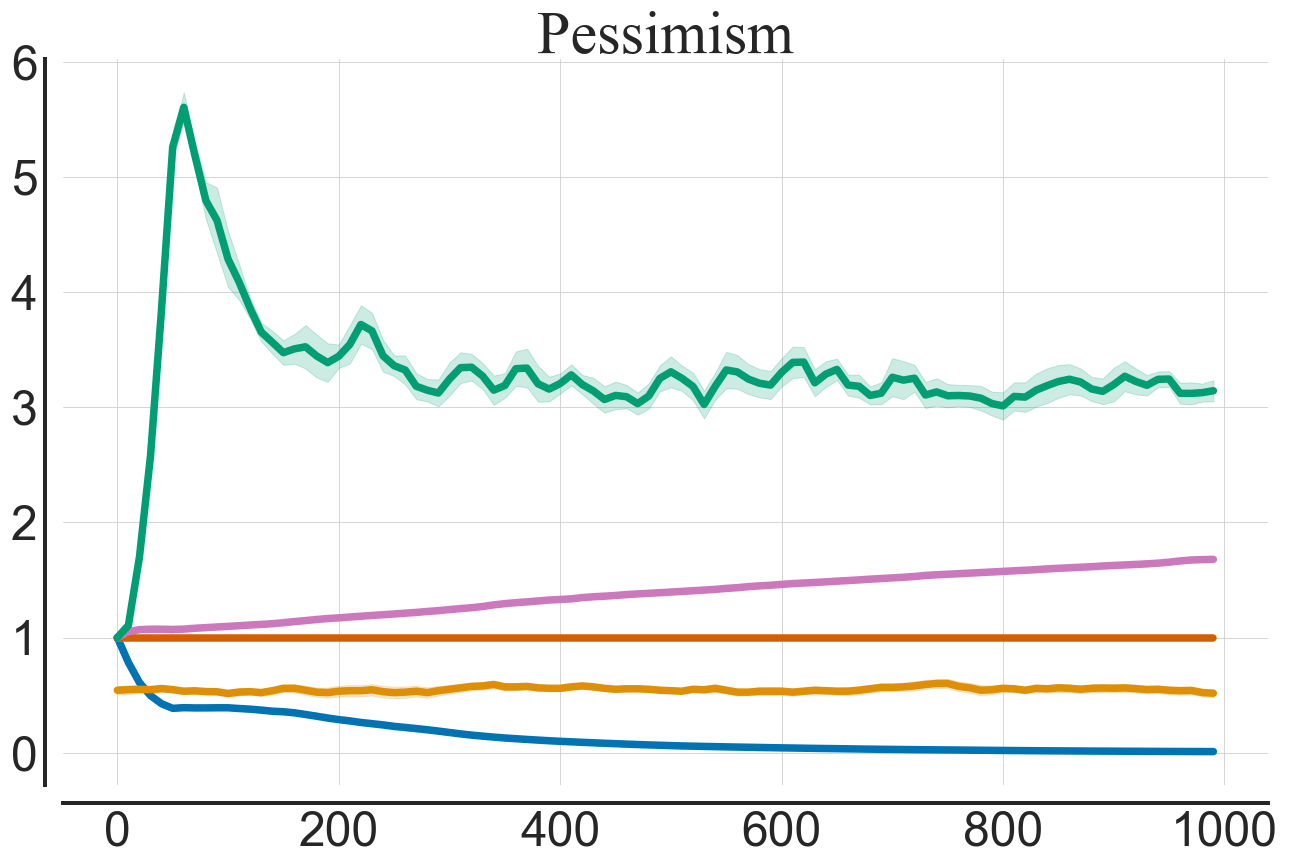}
    \hfill
    \includegraphics[width=0.23\linewidth]{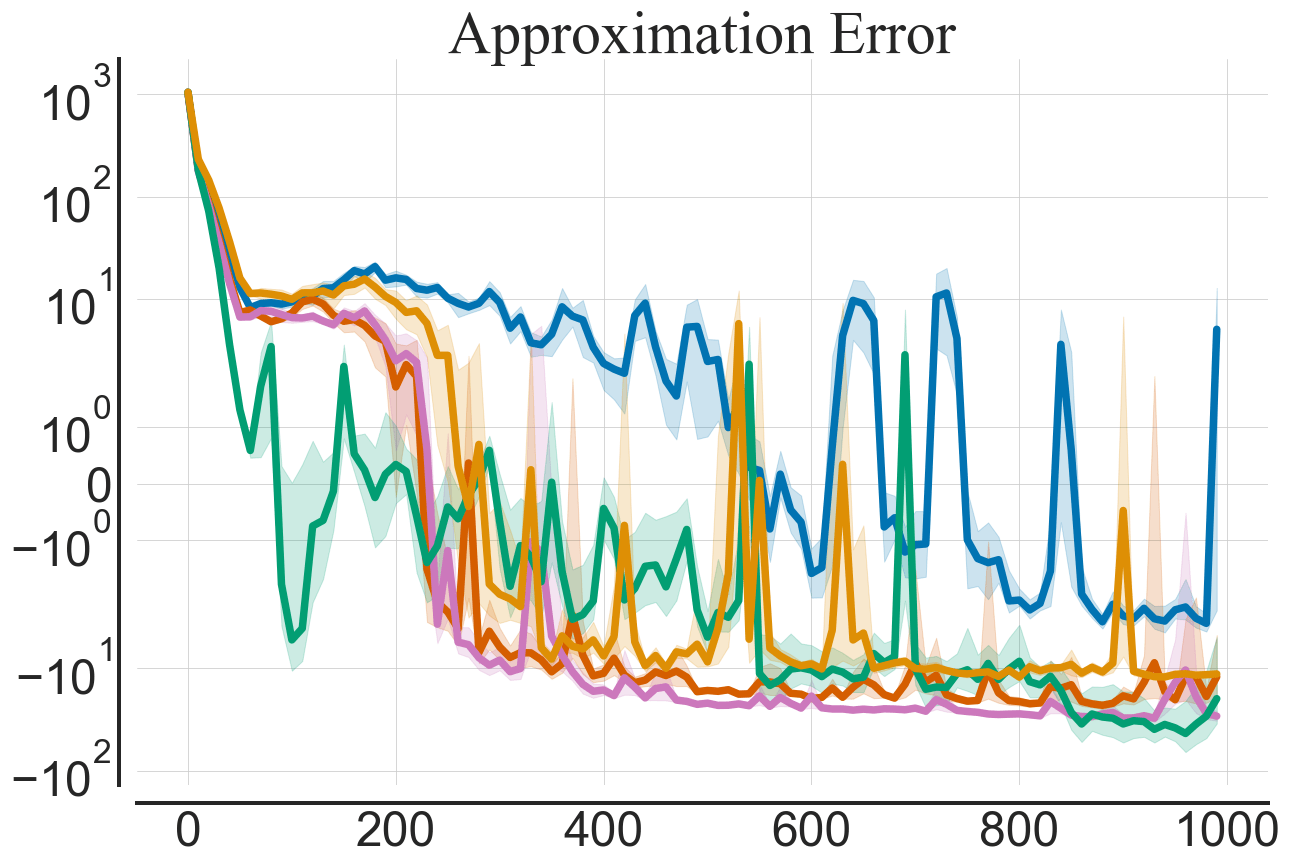}
    \hfill
    \includegraphics[width=0.23\linewidth]{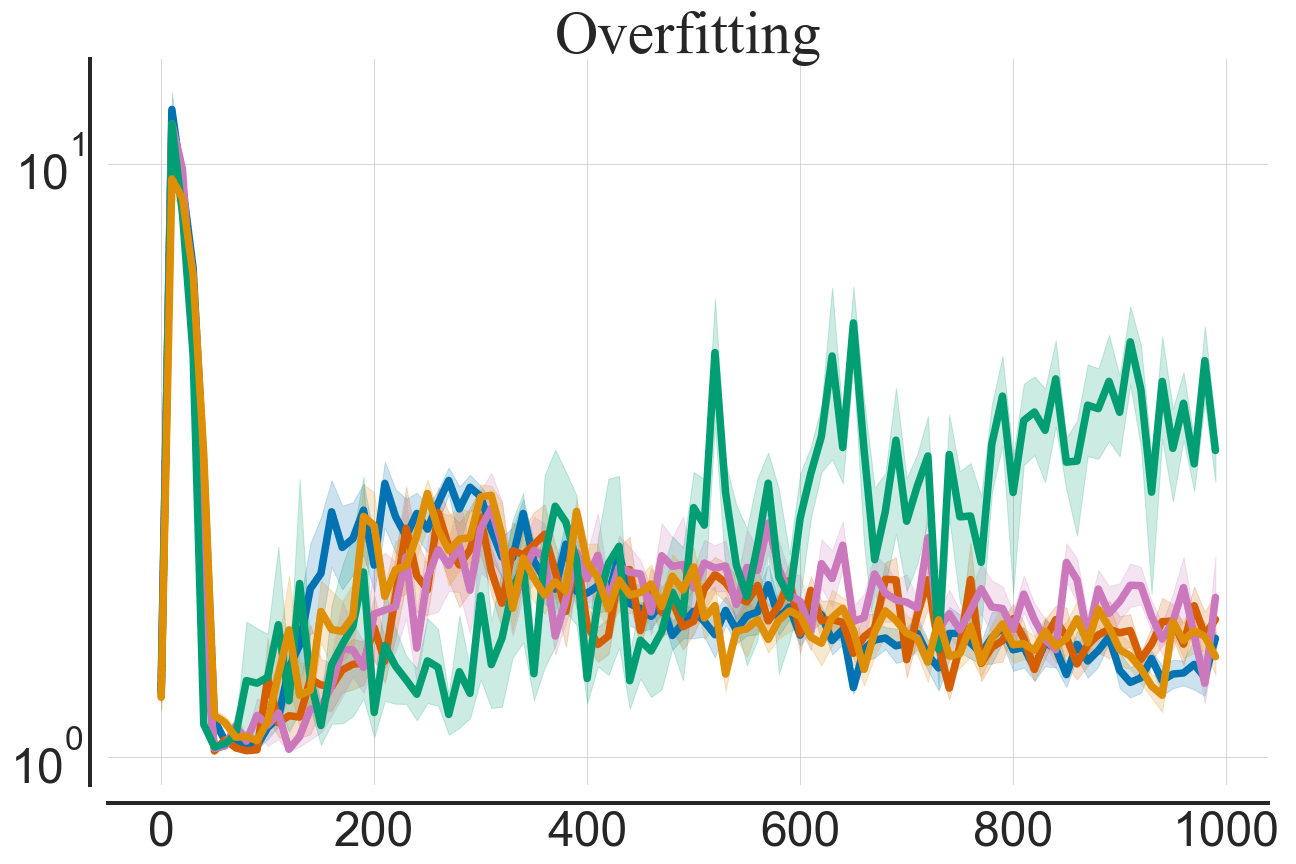}
    \hfill
    \end{subfigure}
    \subcaption{Humanoid Walk}
\end{minipage}
\bigskip
\begin{minipage}[h]{1.0\linewidth}
    \begin{subfigure}{1.0\linewidth}
    \hfill
    \includegraphics[width=0.23\linewidth]{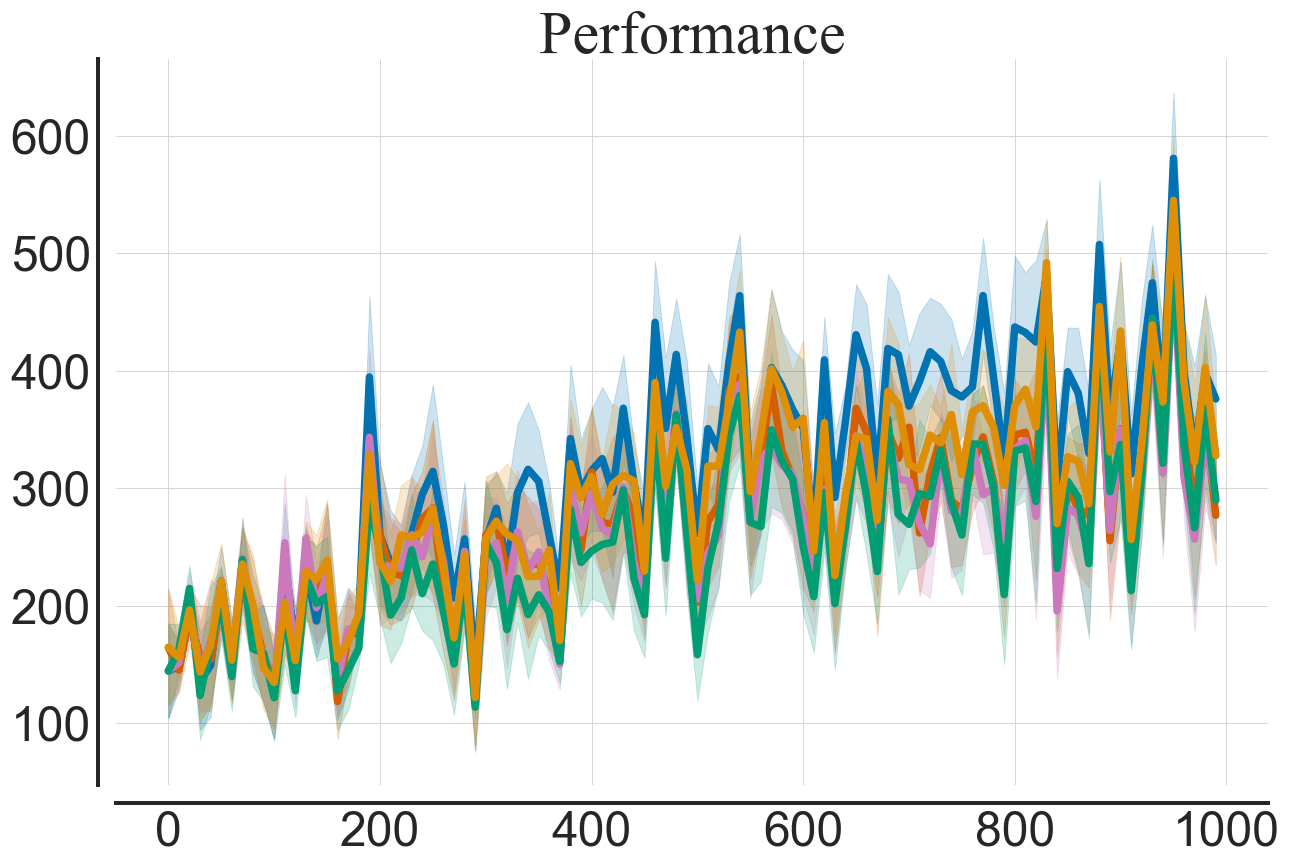}
    \hfill
    \includegraphics[width=0.23\linewidth]{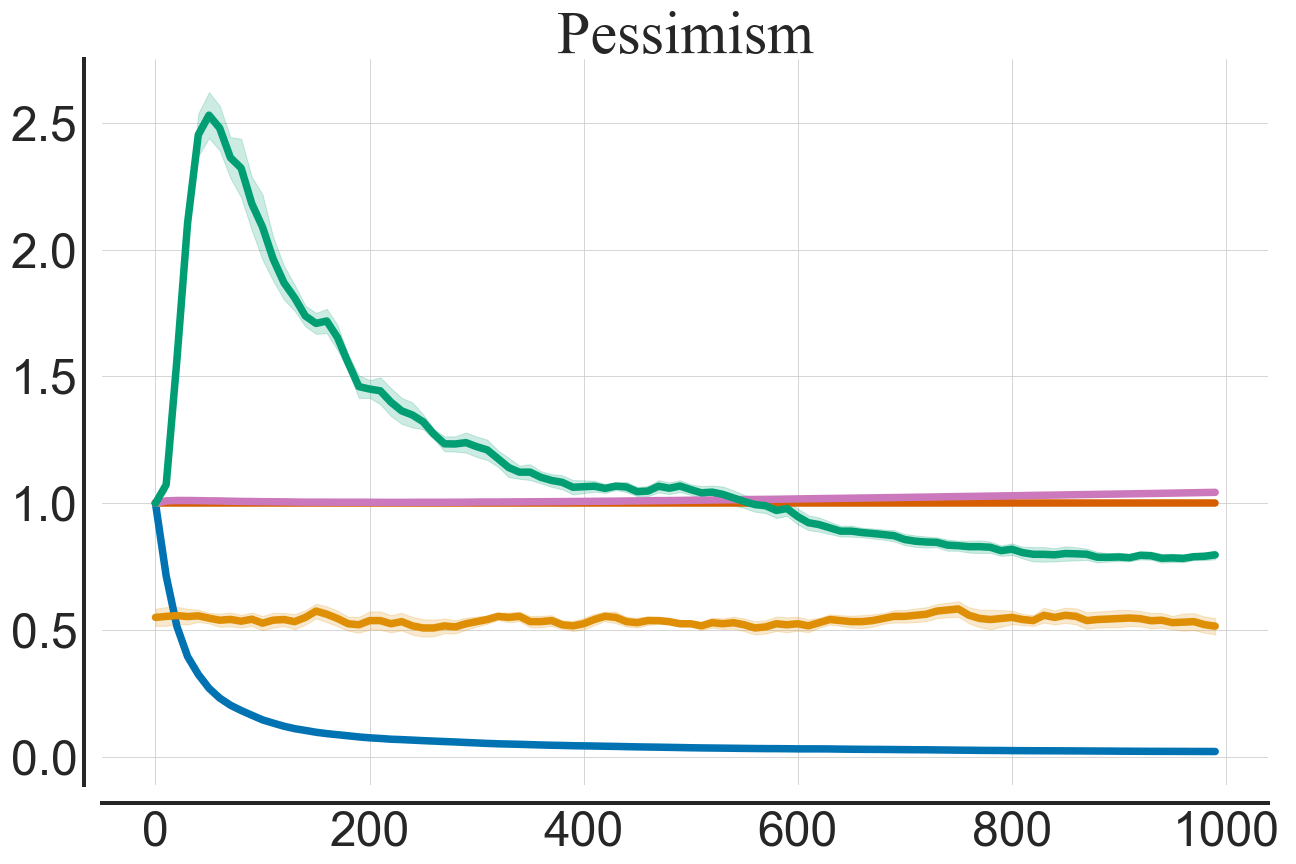}
    \hfill
    \includegraphics[width=0.23\linewidth]{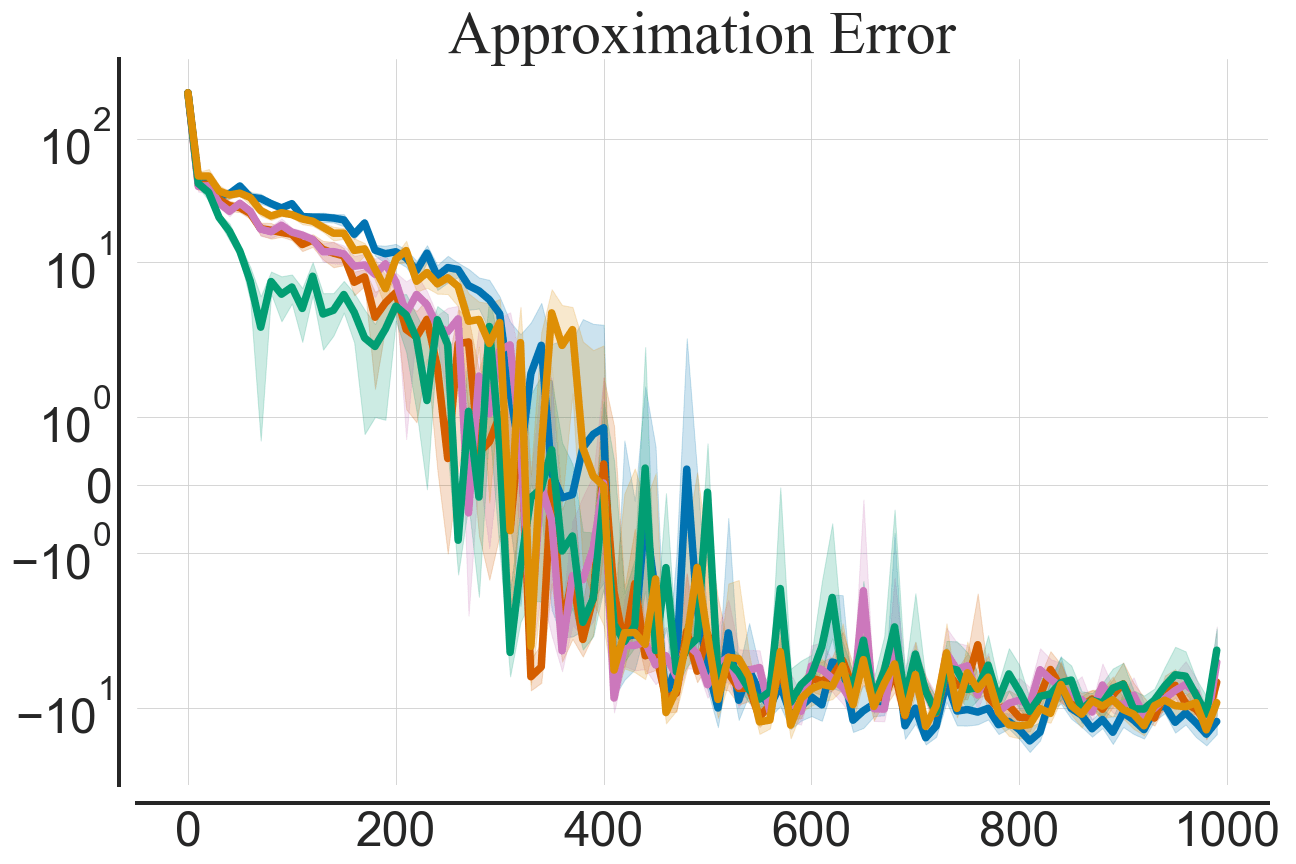}
    \hfill
    \includegraphics[width=0.23\linewidth]{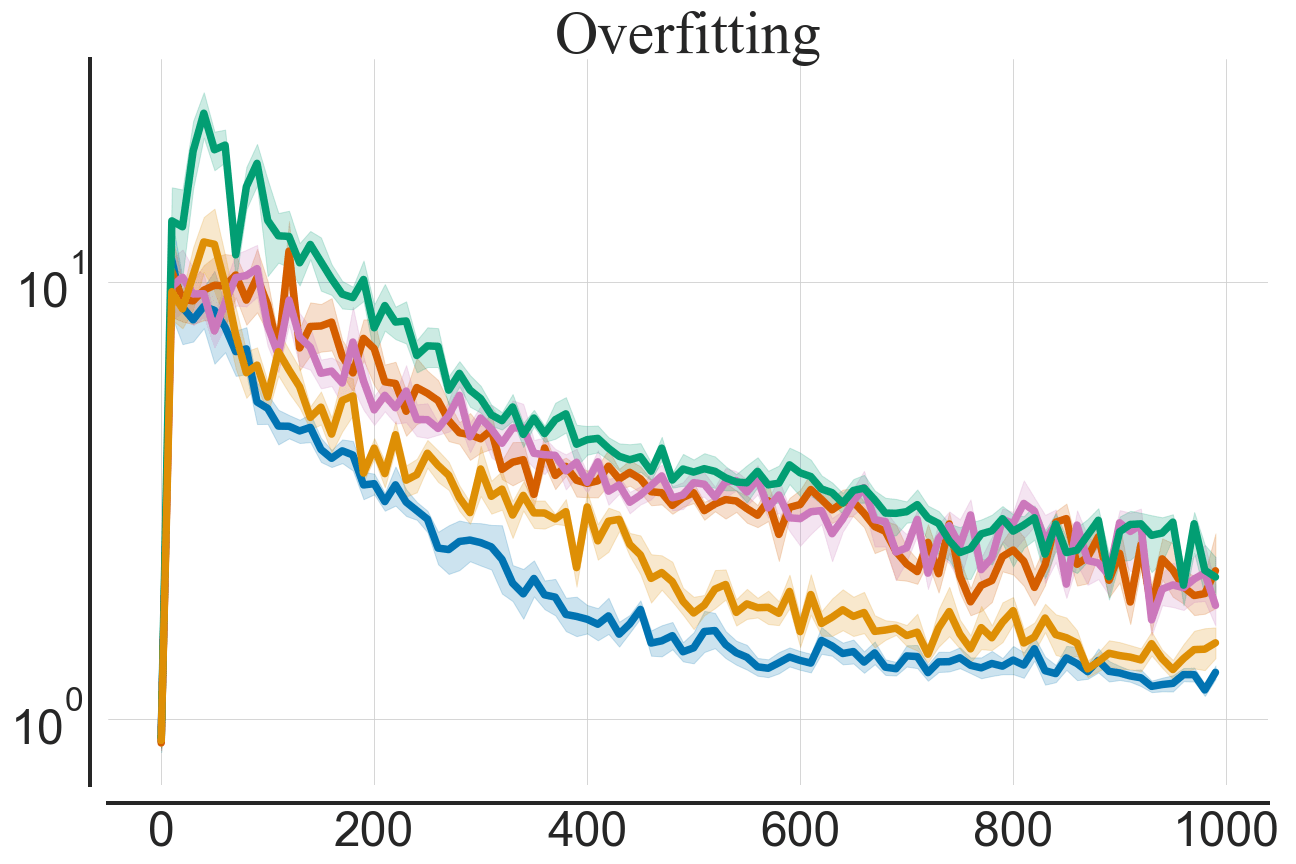}
    \hfill
    \end{subfigure}
    \subcaption{Swimmer Swimmer6}
\end{minipage}
\bigskip
\begin{minipage}[h]{1.0\linewidth}
    \begin{subfigure}{1.0\linewidth}
    \hfill
    \includegraphics[width=0.23\linewidth]{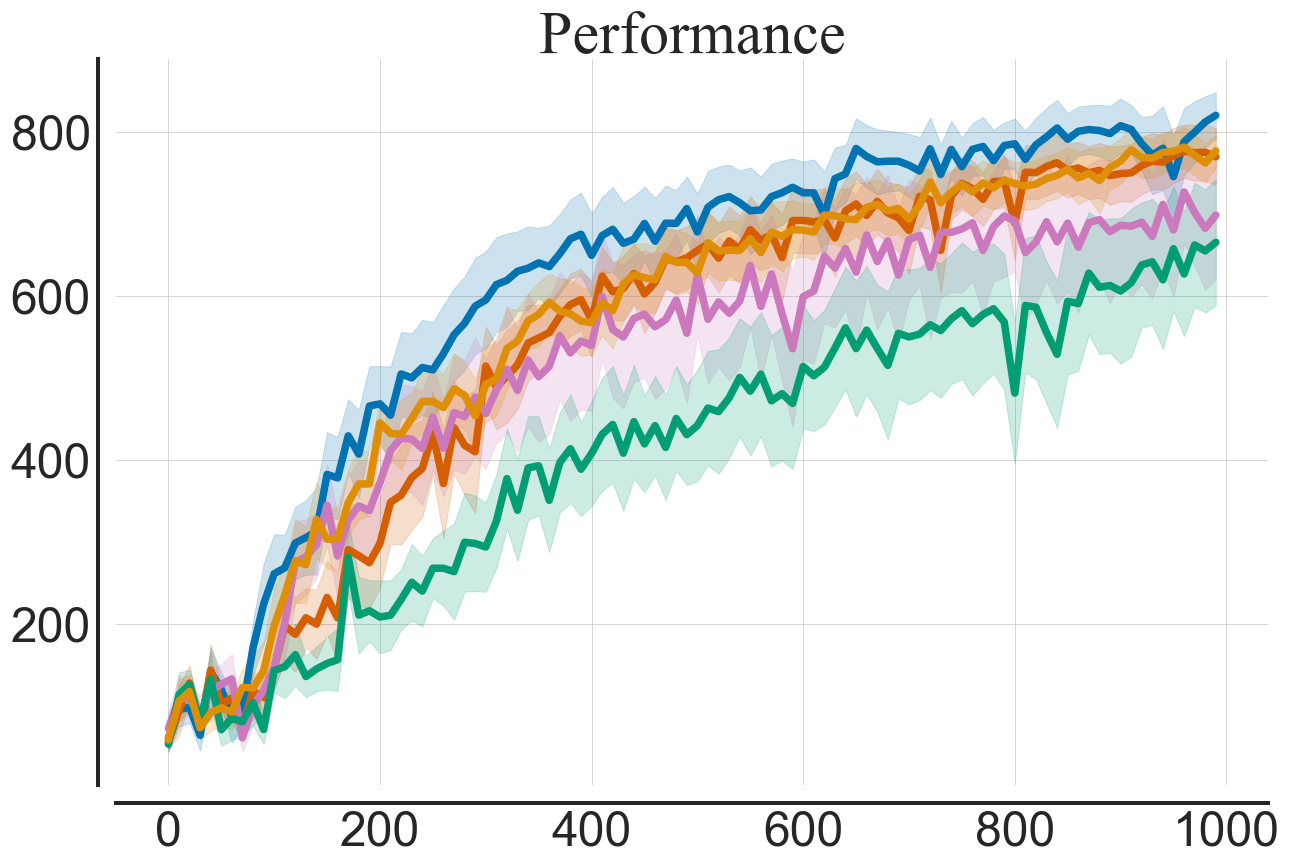}
    \hfill
    \includegraphics[width=0.23\linewidth]{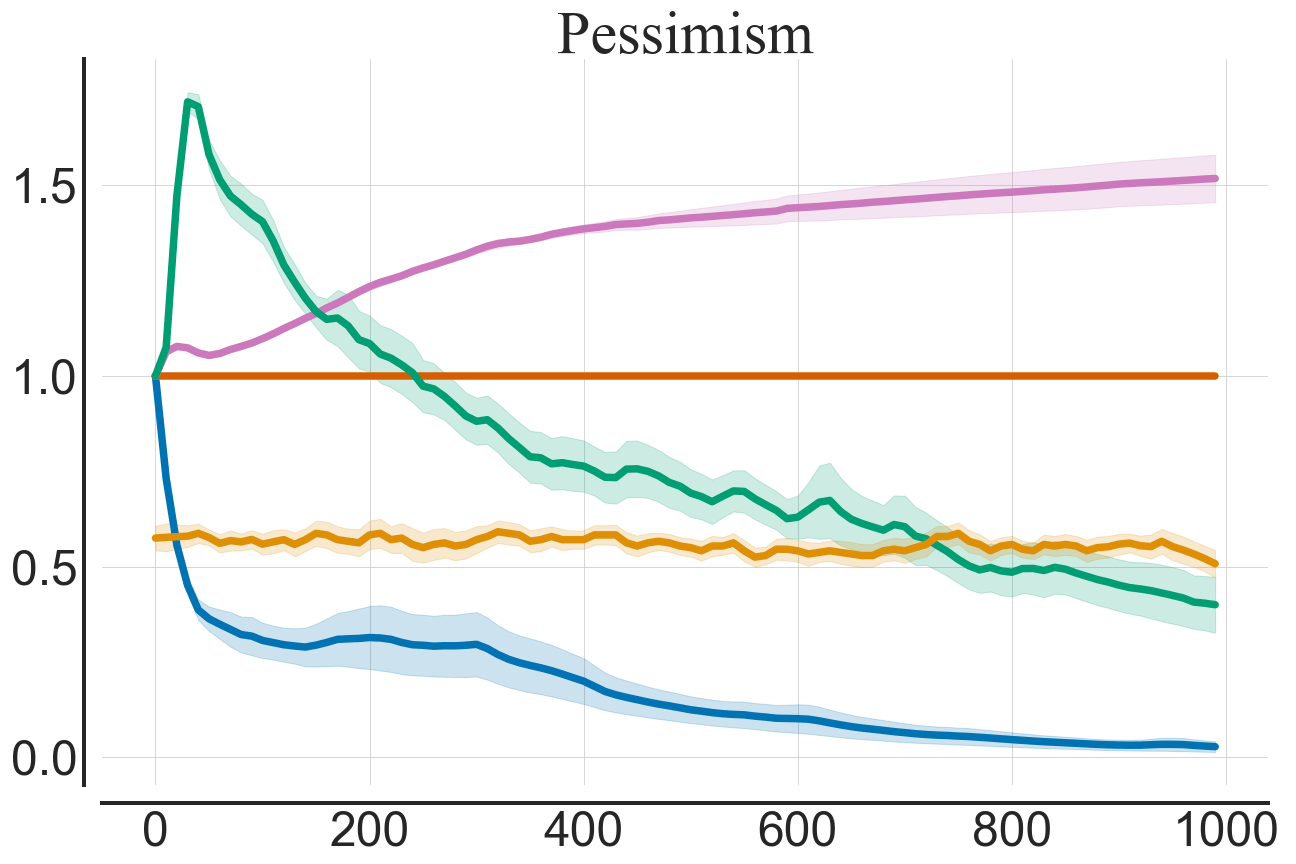}
    \hfill
    \includegraphics[width=0.23\linewidth]{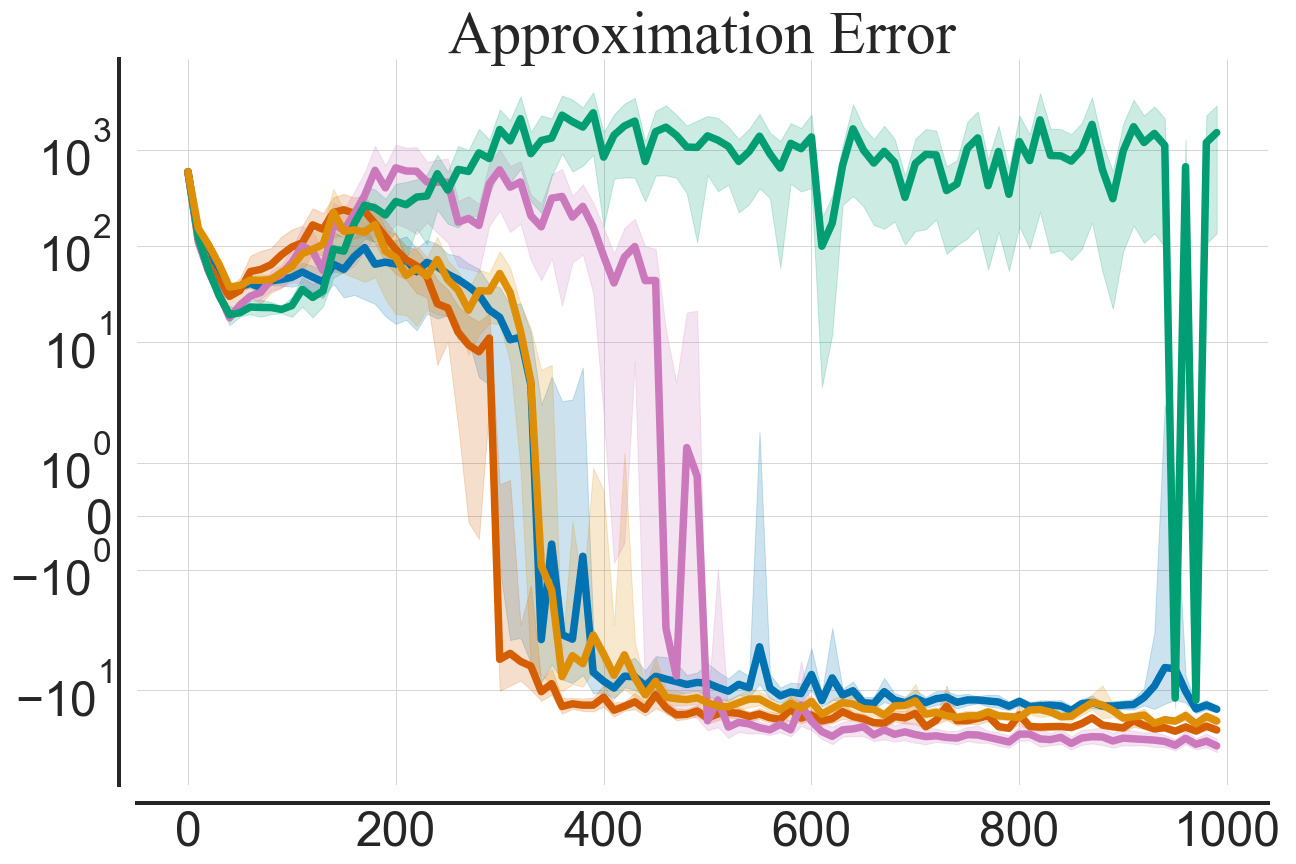}
    \hfill
    \includegraphics[width=0.23\linewidth]{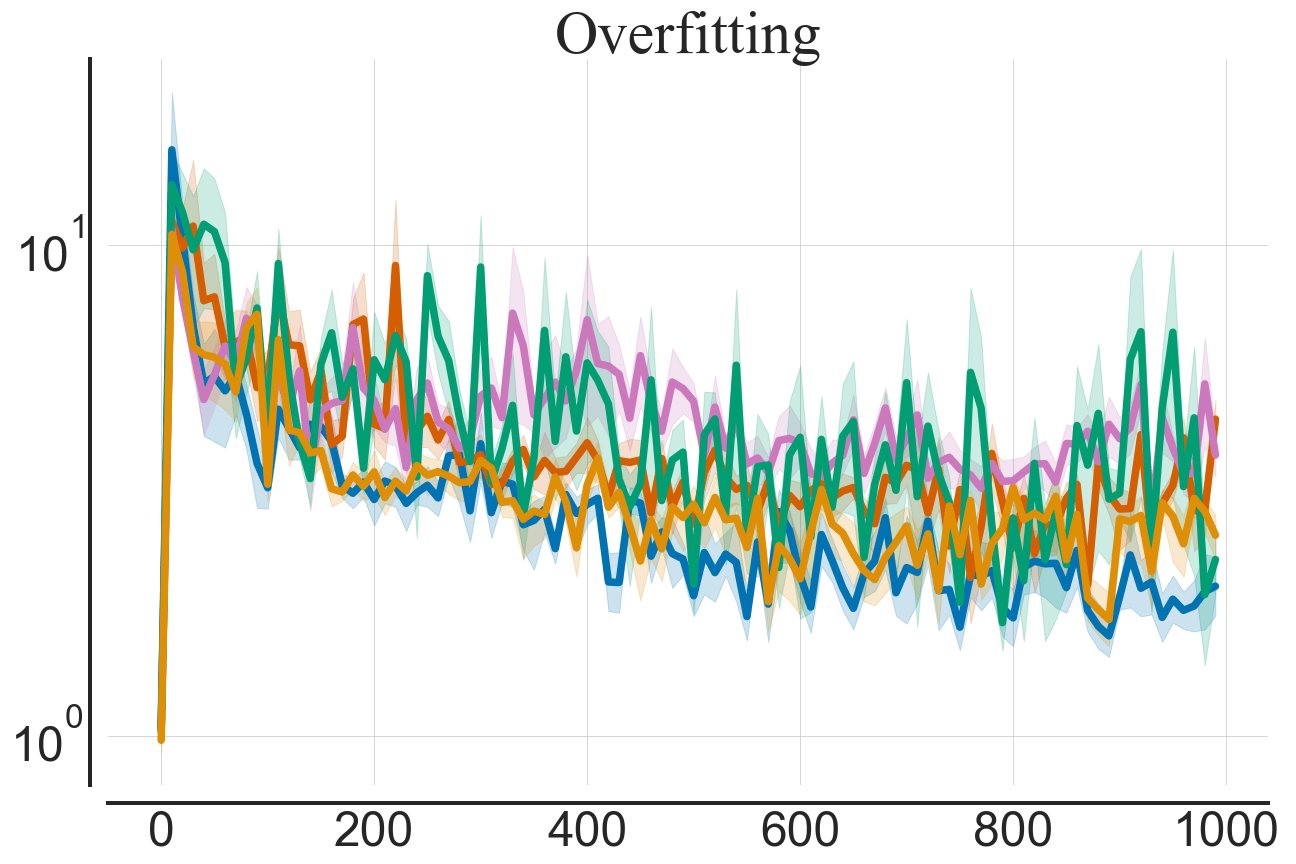}
    \hfill
    \end{subfigure}
    \subcaption{Quadruped Run}
\end{minipage}
\bigskip
\begin{minipage}[h]{1.0\linewidth}
    \begin{subfigure}{1.0\linewidth}
    \hfill
    \includegraphics[width=0.23\linewidth]{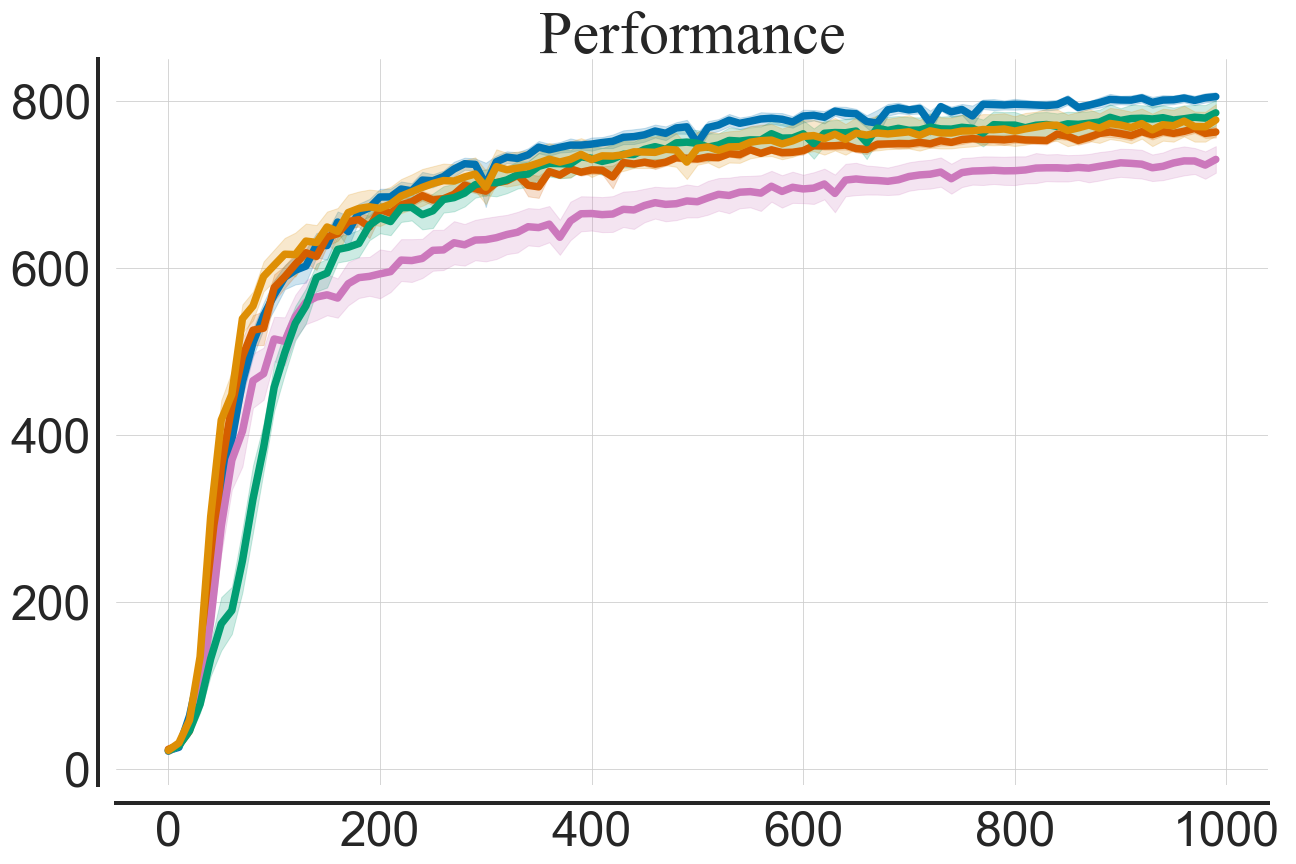}
    \hfill
    \includegraphics[width=0.23\linewidth]{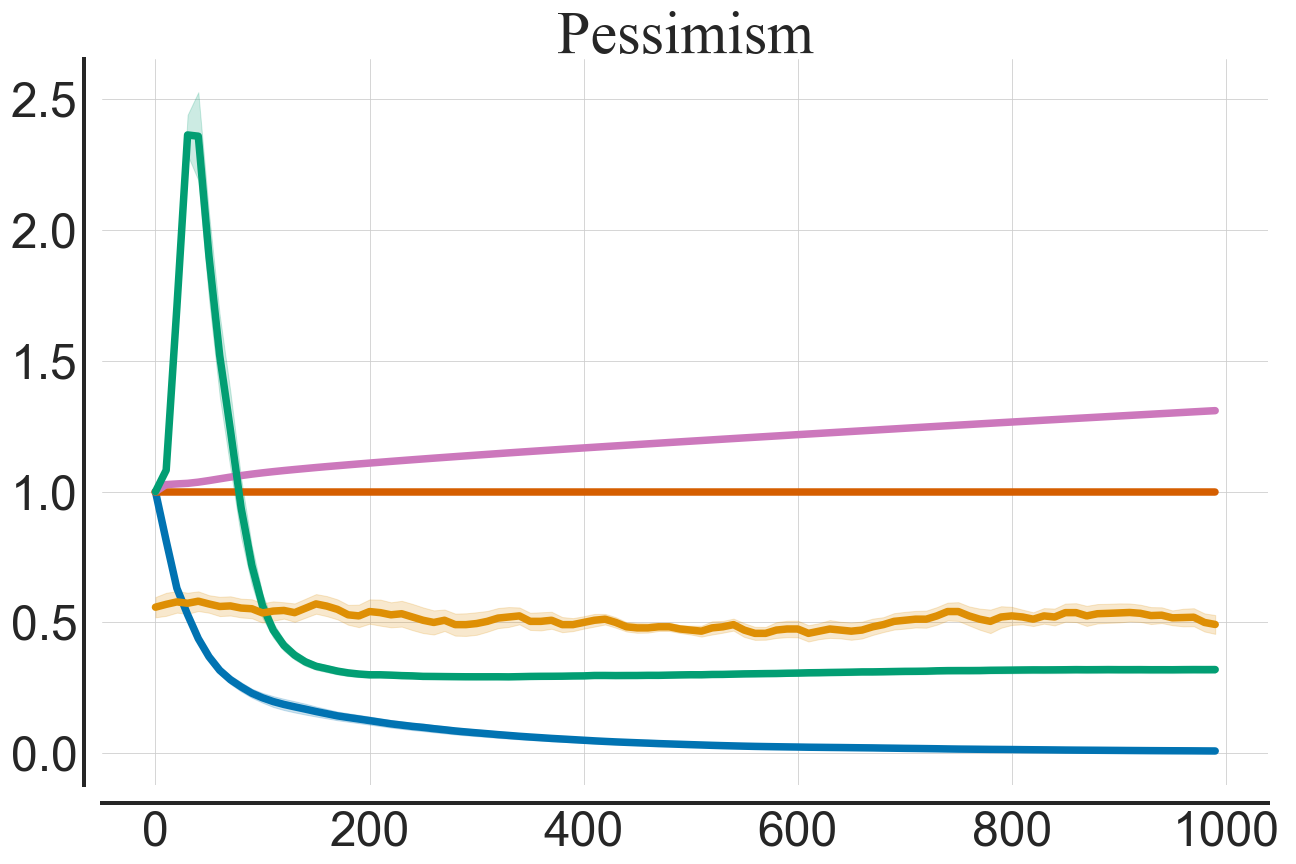}
    \hfill
    \includegraphics[width=0.23\linewidth]{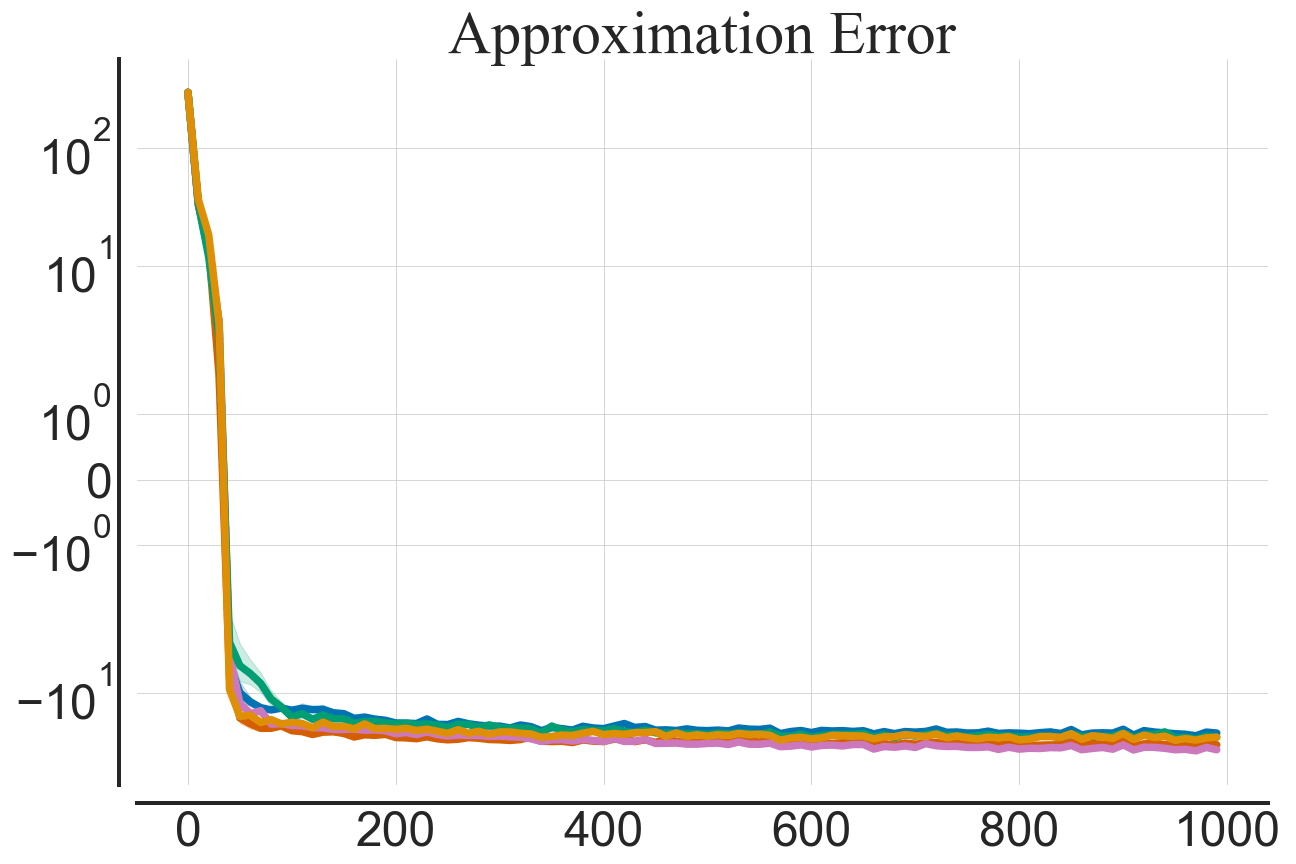}
    \hfill
    \includegraphics[width=0.23\linewidth]{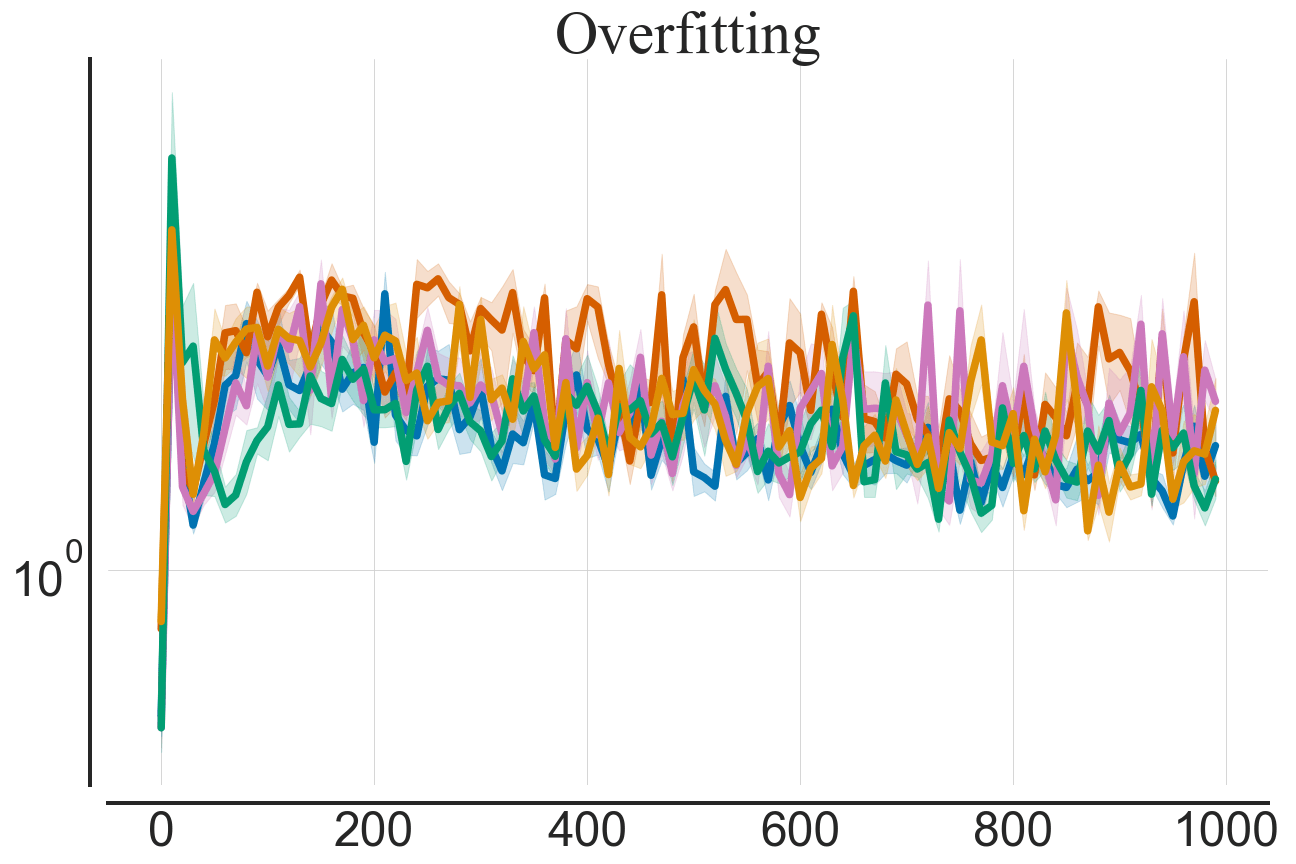}
    \hfill
    \end{subfigure}
    \subcaption{Walker Run}
\end{minipage}
\caption{Low replay regime results for each considered task (2/4). 10 seeds per task, mean and 3 standard deviations.}
\label{fig:learning_curves2}
\end{center}
\end{figure*}

\begin{figure*}[ht!]
\begin{center}
\begin{minipage}[h]{1.0\linewidth}
\centering
    \begin{subfigure}{0.88\linewidth}
    \includegraphics[width=\textwidth]{images/legend_1.png}
    \end{subfigure}
\end{minipage}
\bigskip
\begin{minipage}[h]{1.0\linewidth}
    \begin{subfigure}{1.0\linewidth}
    \hfill
    \includegraphics[width=0.23\linewidth]{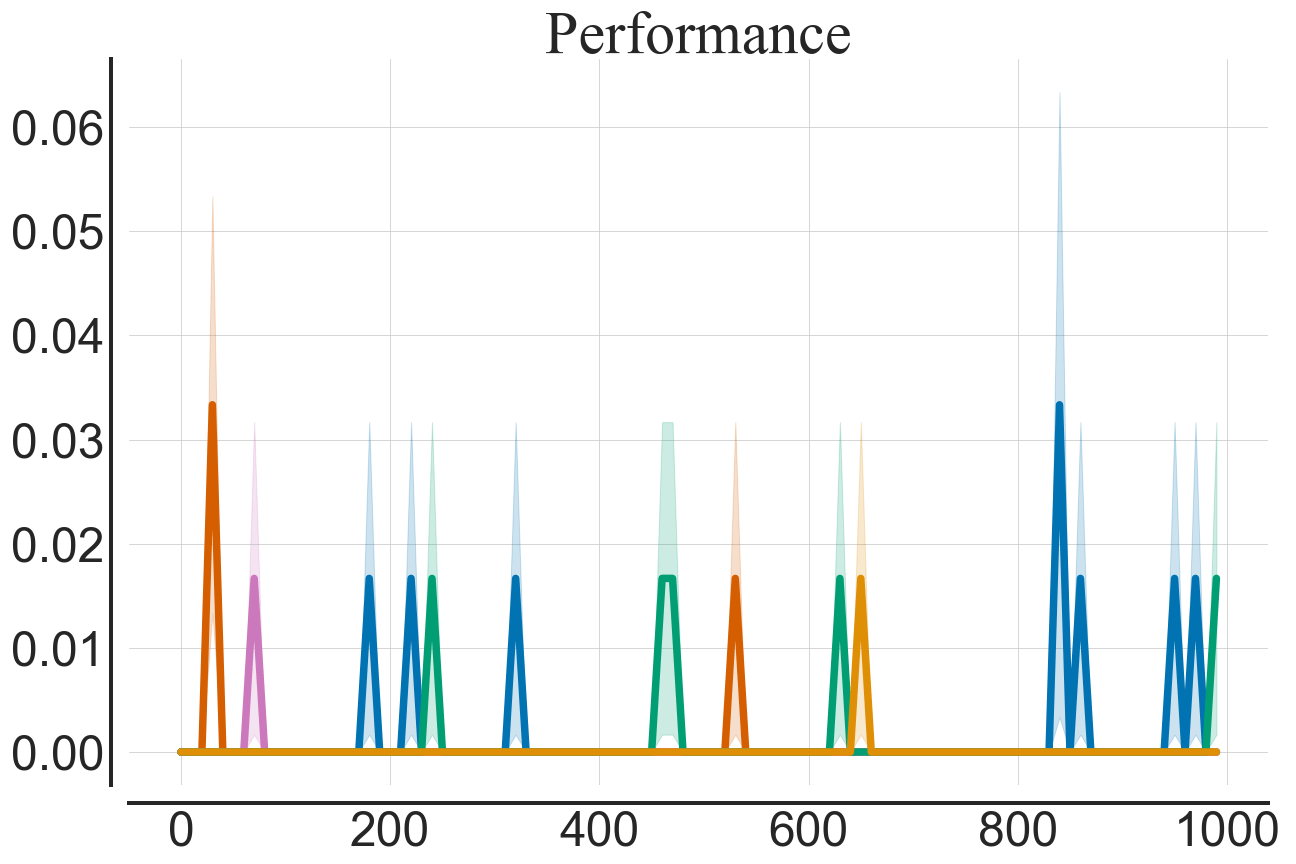}
    \hfill
    \includegraphics[width=0.23\linewidth]{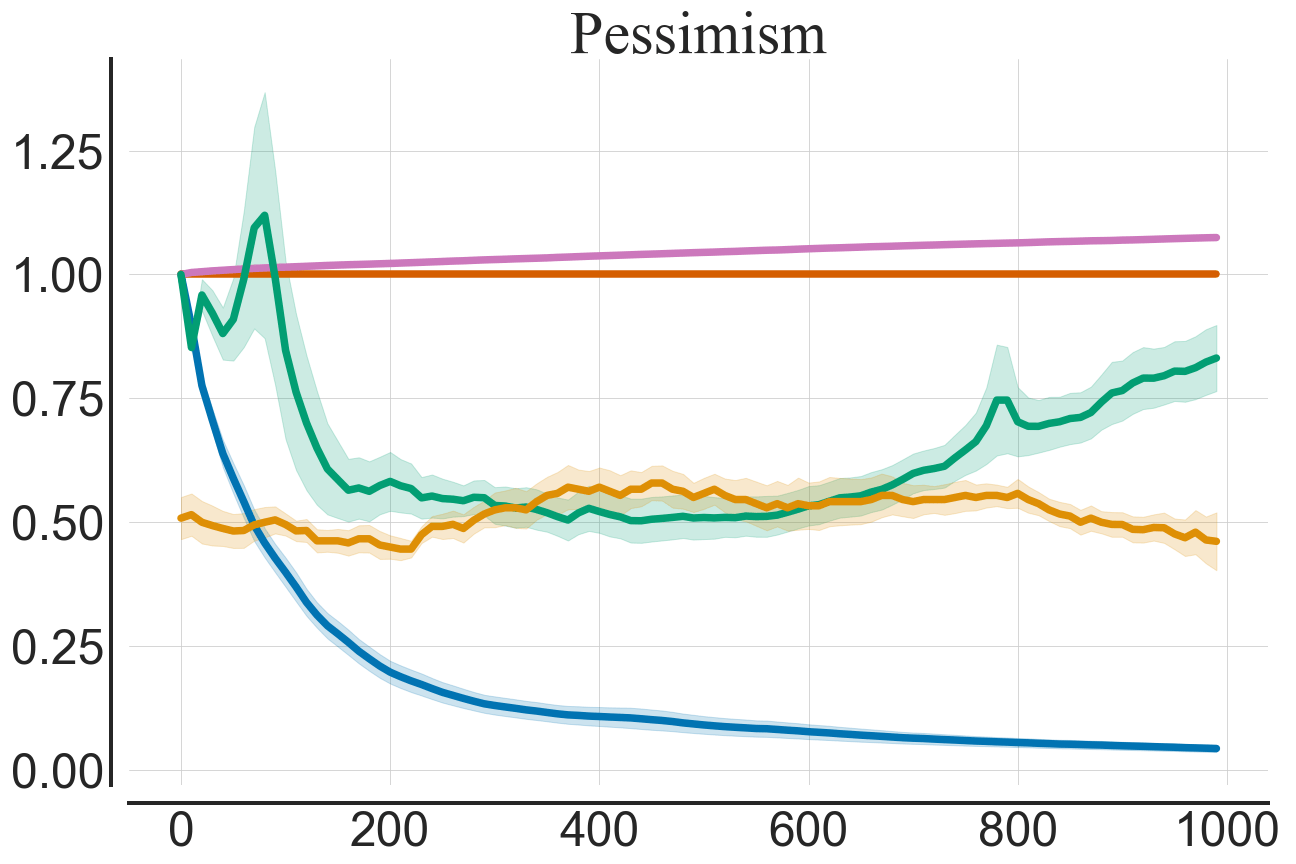}
    \hfill
    \includegraphics[width=0.23\linewidth]{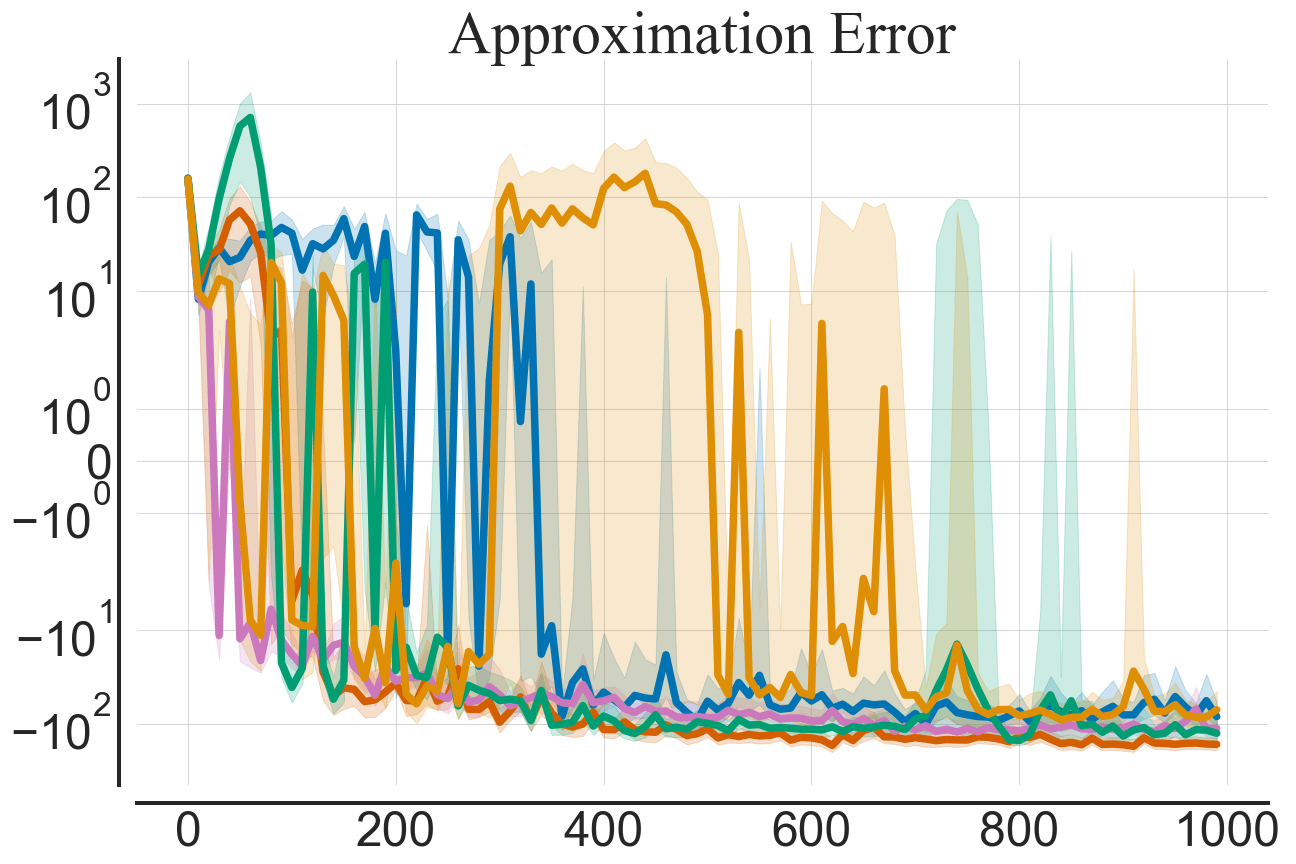}
    \hfill
    \includegraphics[width=0.23\linewidth]{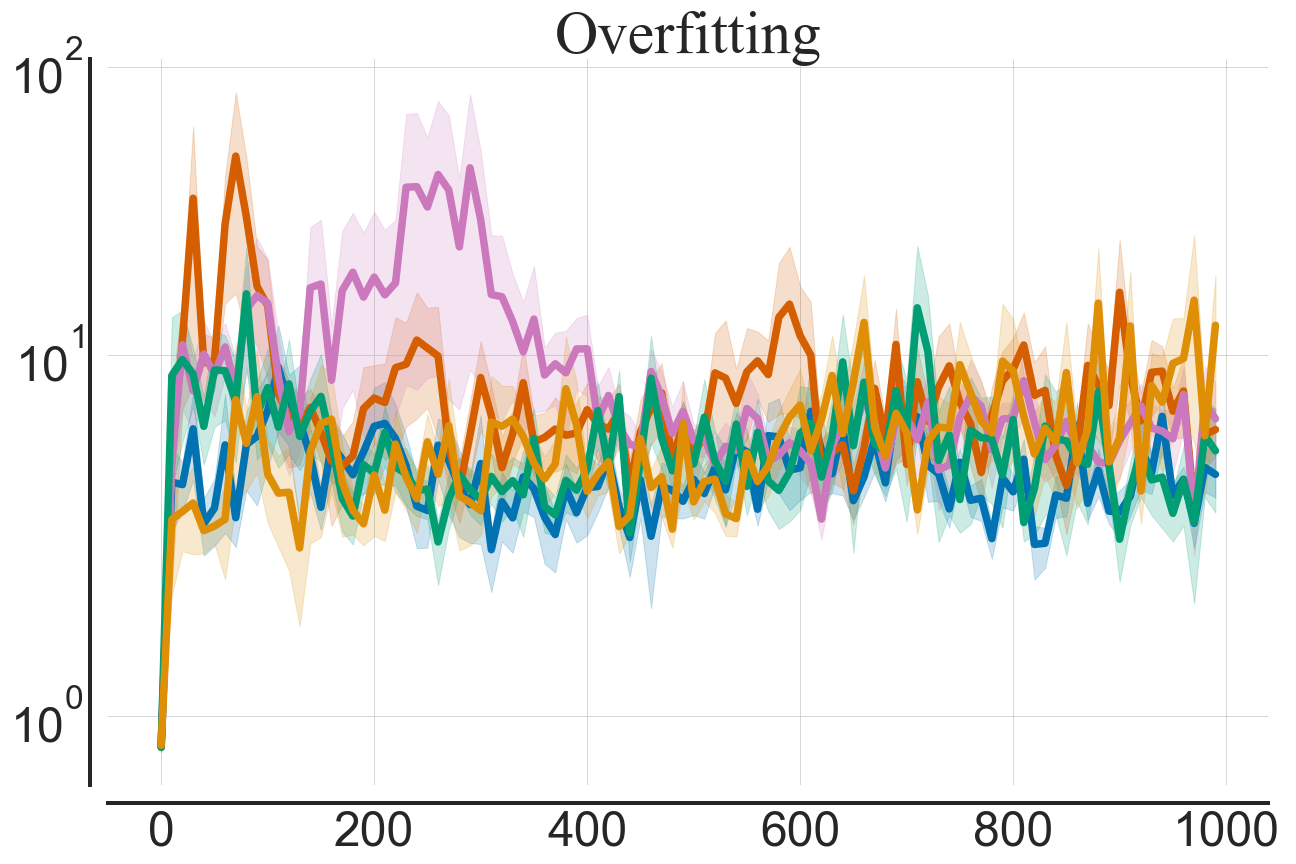}
    \hfill
    \end{subfigure}
    \subcaption{Assembly}
\end{minipage}
\bigskip
\begin{minipage}[h]{1.0\linewidth}
    \begin{subfigure}{1.0\linewidth}
    \hfill
    \includegraphics[width=0.23\linewidth]{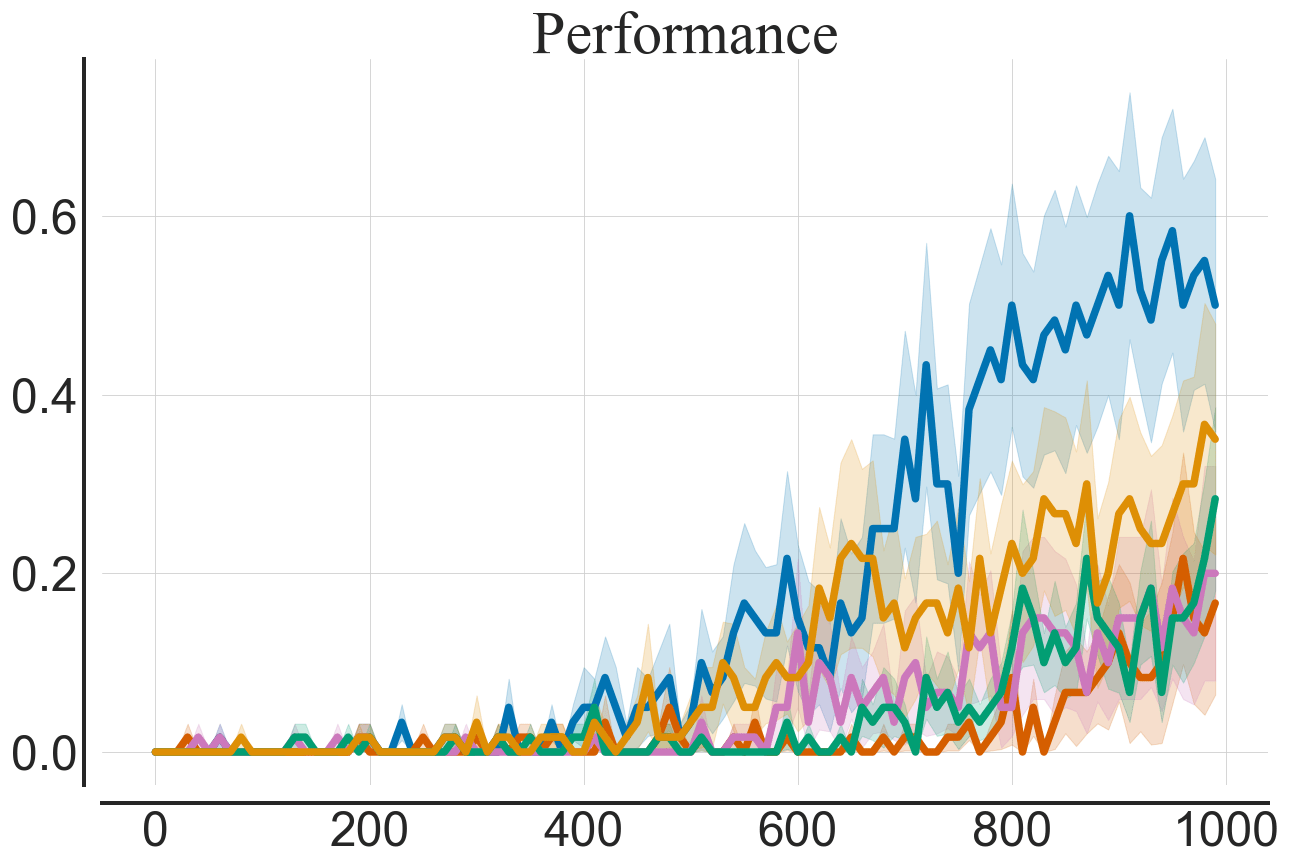}
    \hfill
    \includegraphics[width=0.23\linewidth]{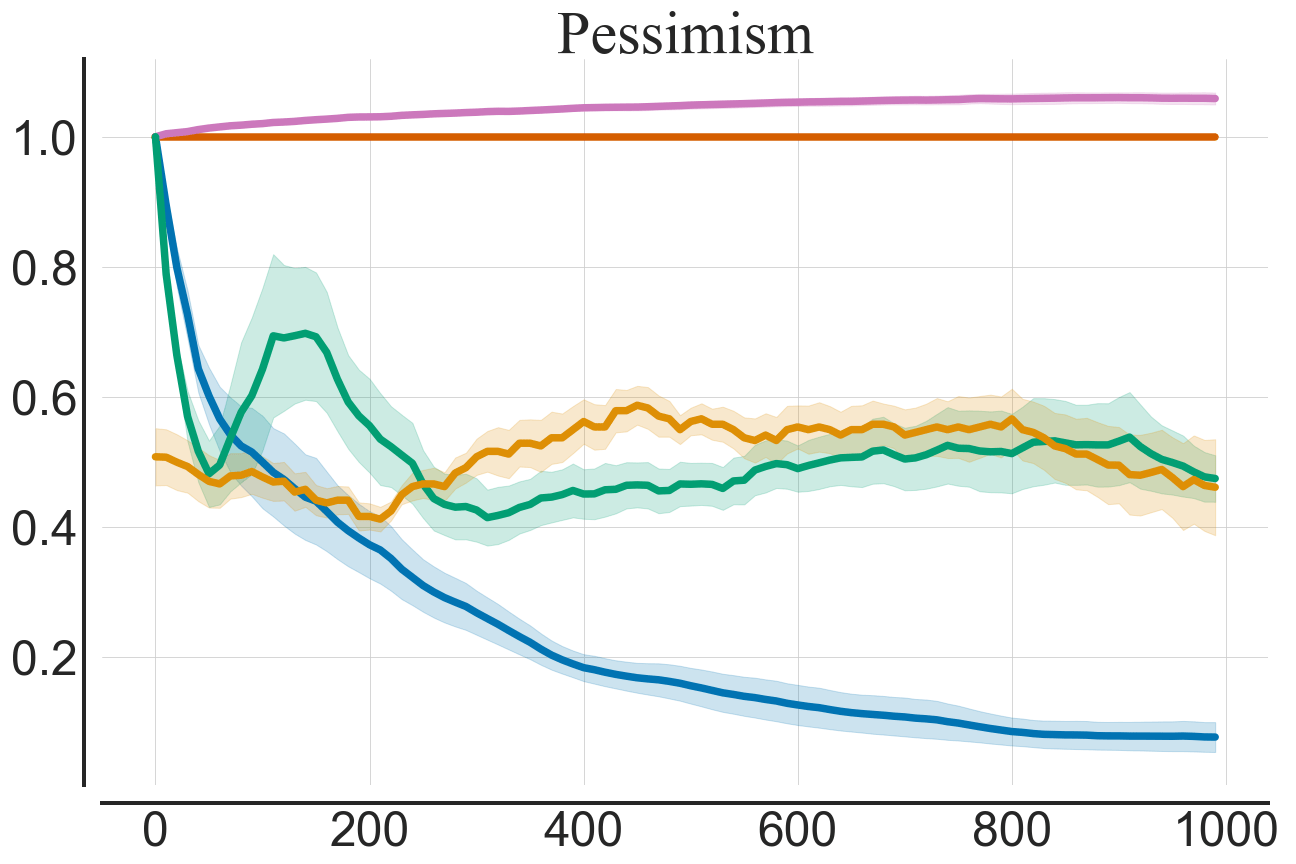}
    \hfill
    \includegraphics[width=0.23\linewidth]{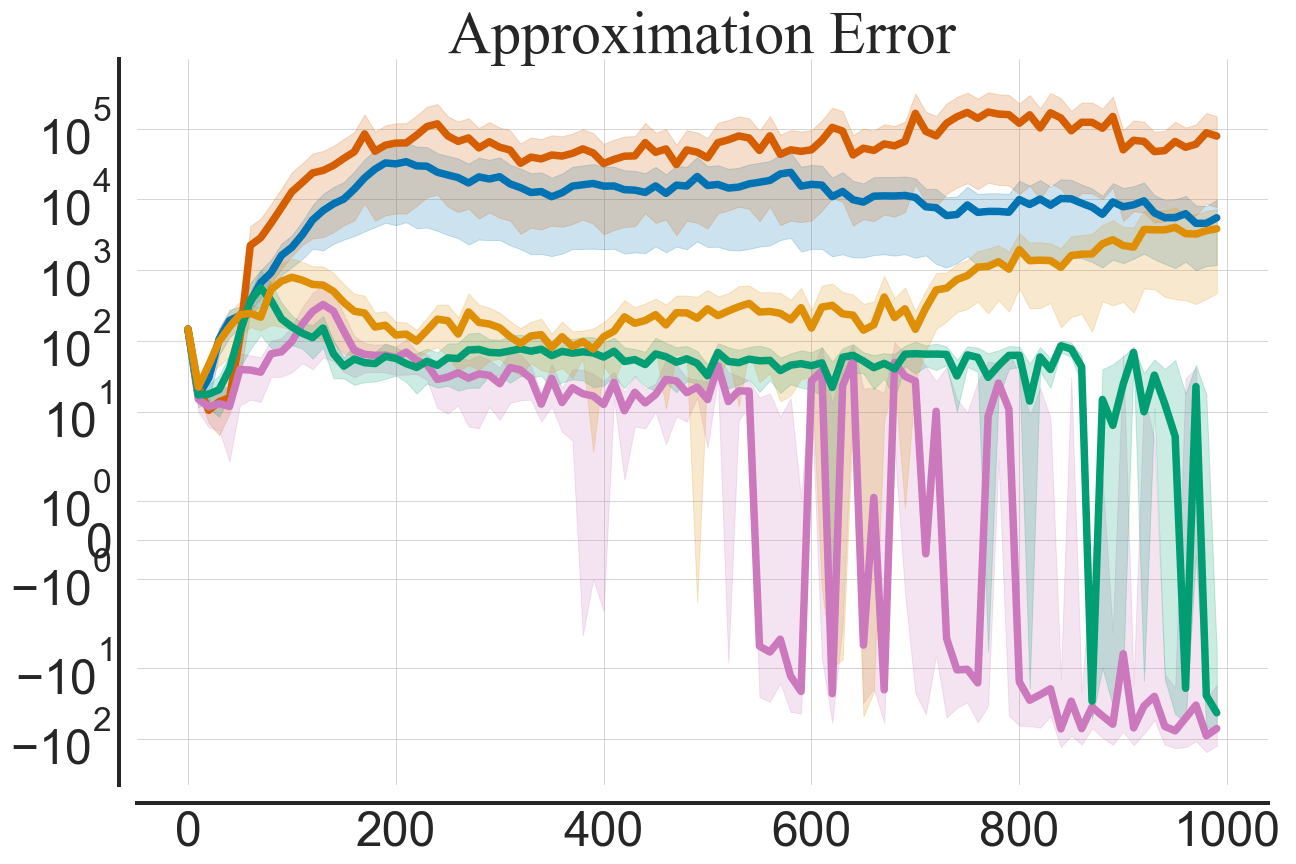}
    \hfill
    \includegraphics[width=0.23\linewidth]{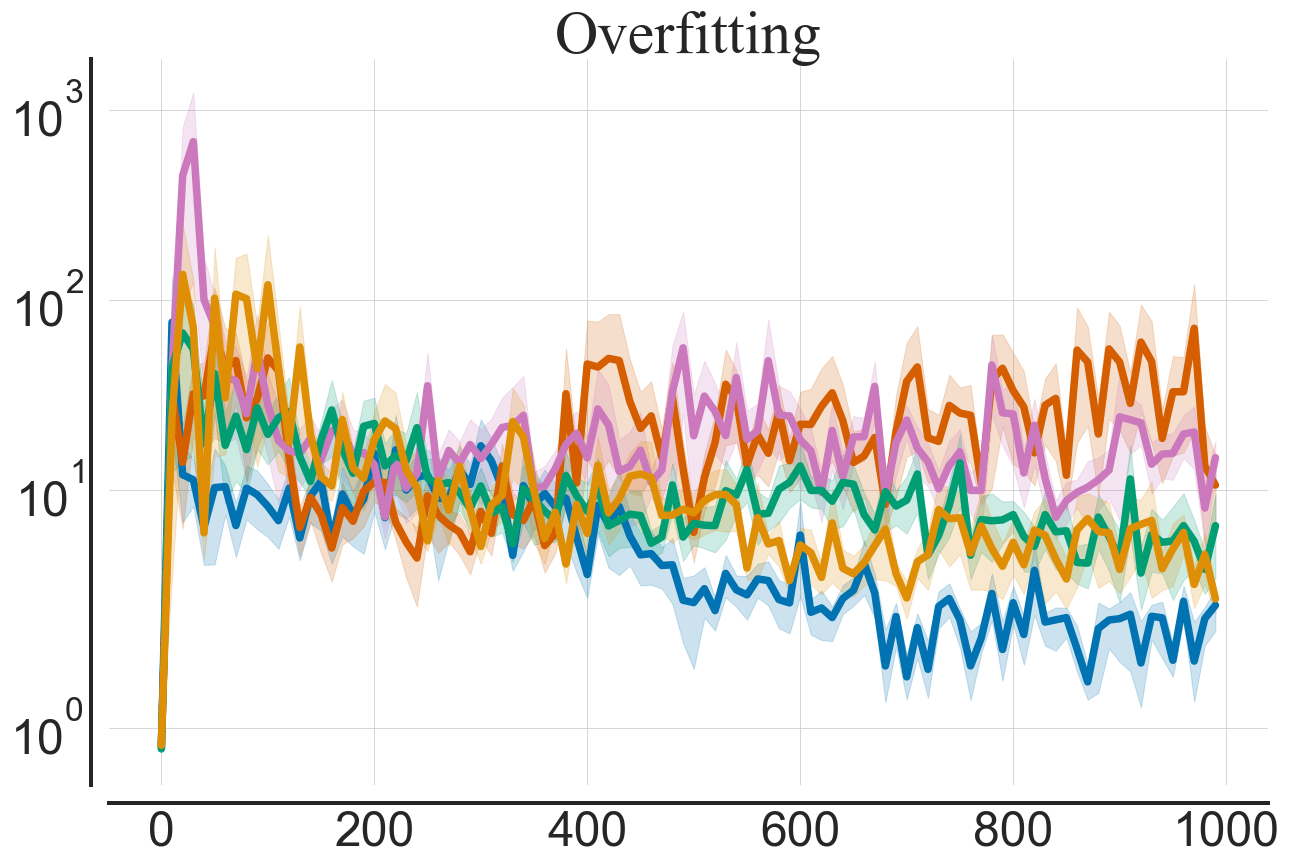}
    \hfill
    \end{subfigure}
    \subcaption{Box Close}
\end{minipage}
\bigskip
\begin{minipage}[h]{1.0\linewidth}
    \begin{subfigure}{1.0\linewidth}
    \hfill
    \includegraphics[width=0.23\linewidth]{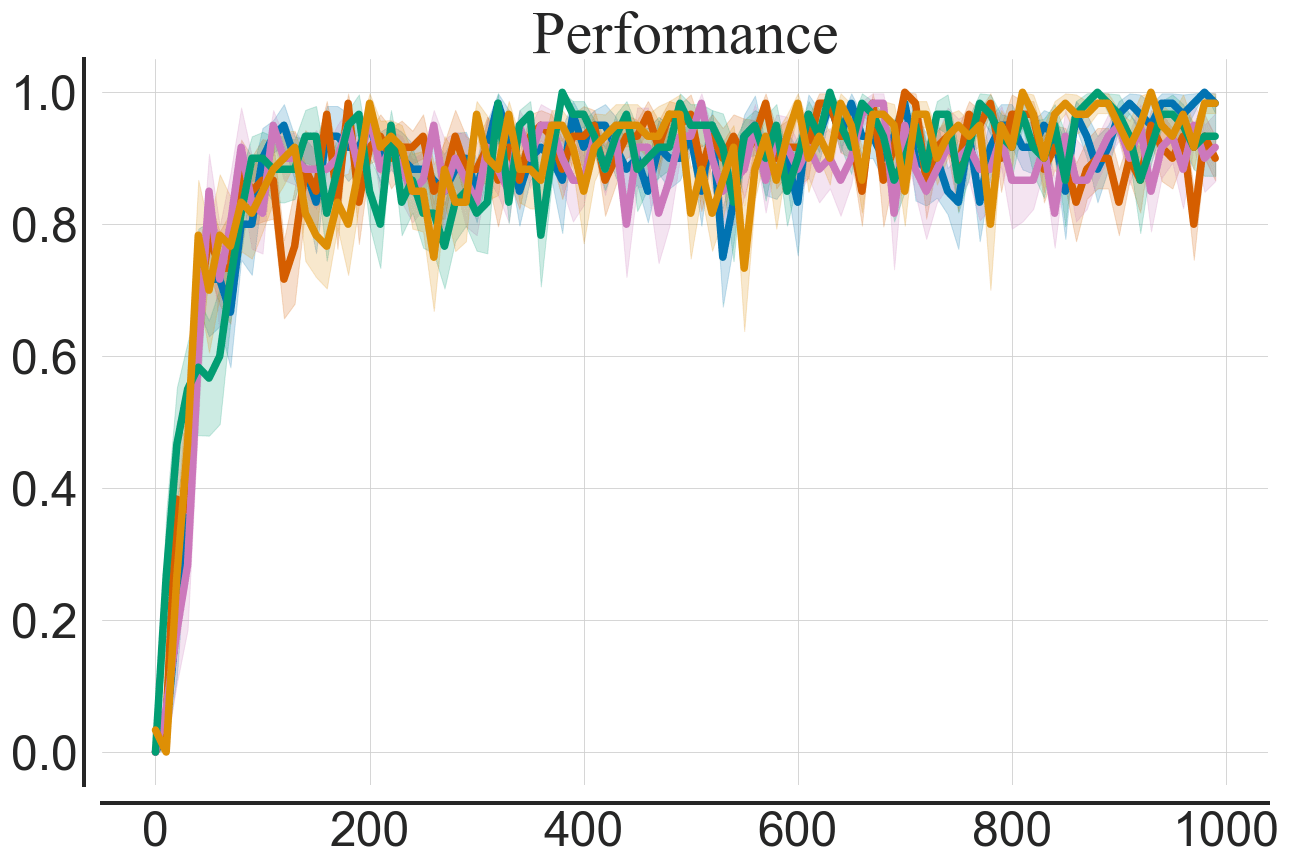}
    \hfill
    \includegraphics[width=0.23\linewidth]{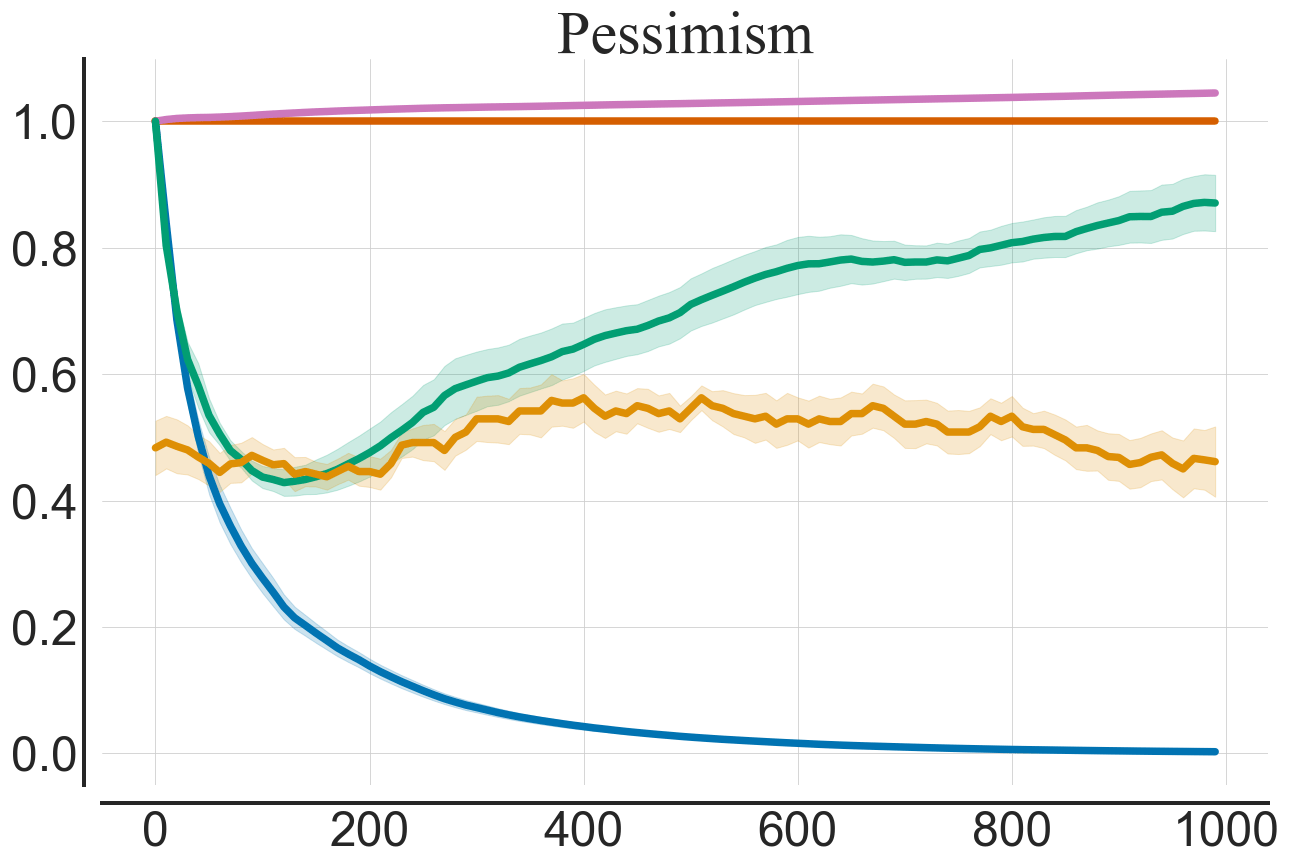}
    \hfill
    \includegraphics[width=0.23\linewidth]{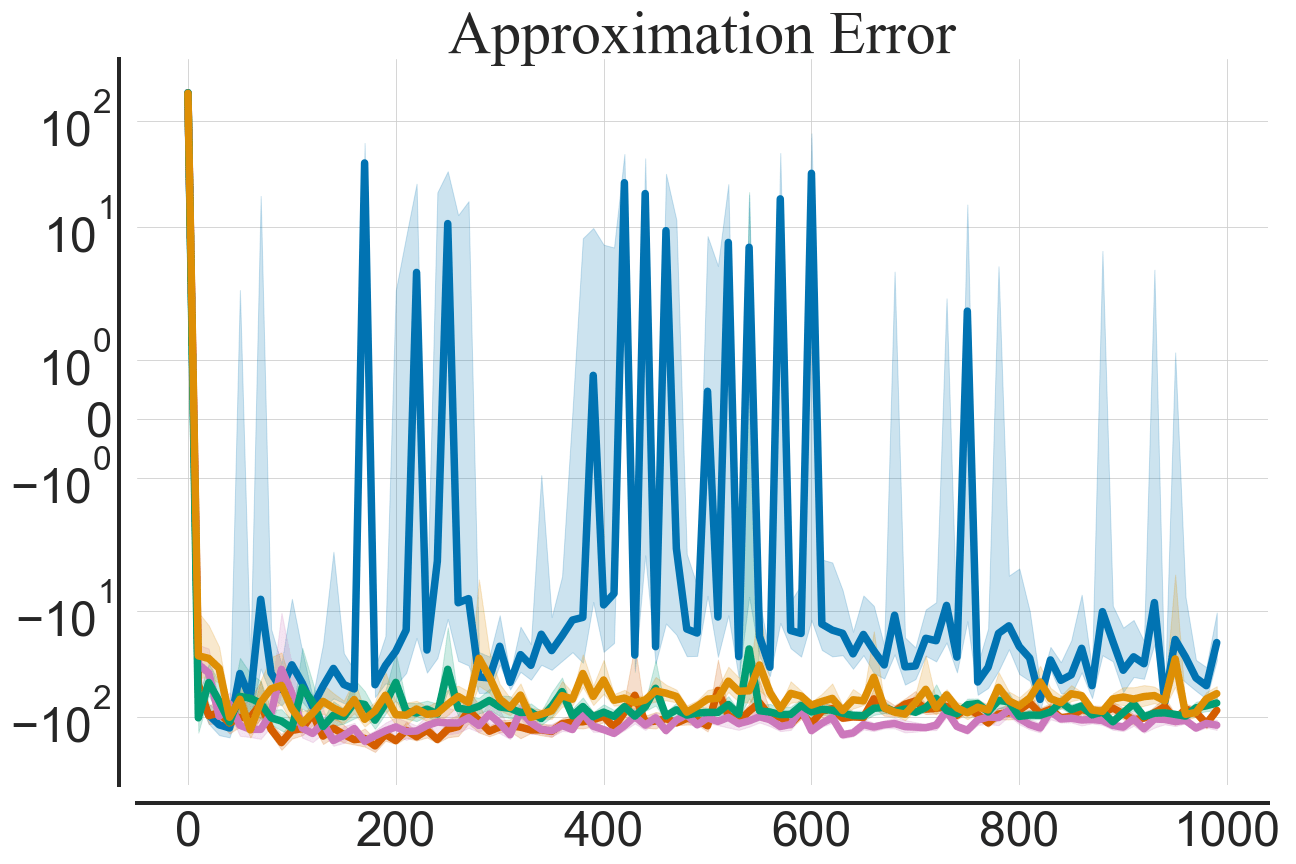}
    \hfill
    \includegraphics[width=0.23\linewidth]{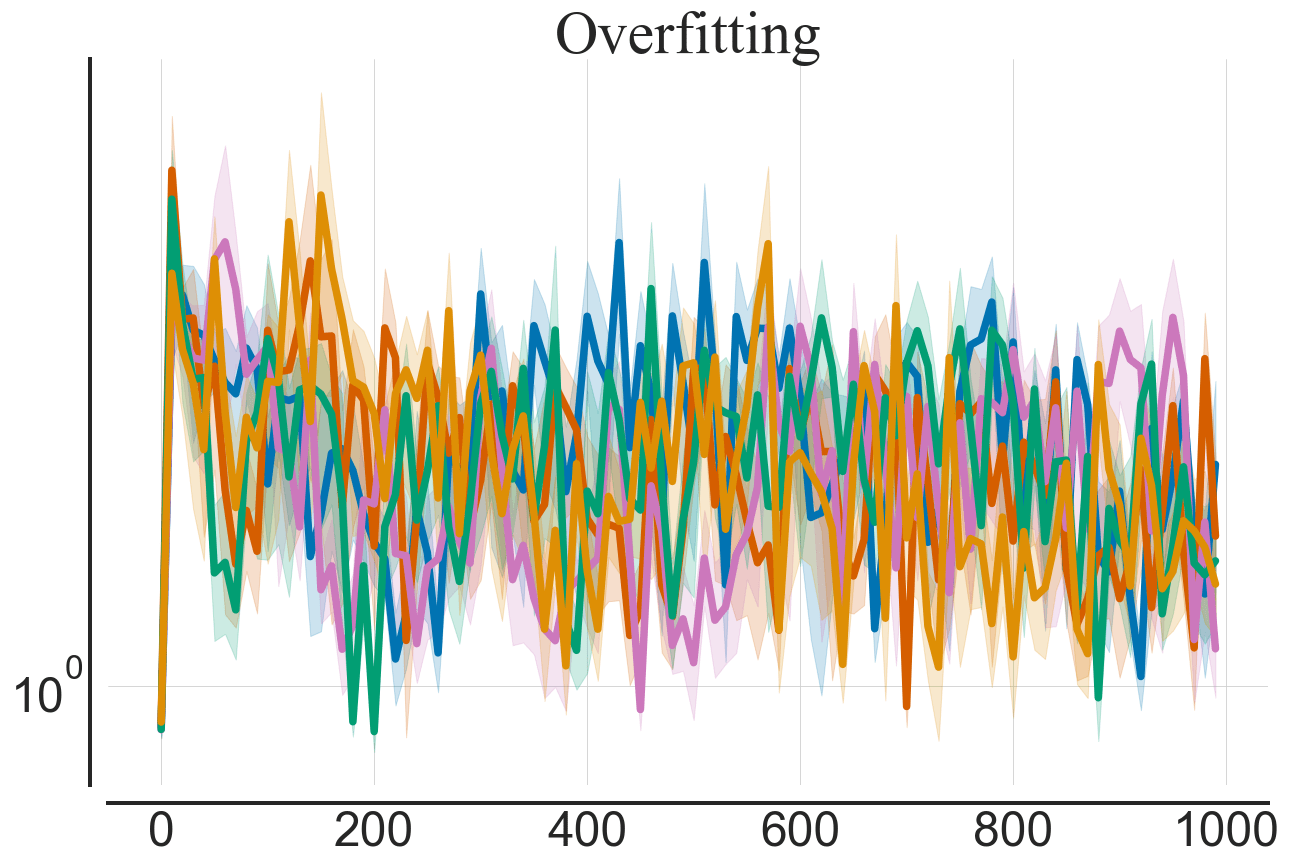}
    \hfill
    \end{subfigure}
    \subcaption{Button Press}
\end{minipage}
\bigskip
\begin{minipage}[h]{1.0\linewidth}
    \begin{subfigure}{1.0\linewidth}
    \hfill
    \includegraphics[width=0.23\linewidth]{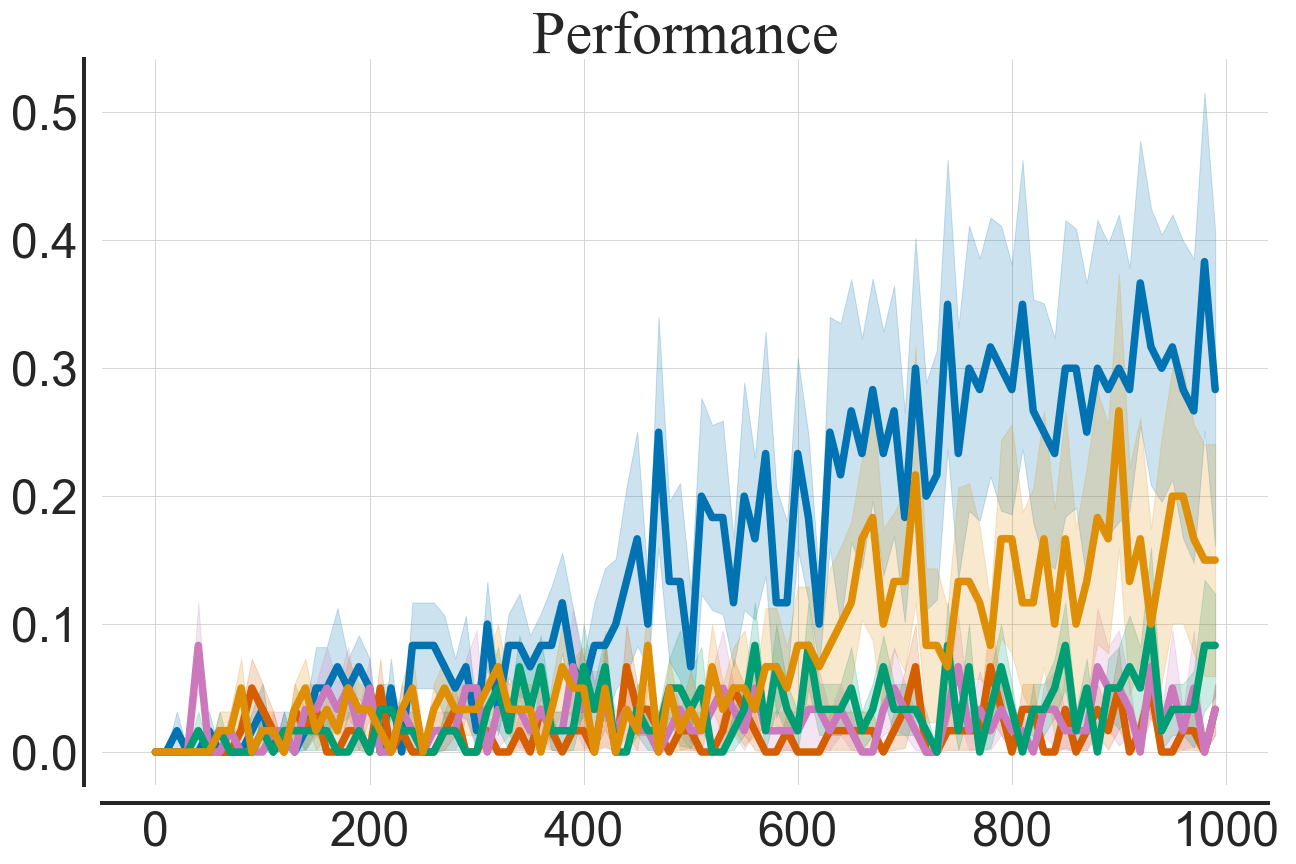}
    \hfill
    \includegraphics[width=0.23\linewidth]{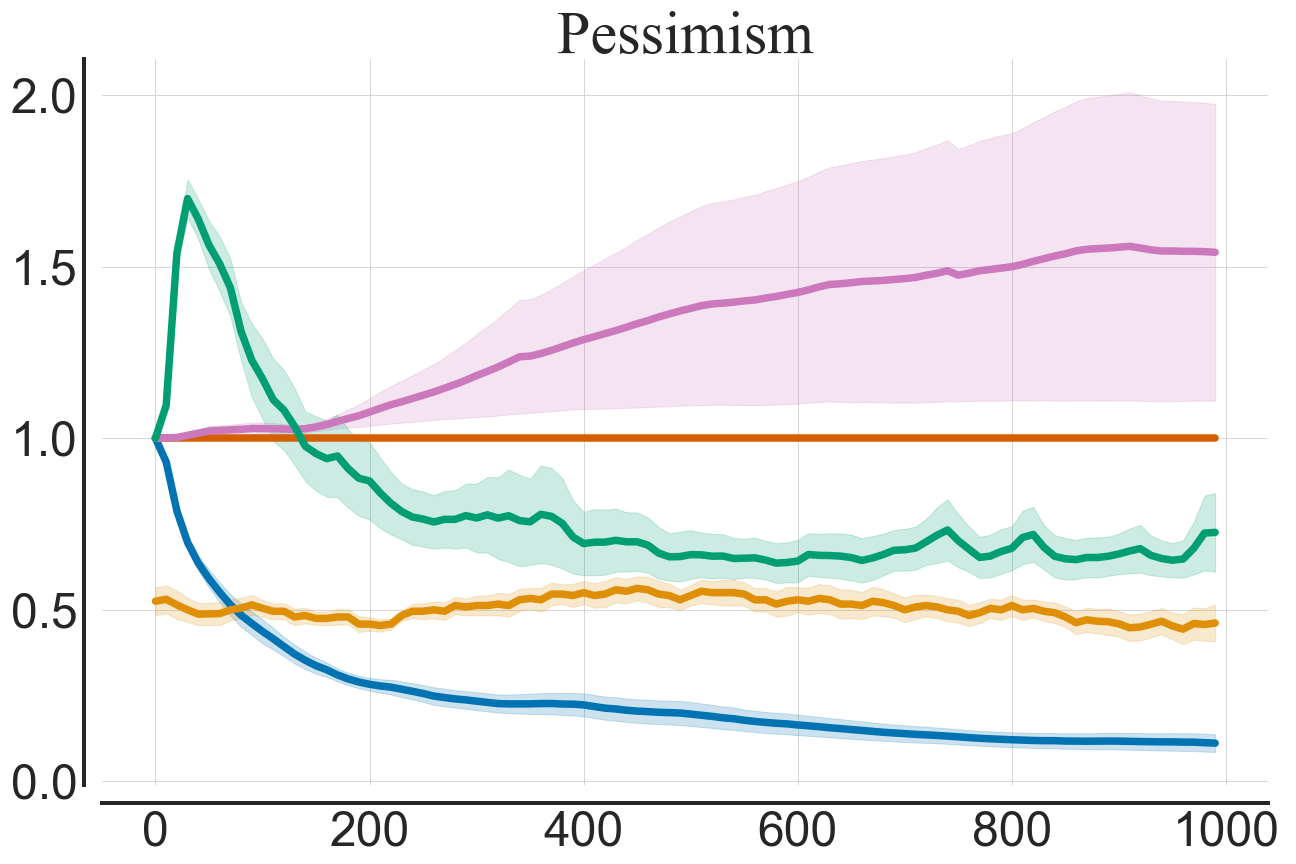}
    \hfill
    \includegraphics[width=0.23\linewidth]{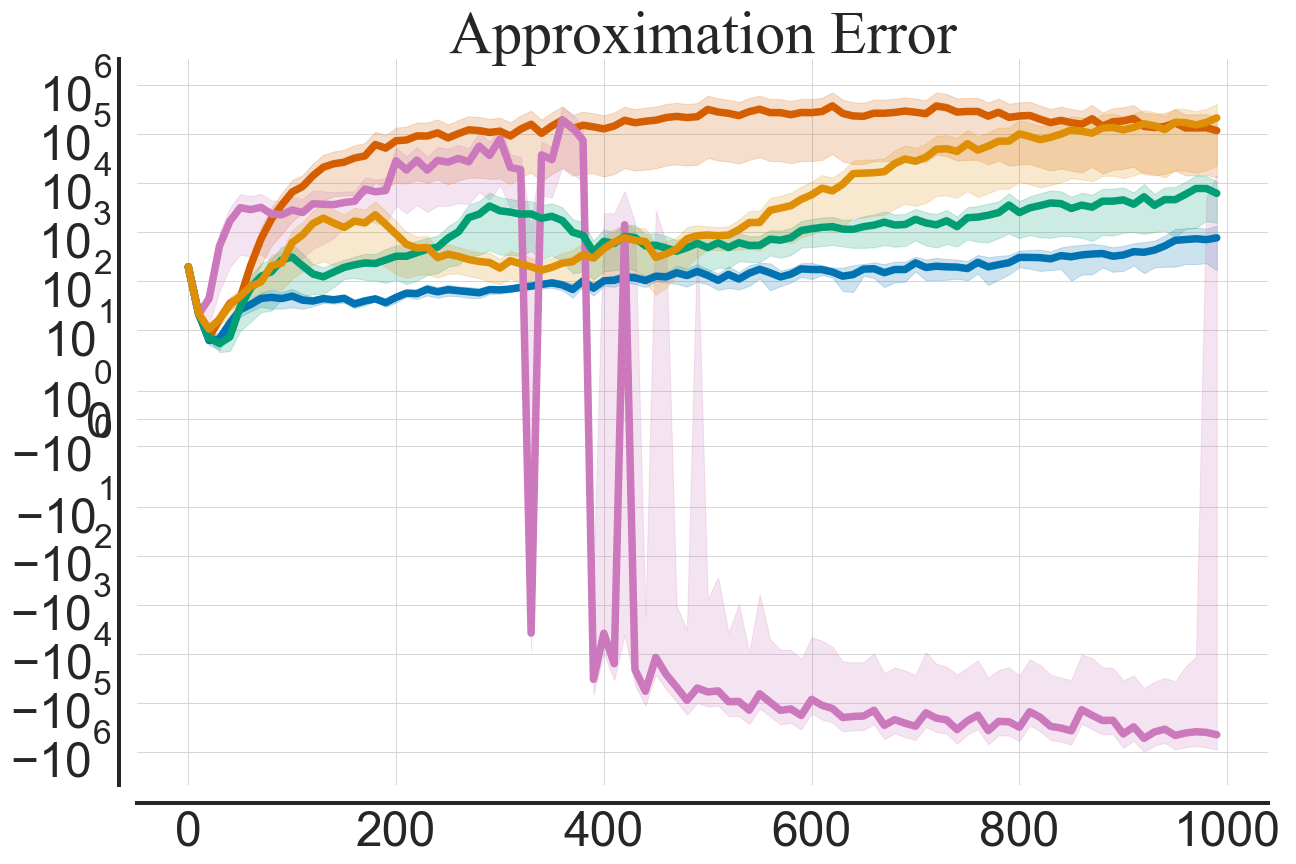}
    \hfill
    \includegraphics[width=0.23\linewidth]{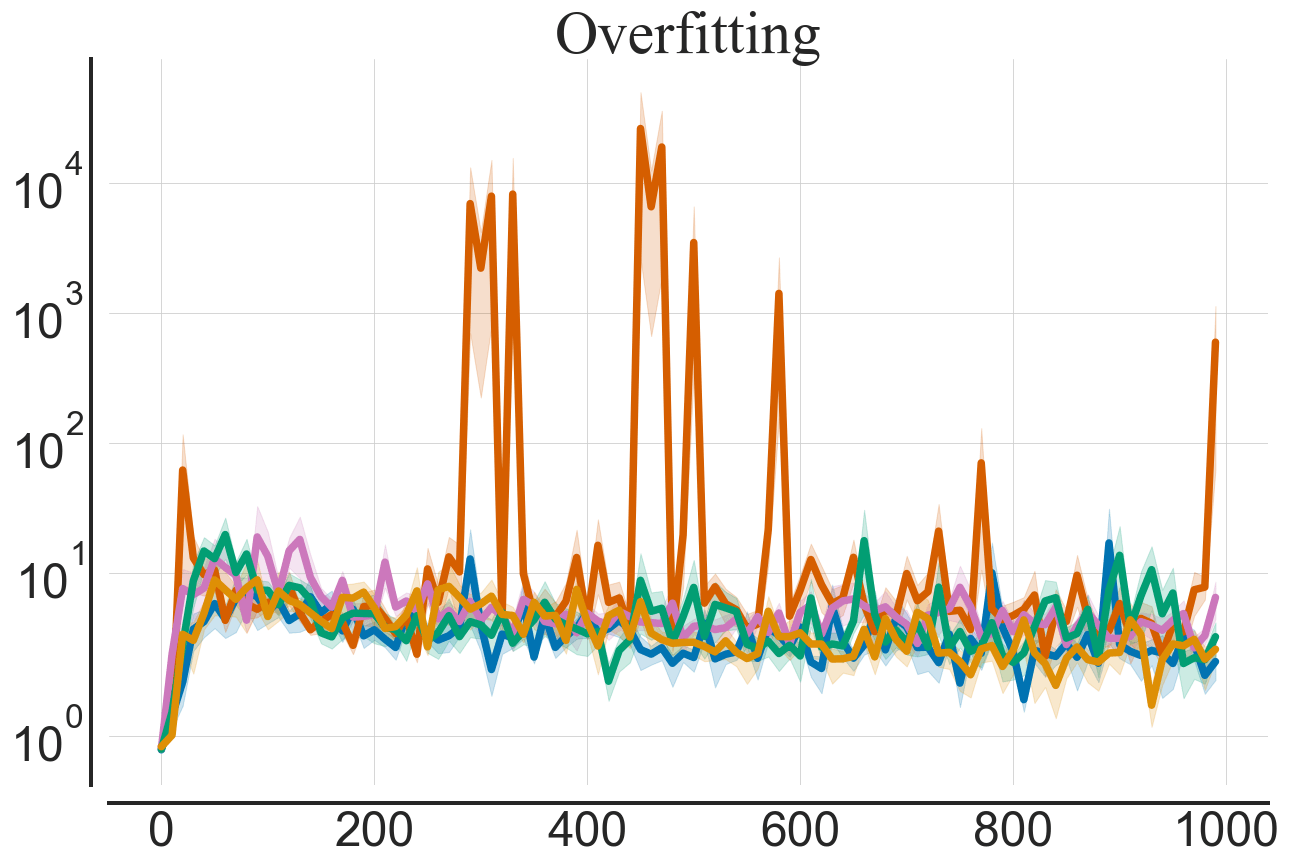}
    \hfill
    \end{subfigure}
    \subcaption{Coffee Pull}
\end{minipage}
\bigskip
\begin{minipage}[h]{1.0\linewidth}
    \begin{subfigure}{1.0\linewidth}
    \hfill
    \includegraphics[width=0.23\linewidth]{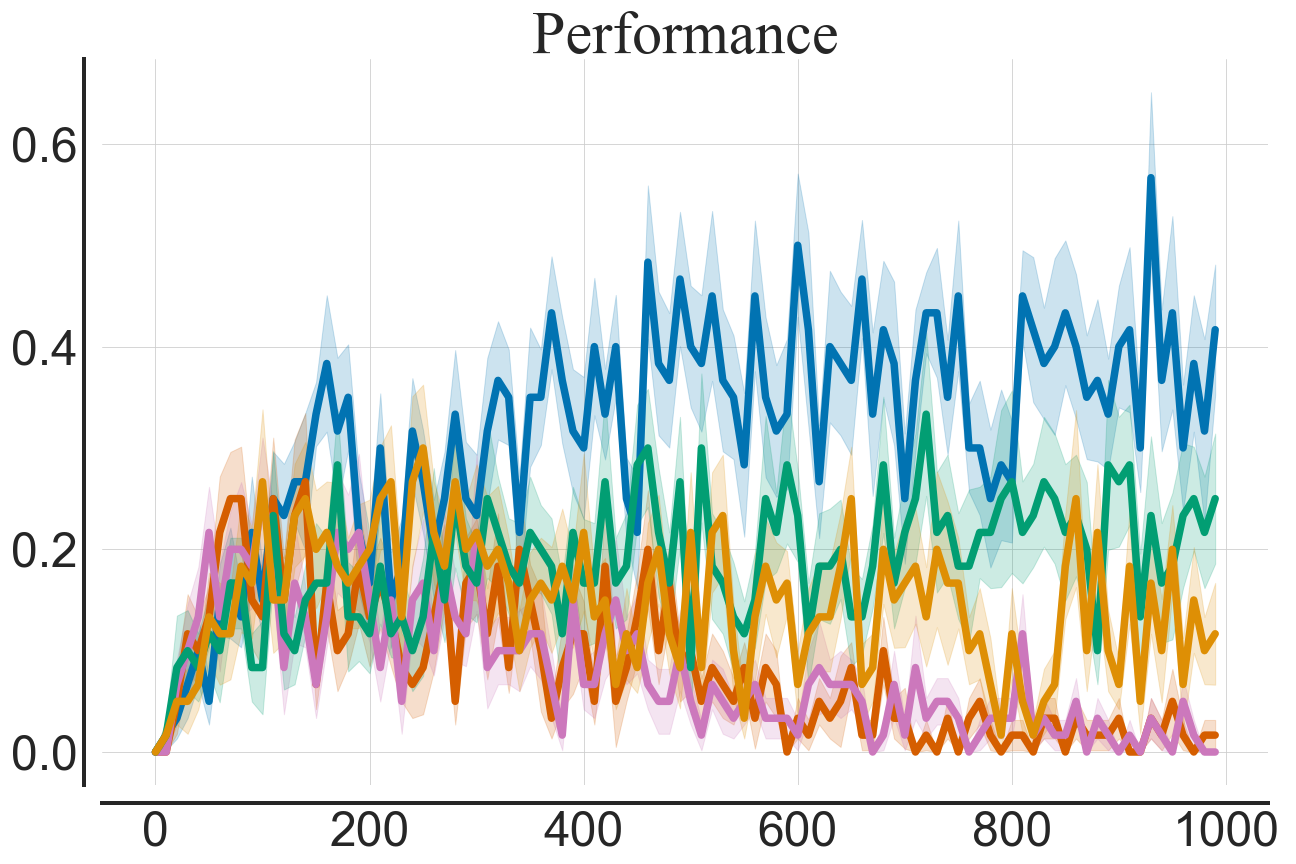}
    \hfill
    \includegraphics[width=0.23\linewidth]{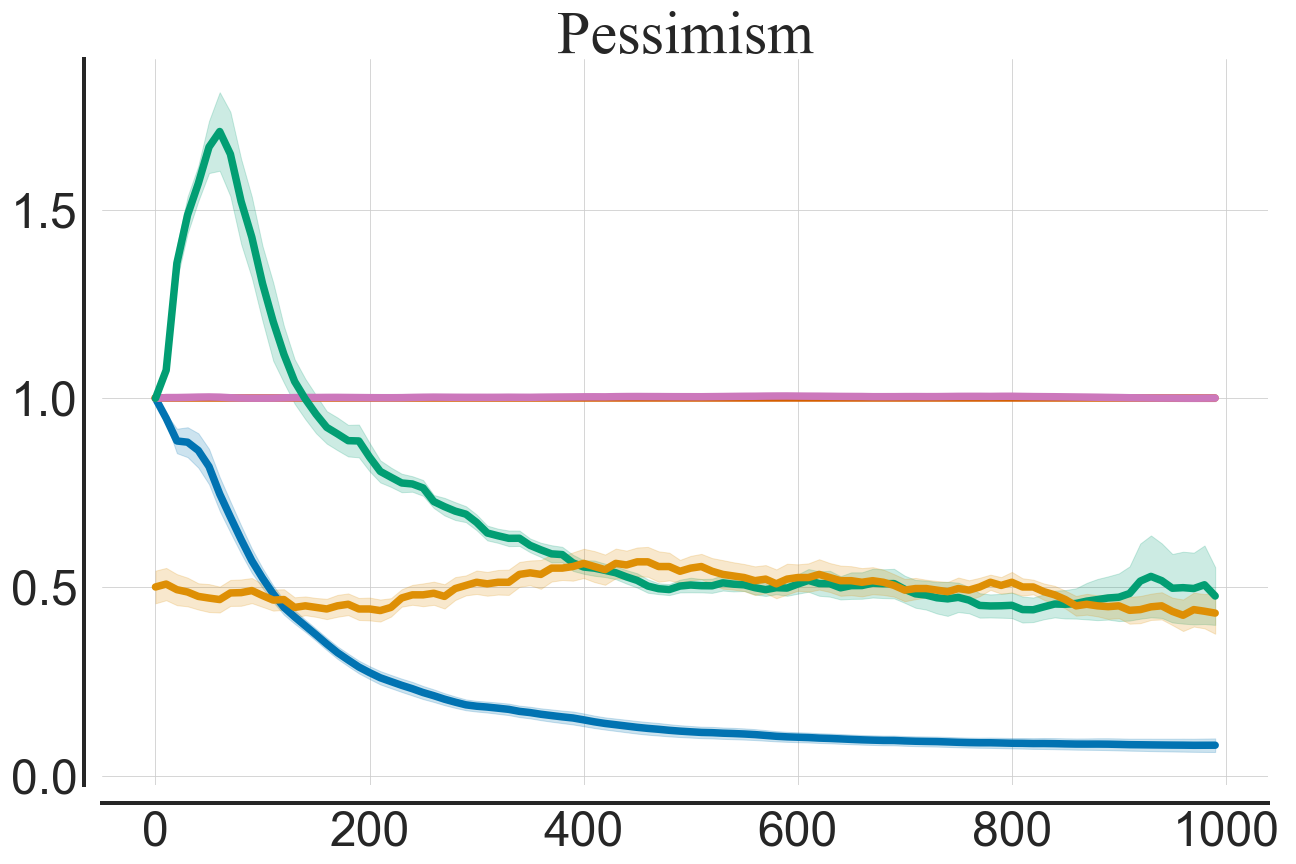}
    \hfill
    \includegraphics[width=0.23\linewidth]{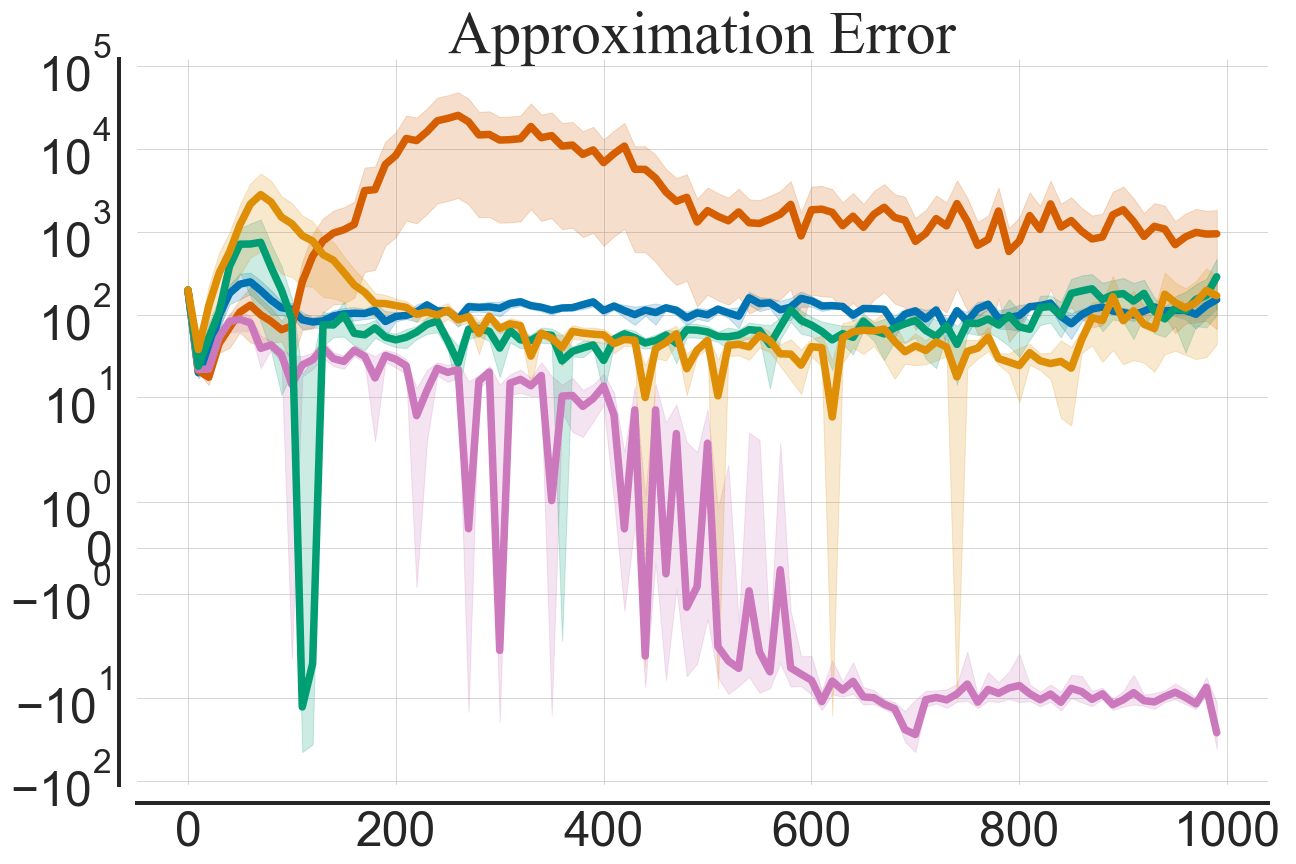}
    \hfill
    \includegraphics[width=0.23\linewidth]{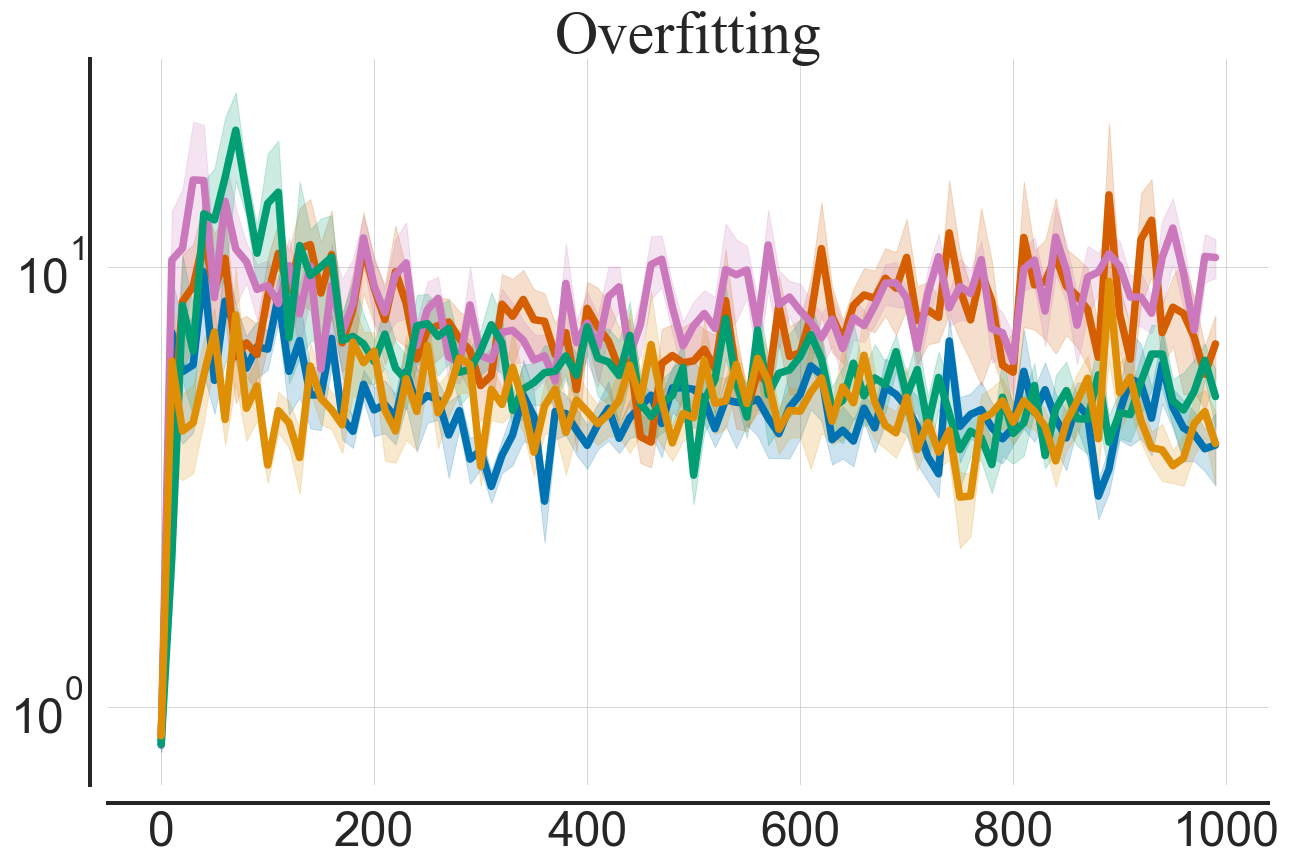}
    \hfill
    \end{subfigure}
    \subcaption{Coffee Push}
\end{minipage}
\caption{Low replay regime results for each considered task (3/4). 10 seeds per task, mean and 3 standard deviations.}
\label{fig:learning_curves3}
\end{center}
\end{figure*}

\begin{figure*}[ht!]
\begin{center}
\begin{minipage}[h]{1.0\linewidth}
\centering
    \begin{subfigure}{0.88\linewidth}
    \includegraphics[width=\textwidth]{images/legend_1.png}
    \end{subfigure}
\end{minipage}
\bigskip
\begin{minipage}[h]{1.0\linewidth}
    \begin{subfigure}{1.0\linewidth}
    \hfill
    \includegraphics[width=0.23\linewidth]{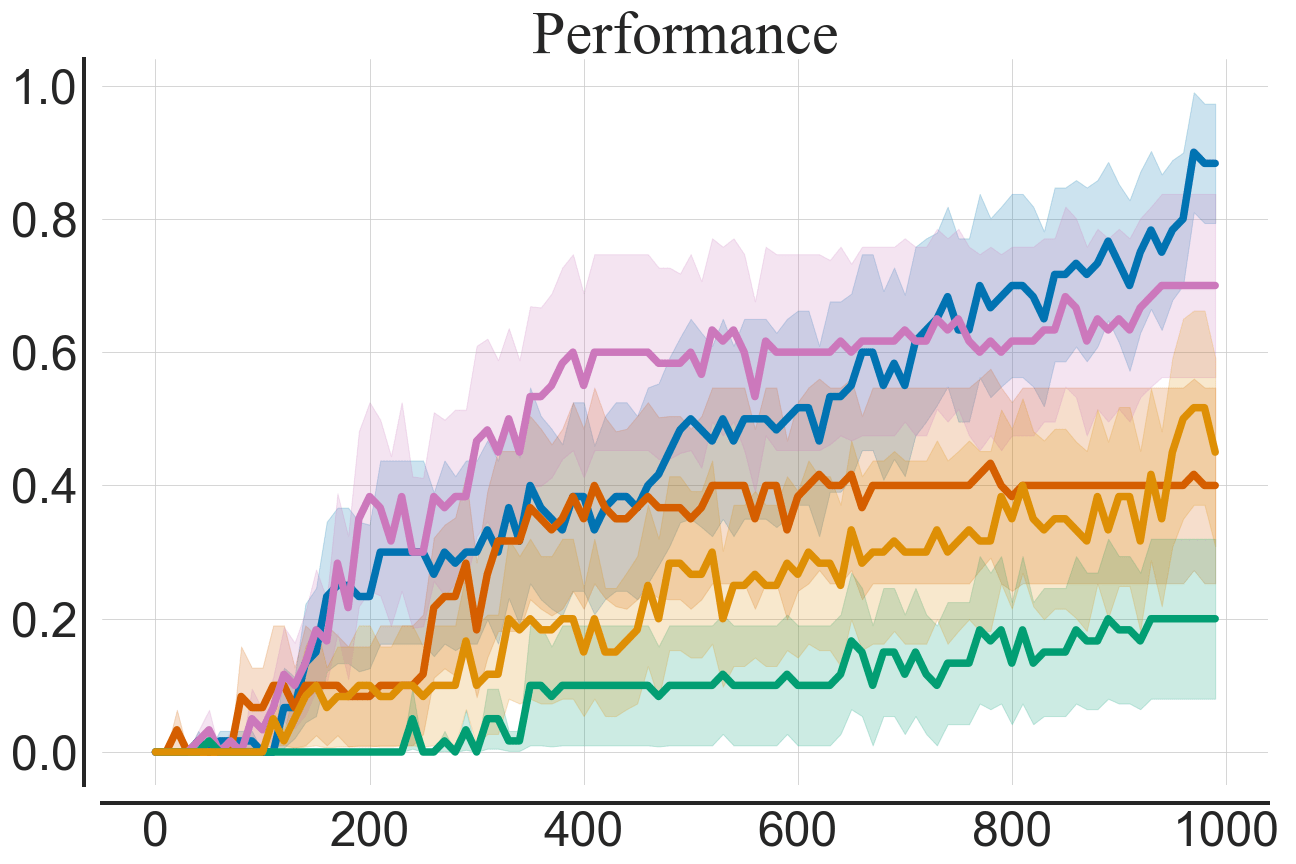}
    \hfill
    \includegraphics[width=0.23\linewidth]{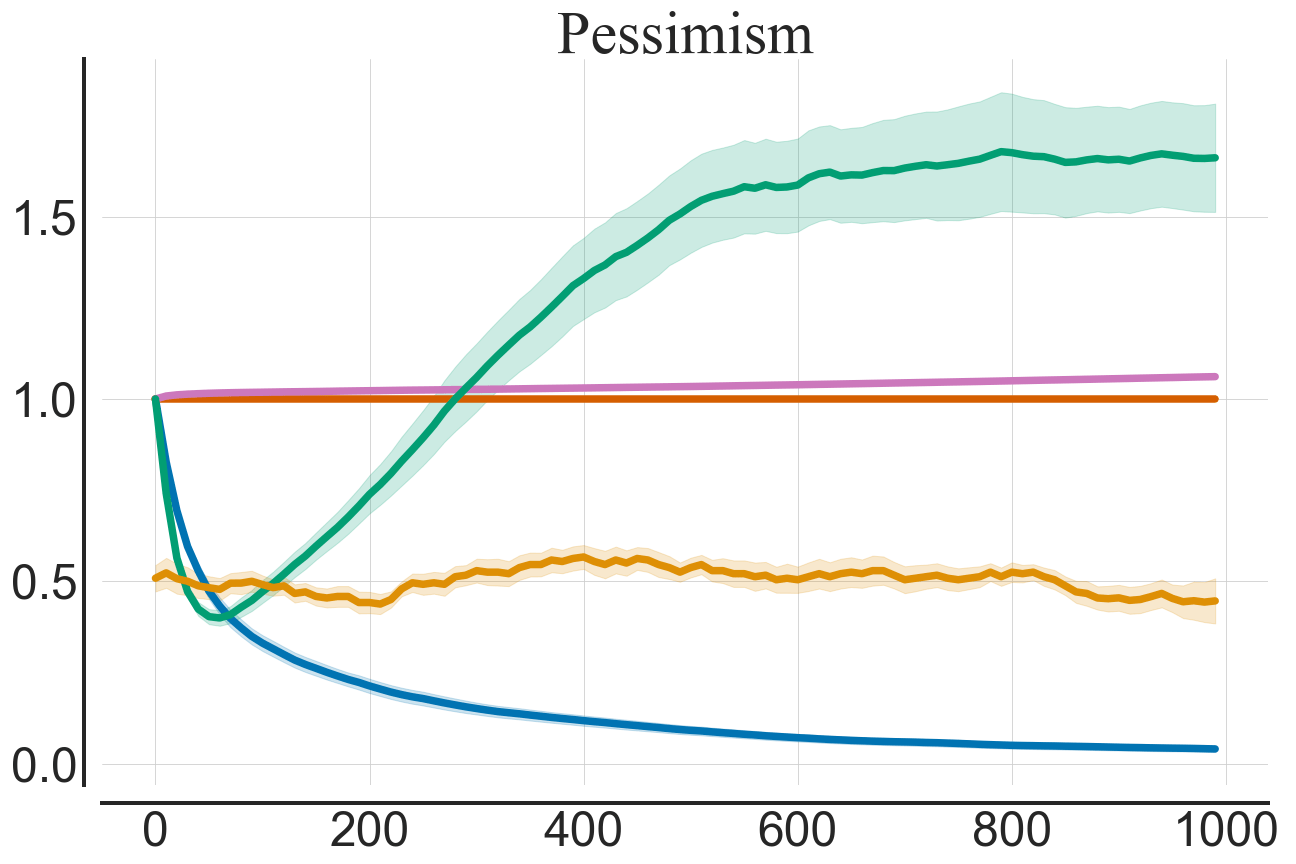}
    \hfill
    \includegraphics[width=0.23\linewidth]{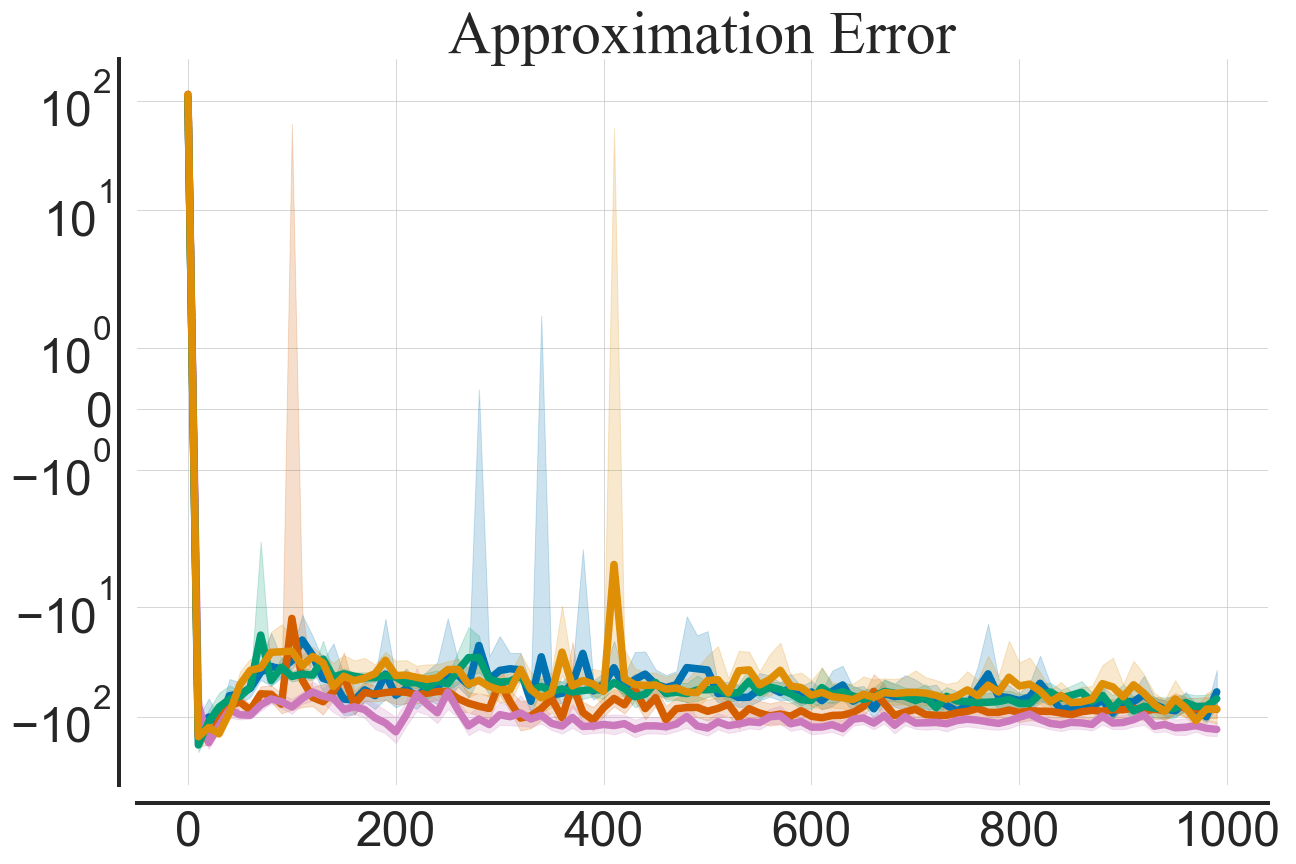}
    \hfill
    \includegraphics[width=0.23\linewidth]{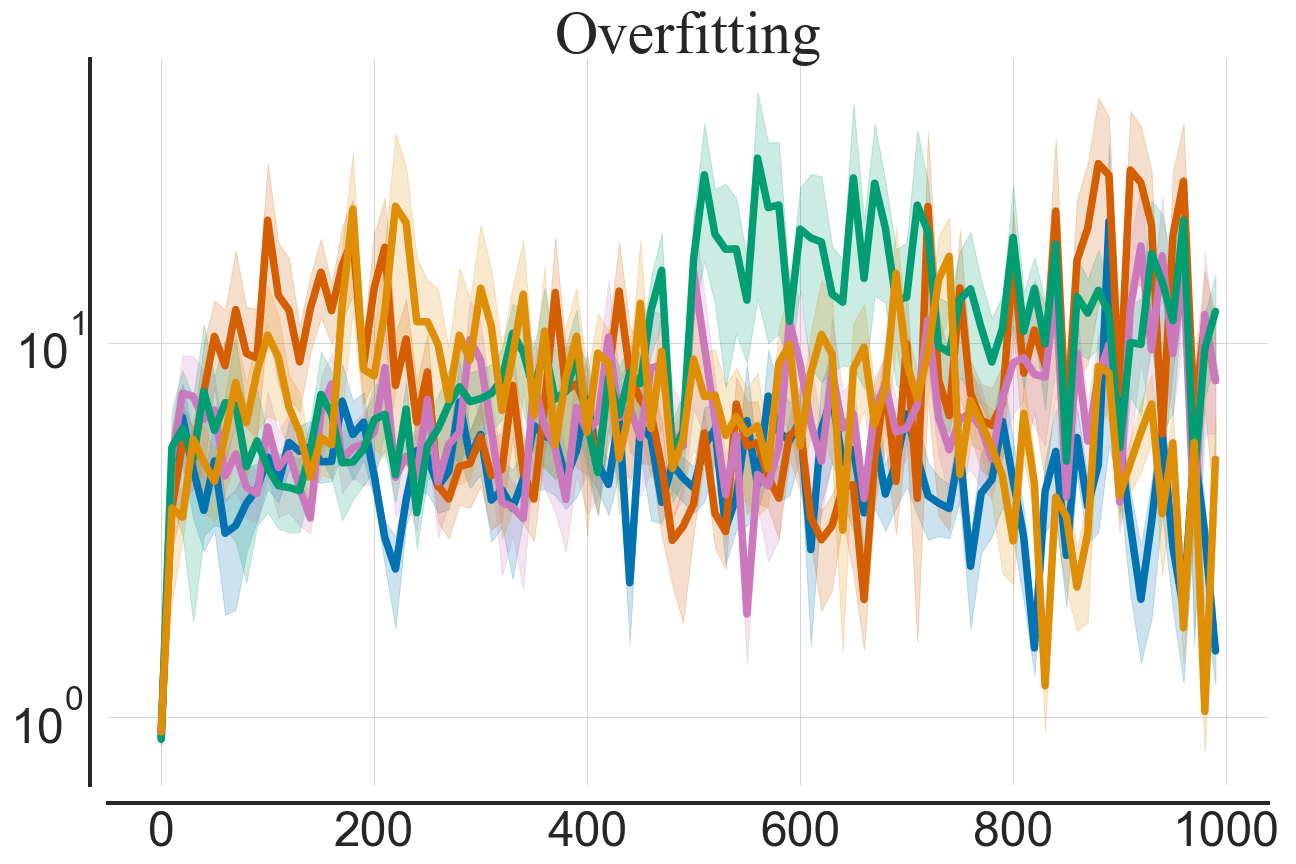}
    \hfill
    \end{subfigure}
    \subcaption{Drawer Open}
\end{minipage}
\bigskip
\begin{minipage}[h]{1.0\linewidth}
    \begin{subfigure}{1.0\linewidth}
    \hfill
    \includegraphics[width=0.23\linewidth]{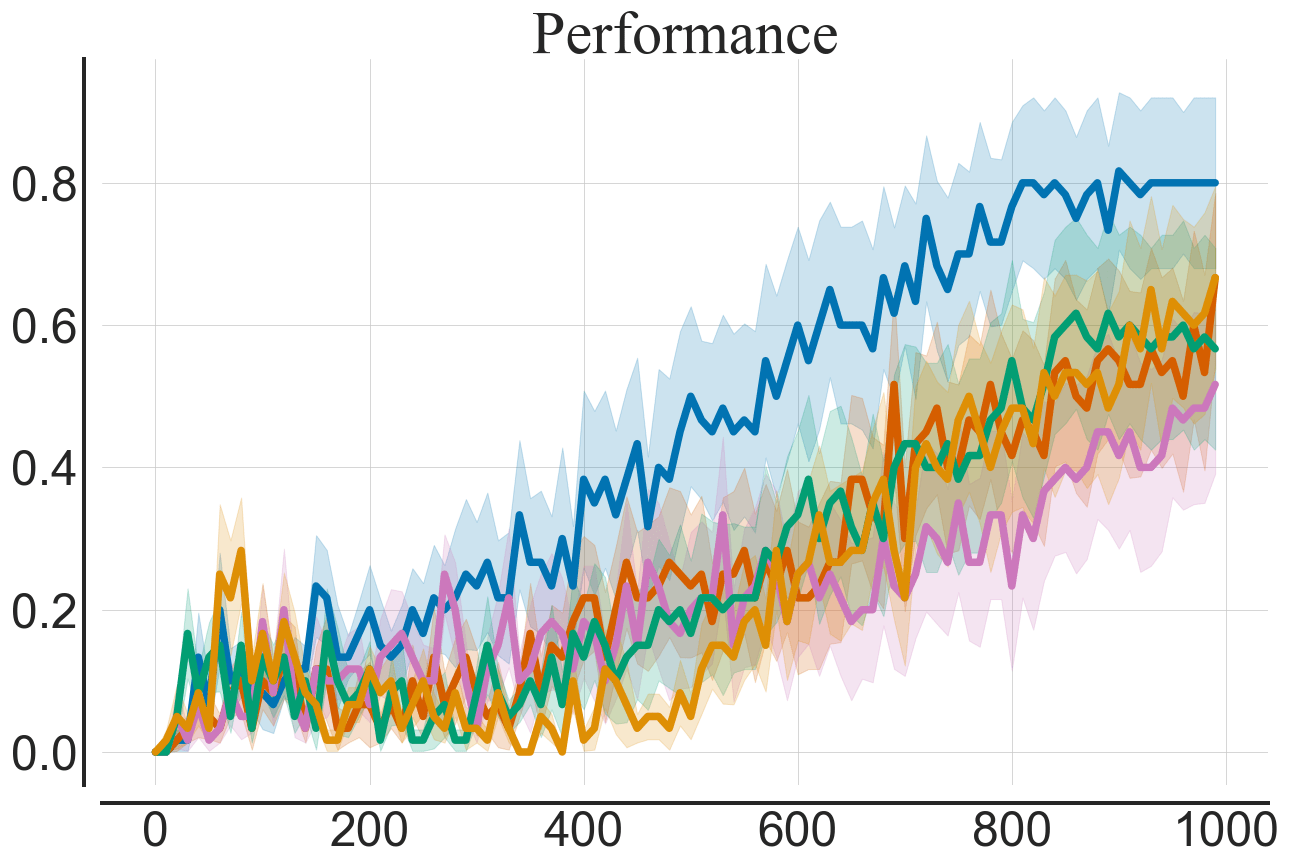}
    \hfill
    \includegraphics[width=0.23\linewidth]{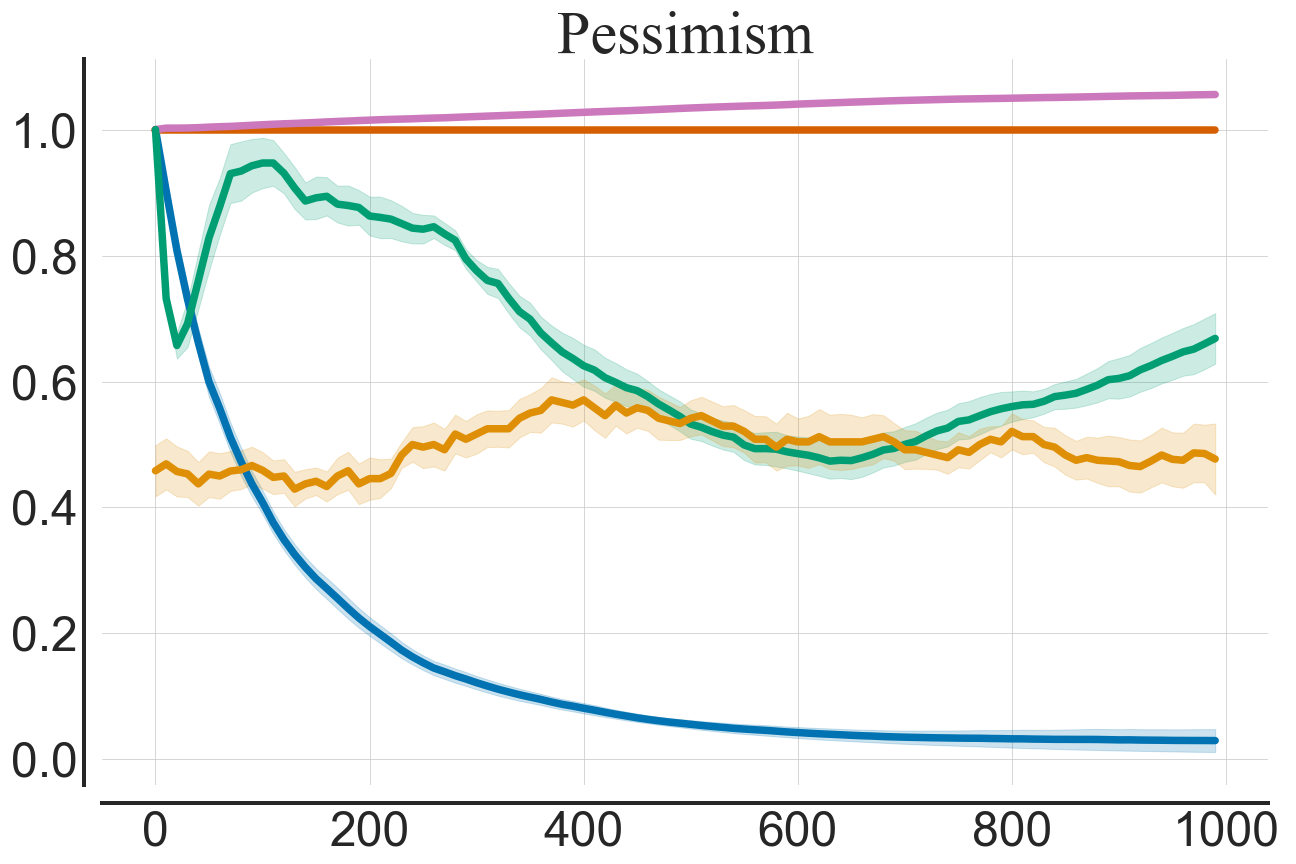}
    \hfill
    \includegraphics[width=0.23\linewidth]{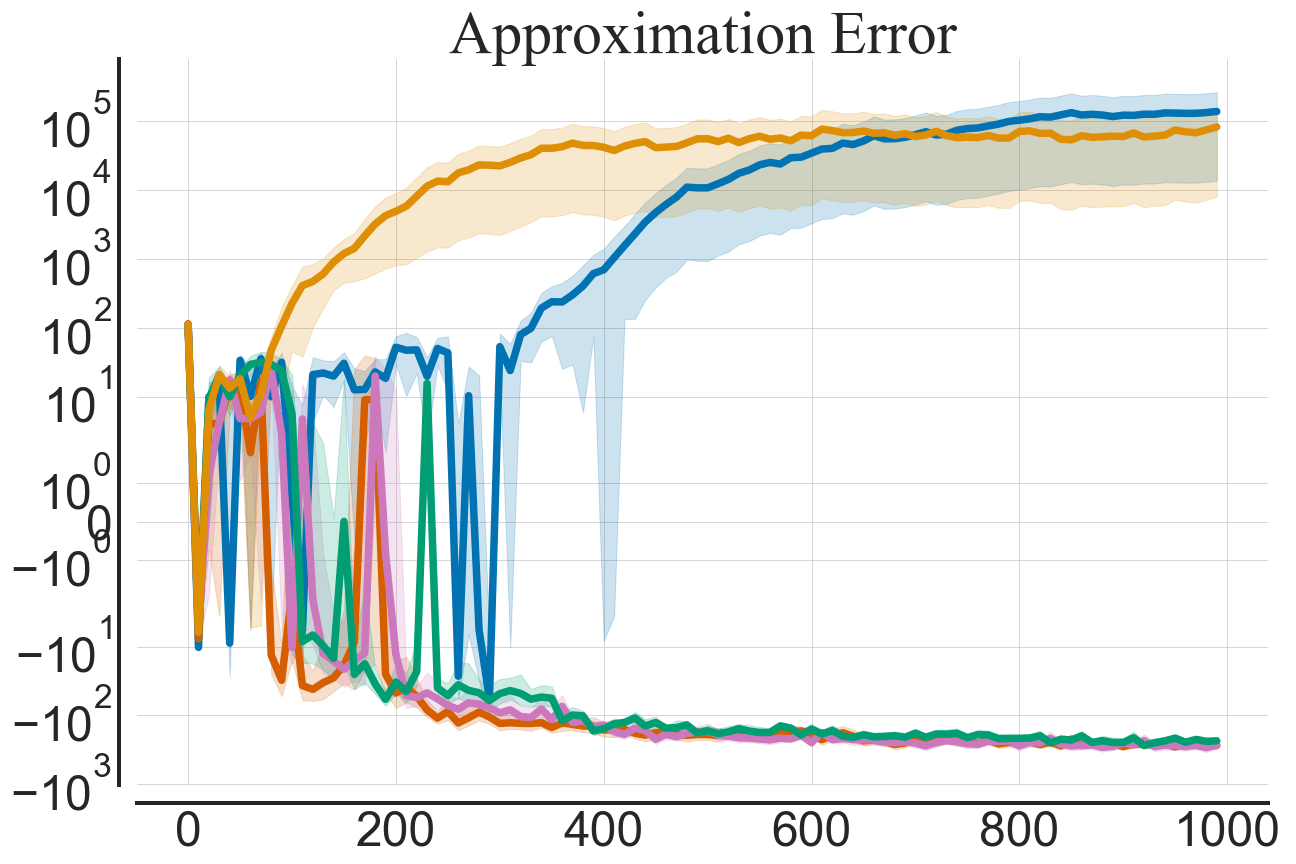}
    \hfill
    \includegraphics[width=0.23\linewidth]{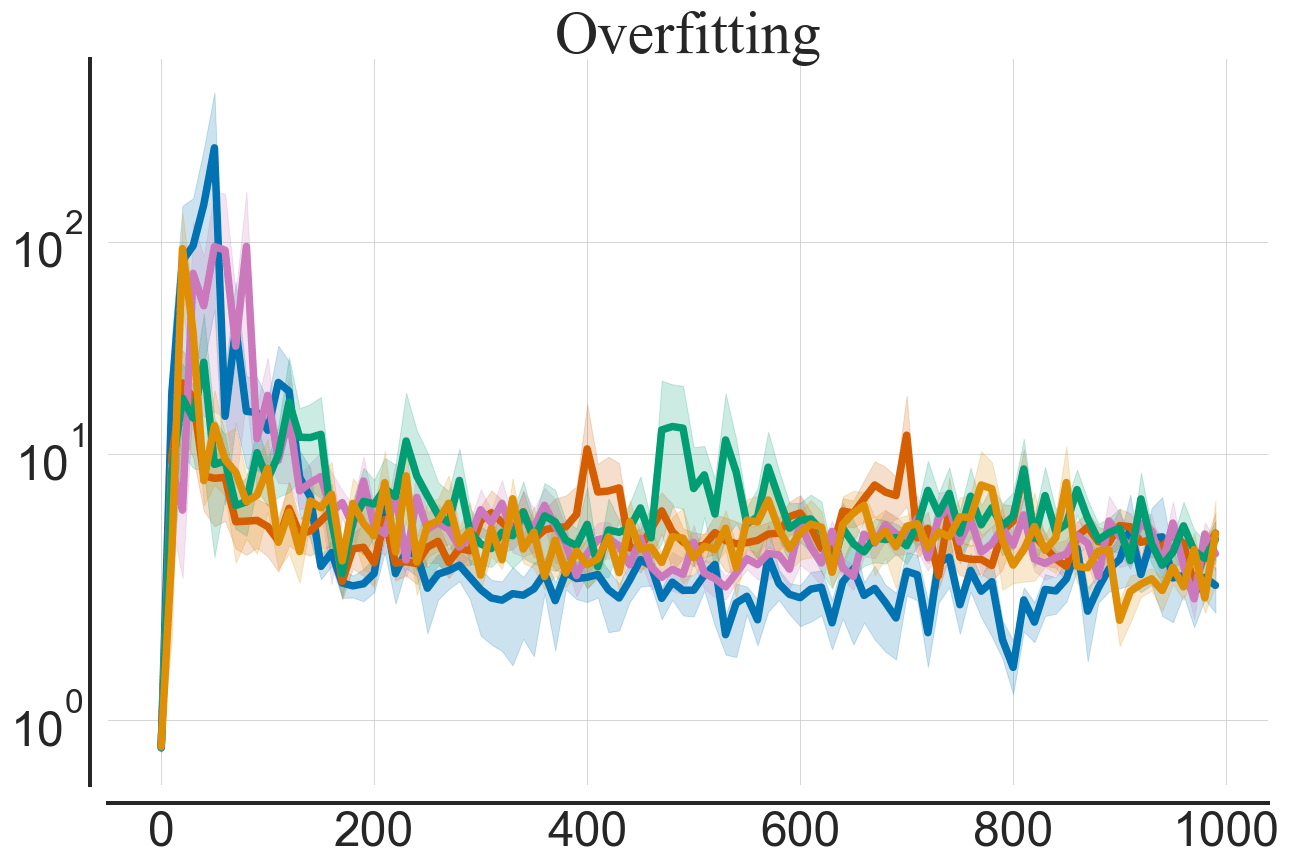}
    \hfill
    \end{subfigure}
    \subcaption{Hammer}
\end{minipage}
\bigskip
\begin{minipage}[h]{1.0\linewidth}
    \begin{subfigure}{1.0\linewidth}
    \hfill
    \includegraphics[width=0.23\linewidth]{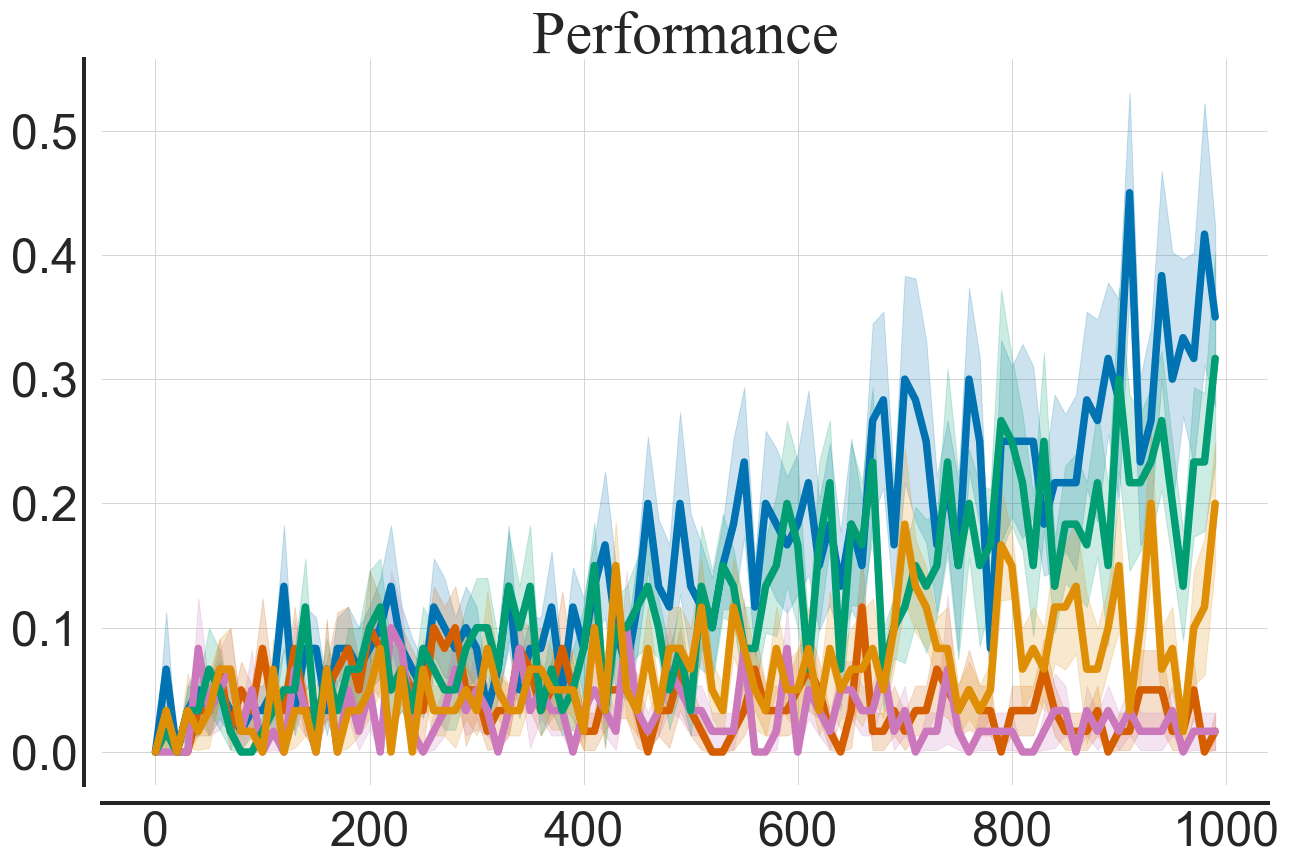}
    \hfill
    \includegraphics[width=0.23\linewidth]{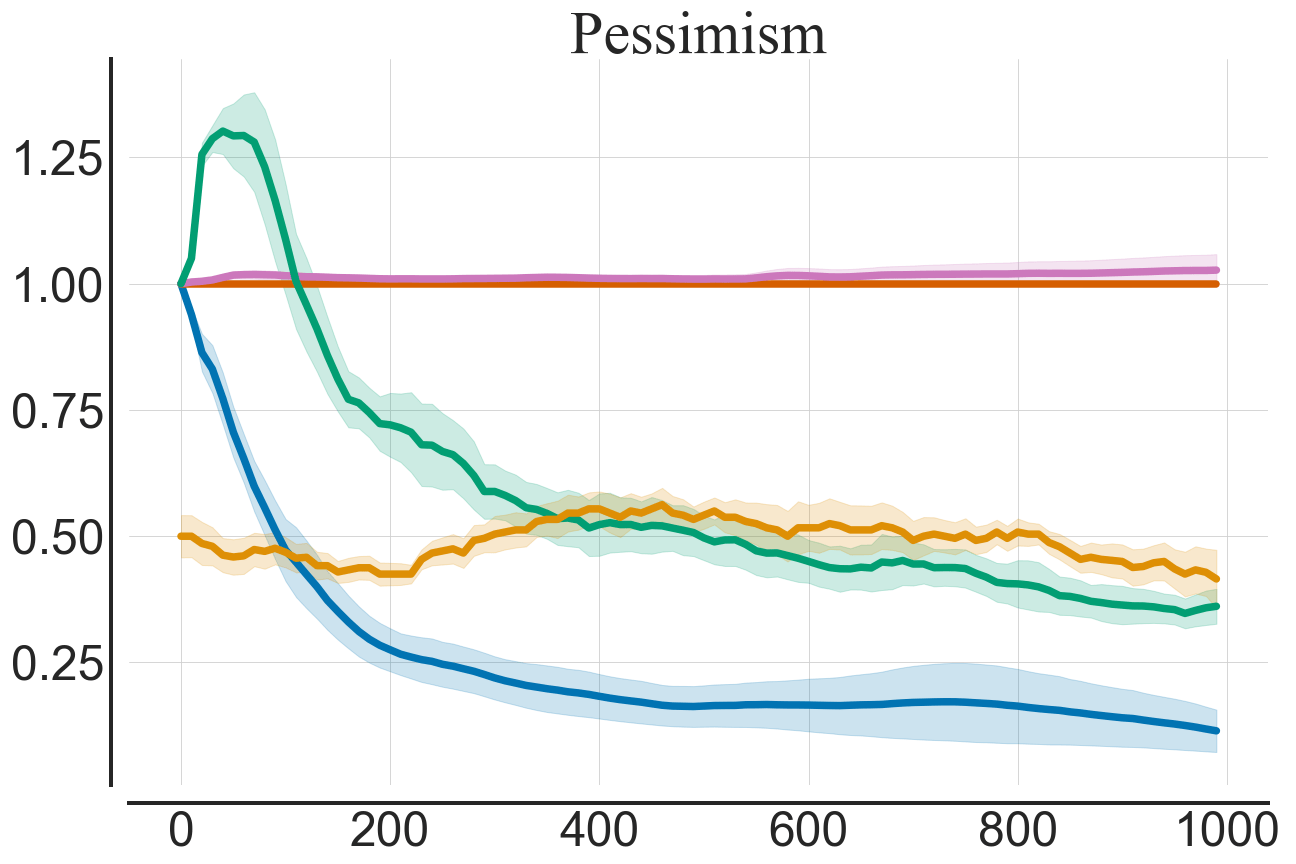}
    \hfill
    \includegraphics[width=0.23\linewidth]{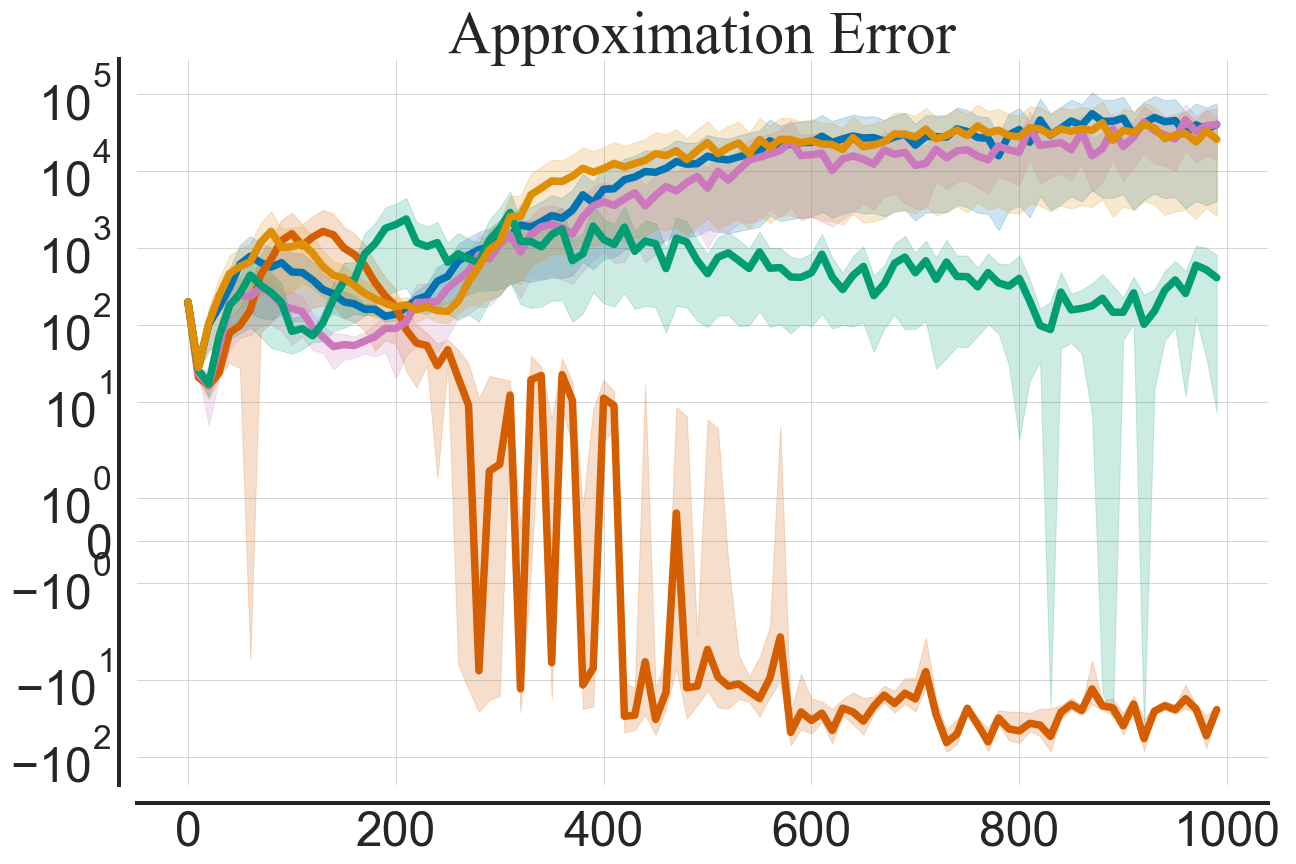}
    \hfill
    \includegraphics[width=0.23\linewidth]{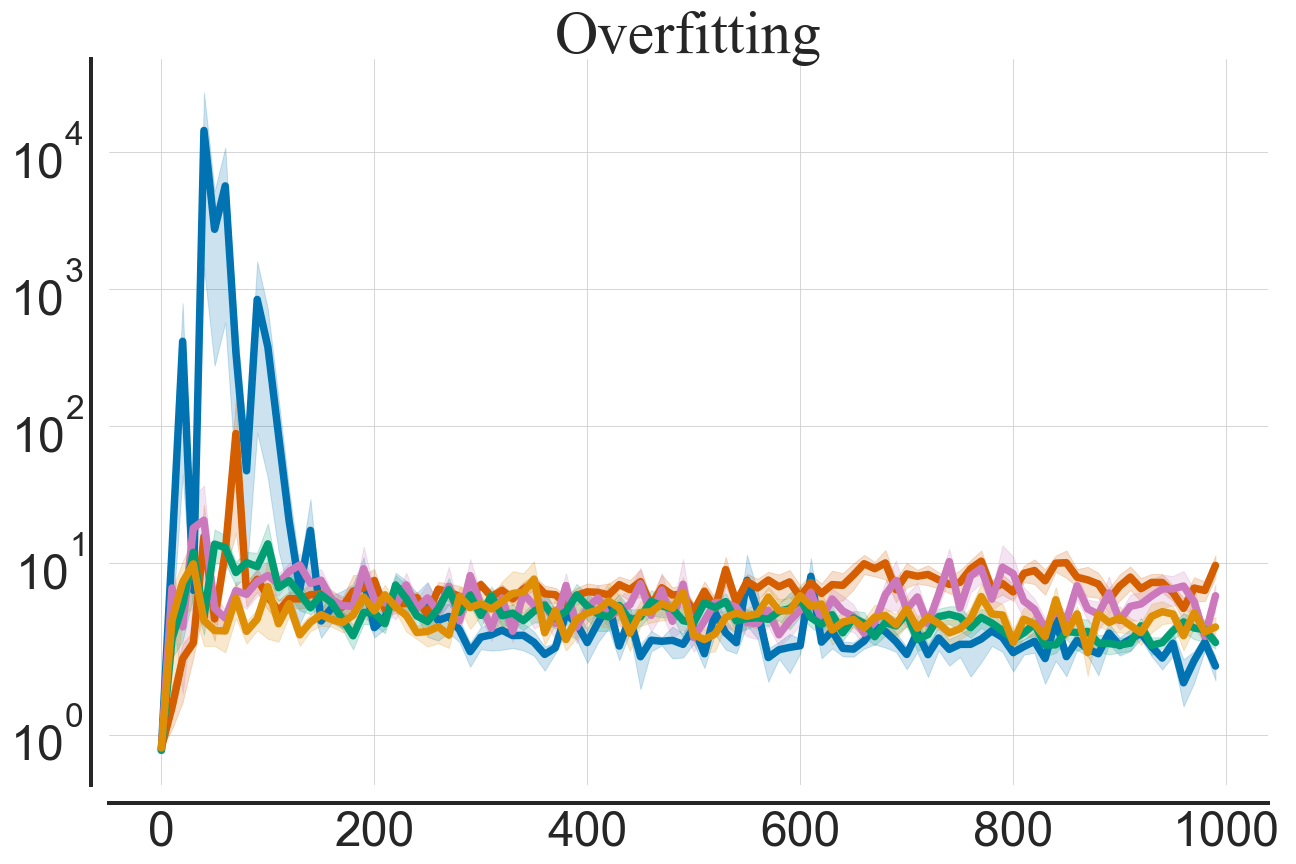}
    \hfill
    \end{subfigure}
    \subcaption{Push}
\end{minipage}
\bigskip
\begin{minipage}[h]{1.0\linewidth}
    \begin{subfigure}{1.0\linewidth}
    \hfill
    \includegraphics[width=0.23\linewidth]{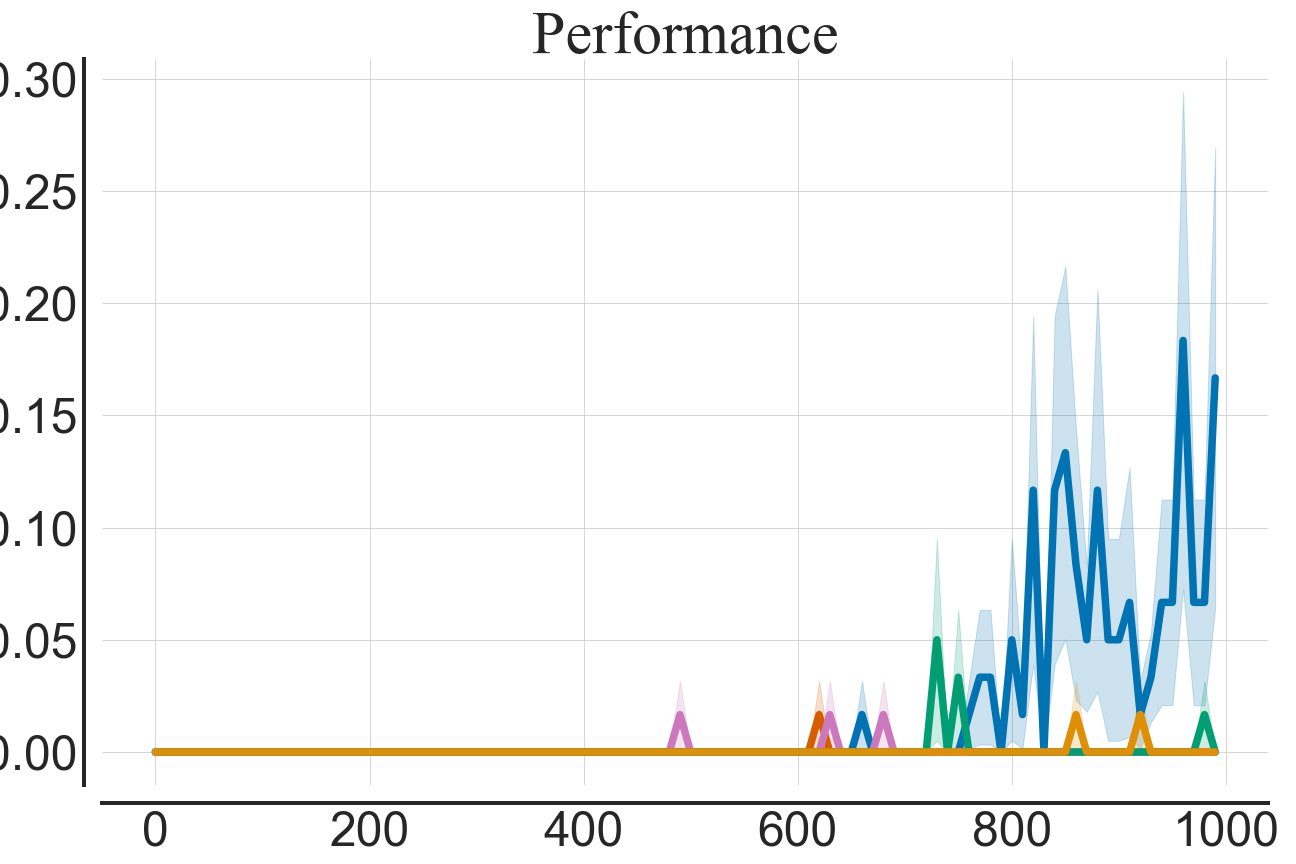}
    \hfill
    \includegraphics[width=0.23\linewidth]{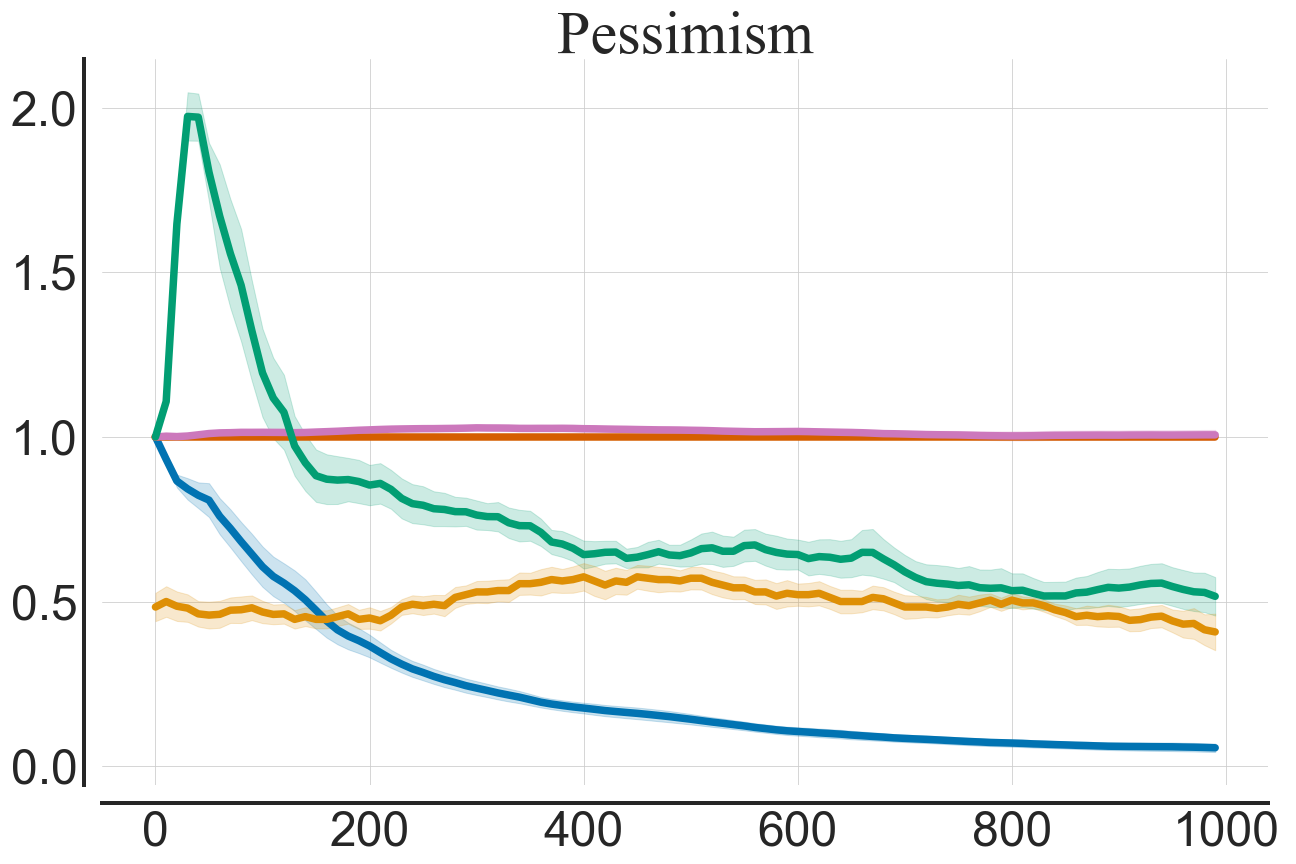}
    \hfill
    \includegraphics[width=0.23\linewidth]{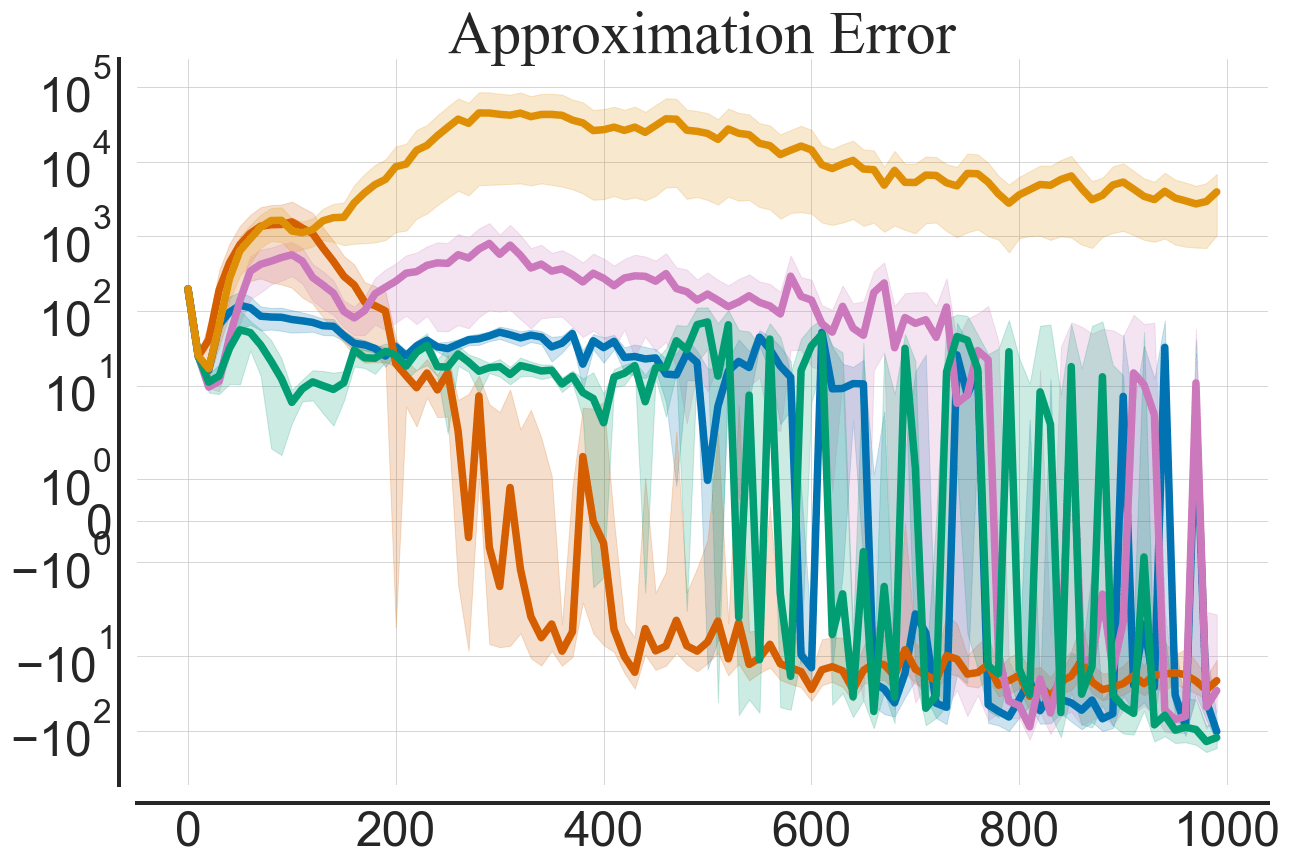}
    \hfill
    \includegraphics[width=0.23\linewidth]{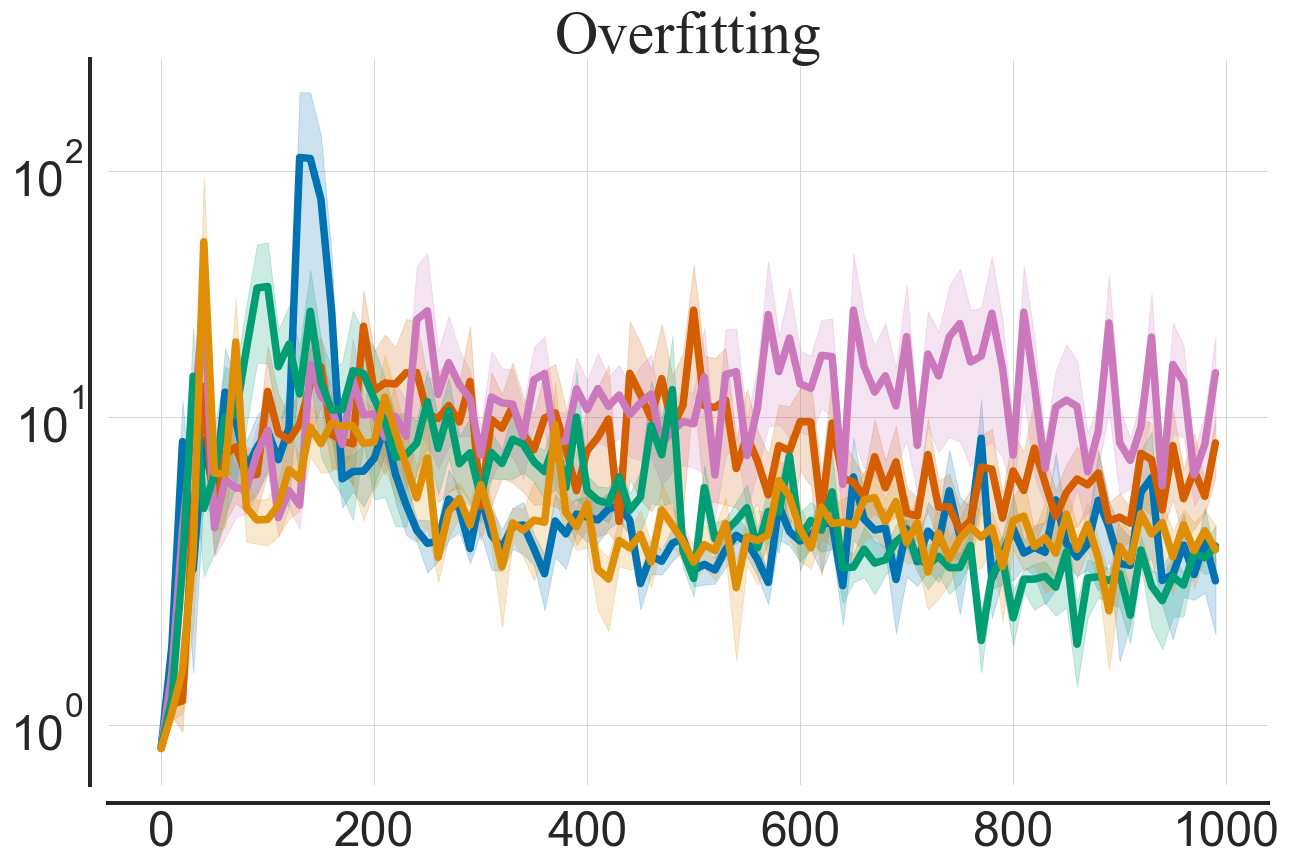}
    \hfill
    \end{subfigure}
    \subcaption{Stick Pull}
\end{minipage}
\bigskip
\begin{minipage}[h]{1.0\linewidth}
    \begin{subfigure}{1.0\linewidth}
    \hfill
    \includegraphics[width=0.23\linewidth]{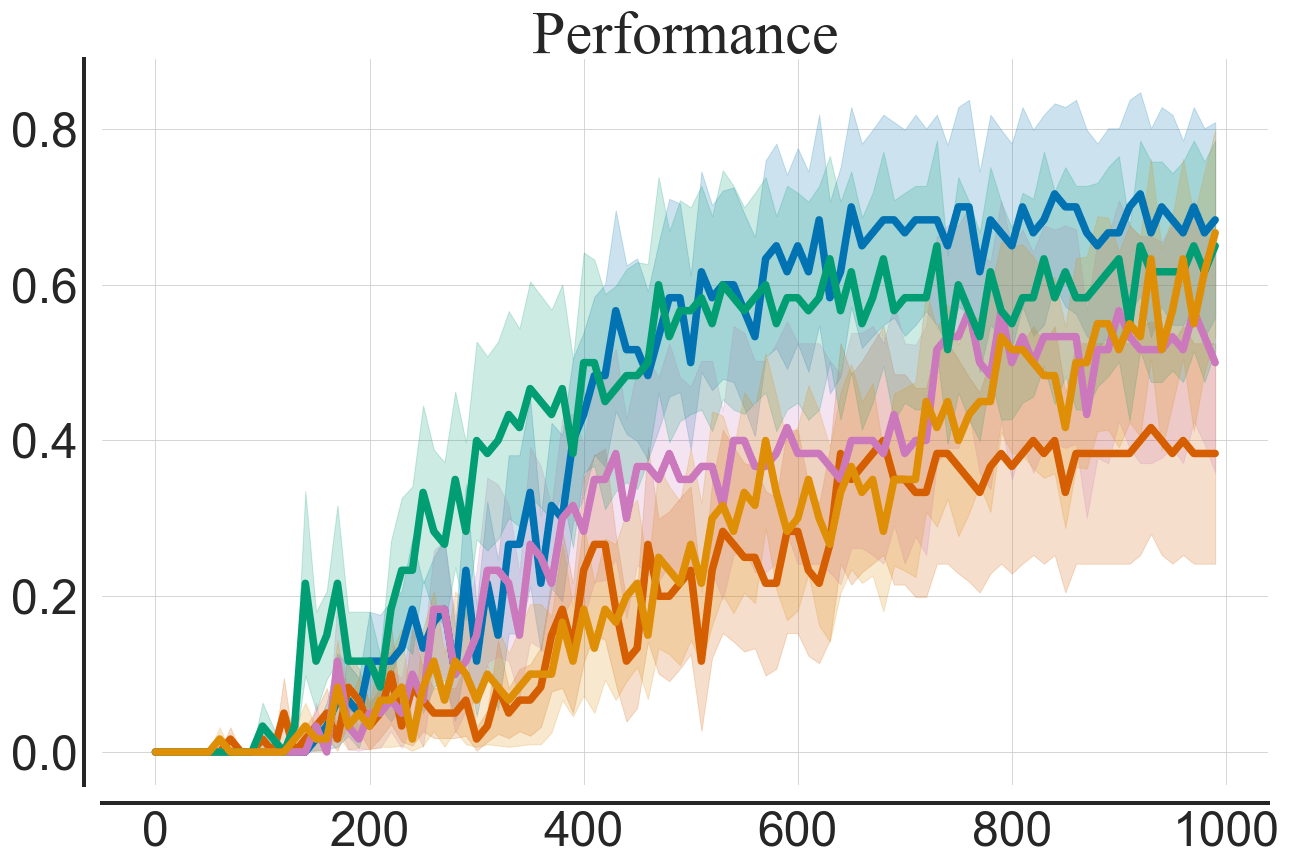}
    \hfill
    \includegraphics[width=0.23\linewidth]{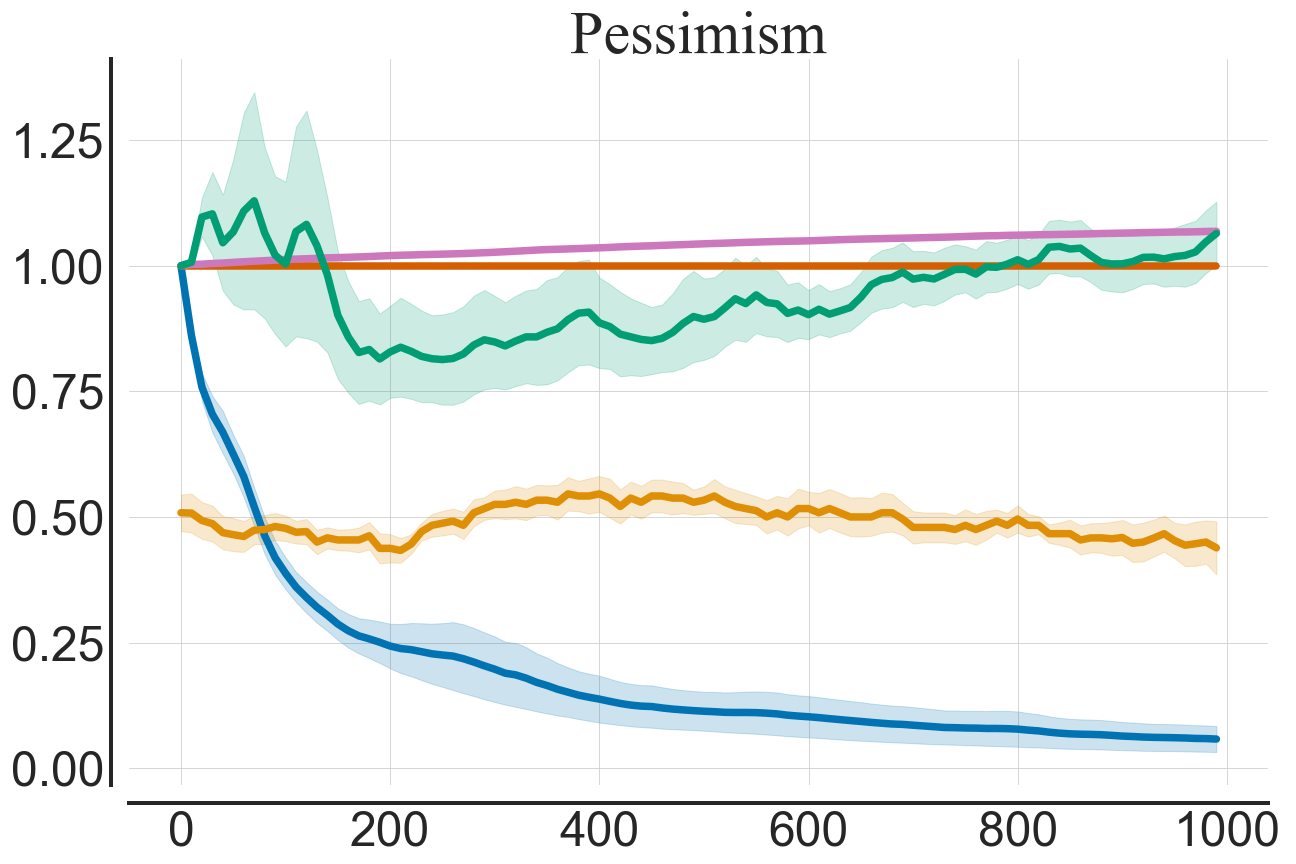}
    \hfill
    \includegraphics[width=0.23\linewidth]{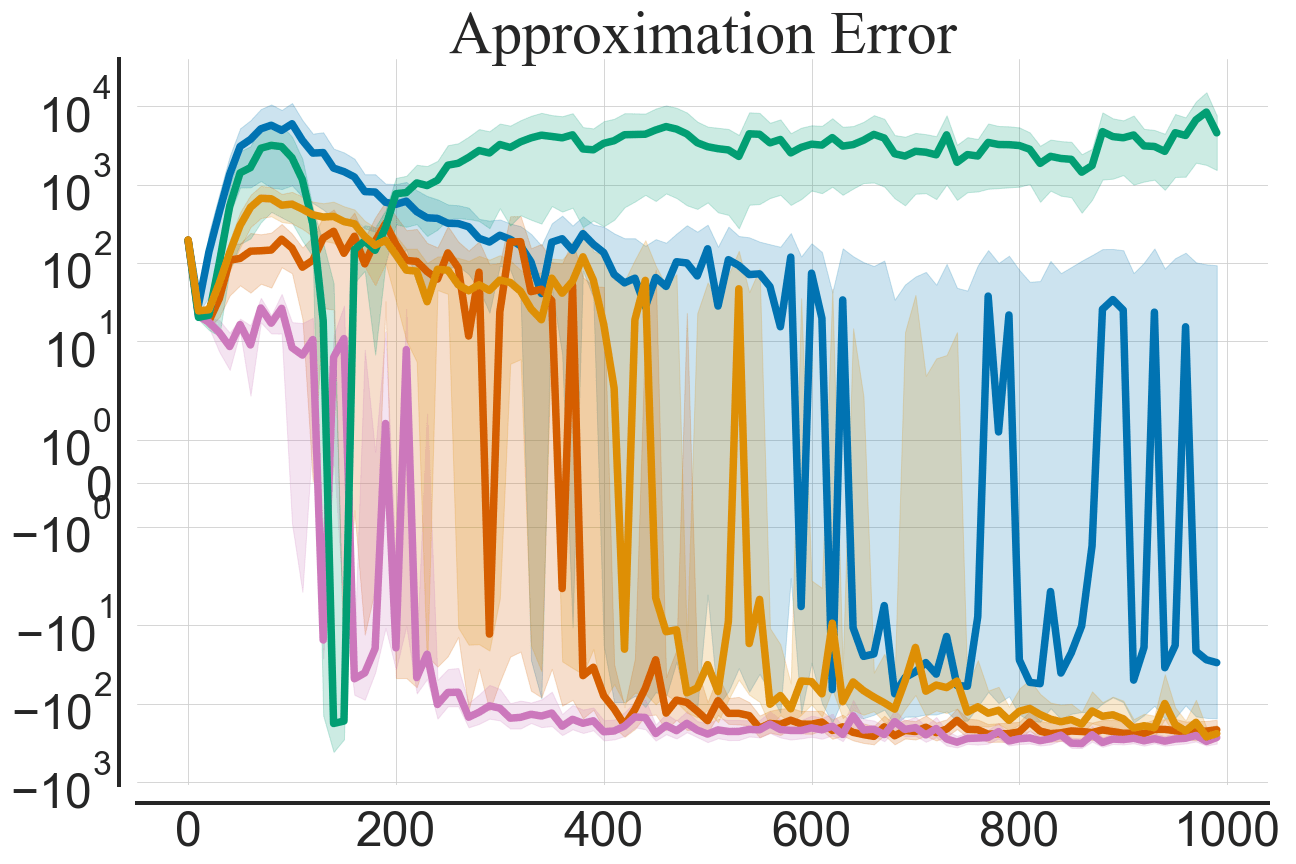}
    \hfill
    \includegraphics[width=0.23\linewidth]{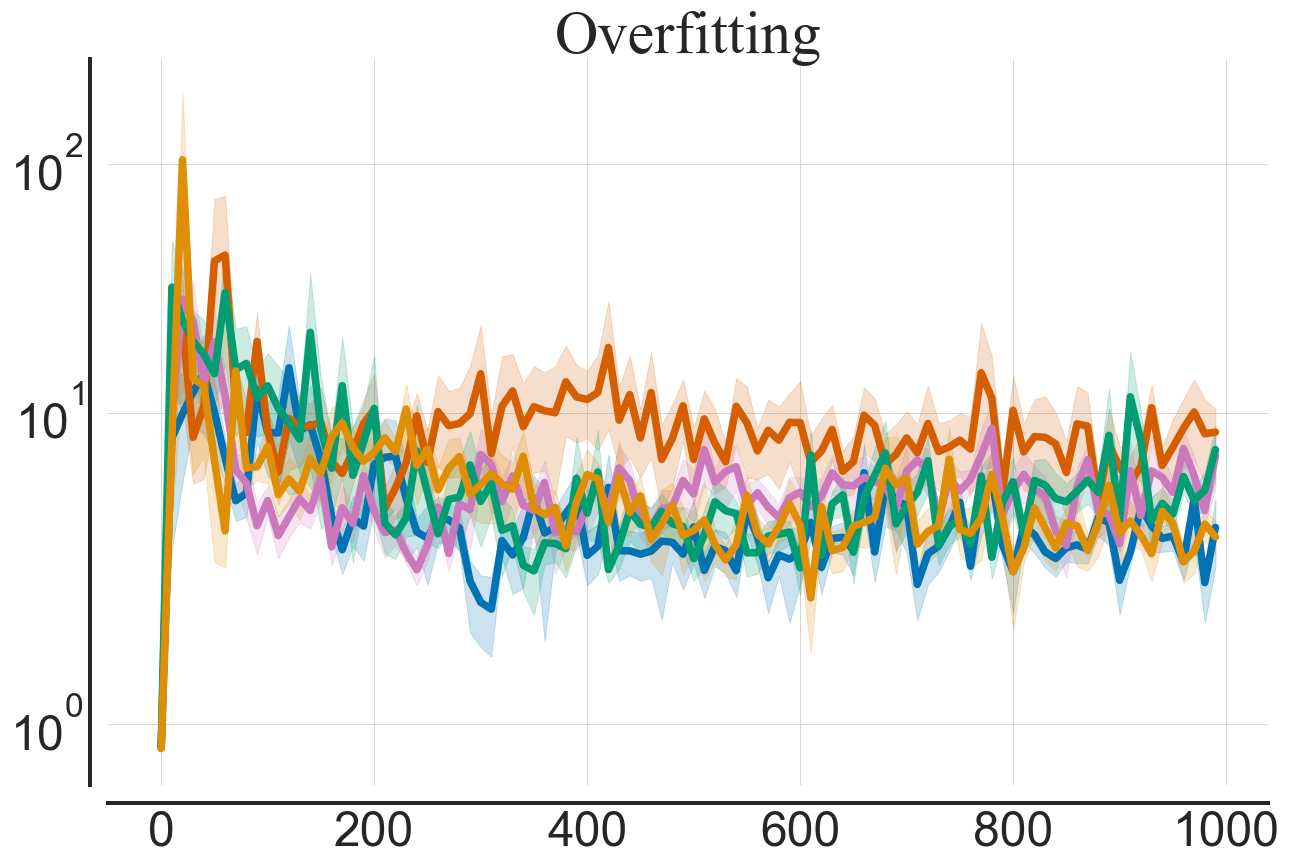}
    \hfill
    \end{subfigure}
    \subcaption{Sweep}
\end{minipage}
\caption{Low replay regime results for each considered task (4/4). 10 seeds per task, mean and 3 standard deviations.}
\label{fig:learning_curves4}
\end{center}
\end{figure*}

\begin{figure*}[ht!]
\begin{center}
\begin{minipage}[h]{1.0\linewidth}
\centering
    \begin{subfigure}{0.88\linewidth}
    \includegraphics[width=\textwidth]{images/legend_1.png}
    \end{subfigure}
\end{minipage}
\bigskip
\begin{minipage}[h]{1.0\linewidth}
    \begin{subfigure}{1.0\linewidth}
    \hfill
    \includegraphics[width=0.23\linewidth]{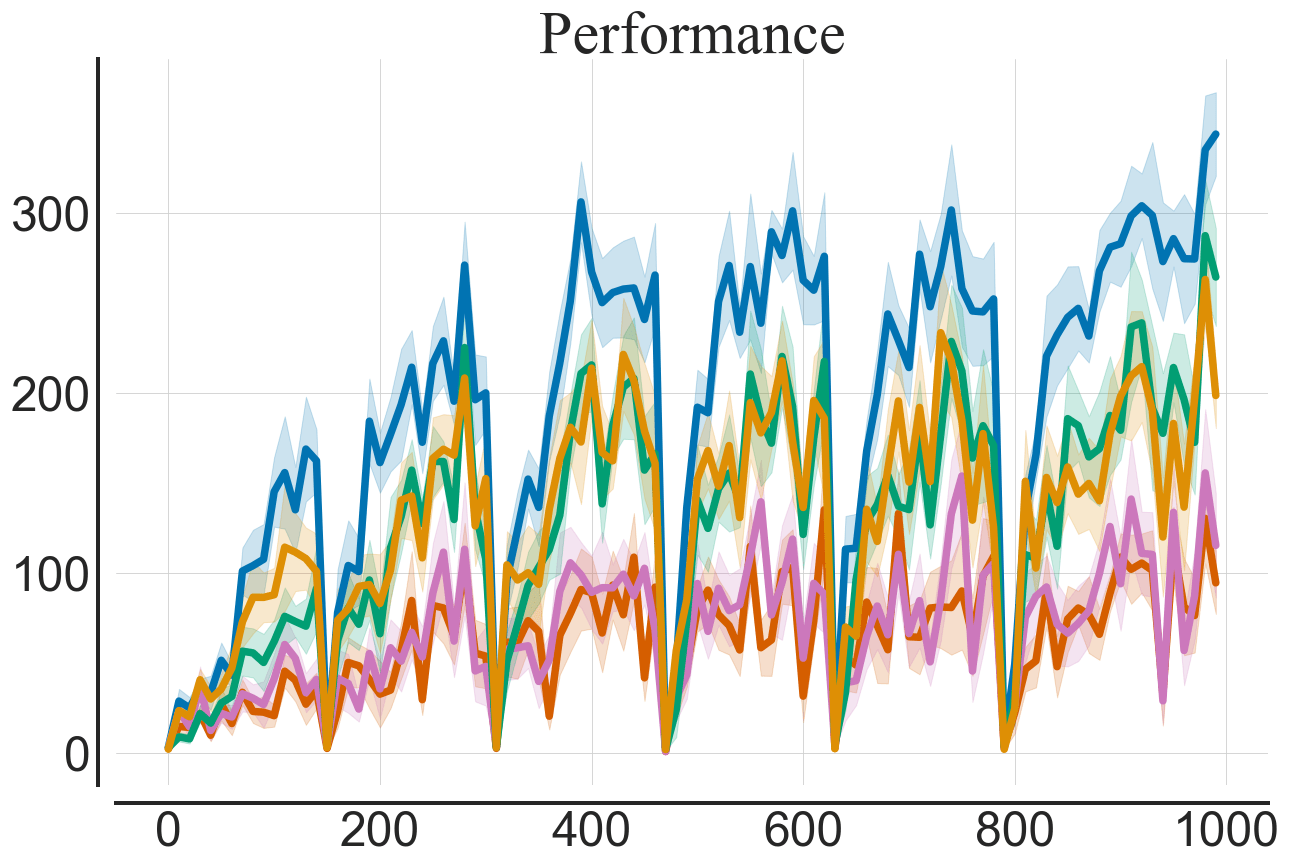}
    \hfill
    \includegraphics[width=0.23\linewidth]{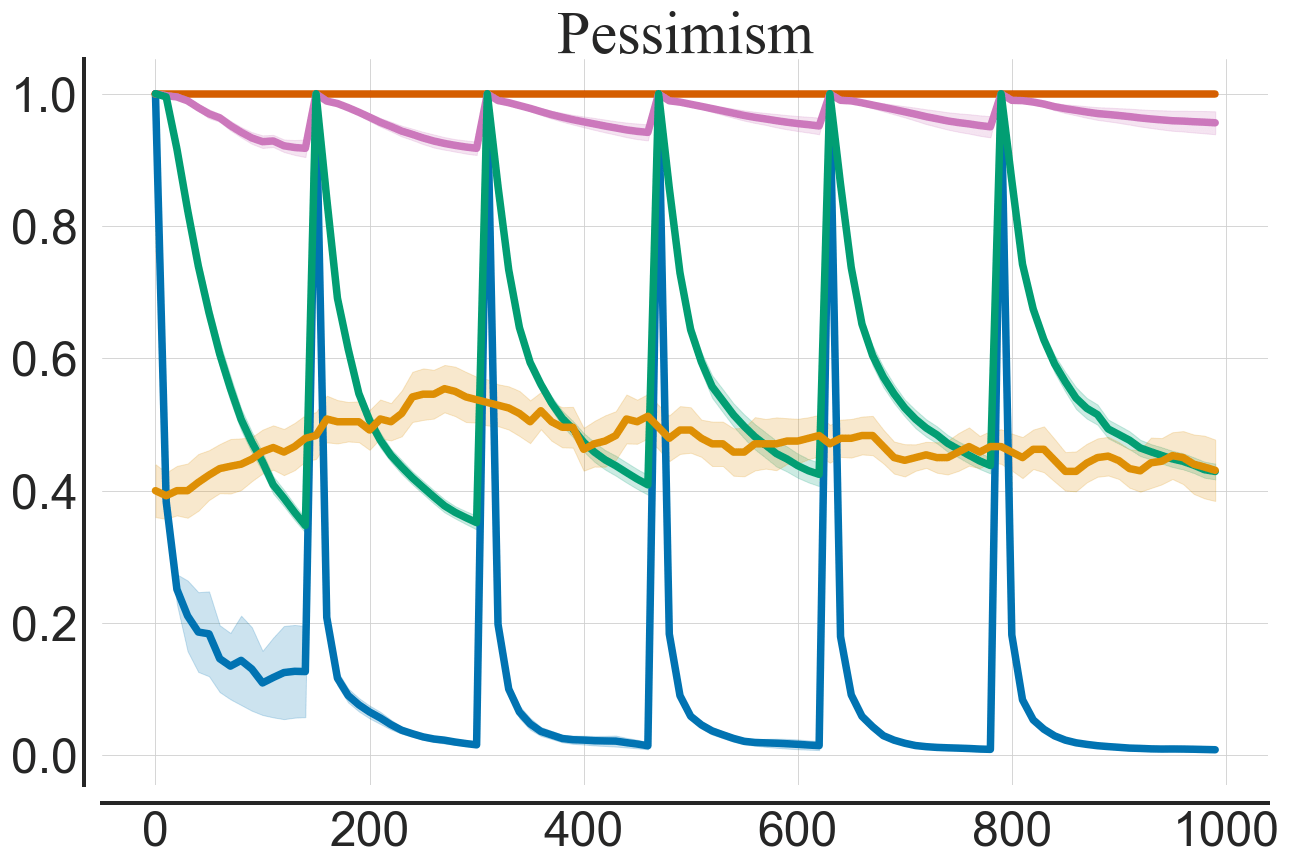}
    \hfill
    \includegraphics[width=0.23\linewidth]{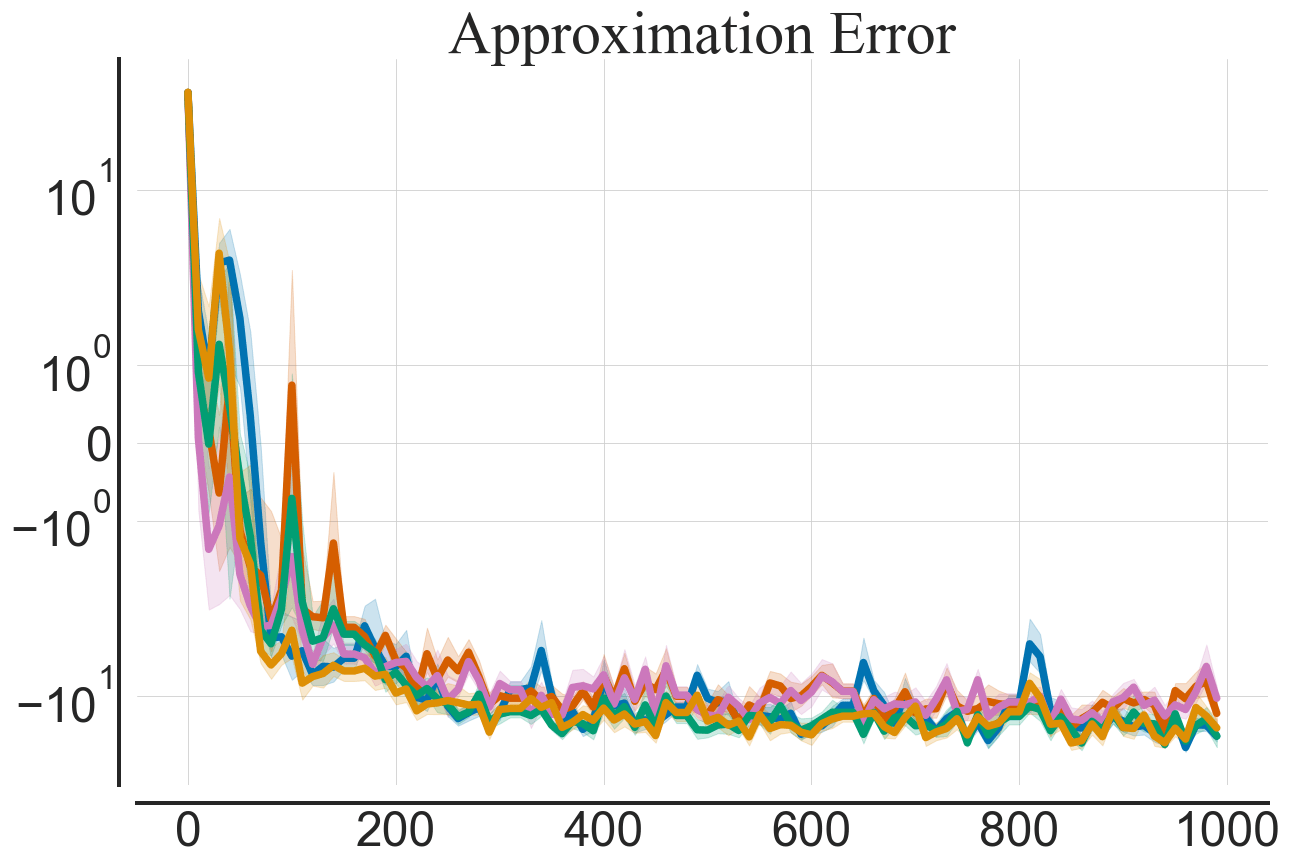}
    \hfill
    \includegraphics[width=0.23\linewidth]{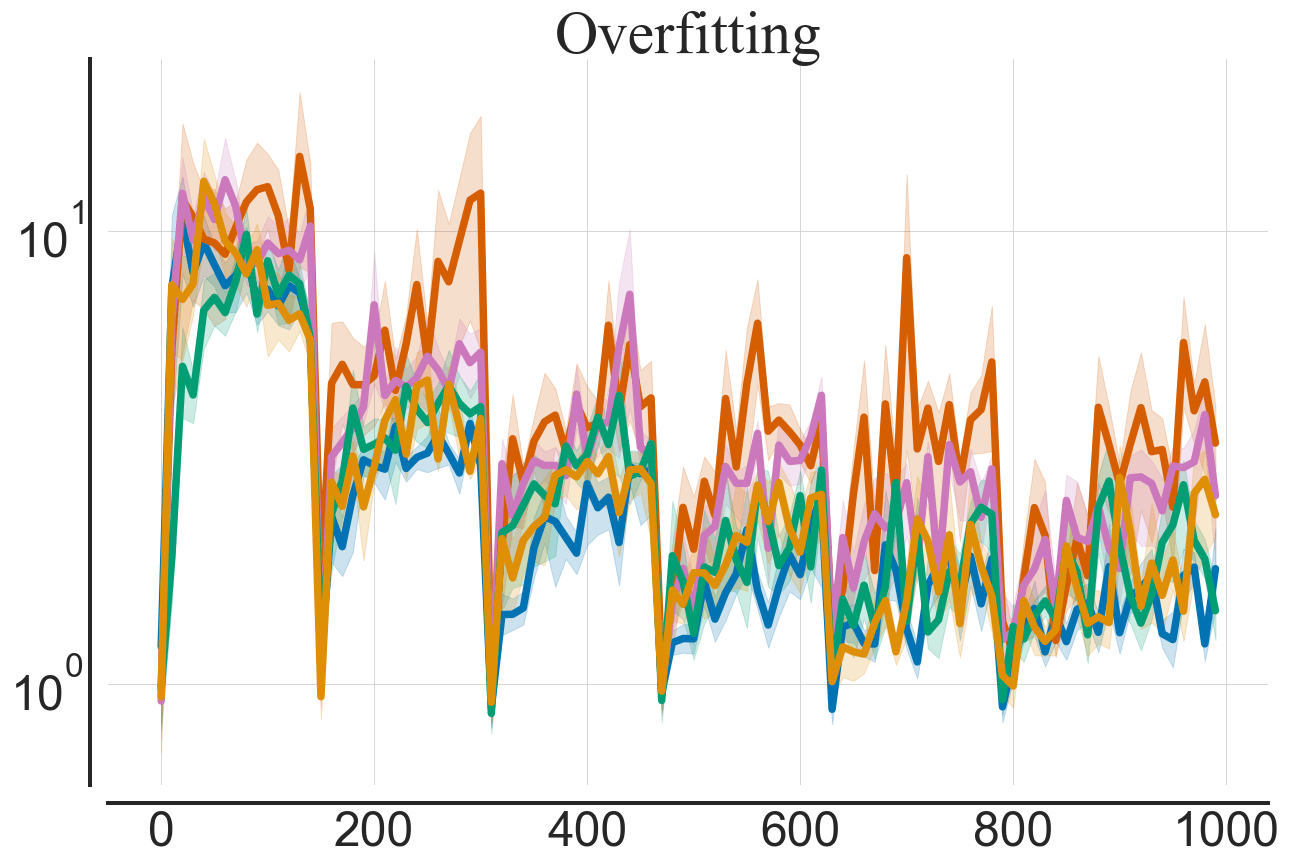}
    \hfill
    \end{subfigure}
    \subcaption{Acrobot Swingup}
\end{minipage}
\bigskip
\begin{minipage}[h]{1.0\linewidth}
    \begin{subfigure}{1.0\linewidth}
    \hfill
    \includegraphics[width=0.23\linewidth]{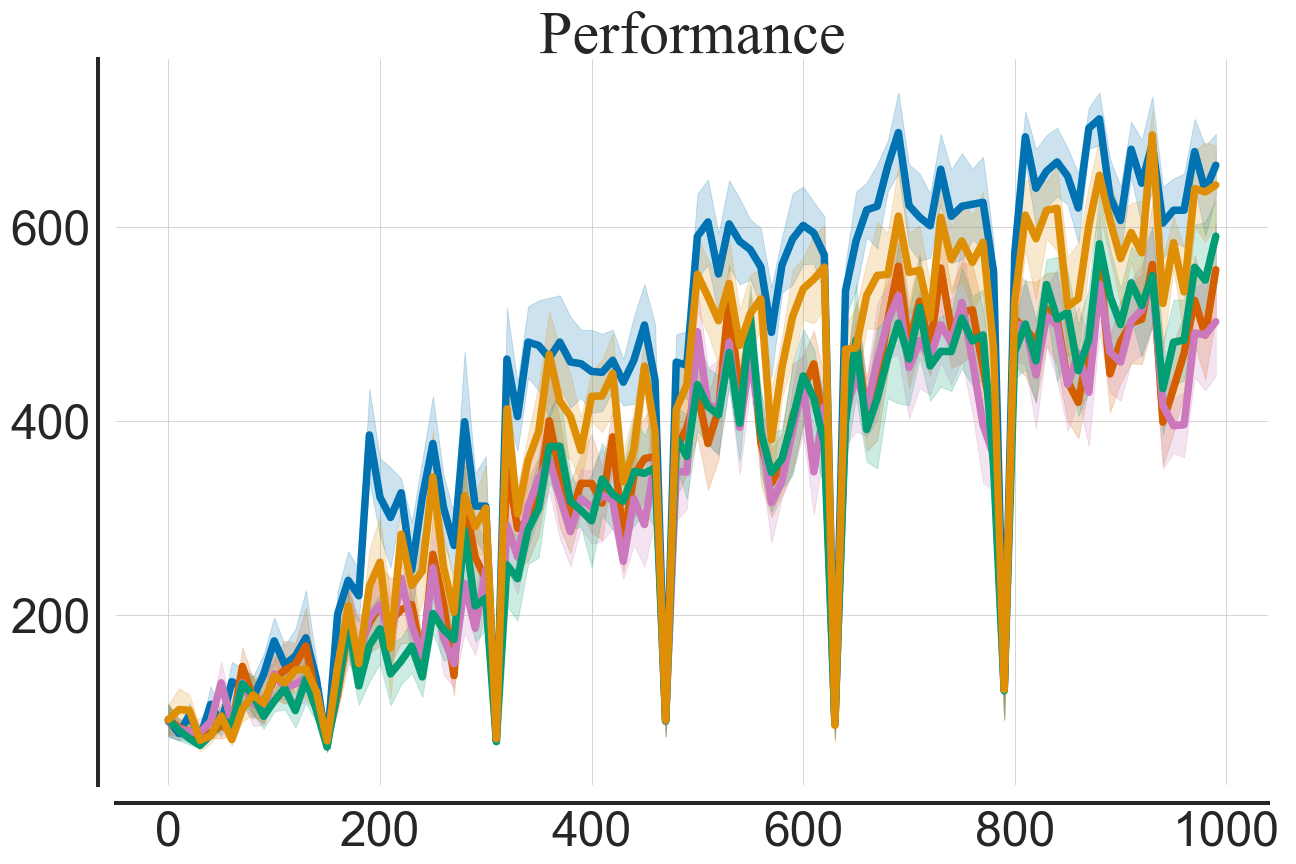}
    \hfill
    \includegraphics[width=0.23\linewidth]{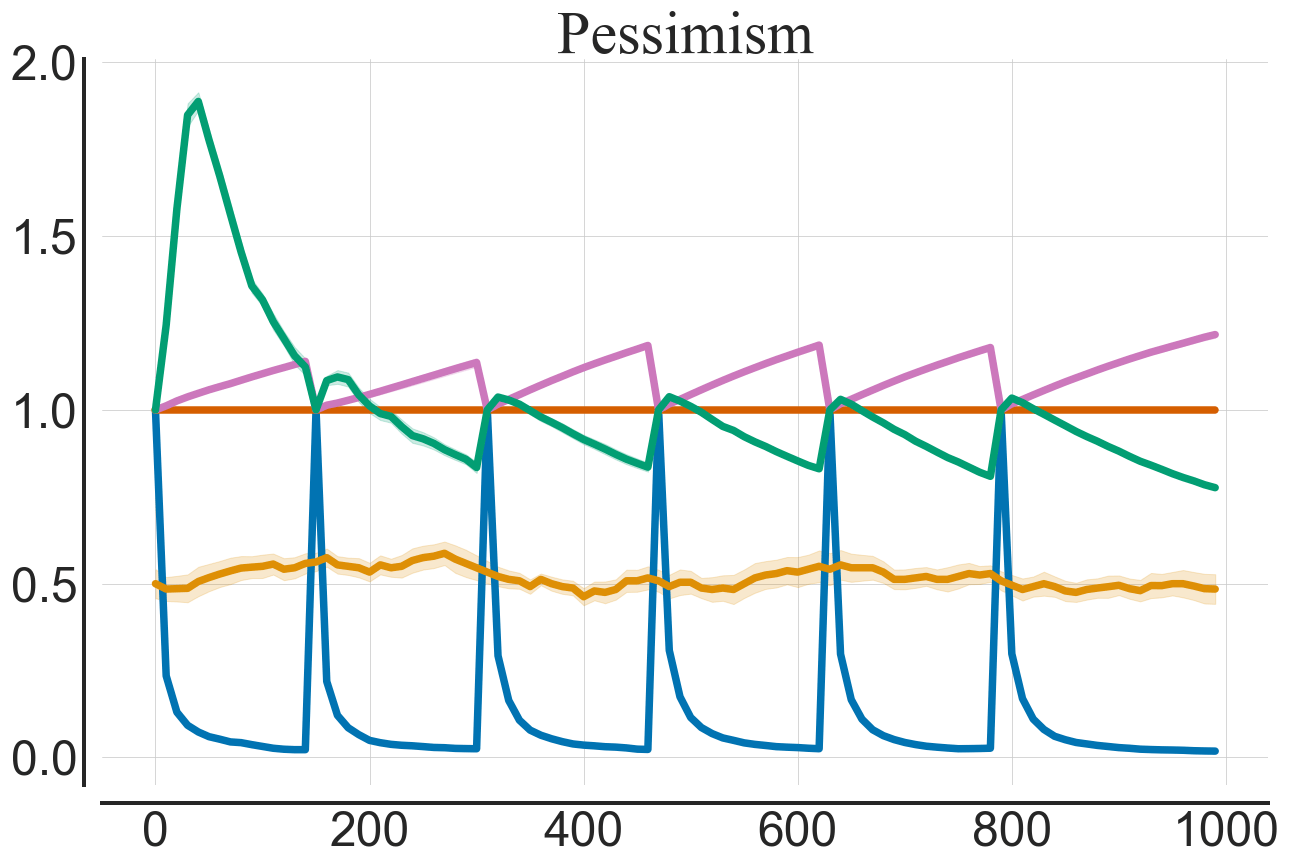}
    \hfill
    \includegraphics[width=0.23\linewidth]{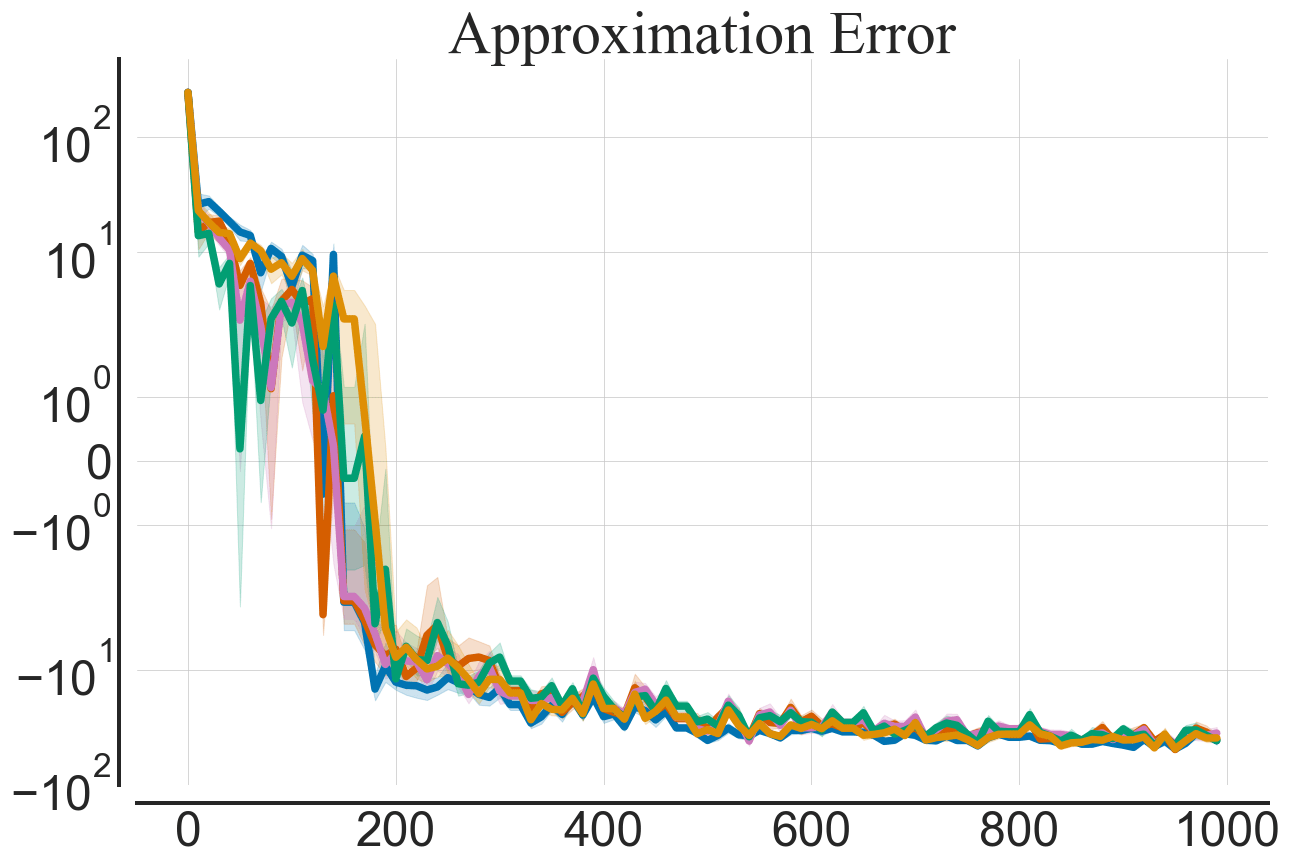}
    \hfill
    \includegraphics[width=0.23\linewidth]{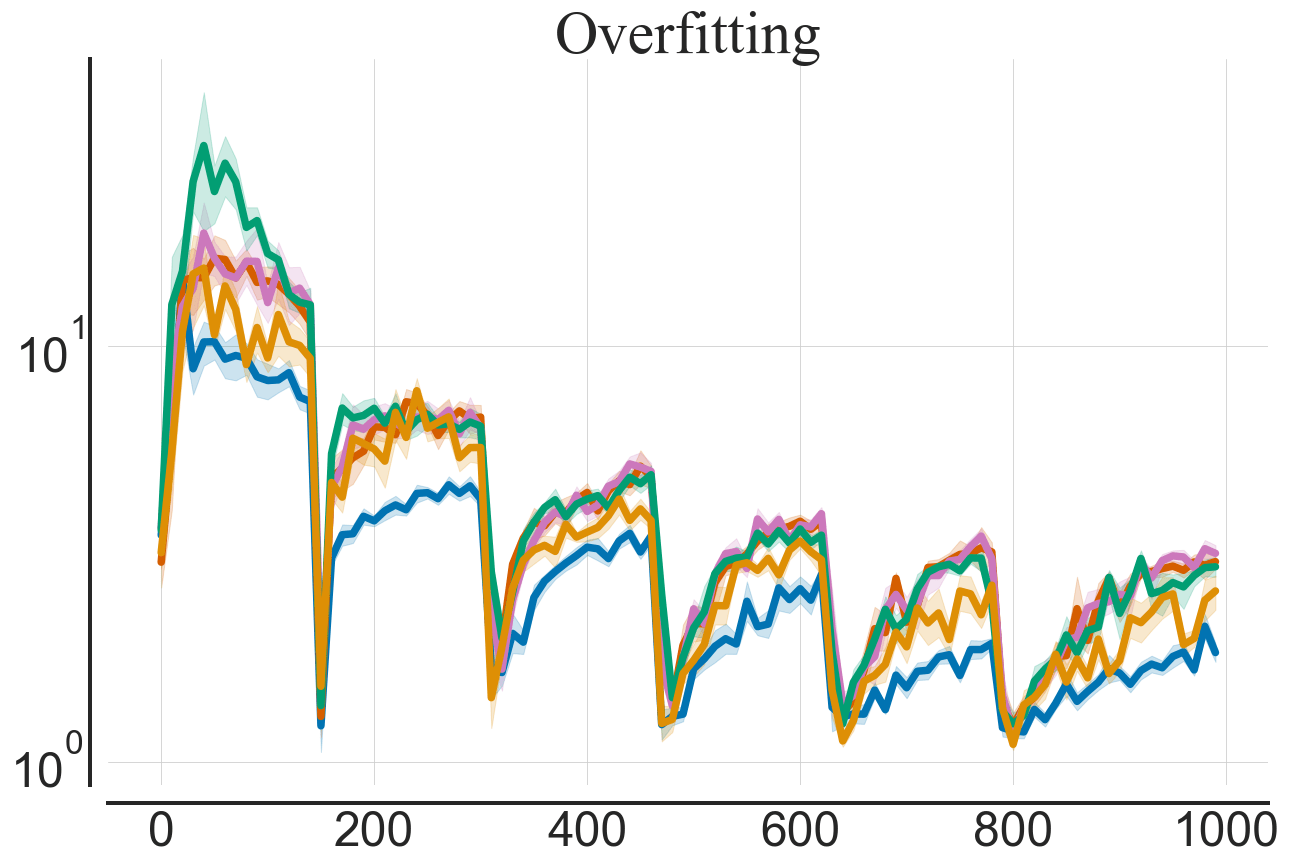}
    \hfill
    \end{subfigure}
    \subcaption{Fish Swim}
\end{minipage}
\bigskip
\begin{minipage}[h]{1.0\linewidth}
    \begin{subfigure}{1.0\linewidth}
    \hfill
    \includegraphics[width=0.23\linewidth]{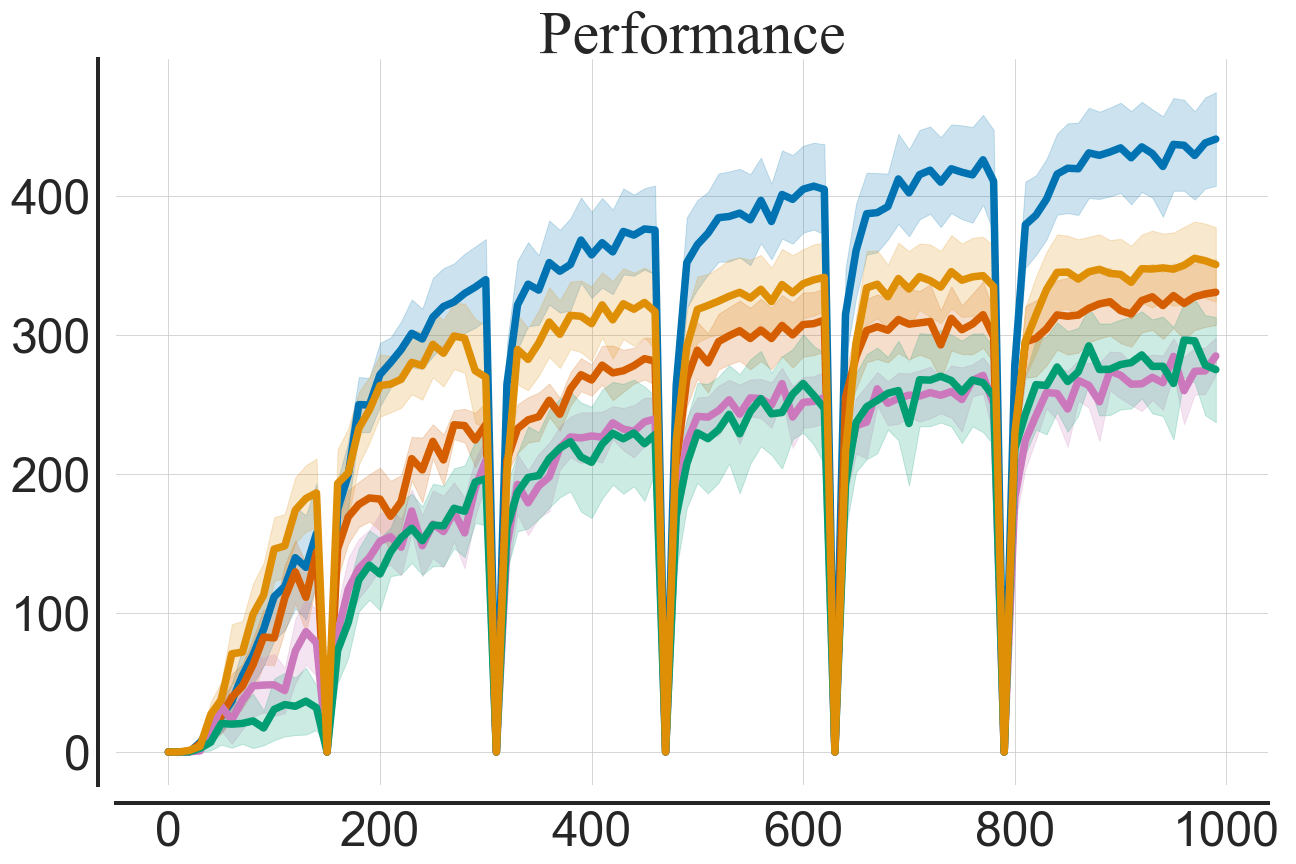}
    \hfill
    \includegraphics[width=0.23\linewidth]{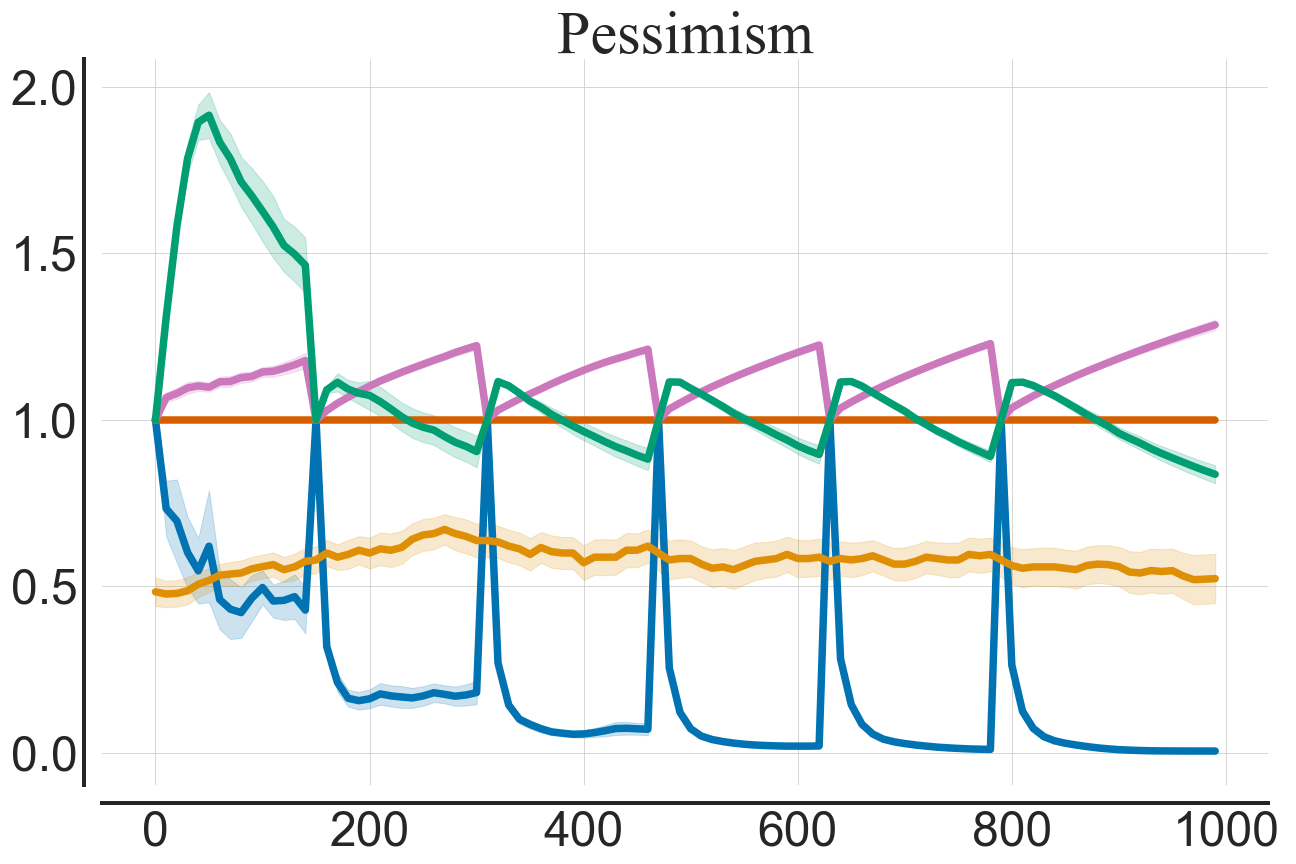}
    \hfill
    \includegraphics[width=0.23\linewidth]{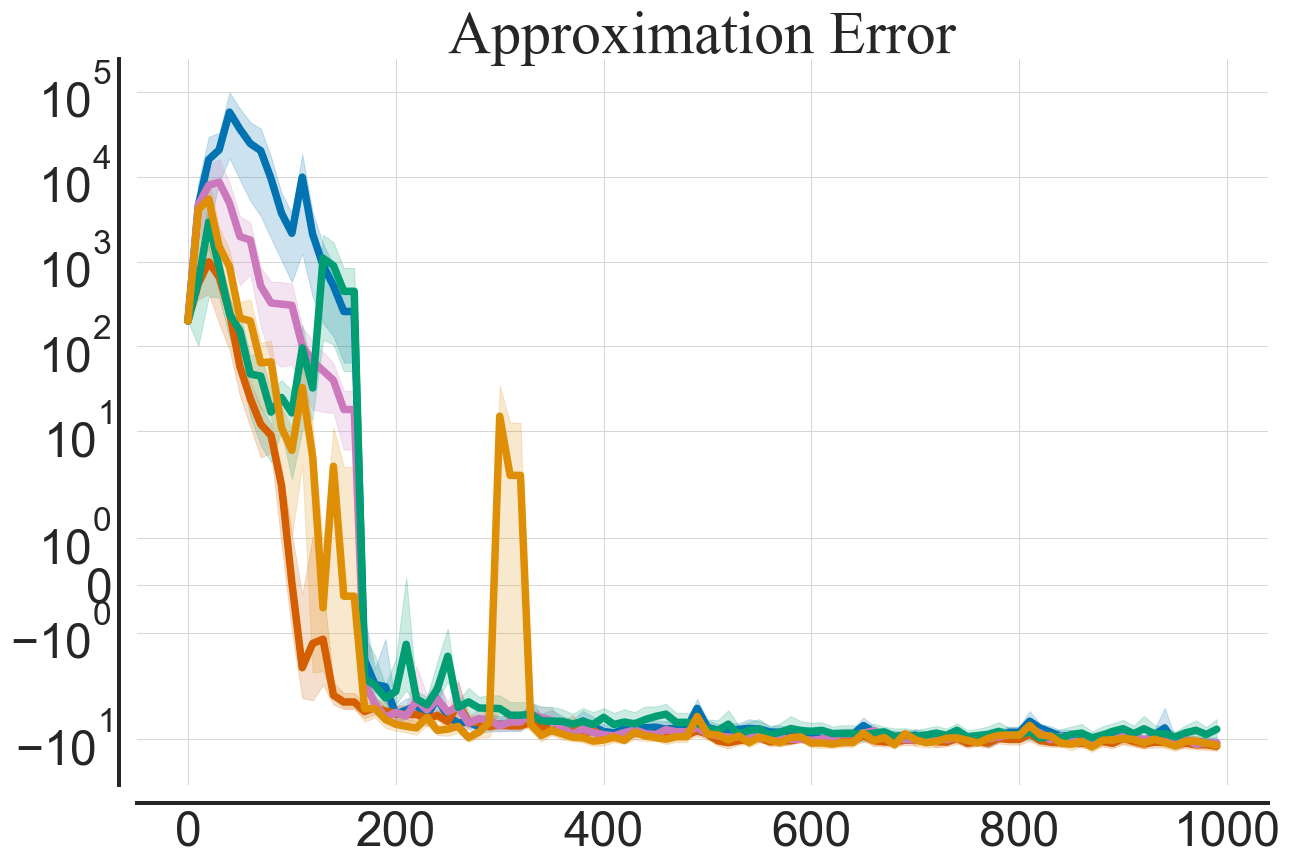}
    \hfill
    \includegraphics[width=0.23\linewidth]{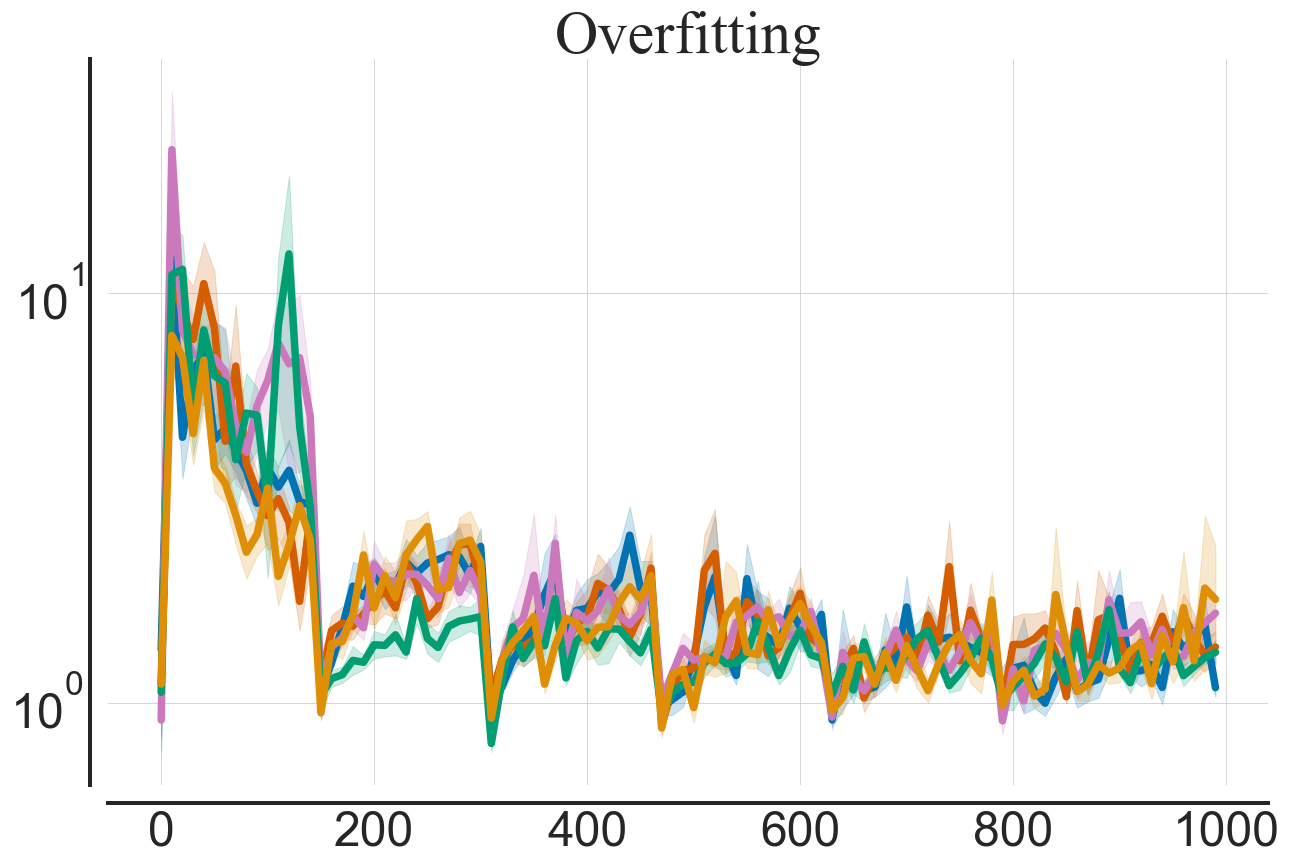}
    \hfill
    \end{subfigure}
    \subcaption{Hopper Hop}
\end{minipage}
\bigskip
\begin{minipage}[h]{1.0\linewidth}
    \begin{subfigure}{1.0\linewidth}
    \hfill
    \includegraphics[width=0.23\linewidth]{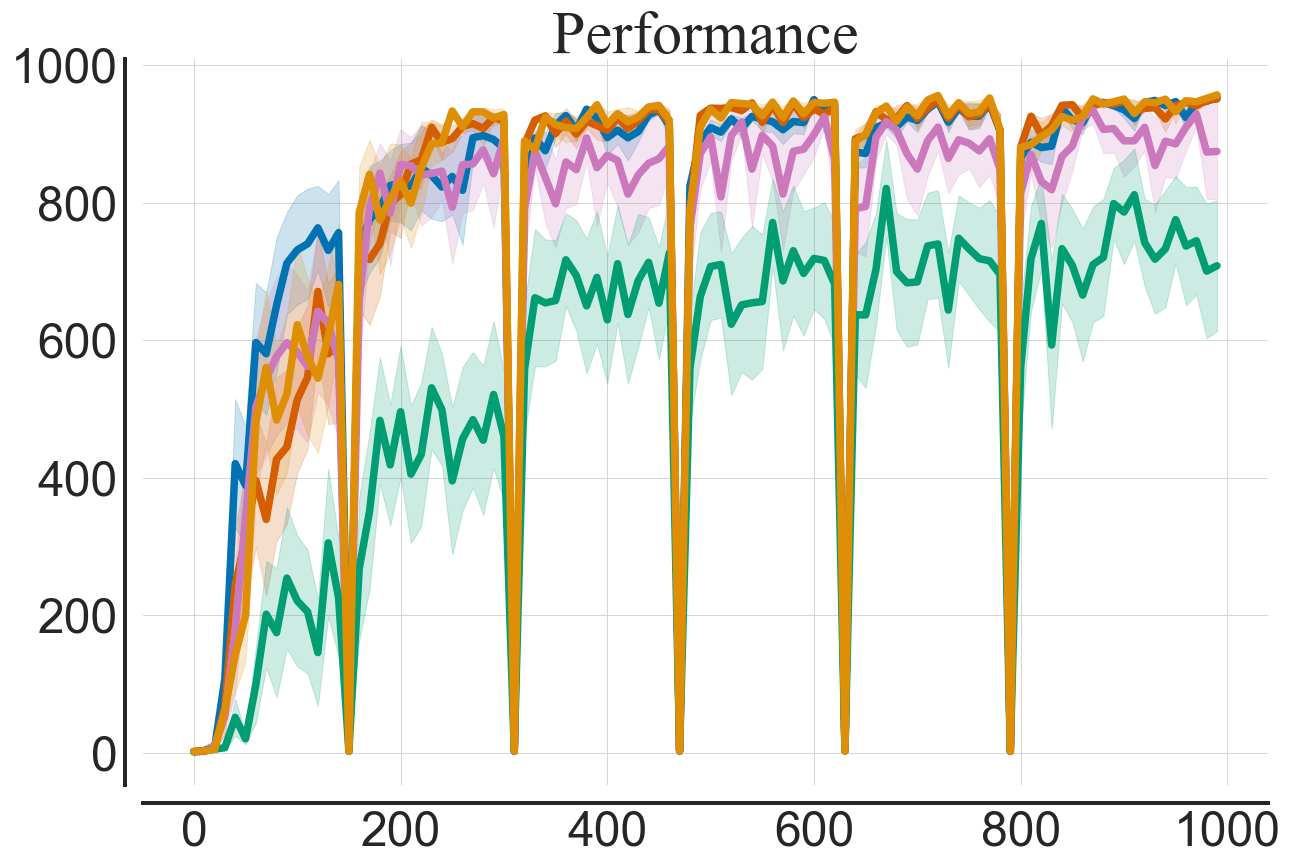}
    \hfill
    \includegraphics[width=0.23\linewidth]{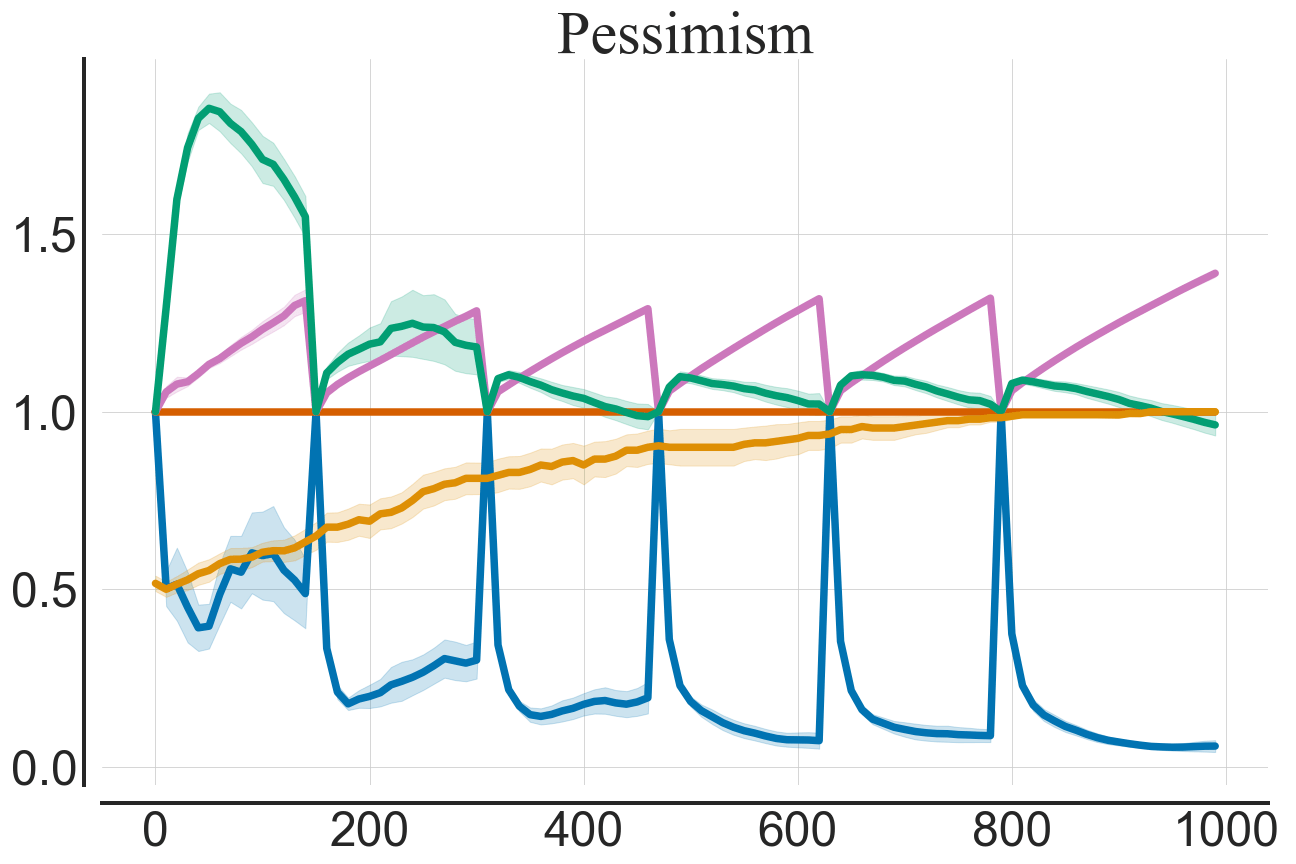}
    \hfill
    \includegraphics[width=0.23\linewidth]{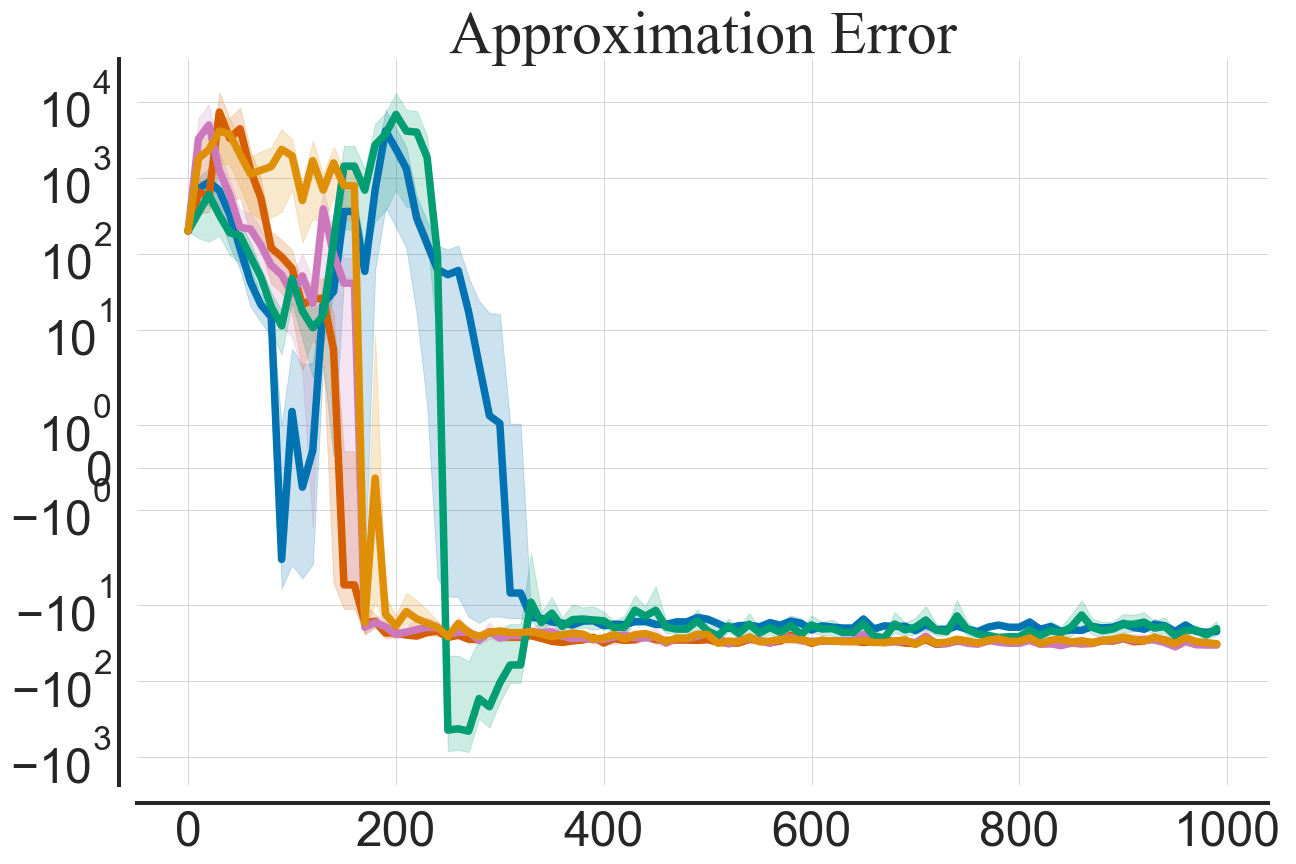}
    \hfill
    \includegraphics[width=0.23\linewidth]{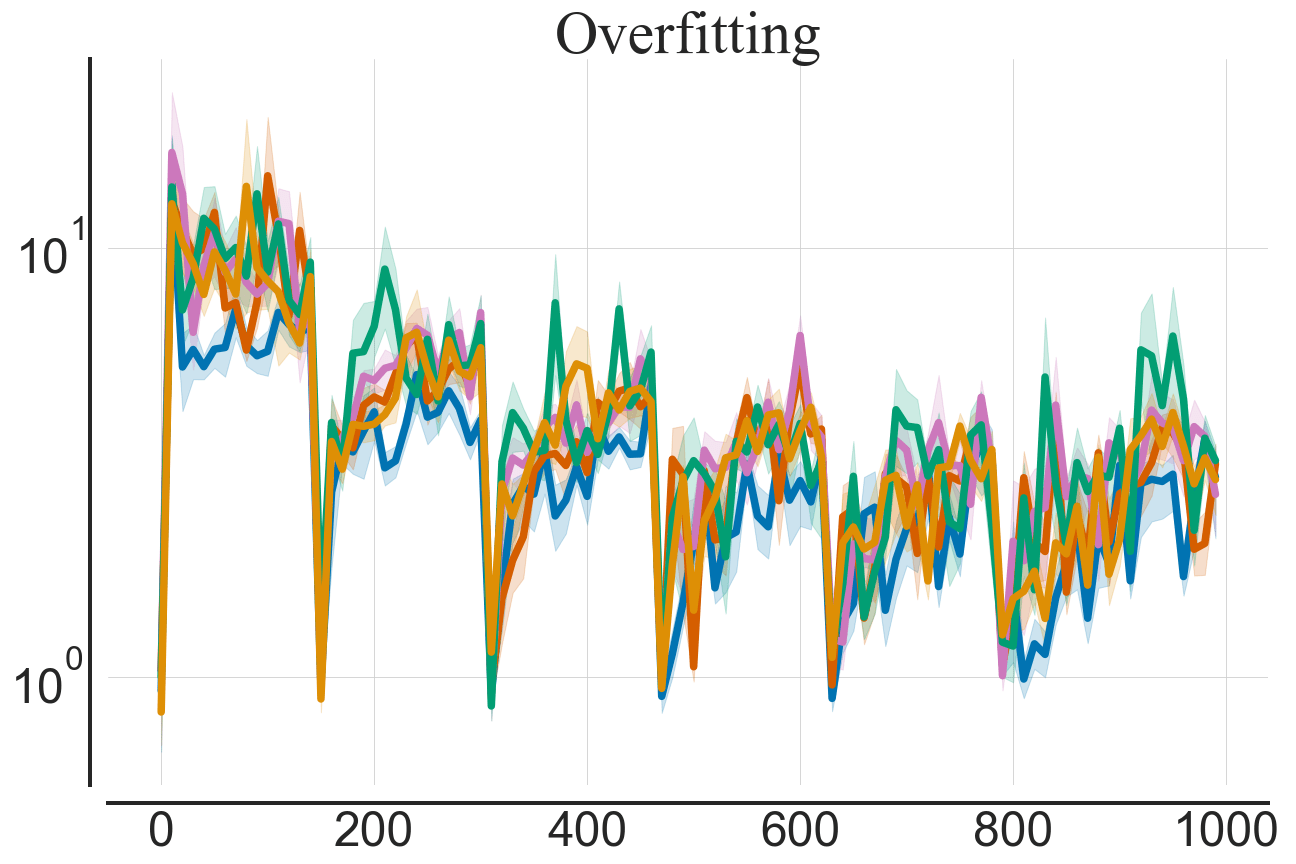}
    \hfill
    \end{subfigure}
    \subcaption{Hopper Stand}
\end{minipage}
\bigskip
\begin{minipage}[h]{1.0\linewidth}
    \begin{subfigure}{1.0\linewidth}
    \hfill
    \includegraphics[width=0.23\linewidth]{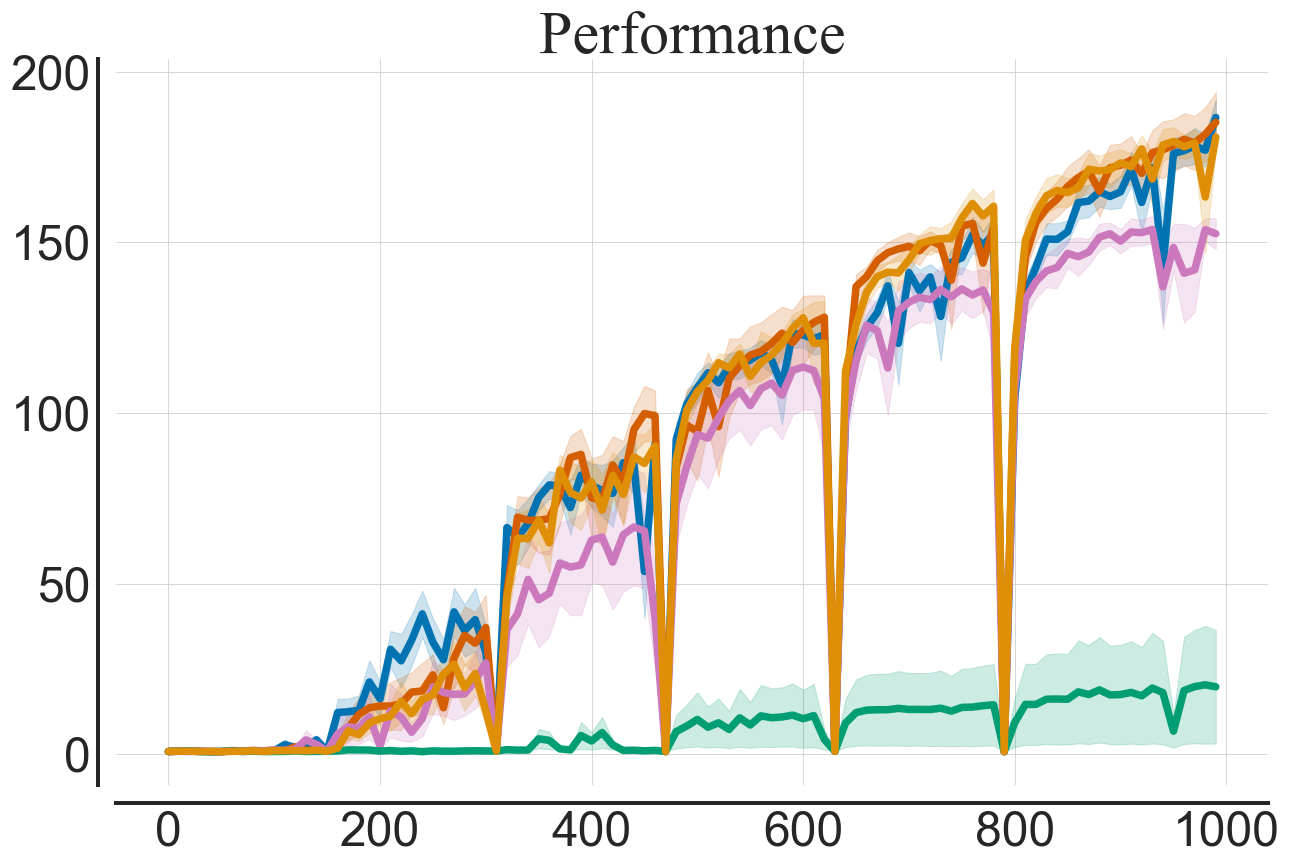}
    \hfill
    \includegraphics[width=0.23\linewidth]{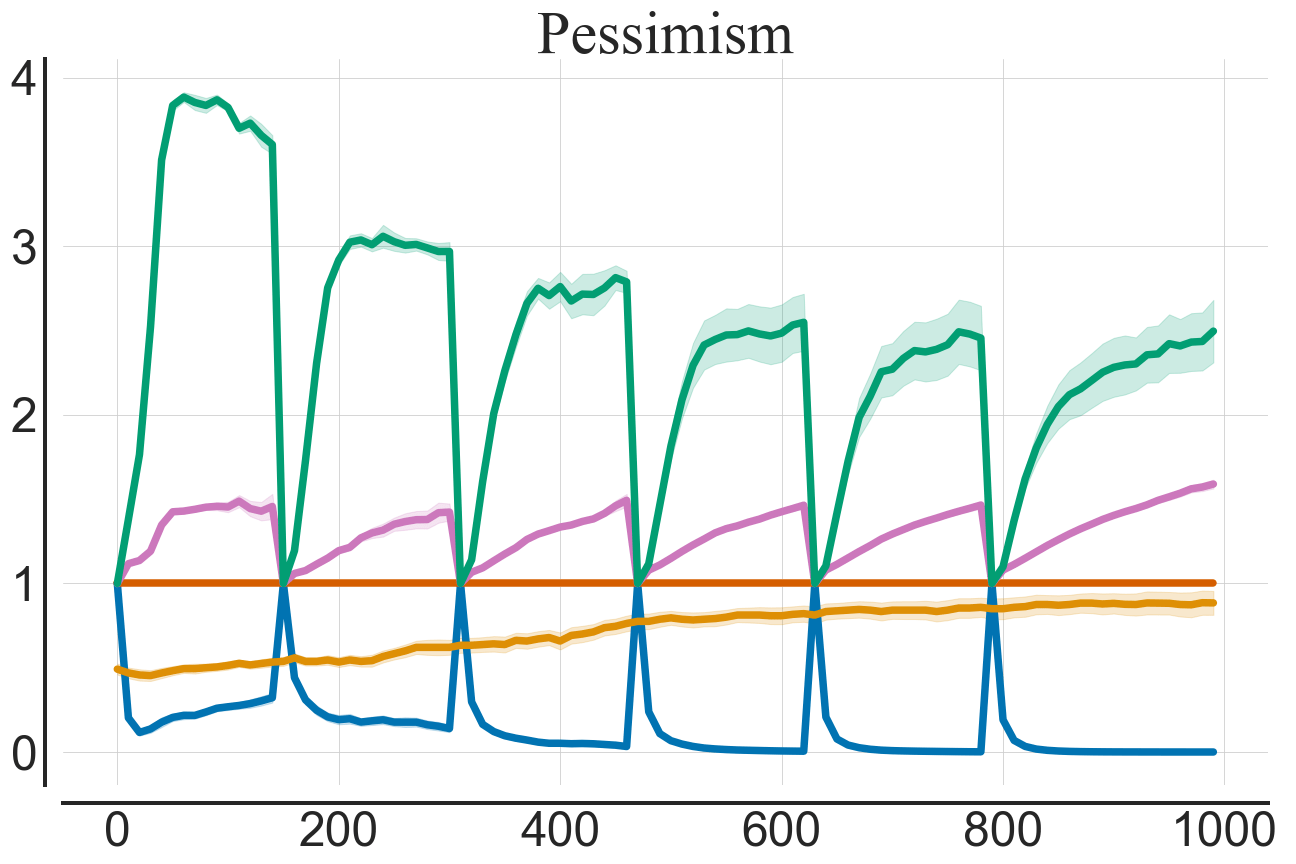}
    \hfill
    \includegraphics[width=0.23\linewidth]{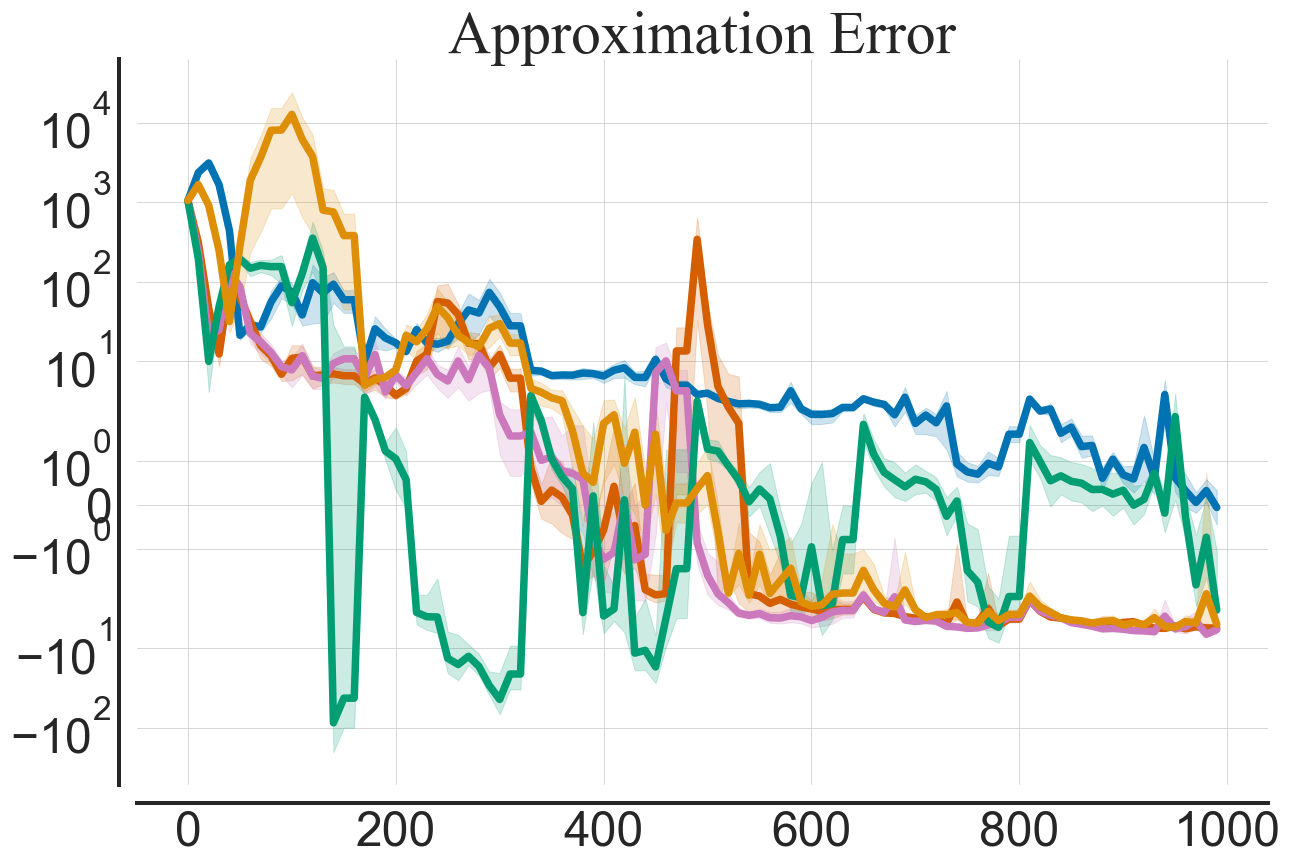}
    \hfill
    \includegraphics[width=0.23\linewidth]{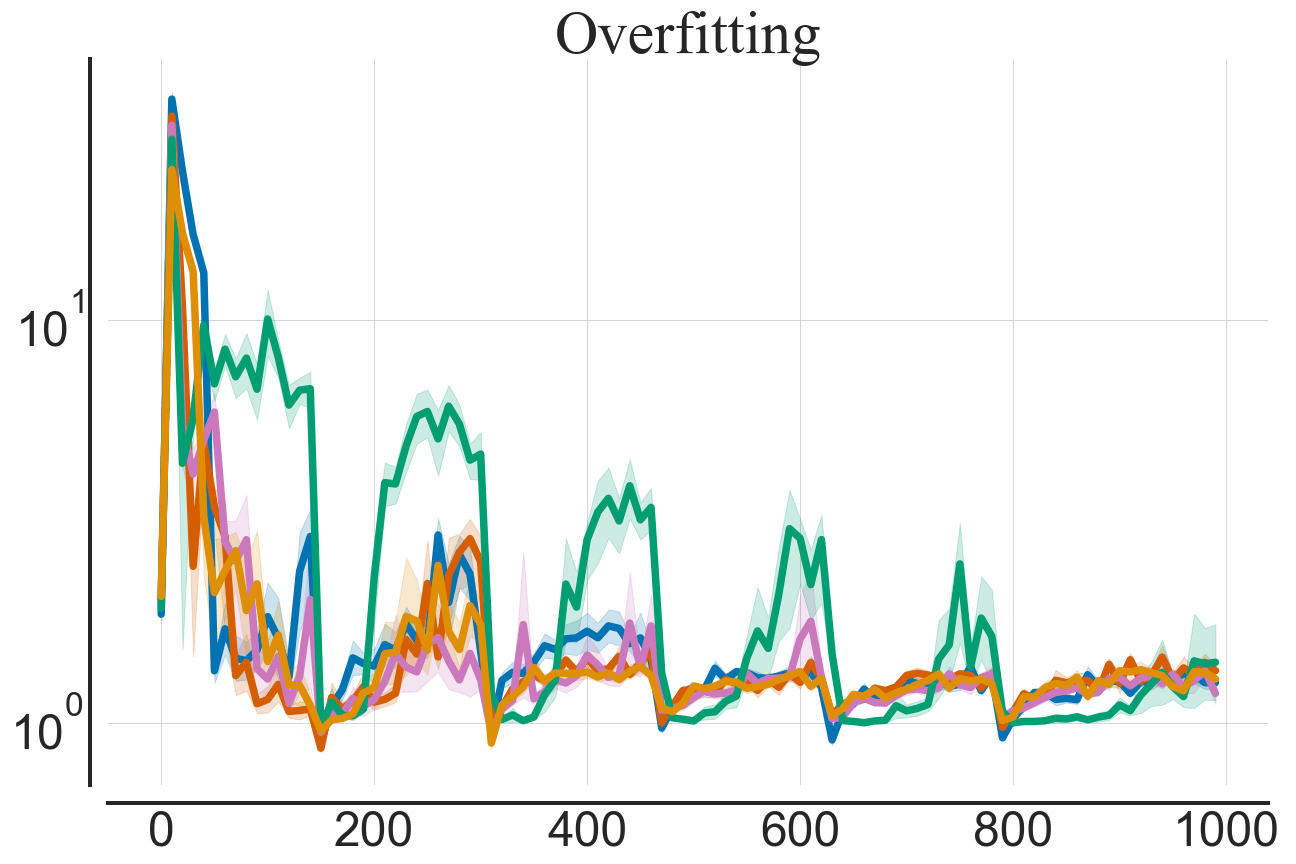}
    \hfill
    \end{subfigure}
    \subcaption{Humanoid Run}
\end{minipage}
\caption{High replay regime results for each considered task (1/4). 10 seeds per task, mean and 3 standard deviations.}
\label{fig:learning_curves5}
\end{center}
\end{figure*}

\begin{figure*}[ht!]
\begin{center}
\begin{minipage}[h]{1.0\linewidth}
\centering
    \begin{subfigure}{0.88\linewidth}
    \includegraphics[width=\textwidth]{images/legend_1.png}
    \end{subfigure}
\end{minipage}
\bigskip
\begin{minipage}[h]{1.0\linewidth}
    \begin{subfigure}{1.0\linewidth}
    \hfill
    \includegraphics[width=0.23\linewidth]{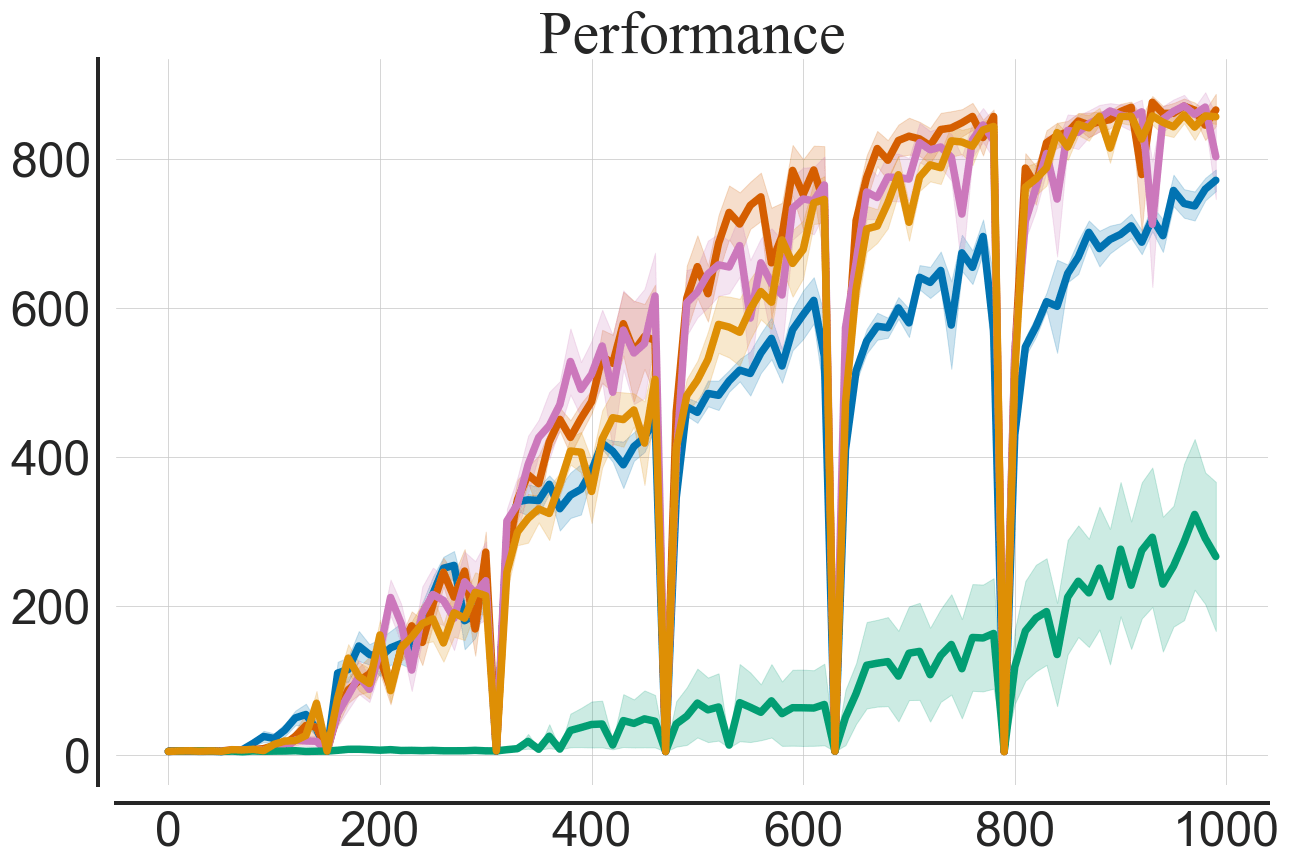}
    \hfill
    \includegraphics[width=0.23\linewidth]{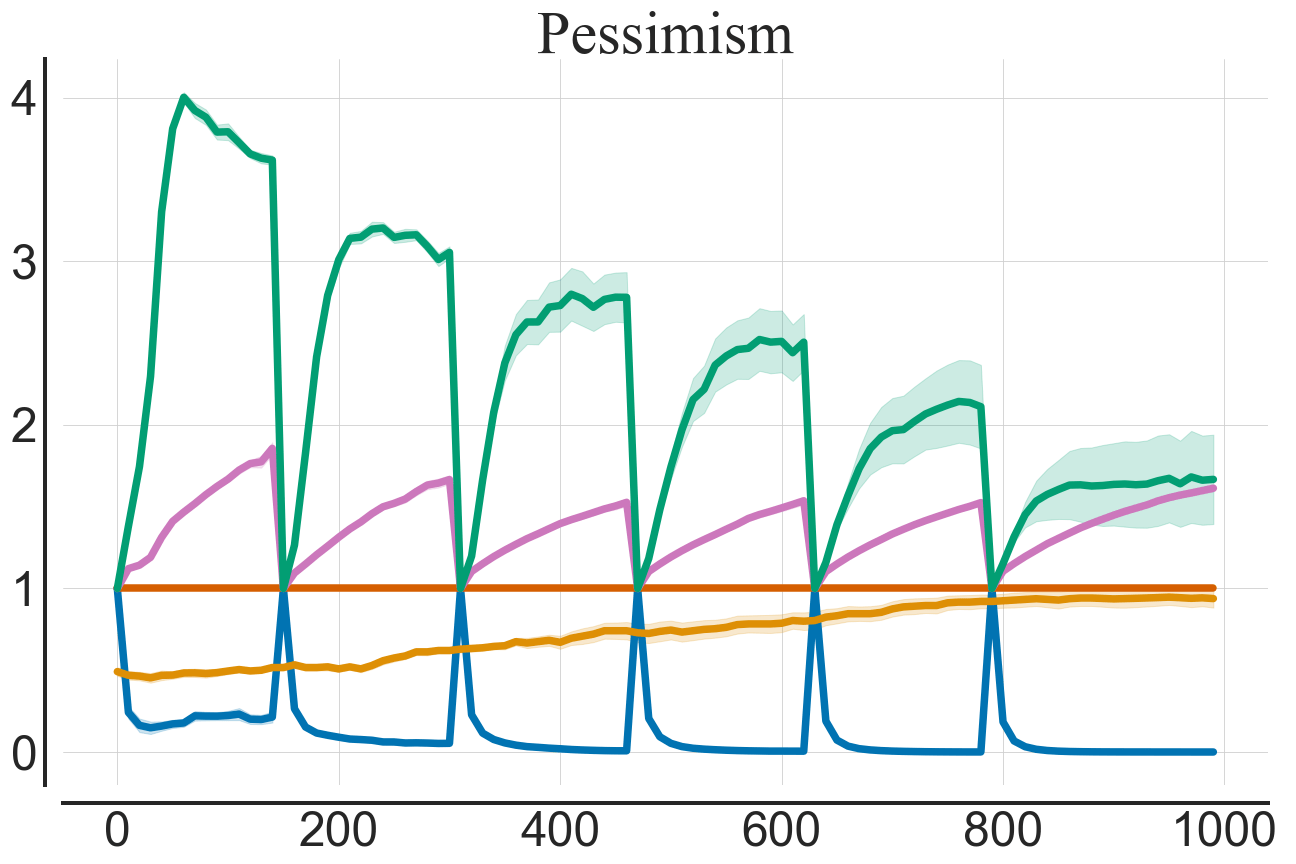}
    \hfill
    \includegraphics[width=0.23\linewidth]{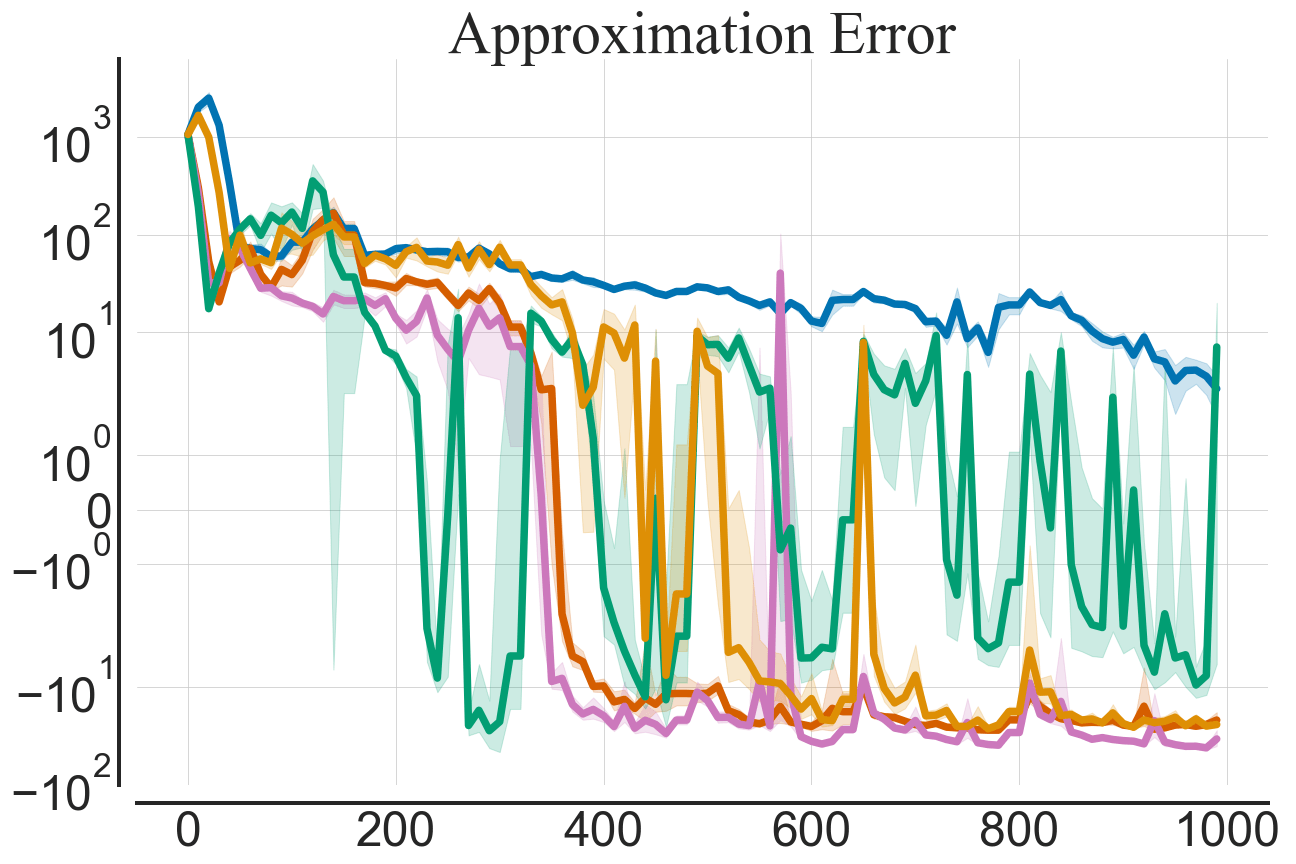}
    \hfill
    \includegraphics[width=0.23\linewidth]{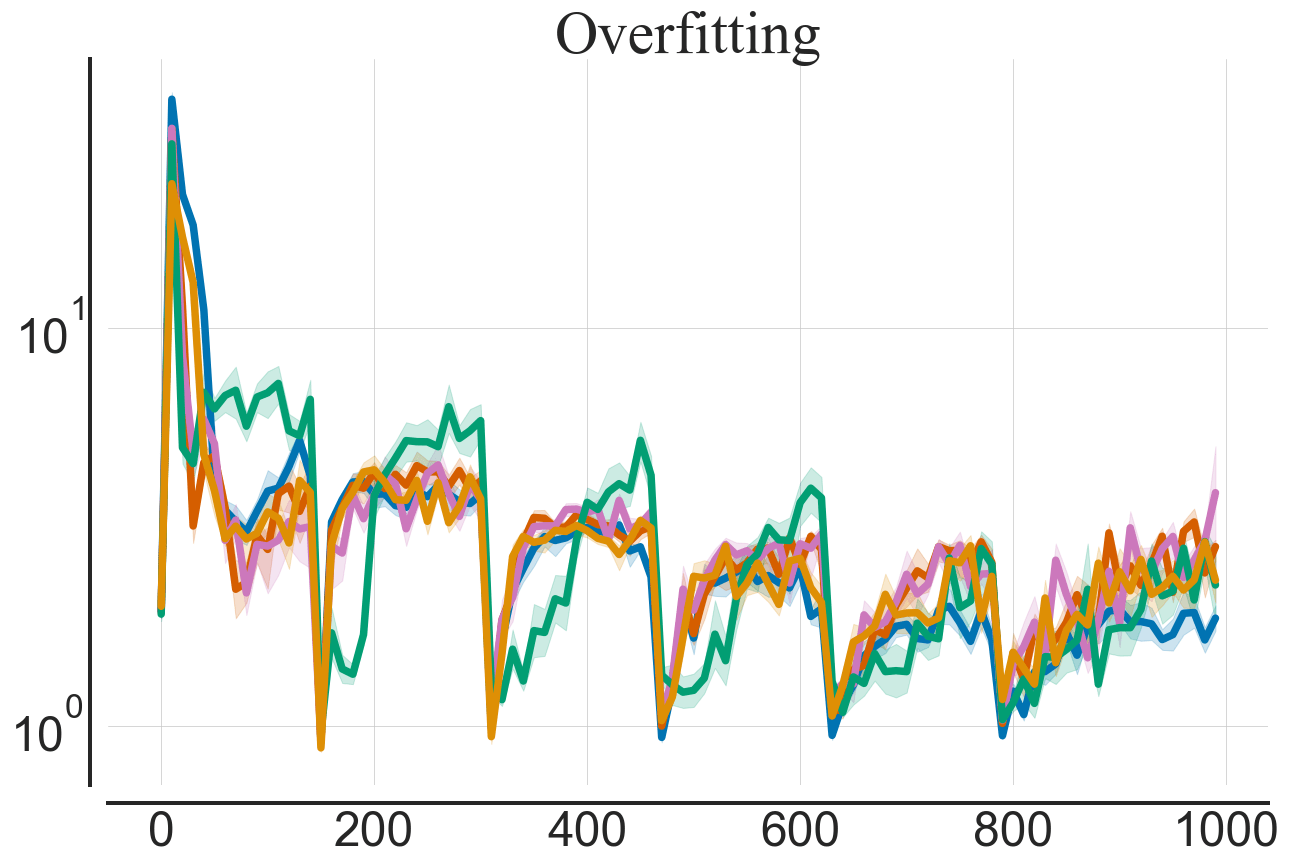}
    \hfill
    \end{subfigure}
    \subcaption{Humanoid Stand}
\end{minipage}
\bigskip
\begin{minipage}[h]{1.0\linewidth}
    \begin{subfigure}{1.0\linewidth}
    \hfill
    \includegraphics[width=0.23\linewidth]{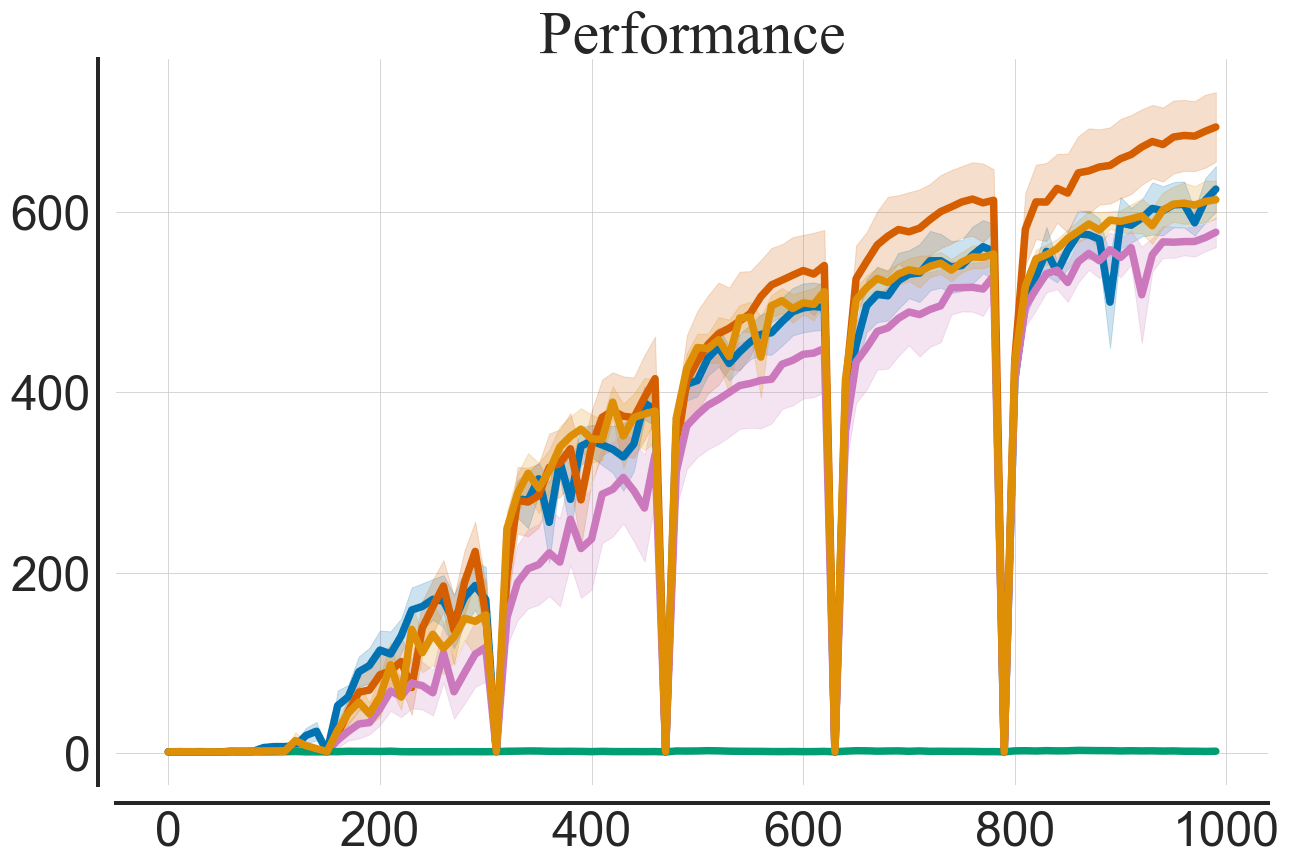}
    \hfill
    \includegraphics[width=0.23\linewidth]{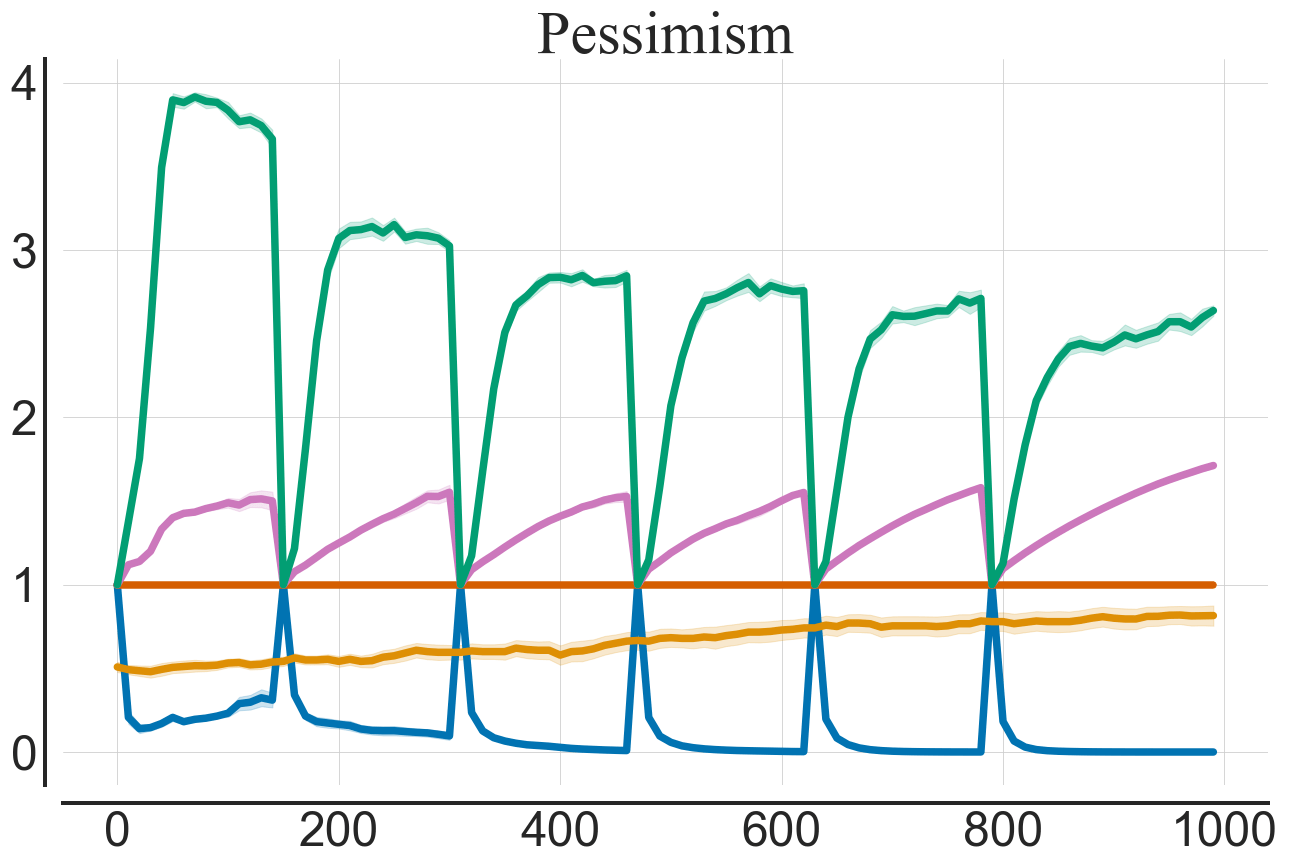}
    \hfill
    \includegraphics[width=0.23\linewidth]{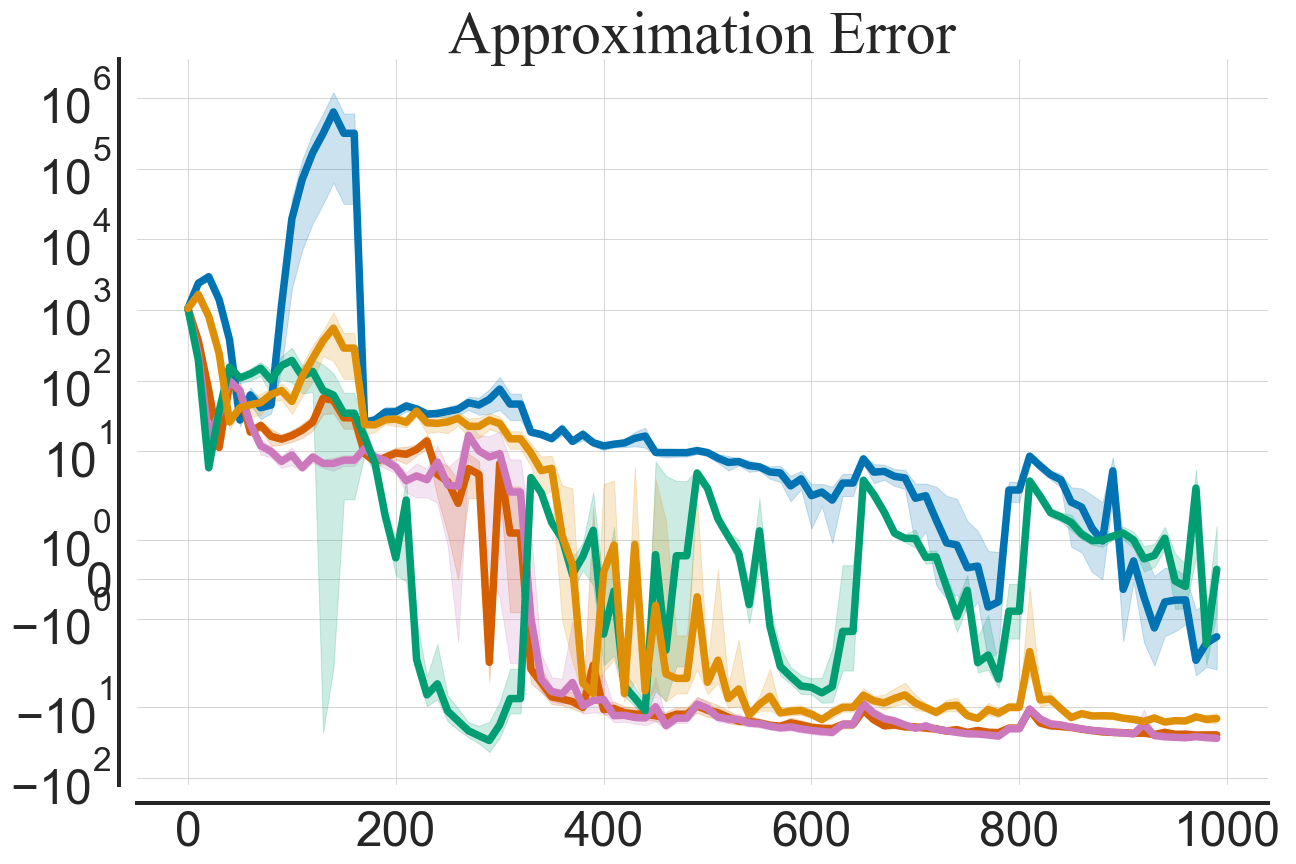}
    \hfill
    \includegraphics[width=0.23\linewidth]{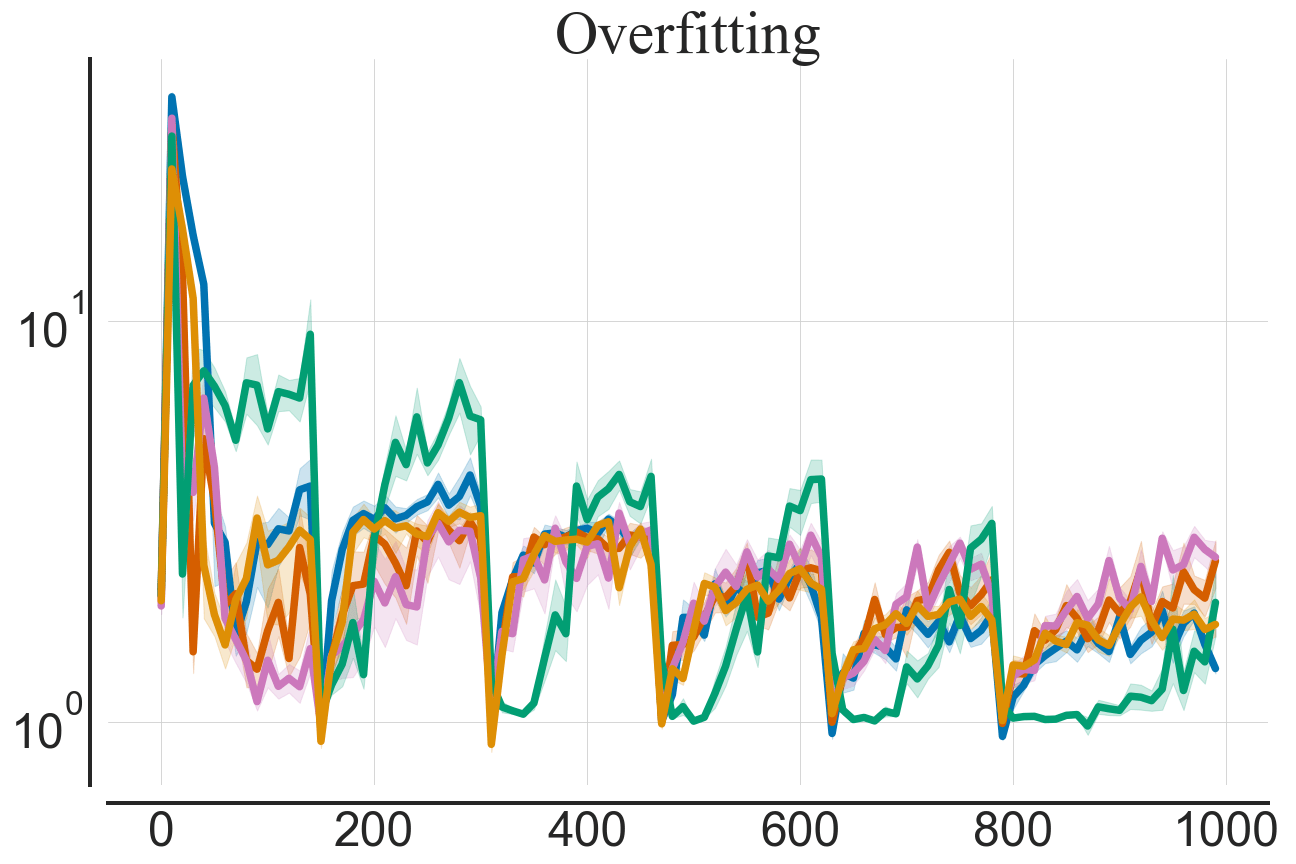}
    \hfill
    \end{subfigure}
    \subcaption{Humanoid Walk}
\end{minipage}
\bigskip
\begin{minipage}[h]{1.0\linewidth}
    \begin{subfigure}{1.0\linewidth}
    \hfill
    \includegraphics[width=0.23\linewidth]{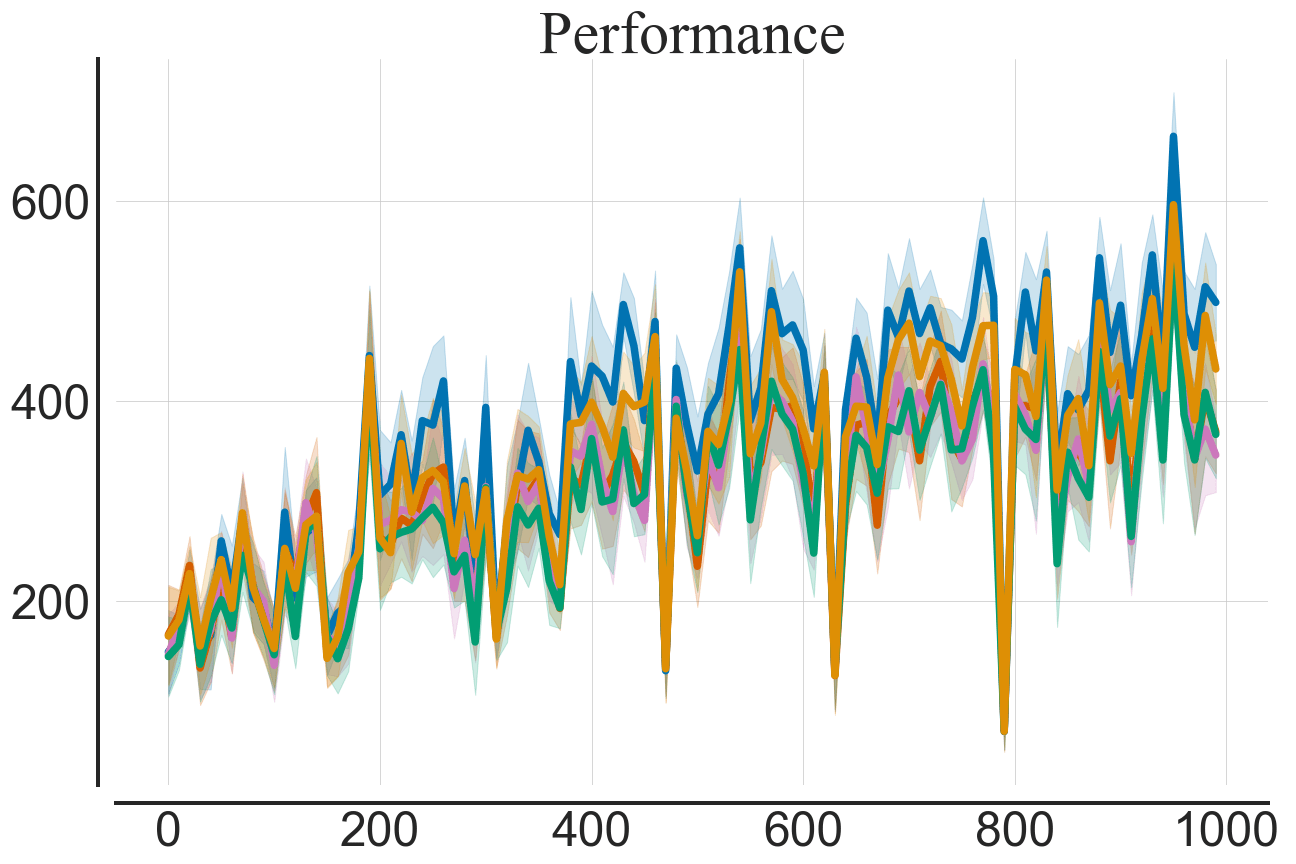}
    \hfill
    \includegraphics[width=0.23\linewidth]{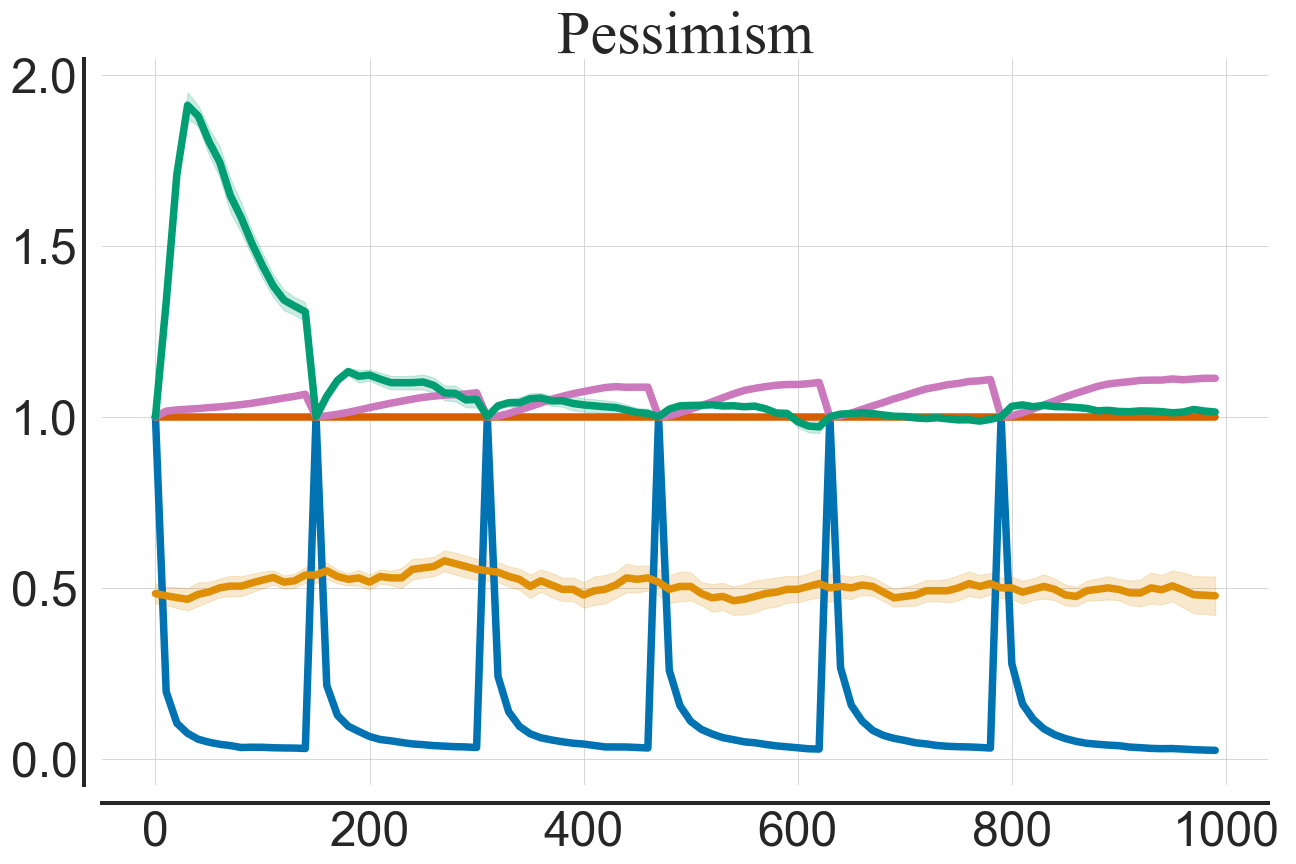}
    \hfill
    \includegraphics[width=0.23\linewidth]{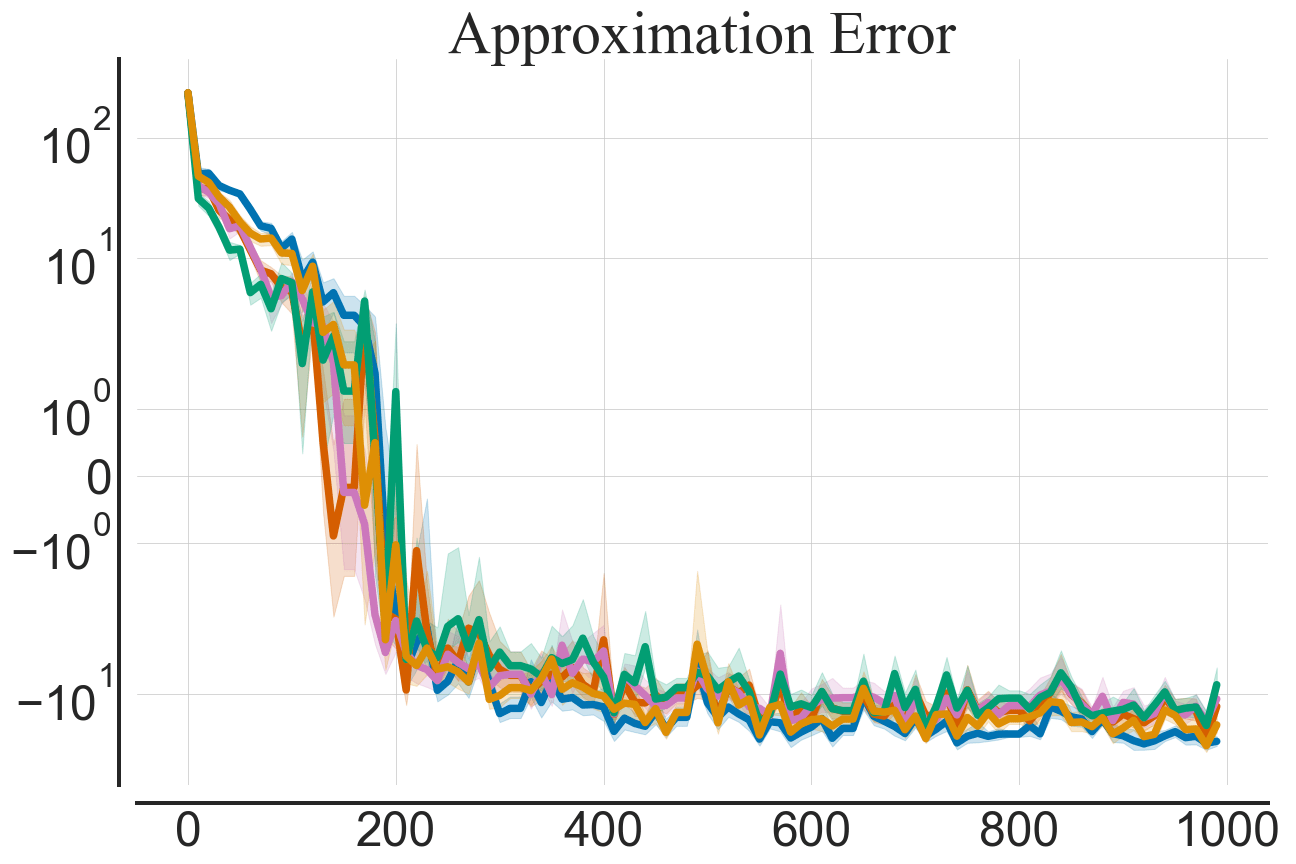}
    \hfill
    \includegraphics[width=0.23\linewidth]{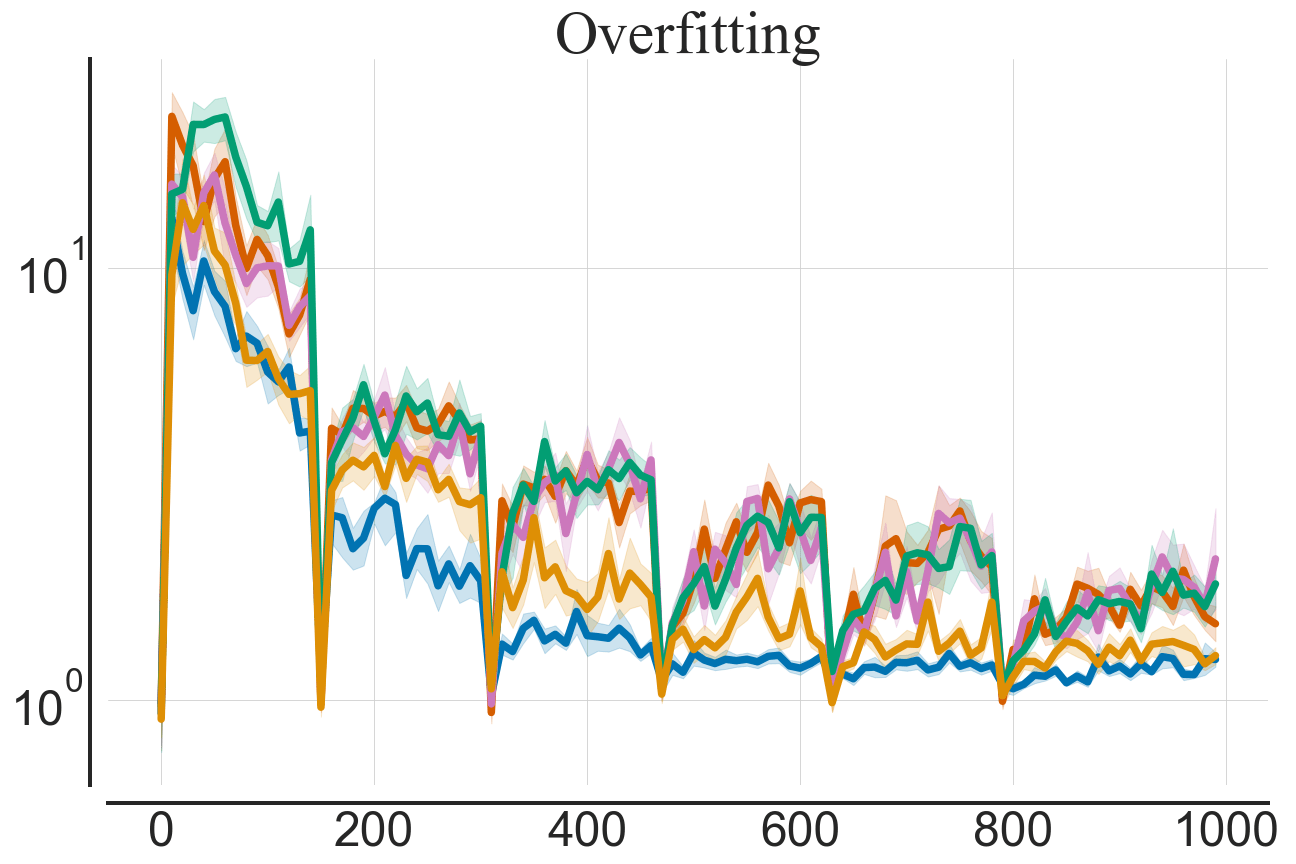}
    \hfill
    \end{subfigure}
    \subcaption{Swimmer Swimmer6}
\end{minipage}
\bigskip
\begin{minipage}[h]{1.0\linewidth}
    \begin{subfigure}{1.0\linewidth}
    \hfill
    \includegraphics[width=0.23\linewidth]{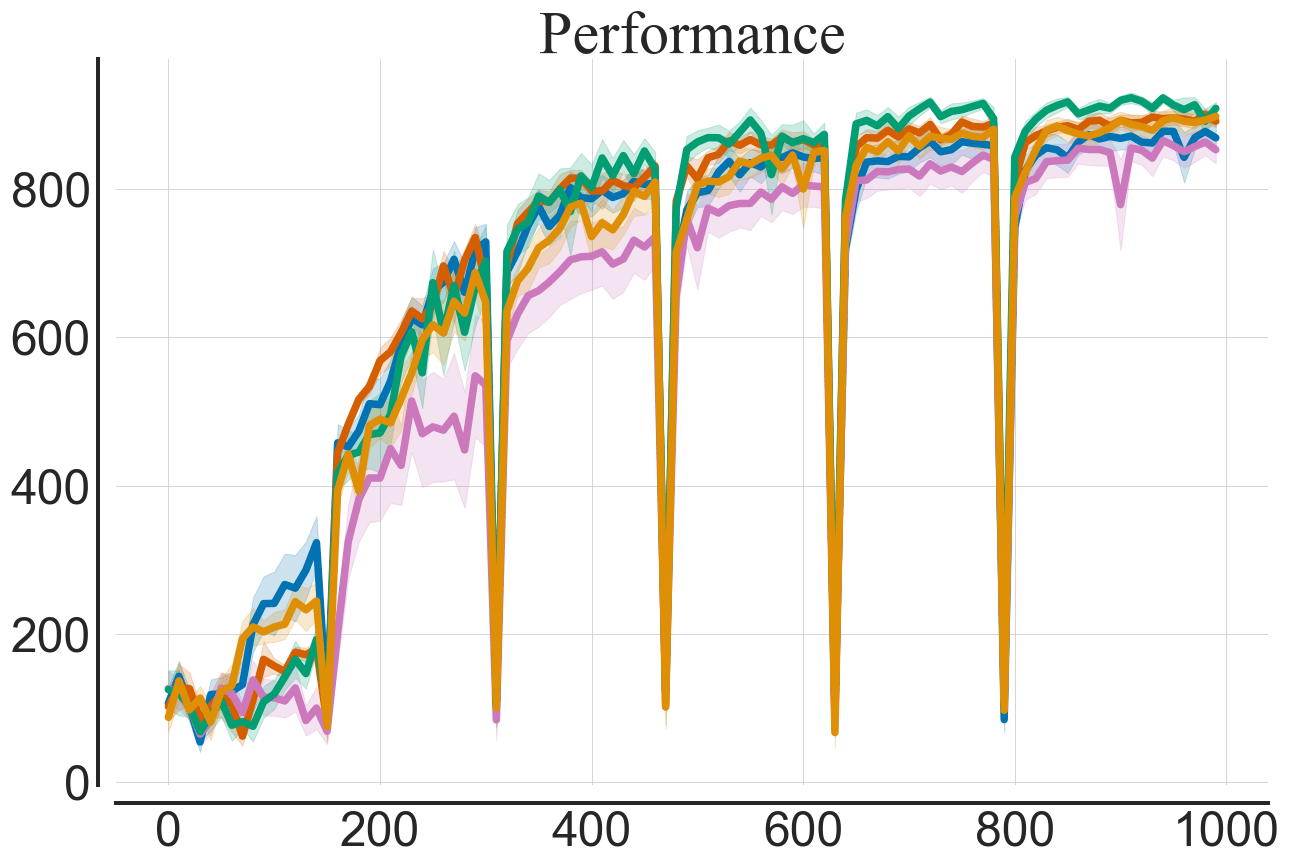}
    \hfill
    \includegraphics[width=0.23\linewidth]{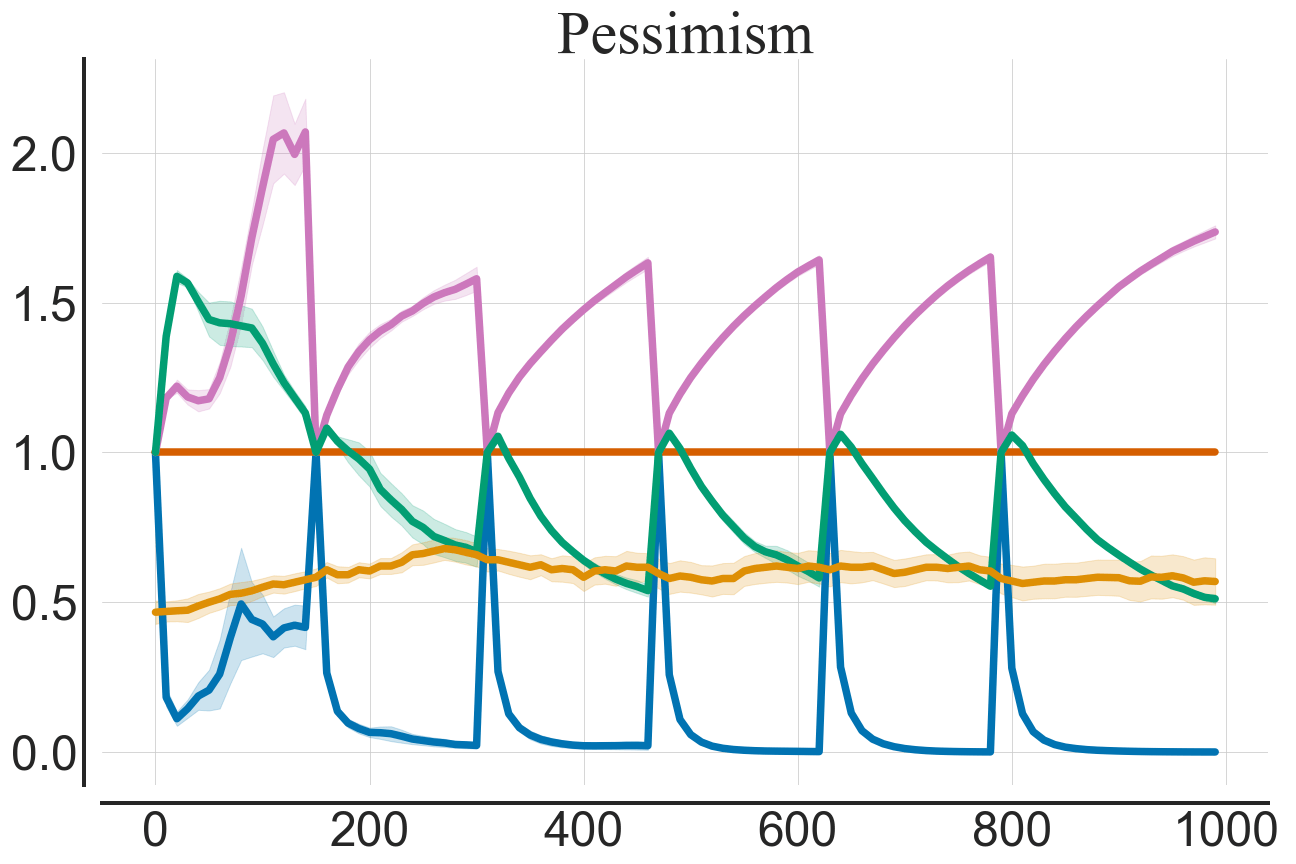}
    \hfill
    \includegraphics[width=0.23\linewidth]{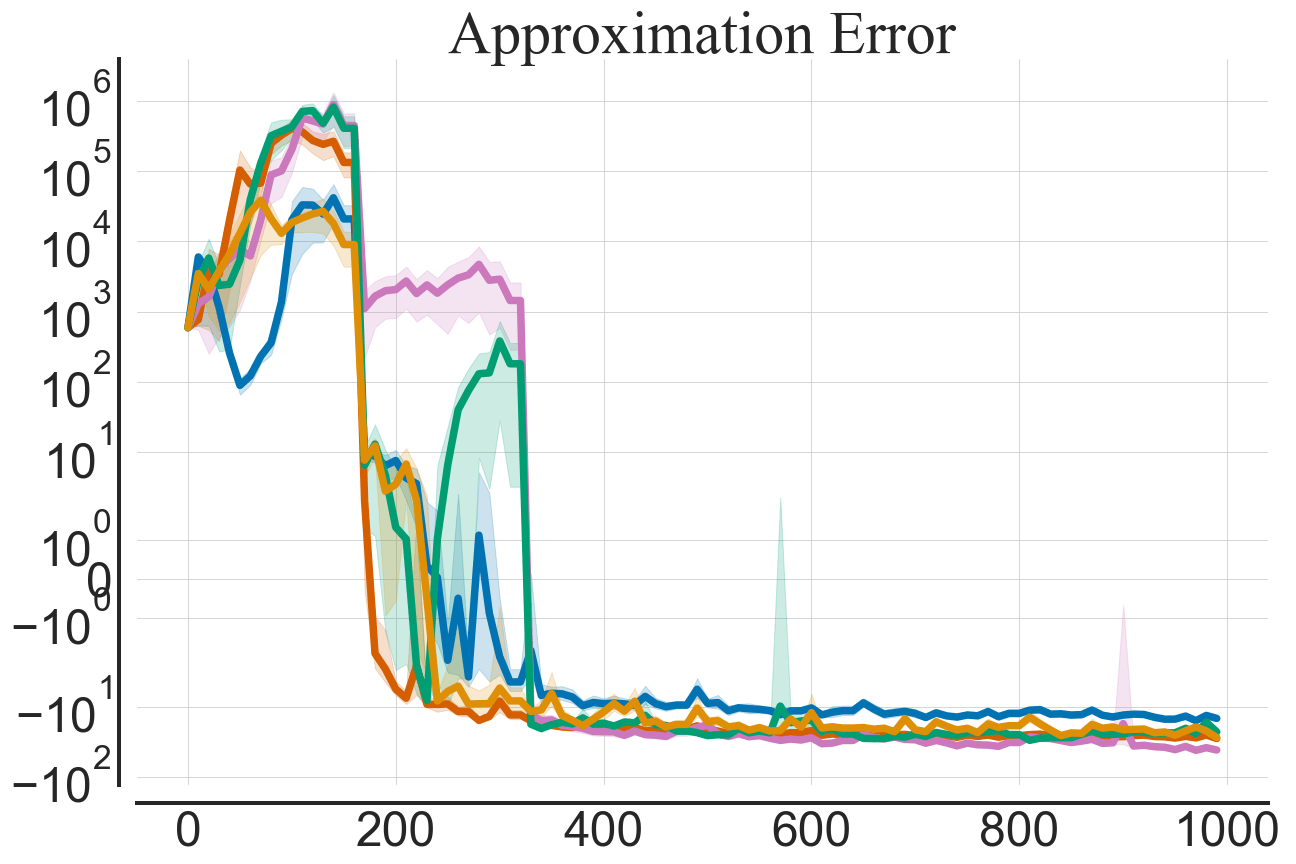}
    \hfill
    \includegraphics[width=0.23\linewidth]{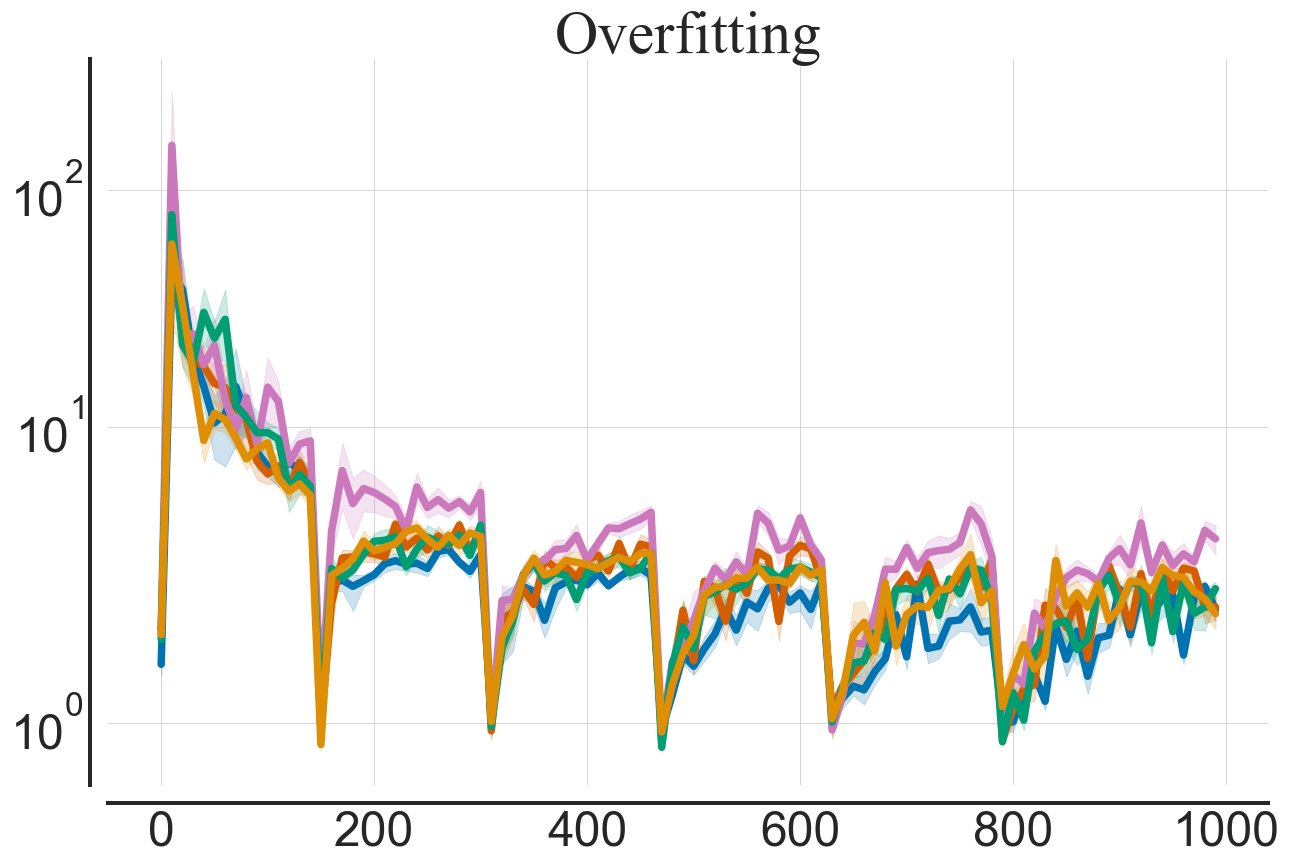}
    \hfill
    \end{subfigure}
    \subcaption{Quadruped Run}
\end{minipage}
\bigskip
\begin{minipage}[h]{1.0\linewidth}
    \begin{subfigure}{1.0\linewidth}
    \hfill
    \includegraphics[width=0.23\linewidth]{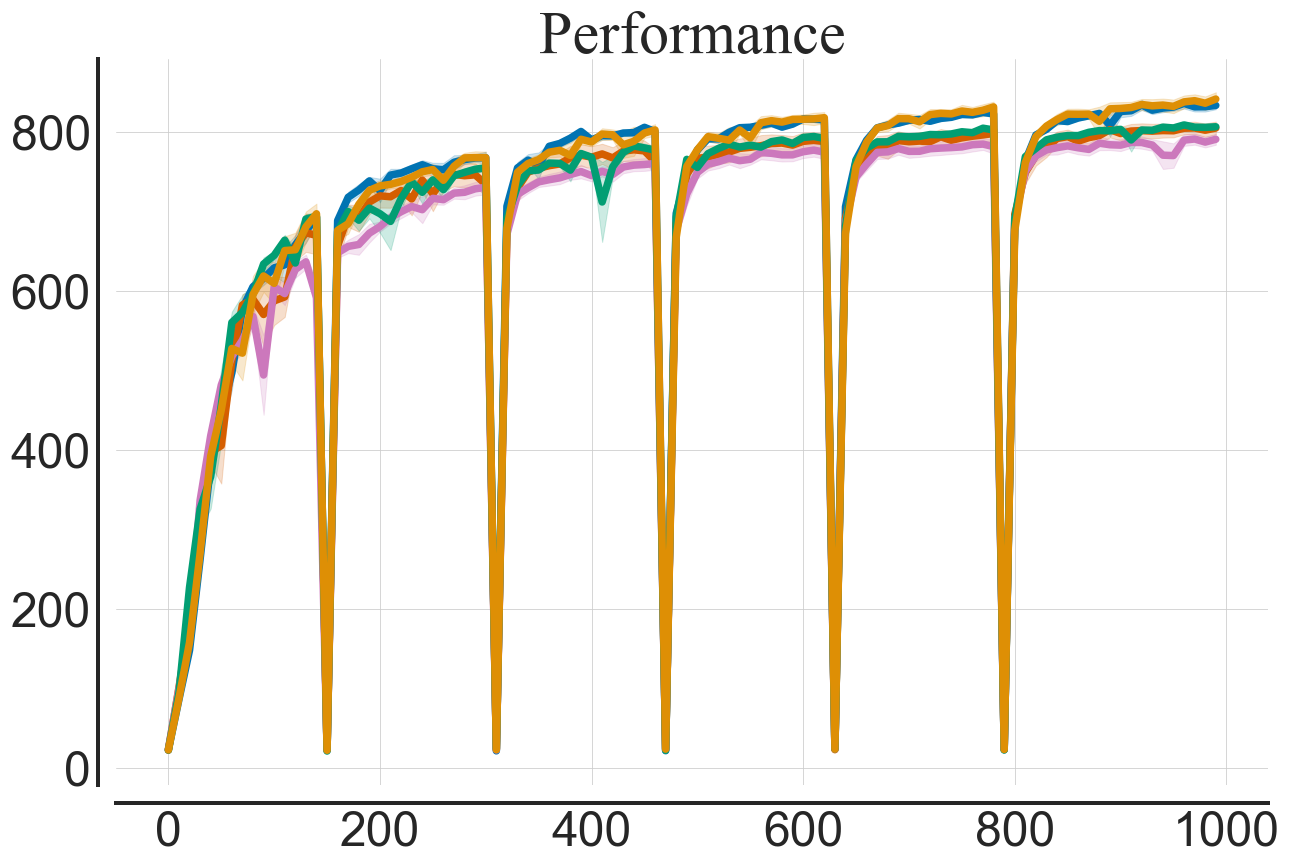}
    \hfill
    \includegraphics[width=0.23\linewidth]{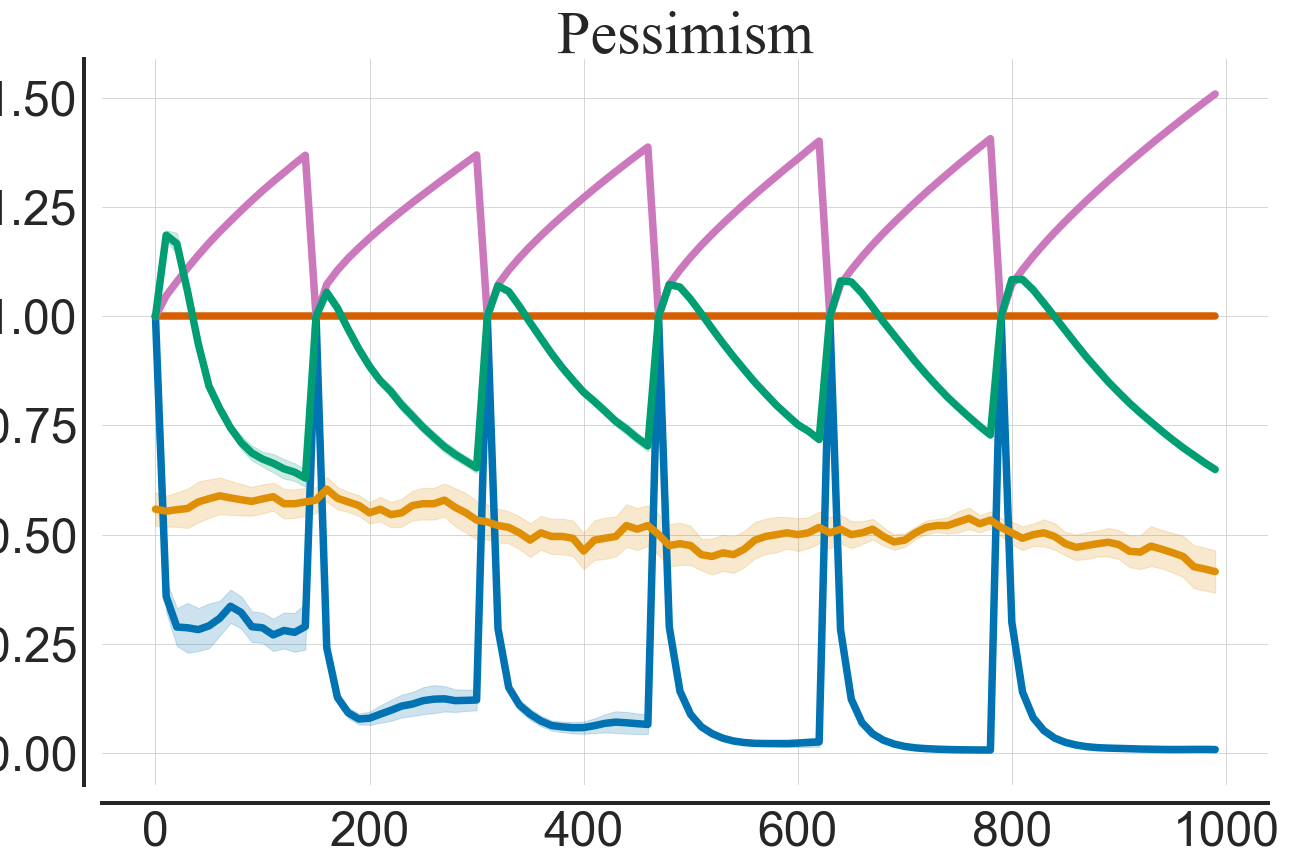}
    \hfill
    \includegraphics[width=0.23\linewidth]{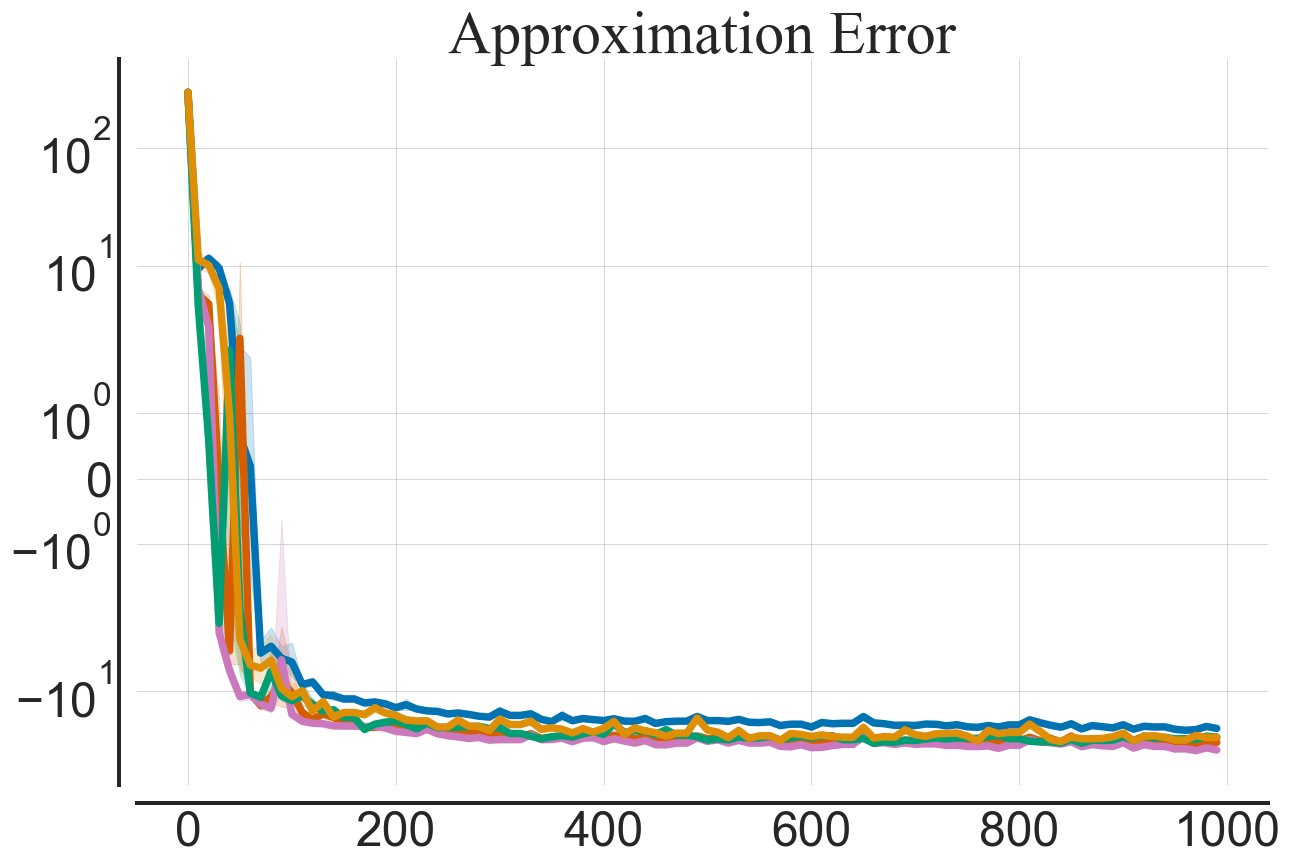}
    \hfill
    \includegraphics[width=0.23\linewidth]{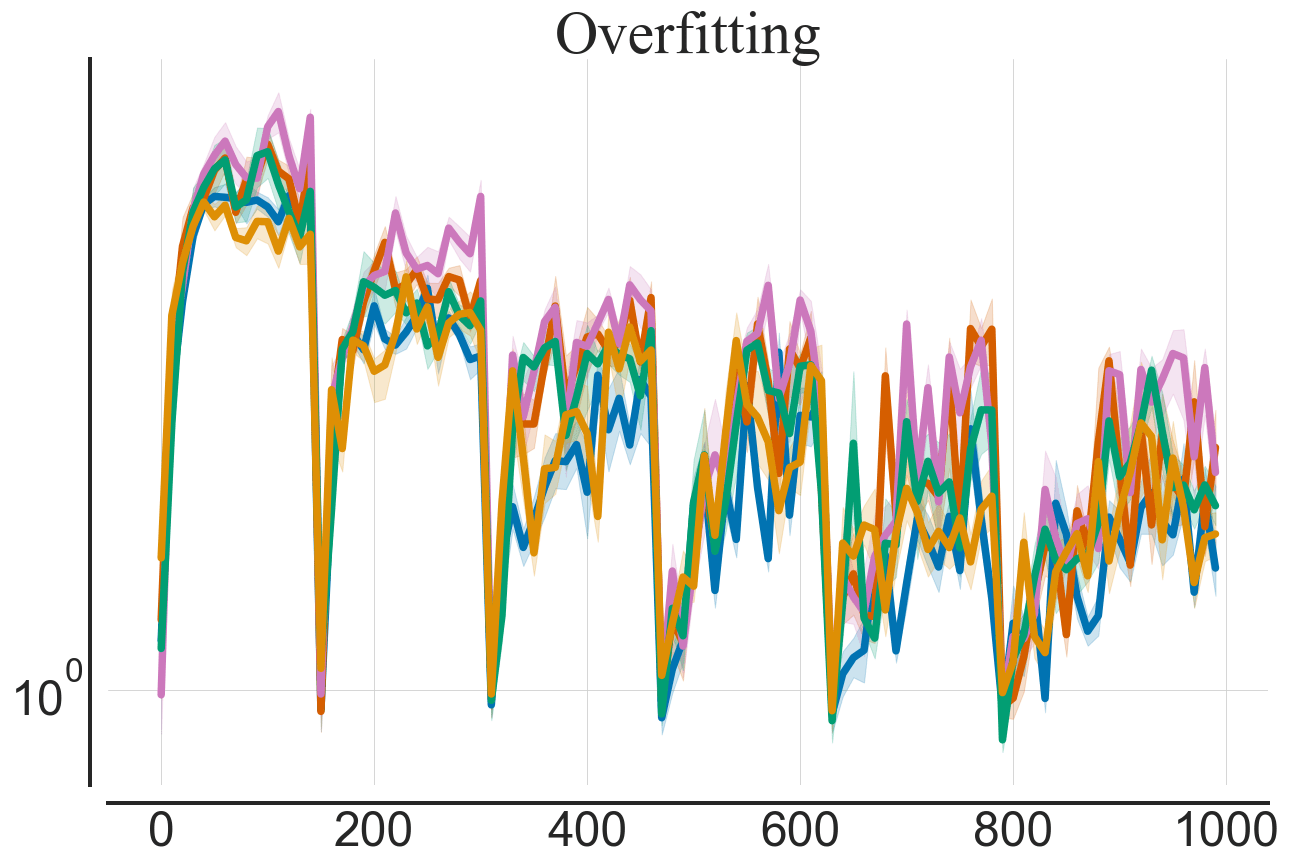}
    \hfill
    \end{subfigure}
    \subcaption{Walker Run}
\end{minipage}
\caption{High replay regime results for each considered task (2/4). 10 seeds per task, mean and 3 standard deviations.}
\label{fig:learning_curves6}
\end{center}
\end{figure*}

\begin{figure*}[ht!]
\begin{center}
\begin{minipage}[h]{1.0\linewidth}
\centering
    \begin{subfigure}{0.88\linewidth}
    \includegraphics[width=\textwidth]{images/legend_1.png}
    \end{subfigure}
\end{minipage}
\bigskip
\begin{minipage}[h]{1.0\linewidth}
    \begin{subfigure}{1.0\linewidth}
    \hfill
    \includegraphics[width=0.23\linewidth]{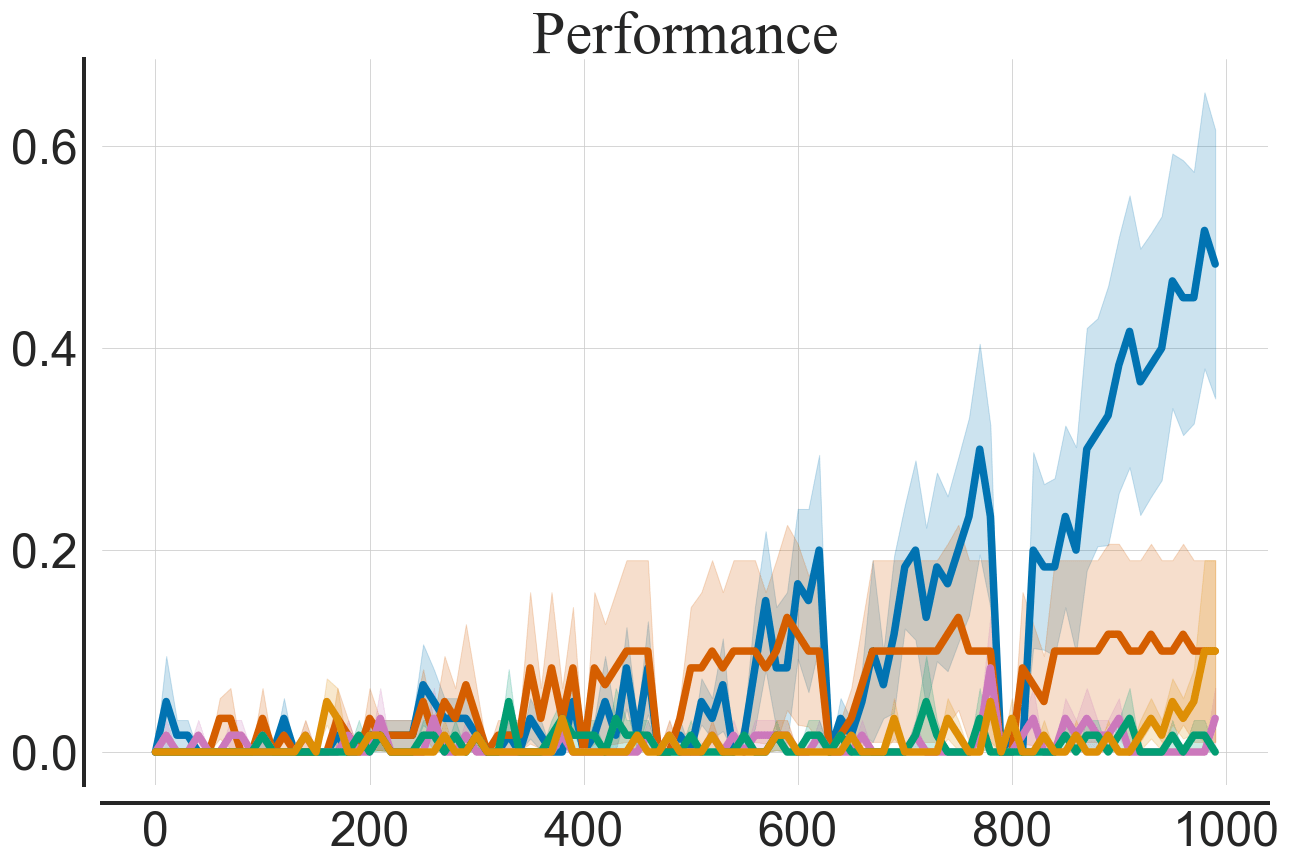}
    \hfill
    \includegraphics[width=0.23\linewidth]{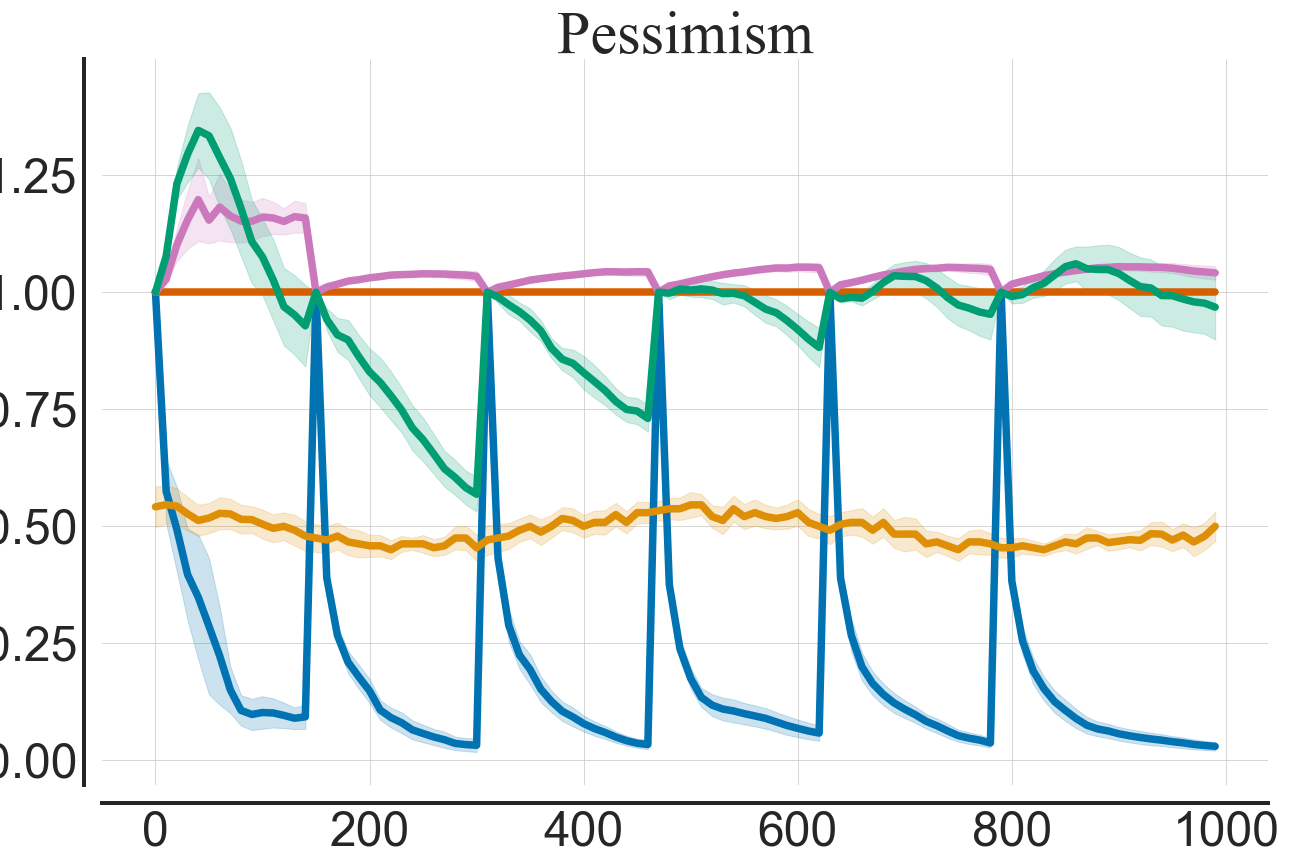}
    \hfill
    \includegraphics[width=0.23\linewidth]{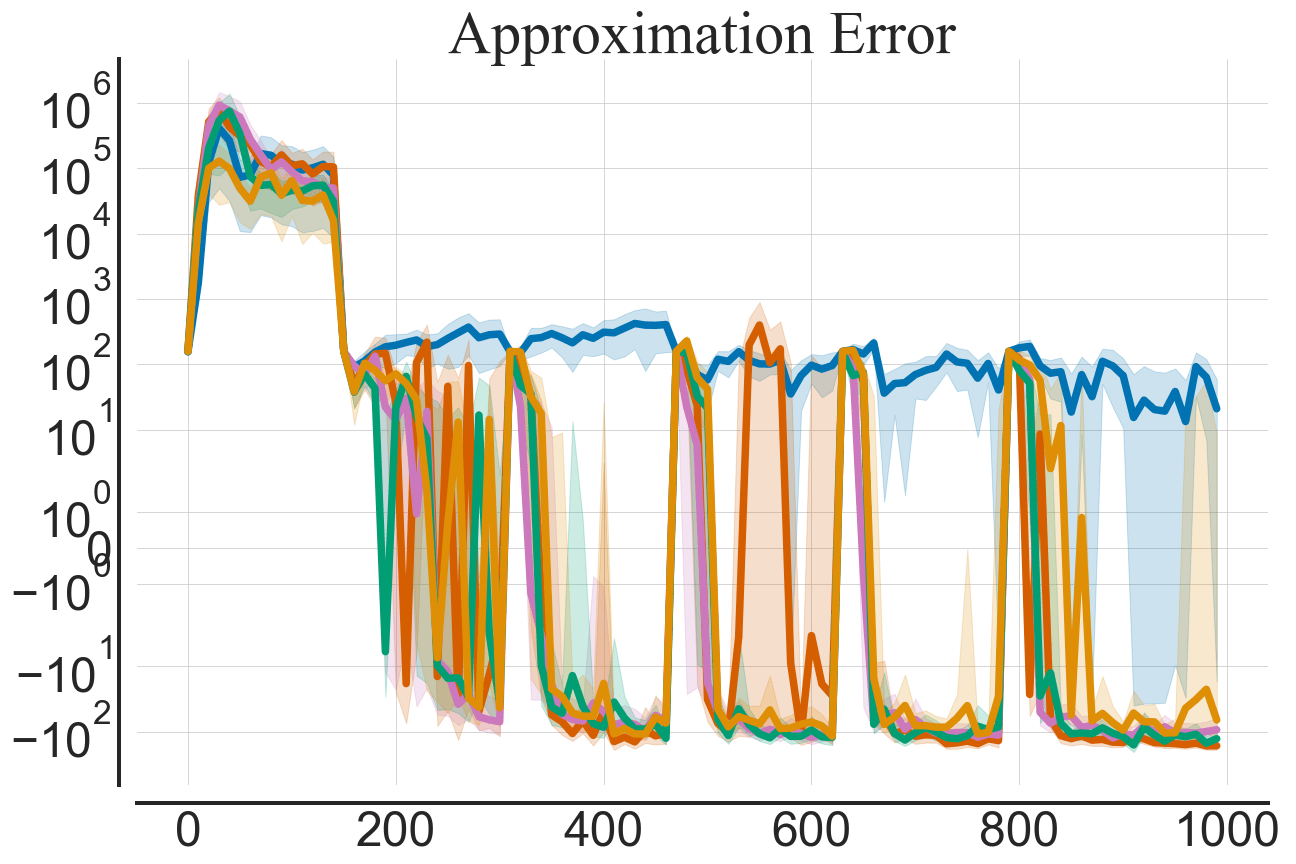}
    \hfill
    \includegraphics[width=0.23\linewidth]{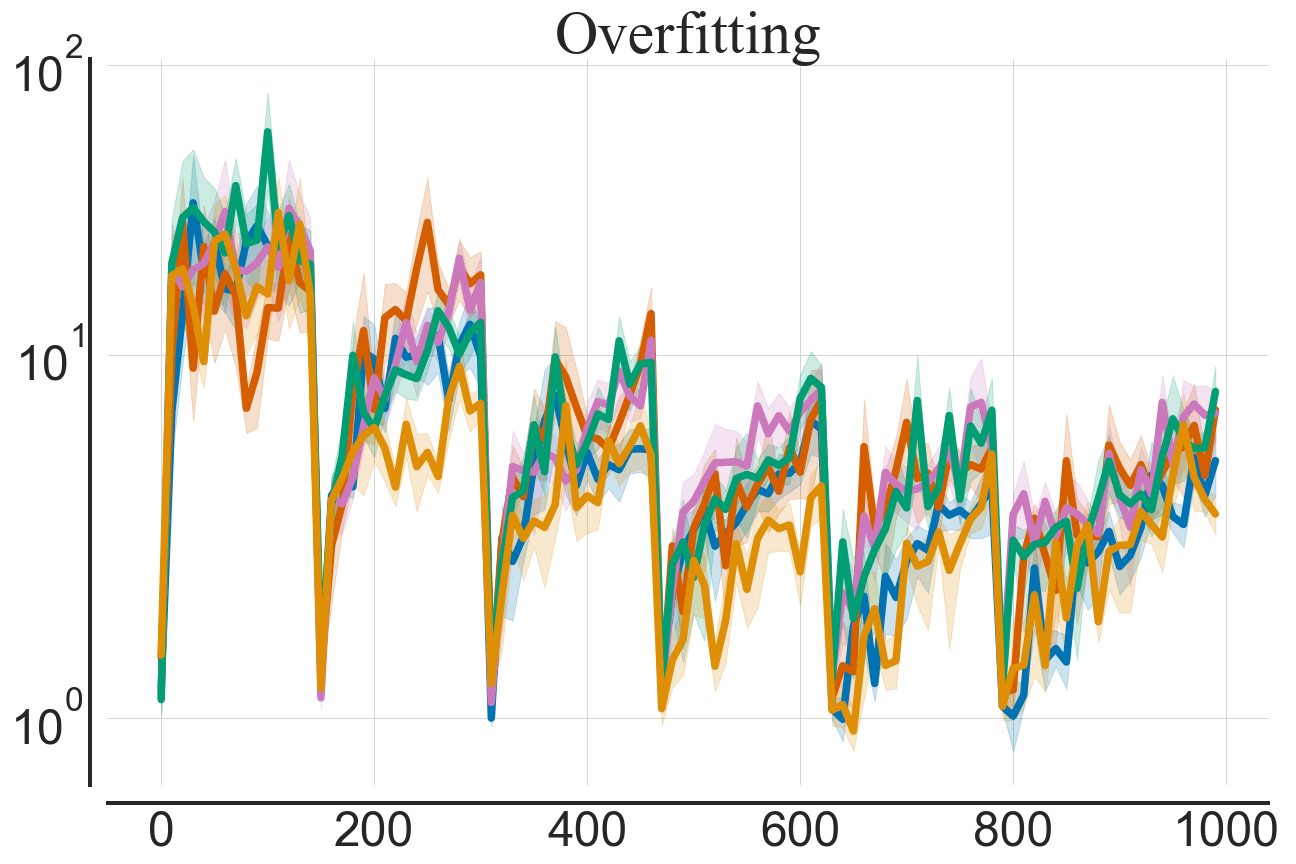}
    \hfill
    \end{subfigure}
    \subcaption{Assembly}
\end{minipage}
\bigskip
\begin{minipage}[h]{1.0\linewidth}
    \begin{subfigure}{1.0\linewidth}
    \hfill
    \includegraphics[width=0.23\linewidth]{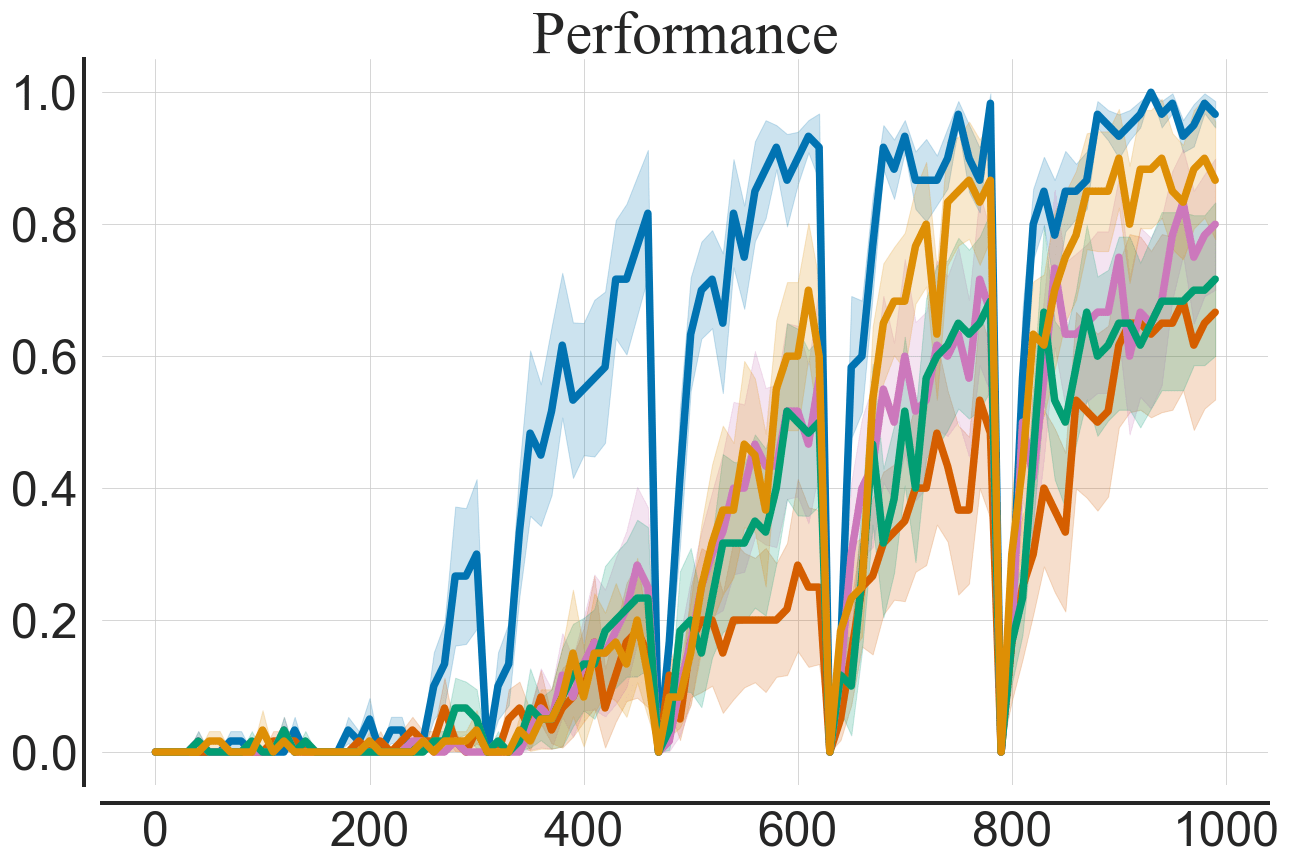}
    \hfill
    \includegraphics[width=0.23\linewidth]{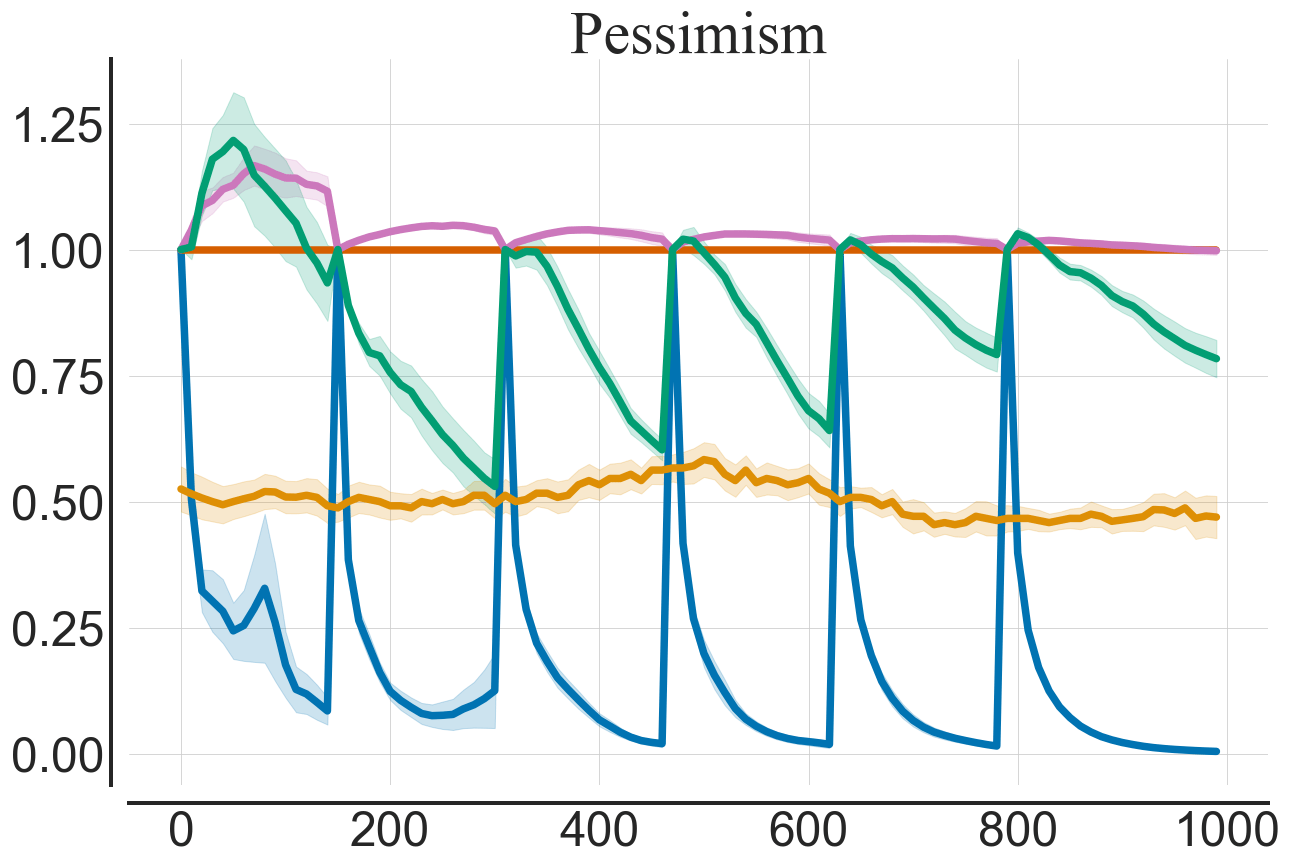}
    \hfill
    \includegraphics[width=0.23\linewidth]{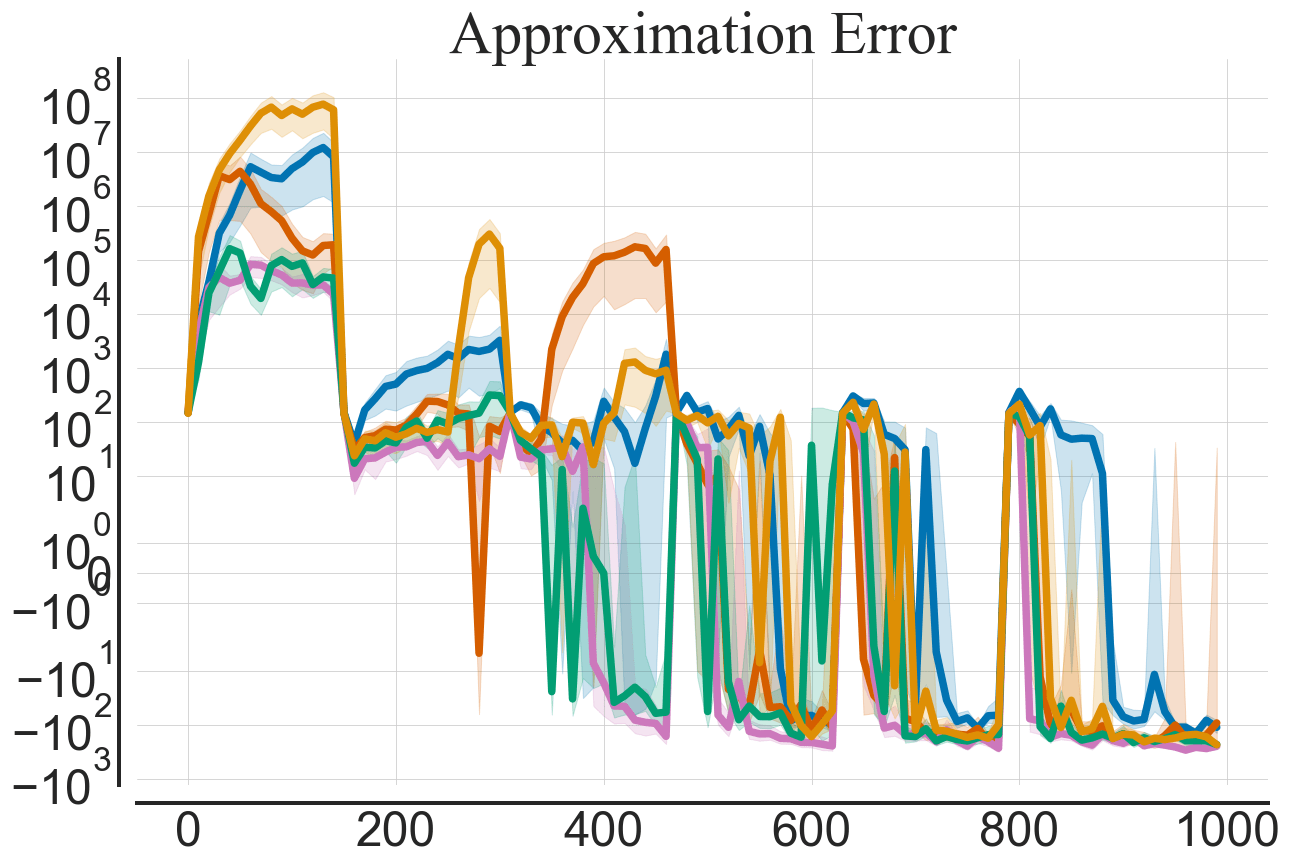}
    \hfill
    \includegraphics[width=0.23\linewidth]{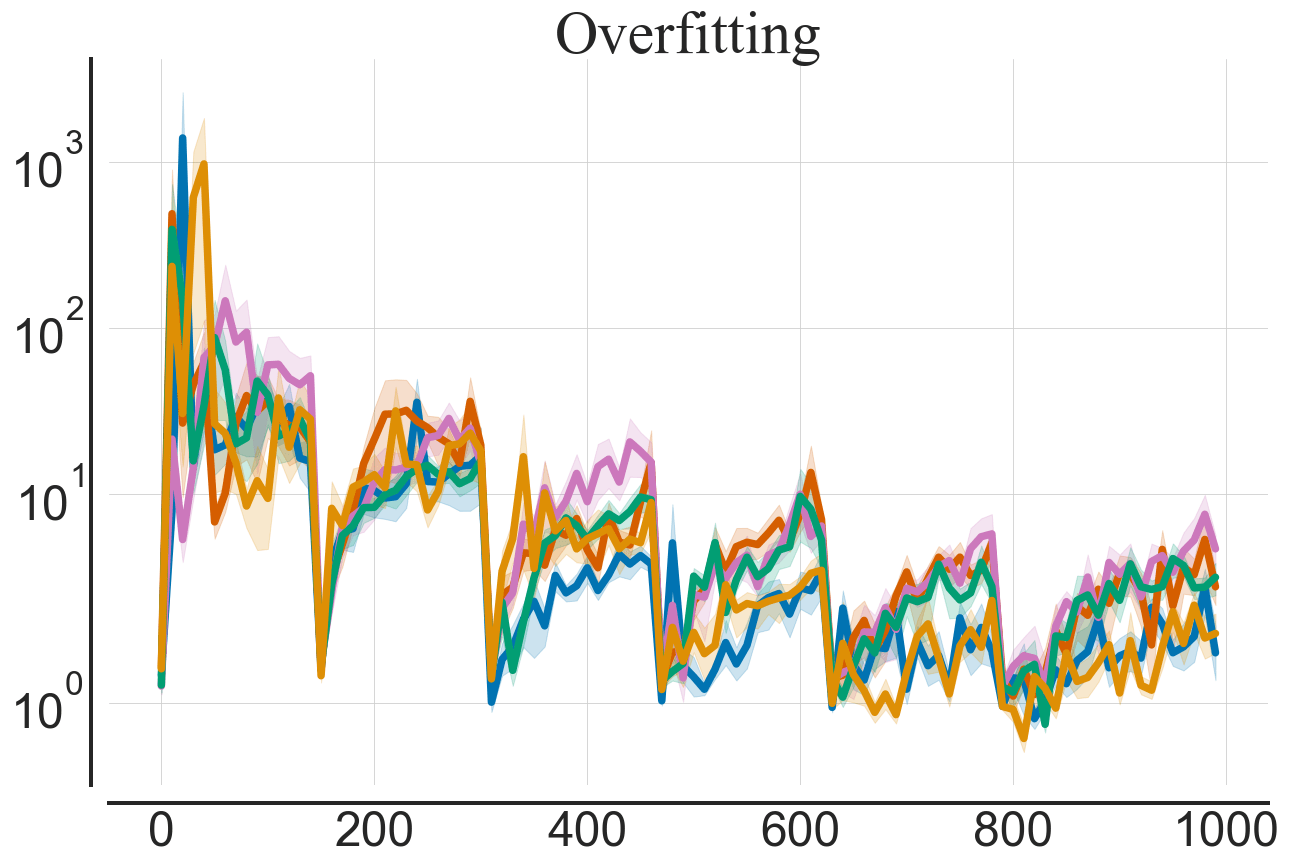}
    \hfill
    \end{subfigure}
    \subcaption{Box Close}
\end{minipage}
\bigskip
\begin{minipage}[h]{1.0\linewidth}
    \begin{subfigure}{1.0\linewidth}
    \hfill
    \includegraphics[width=0.23\linewidth]{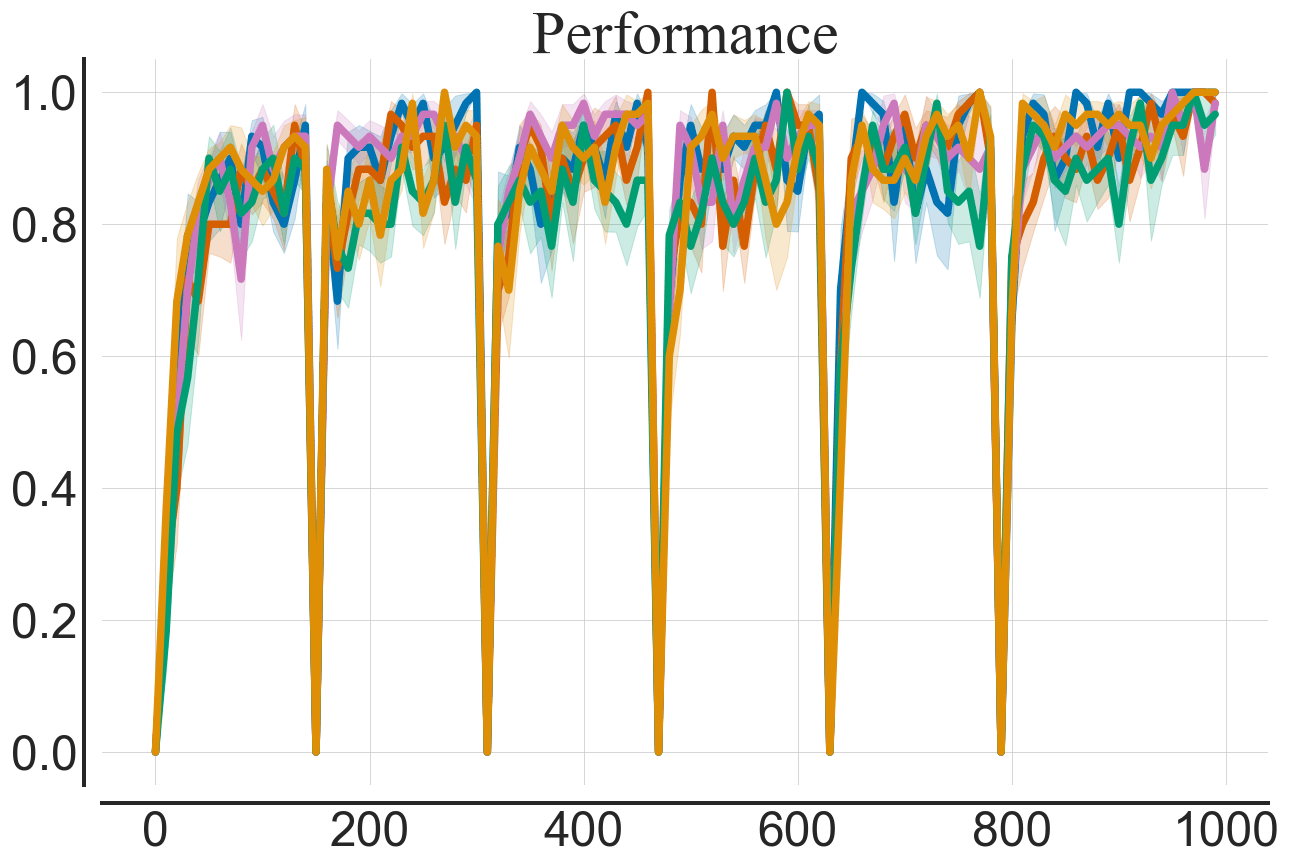}
    \hfill
    \includegraphics[width=0.23\linewidth]{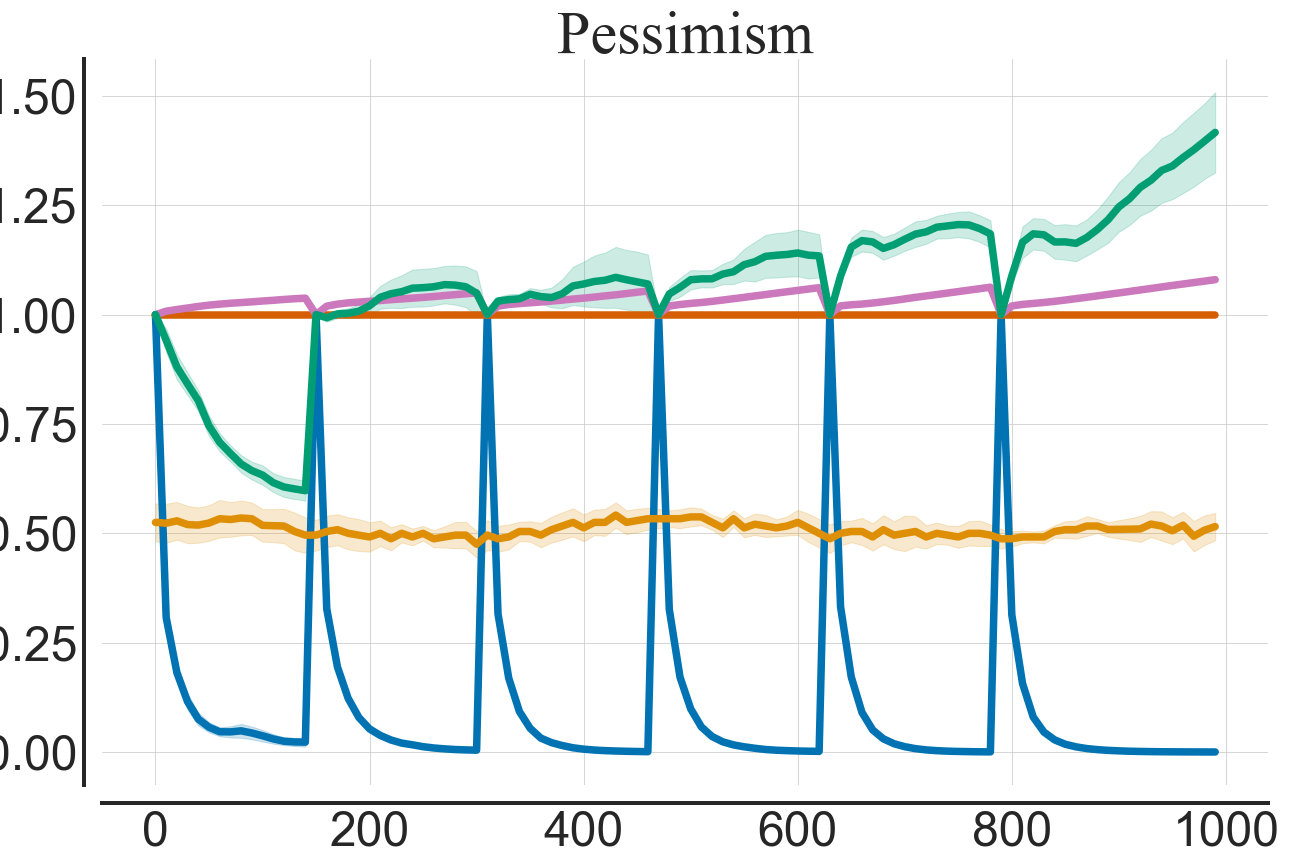}
    \hfill
    \includegraphics[width=0.23\linewidth]{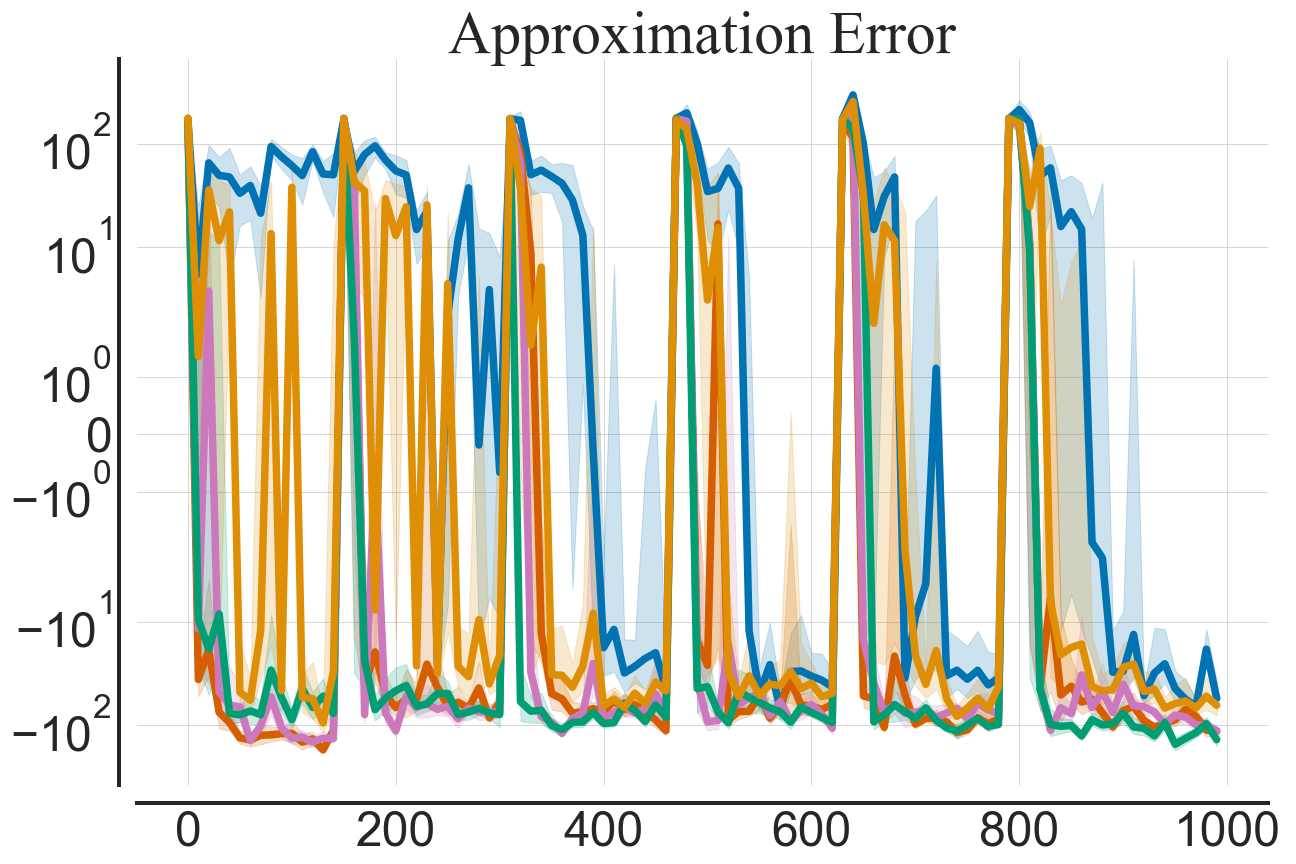}
    \hfill
    \includegraphics[width=0.23\linewidth]{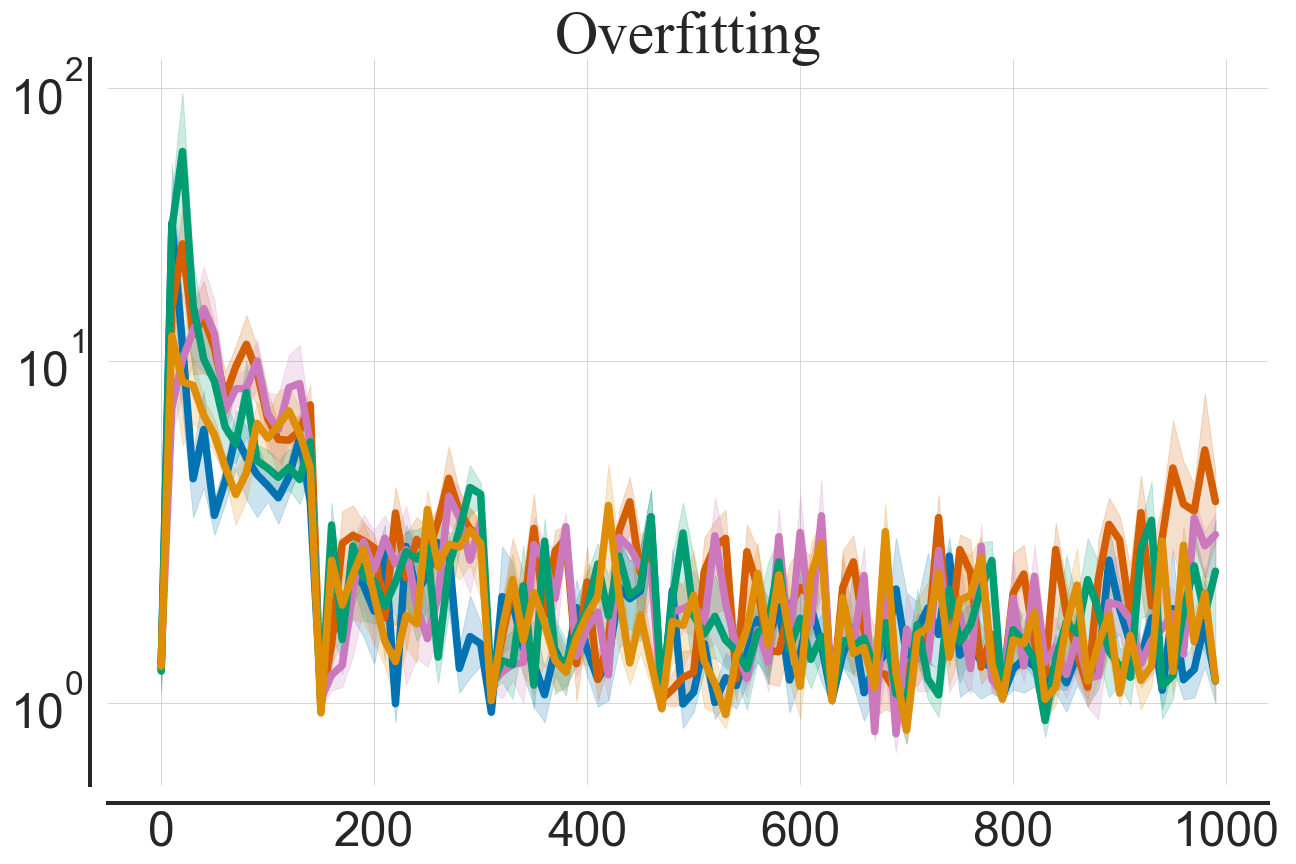}
    \hfill
    \end{subfigure}
    \subcaption{Button Press}
\end{minipage}
\bigskip
\begin{minipage}[h]{1.0\linewidth}
    \begin{subfigure}{1.0\linewidth}
    \hfill
    \includegraphics[width=0.23\linewidth]{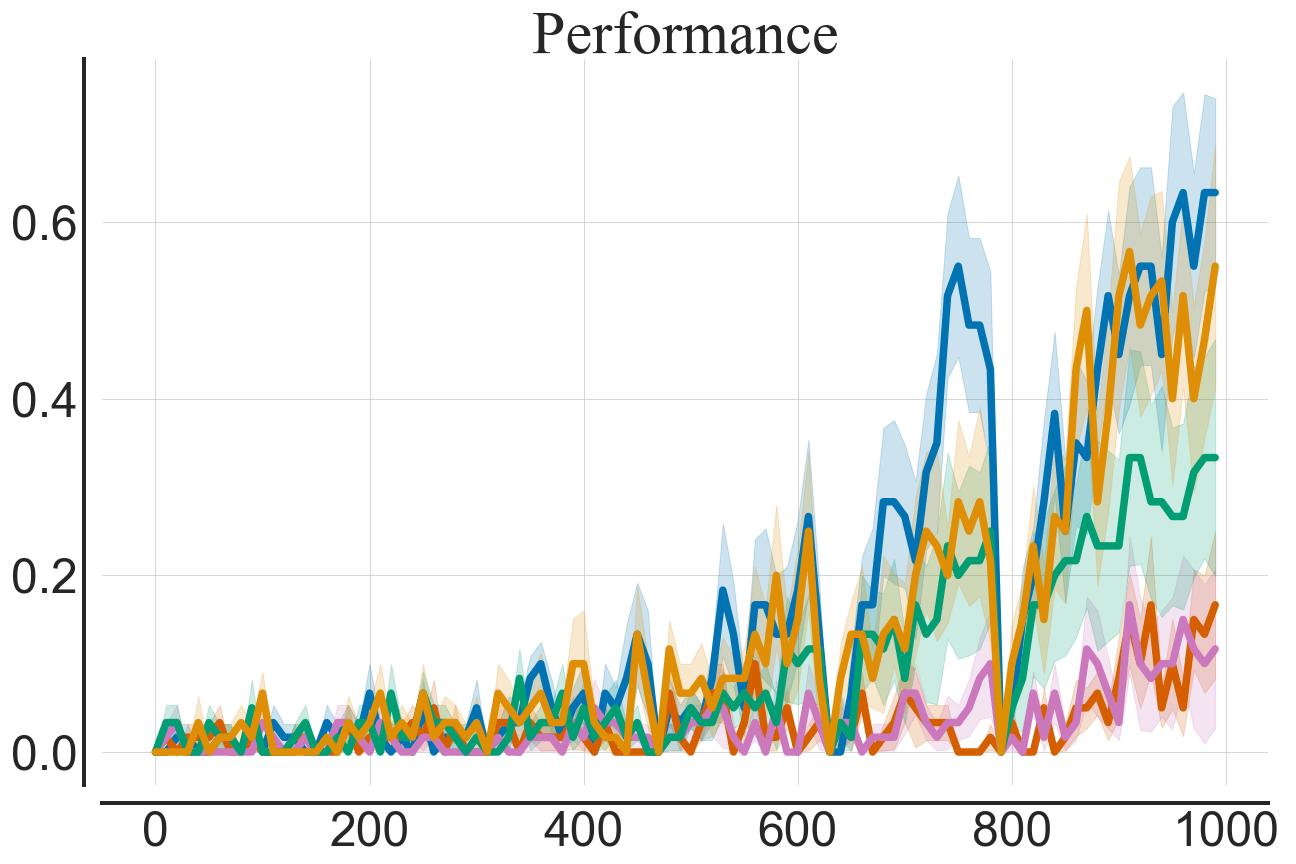}
    \hfill
    \includegraphics[width=0.23\linewidth]{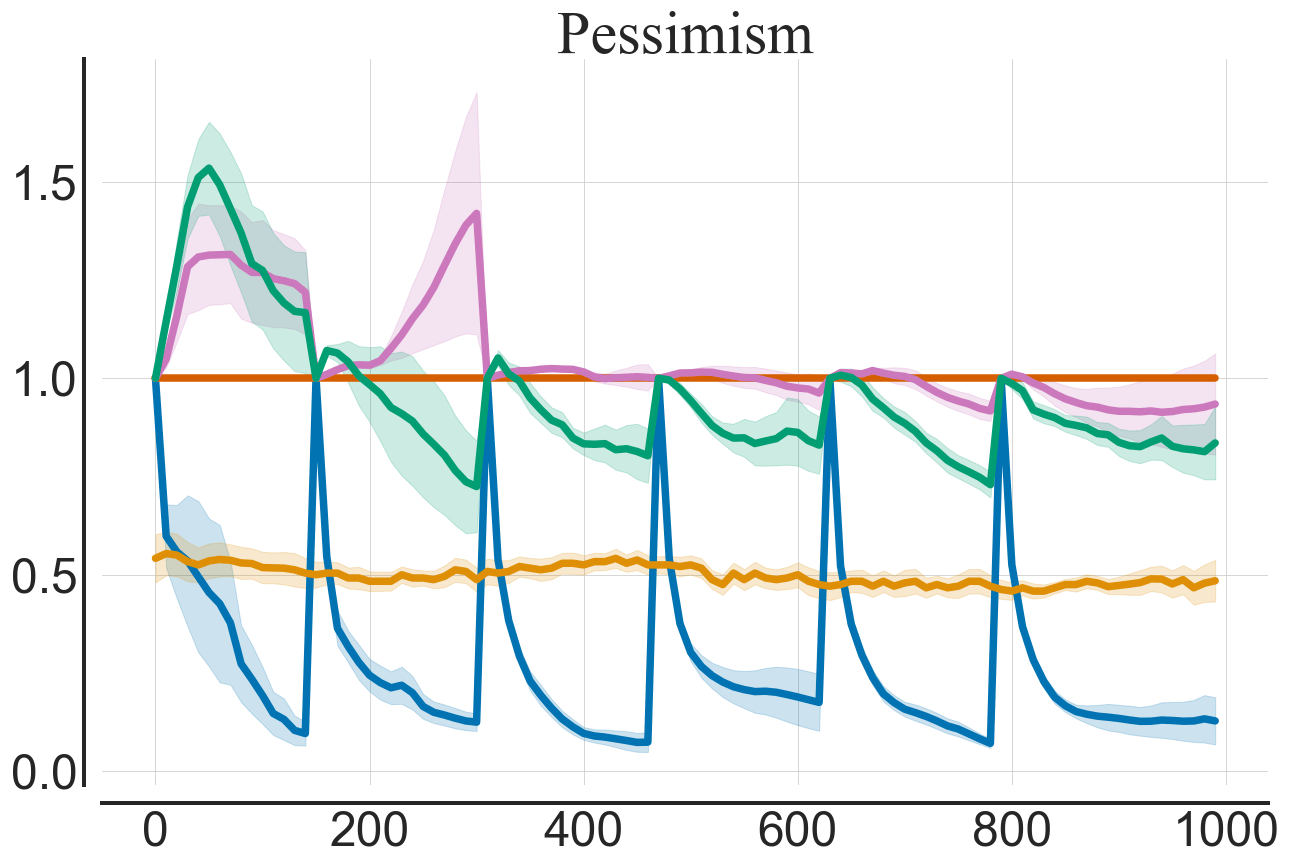}
    \hfill
    \includegraphics[width=0.23\linewidth]{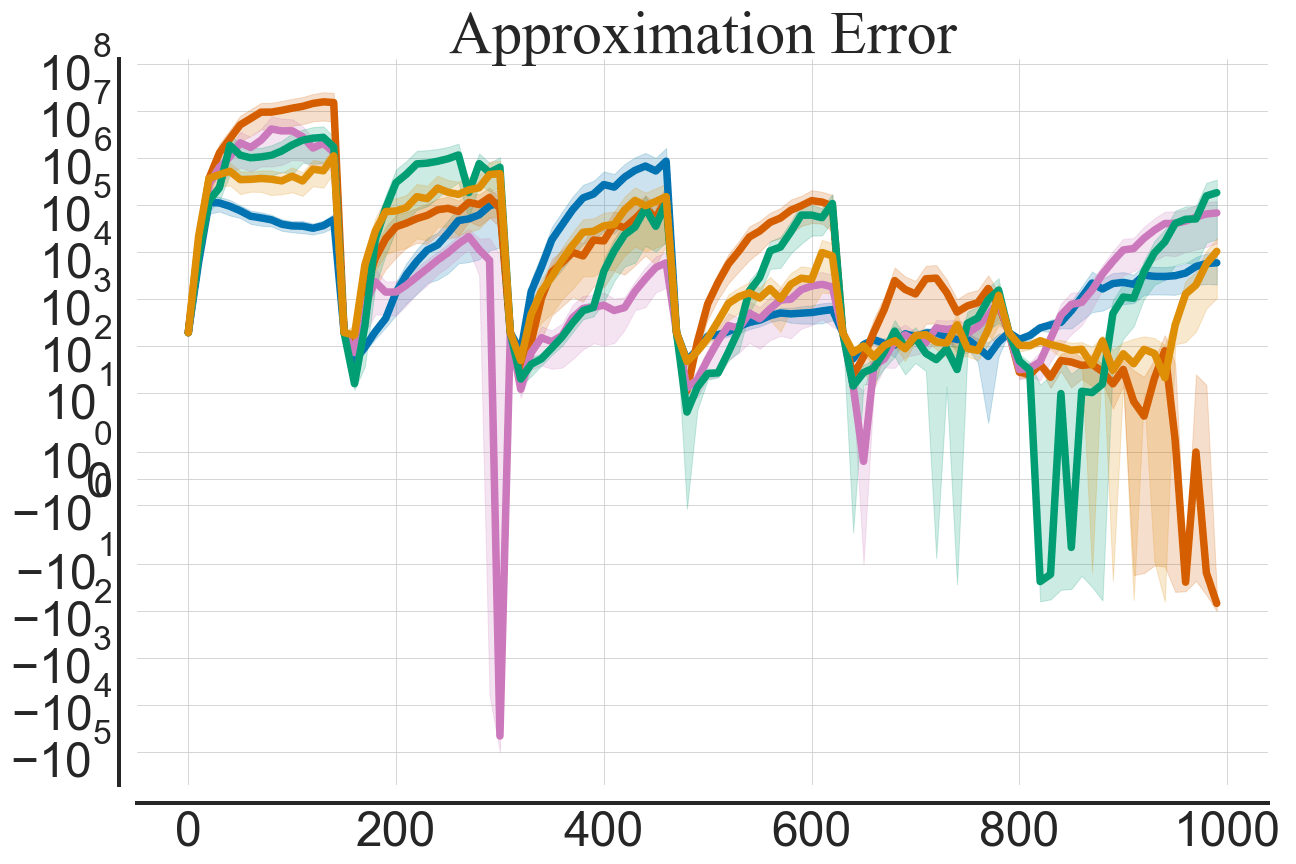}
    \hfill
    \includegraphics[width=0.23\linewidth]{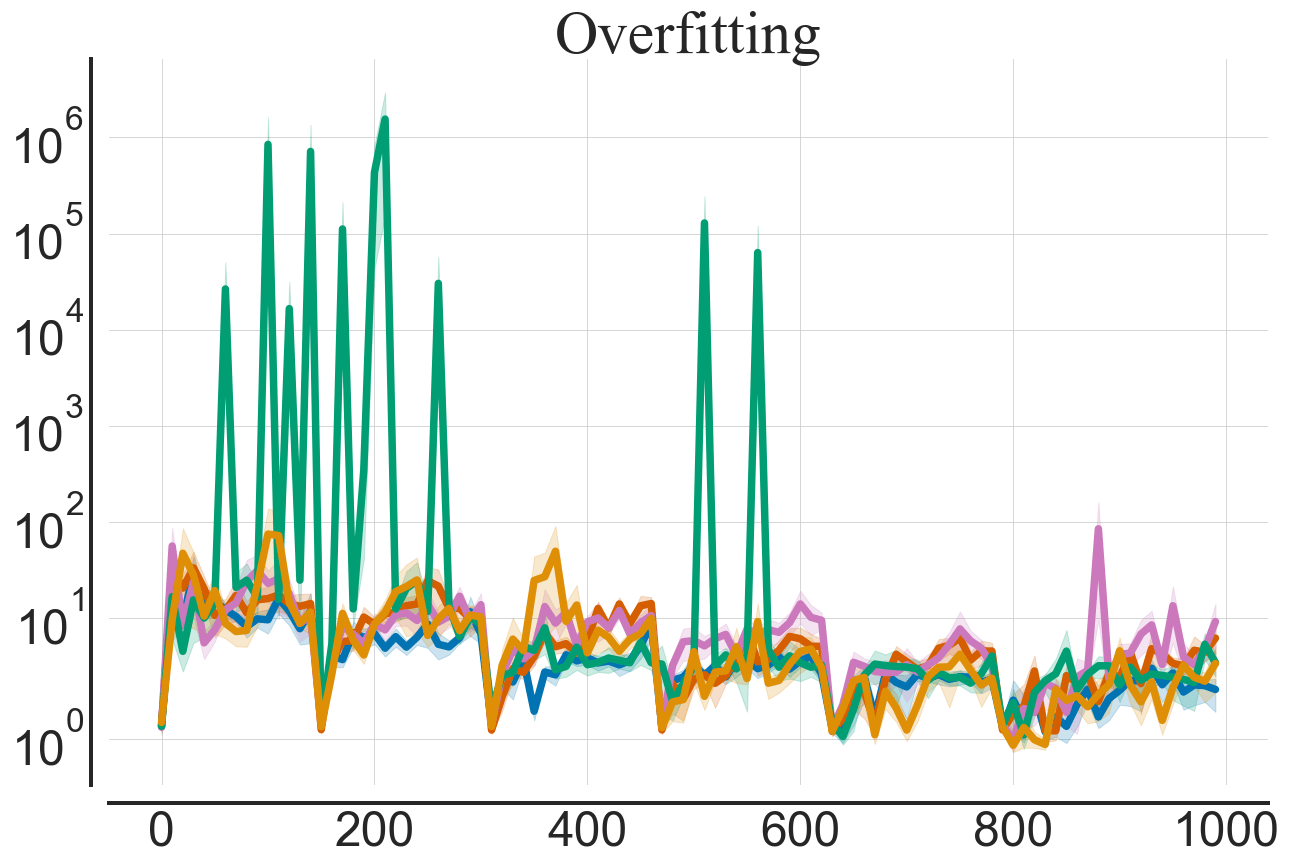}
    \hfill
    \end{subfigure}
    \subcaption{Coffee Pull}
\end{minipage}
\bigskip
\begin{minipage}[h]{1.0\linewidth}
    \begin{subfigure}{1.0\linewidth}
    \hfill
    \includegraphics[width=0.23\linewidth]{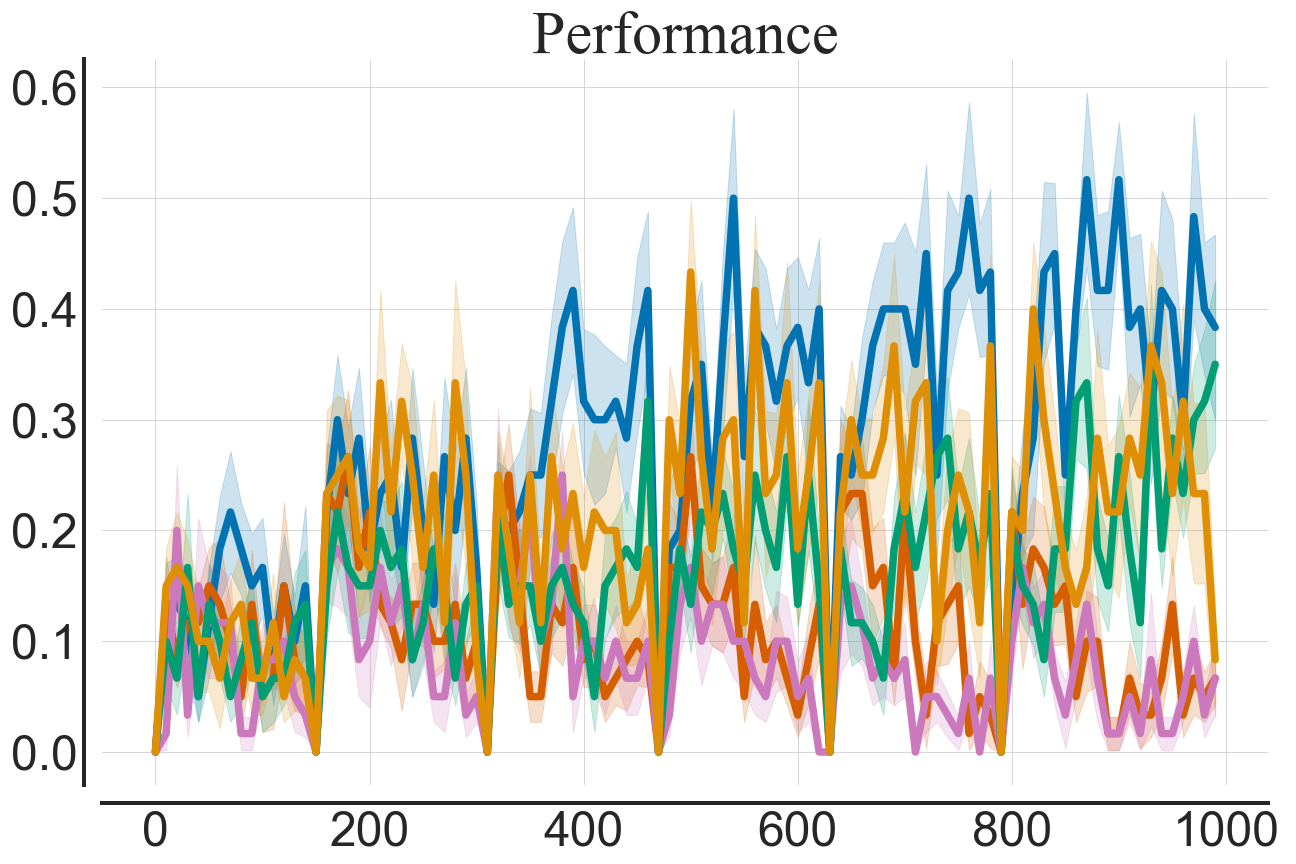}
    \hfill
    \includegraphics[width=0.23\linewidth]{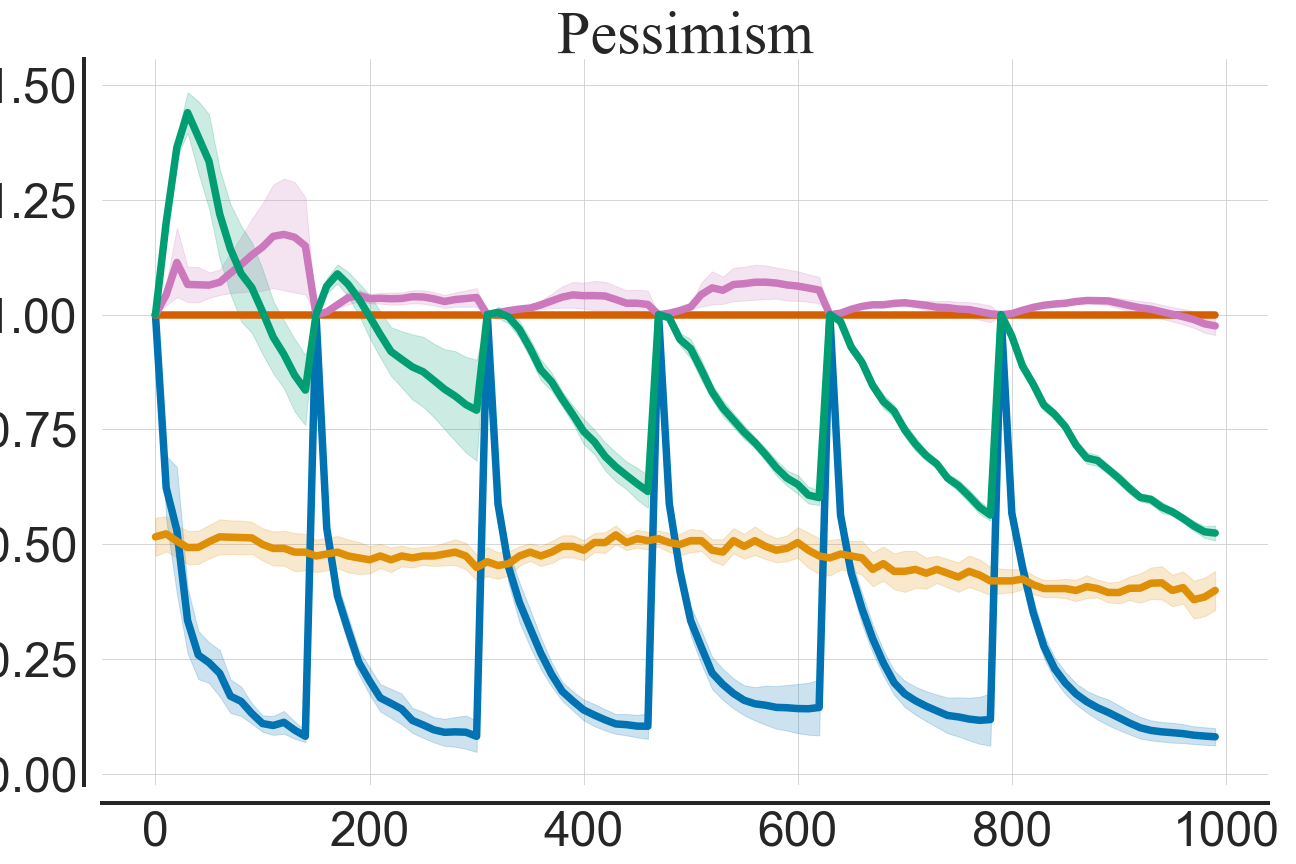}
    \hfill
    \includegraphics[width=0.23\linewidth]{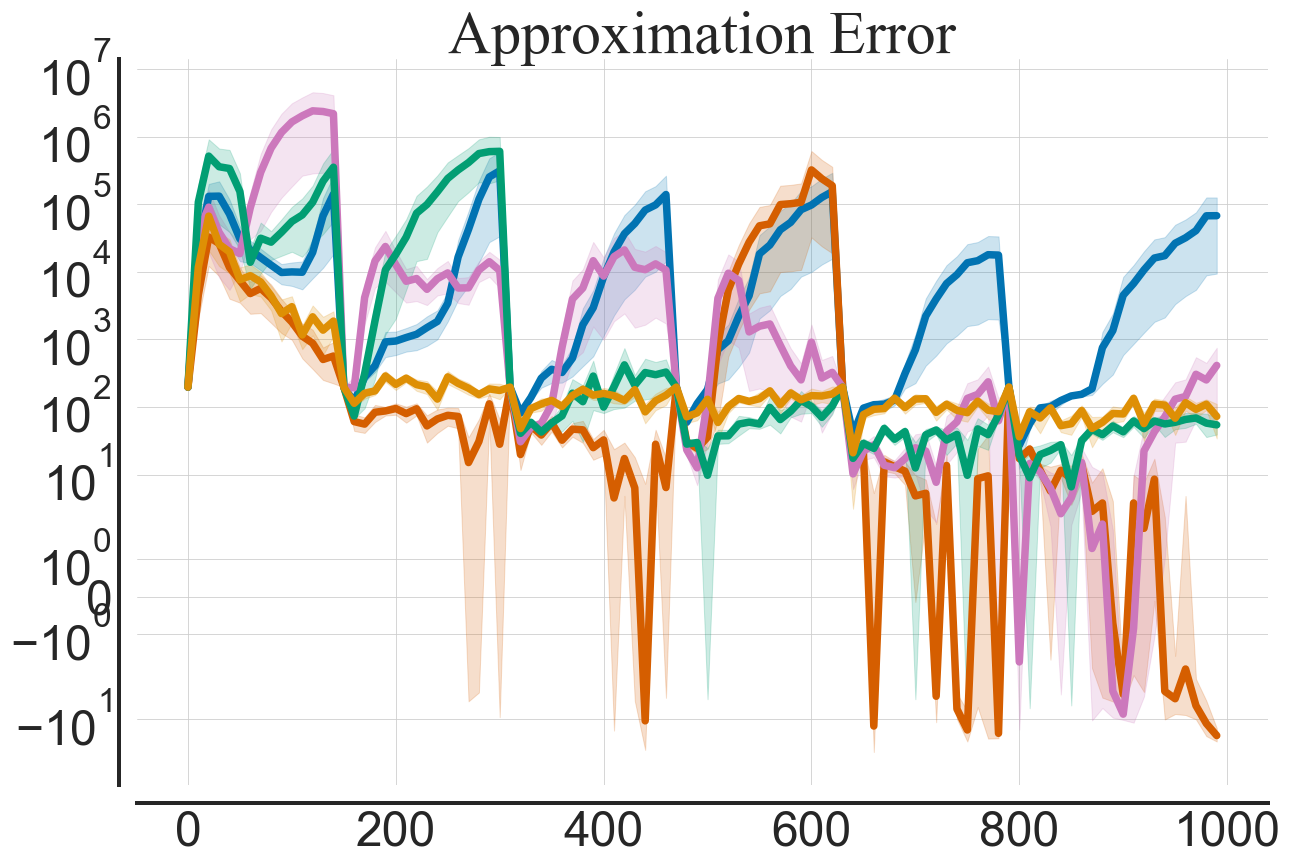}
    \hfill
    \includegraphics[width=0.23\linewidth]{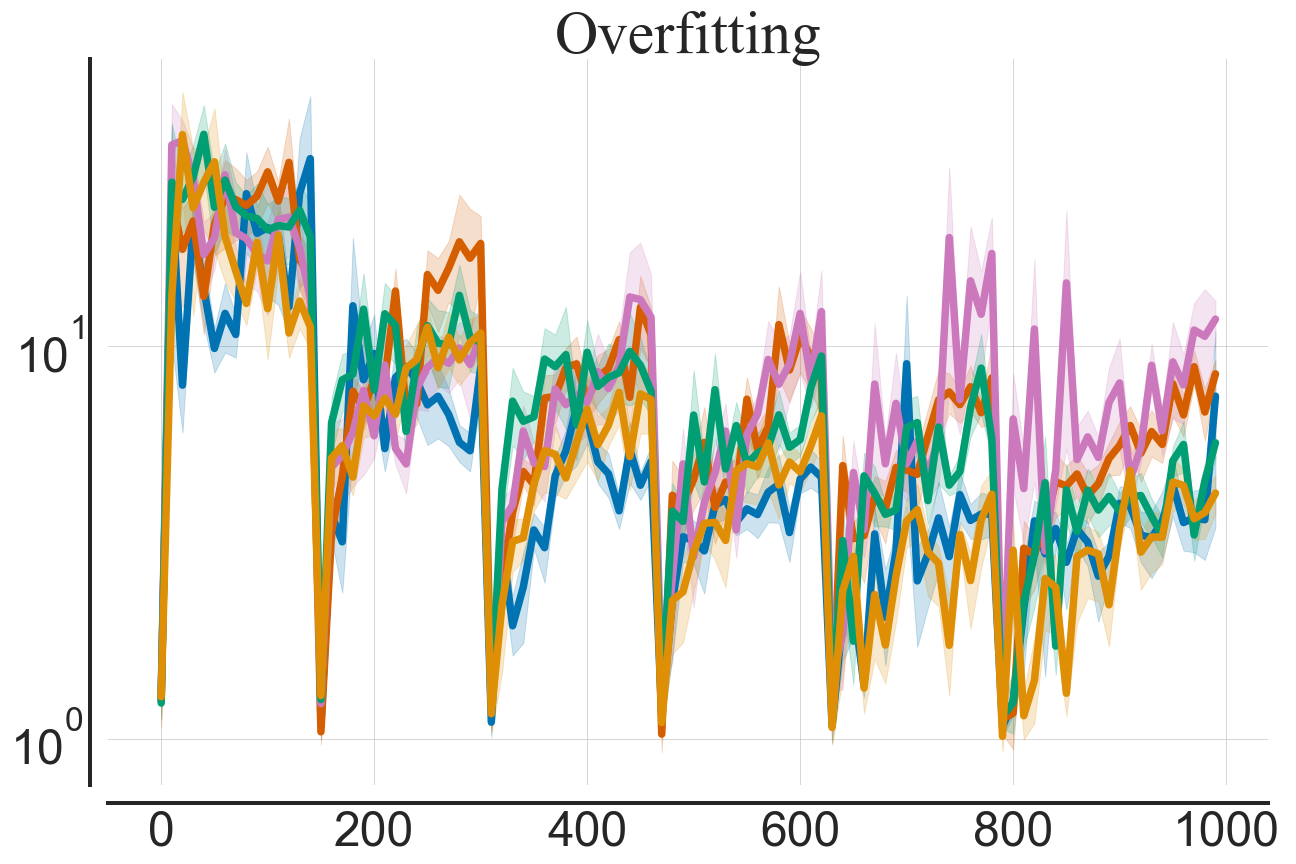}
    \hfill
    \end{subfigure}
    \subcaption{Coffee Push}
\end{minipage}
\caption{High replay regime results for each considered task (3/4). 10 seeds per task, mean and 3 standard deviations.}
\label{fig:learning_curves7}
\end{center}
\end{figure*}

\begin{figure*}[ht!]
\begin{center}
\begin{minipage}[h]{1.0\linewidth}
\centering
    \begin{subfigure}{0.88\linewidth}
    \includegraphics[width=\textwidth]{images/legend_1.png}
    \end{subfigure}
\end{minipage}
\bigskip
\begin{minipage}[h]{1.0\linewidth}
    \begin{subfigure}{1.0\linewidth}
    \hfill
    \includegraphics[width=0.23\linewidth]{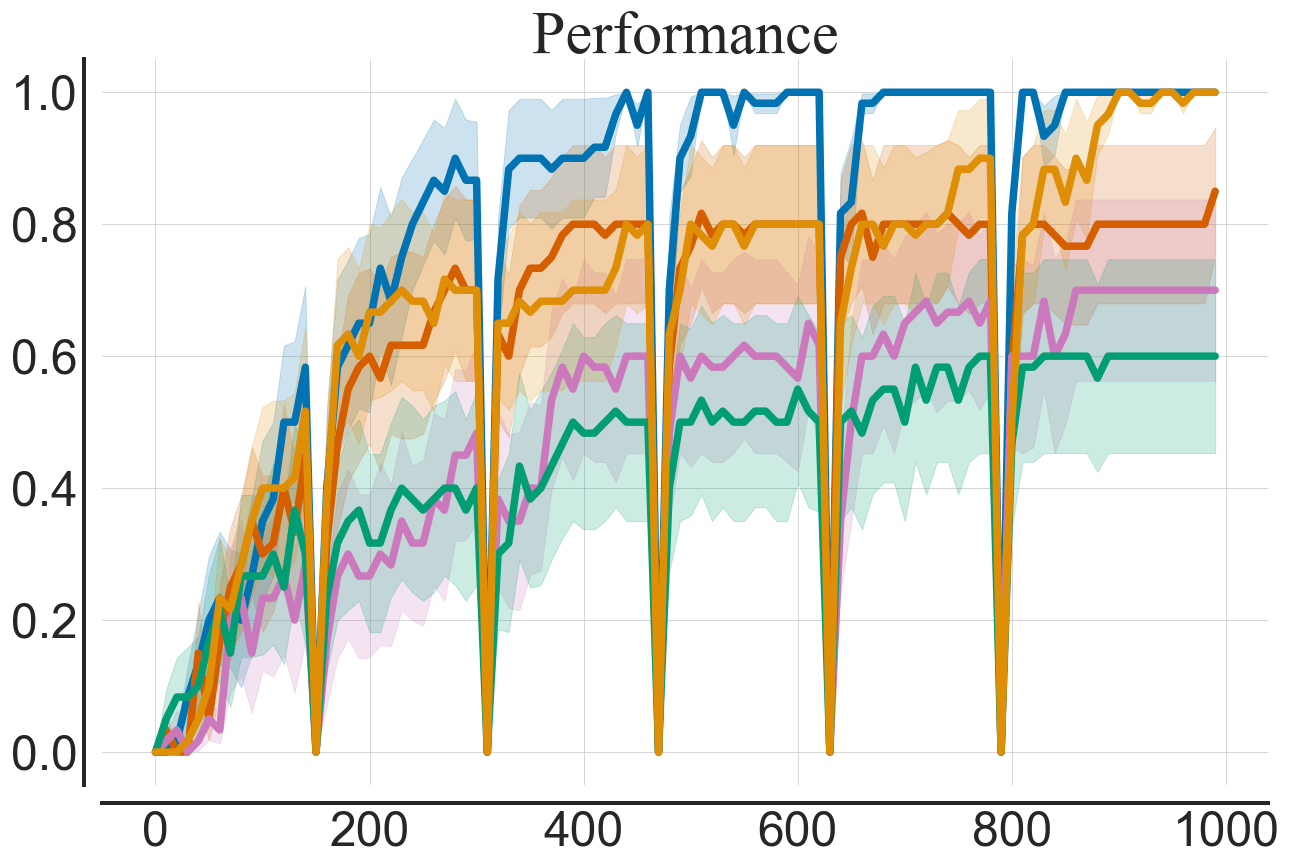}
    \hfill
    \includegraphics[width=0.23\linewidth]{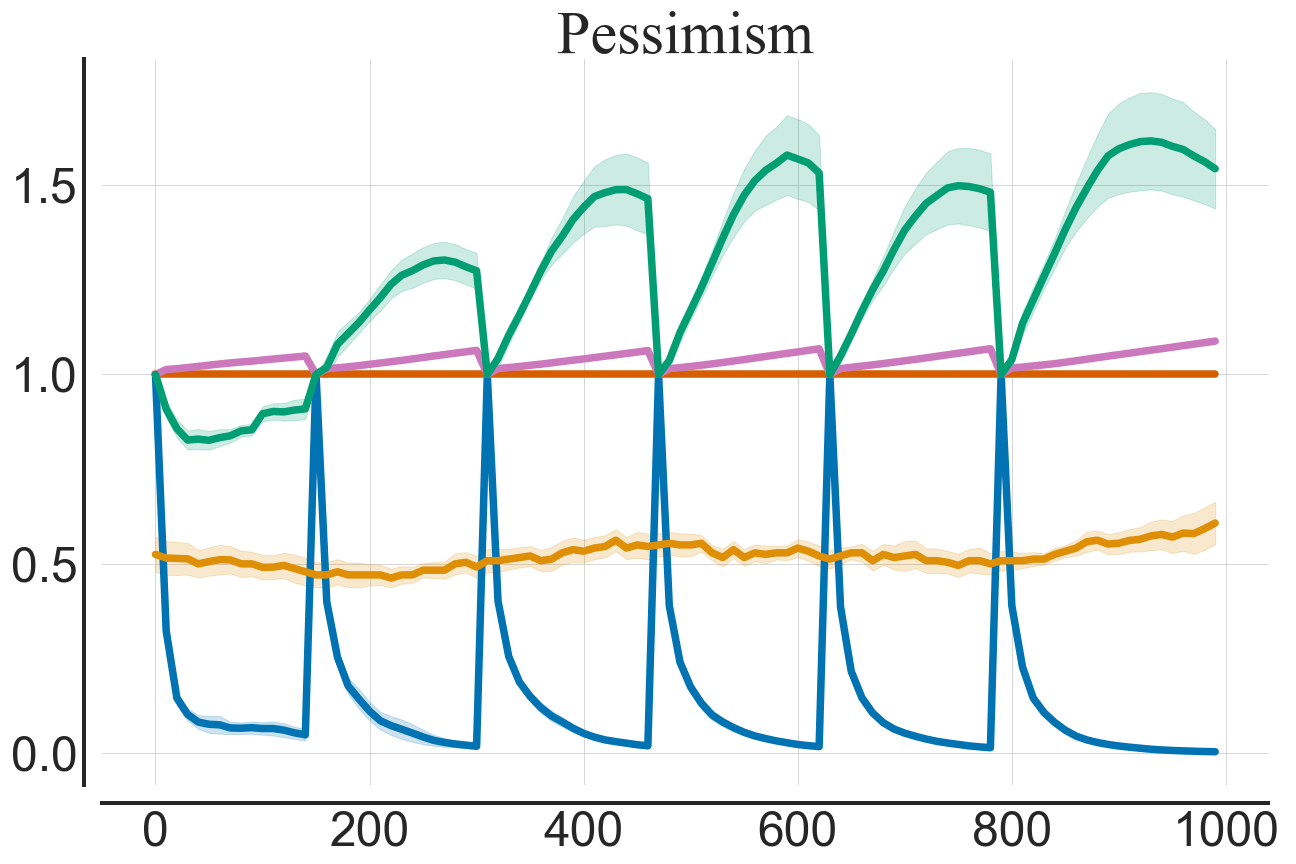}
    \hfill
    \includegraphics[width=0.23\linewidth]{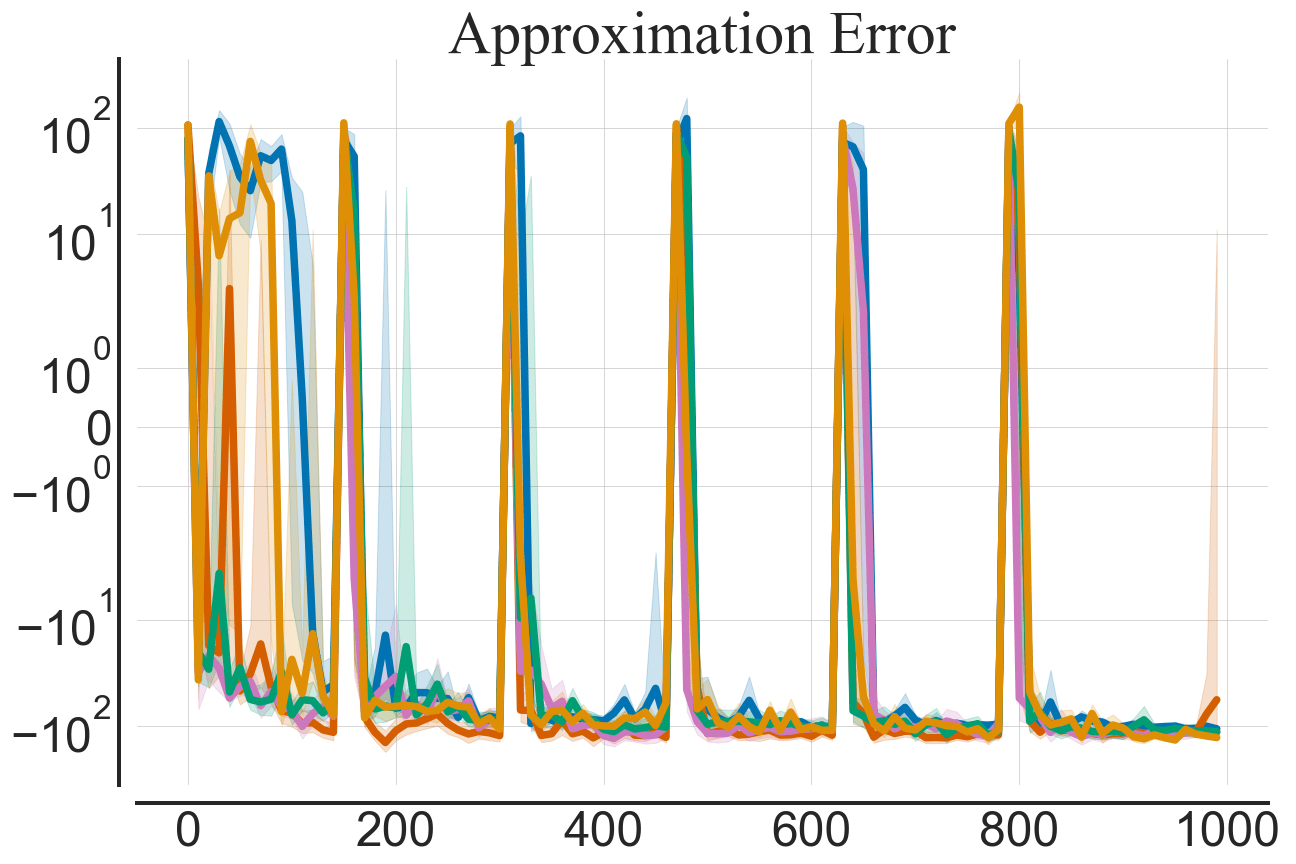}
    \hfill
    \includegraphics[width=0.23\linewidth]{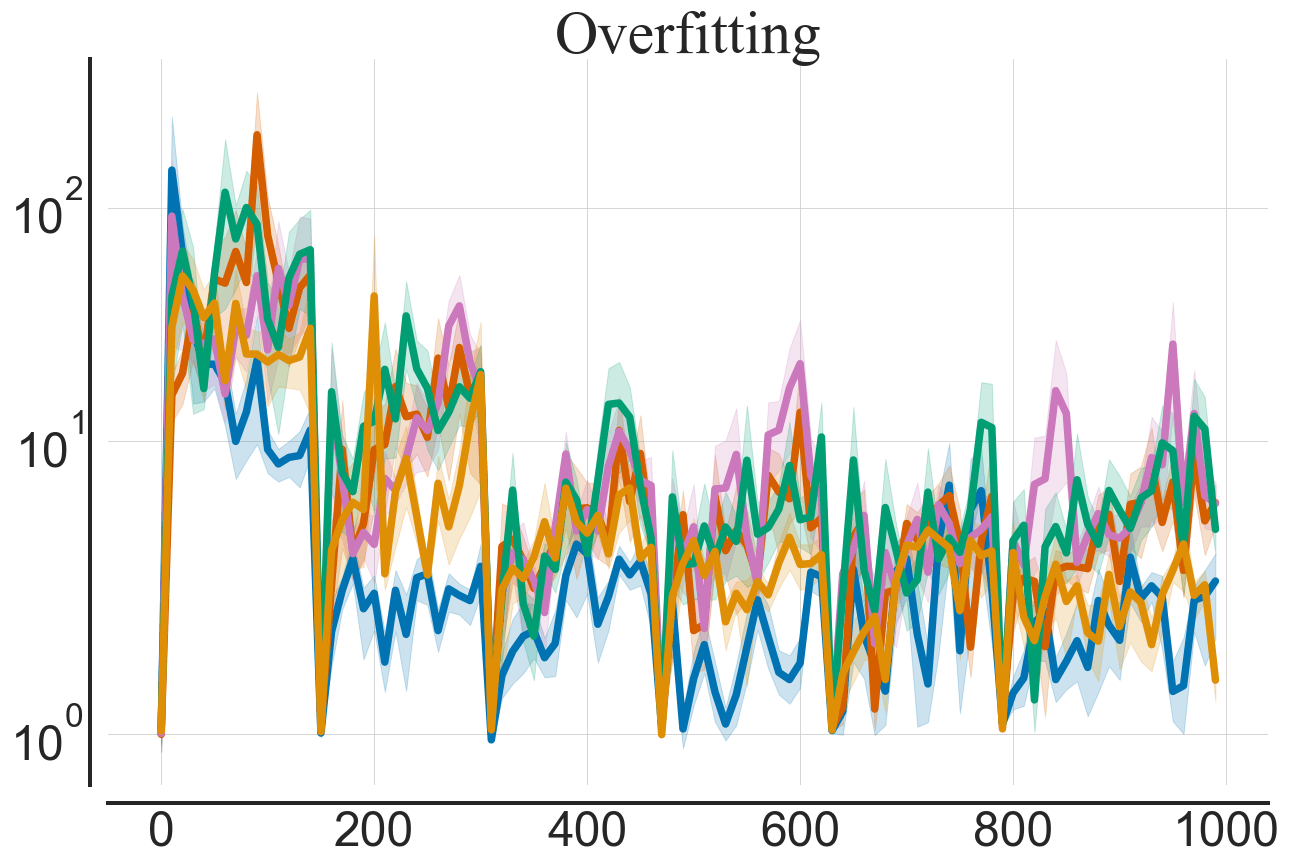}
    \hfill
    \end{subfigure}
    \subcaption{Drawer Open}
\end{minipage}
\bigskip
\begin{minipage}[h]{1.0\linewidth}
    \begin{subfigure}{1.0\linewidth}
    \hfill
    \includegraphics[width=0.23\linewidth]{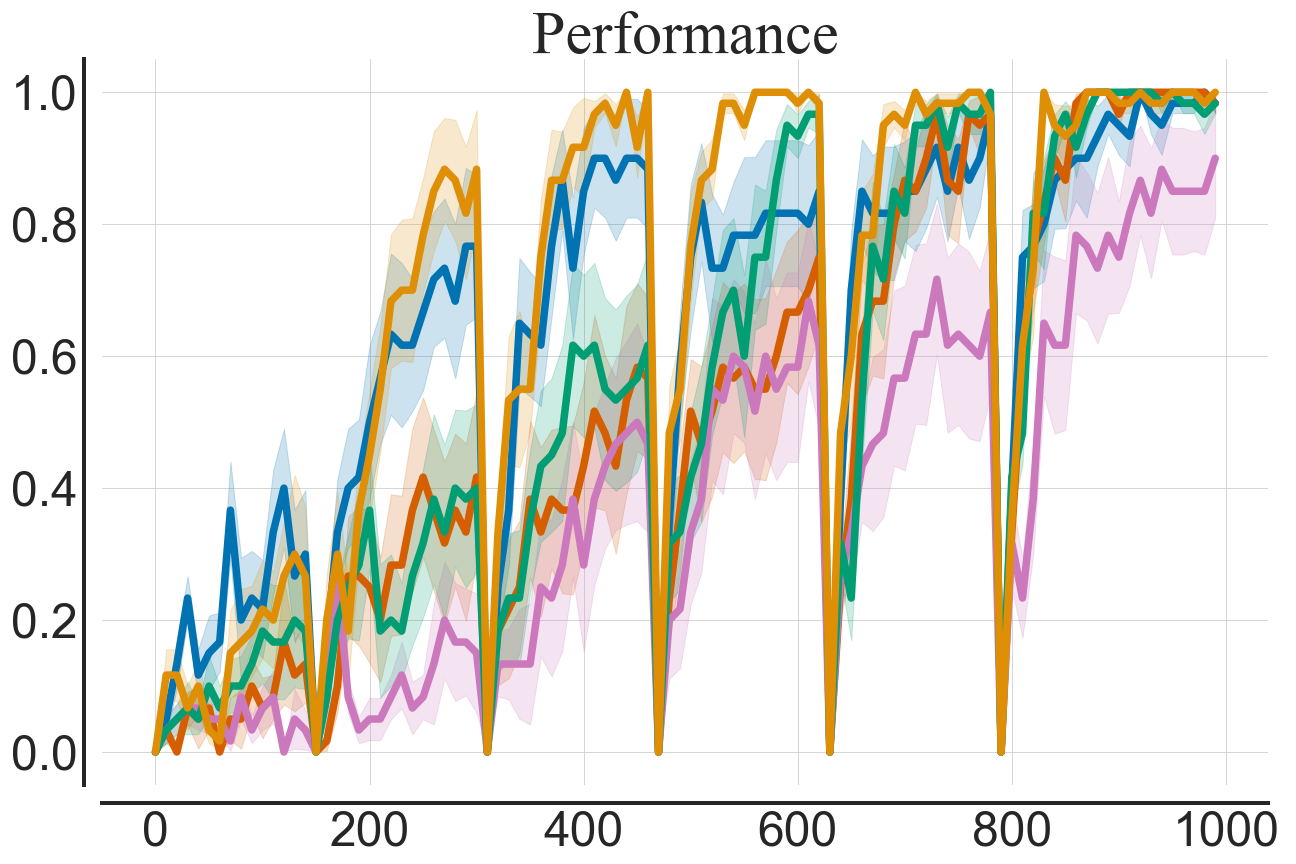}
    \hfill
    \includegraphics[width=0.23\linewidth]{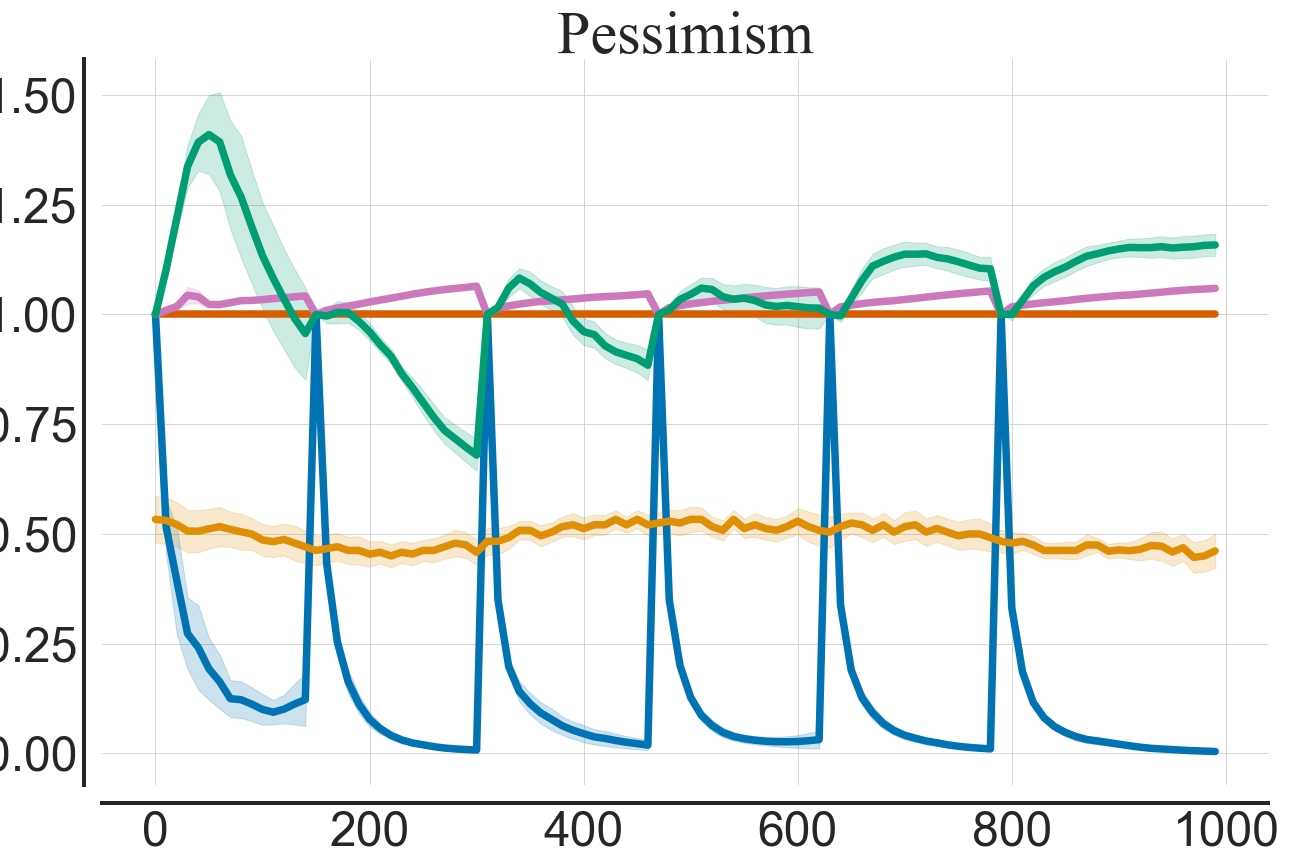}
    \hfill
    \includegraphics[width=0.23\linewidth]{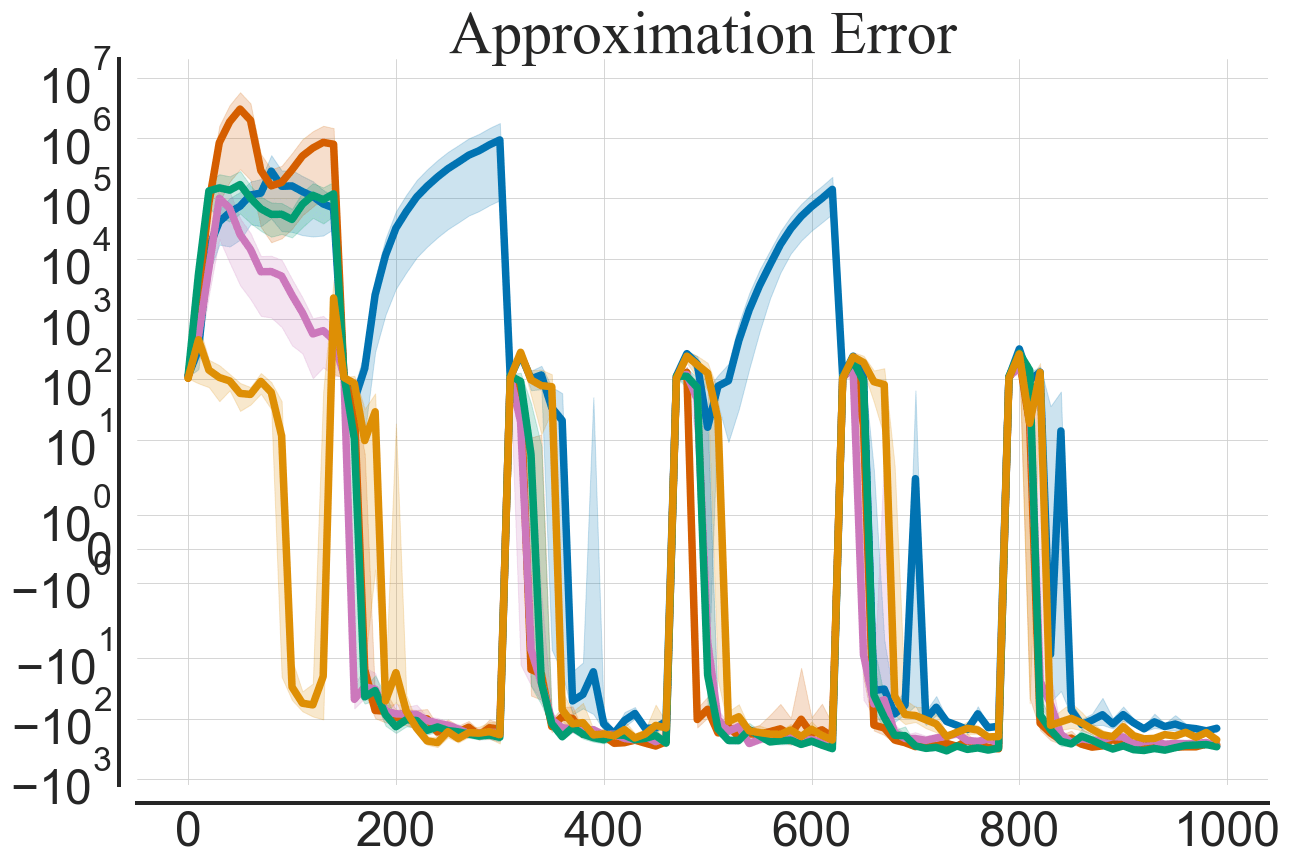}
    \hfill
    \includegraphics[width=0.23\linewidth]{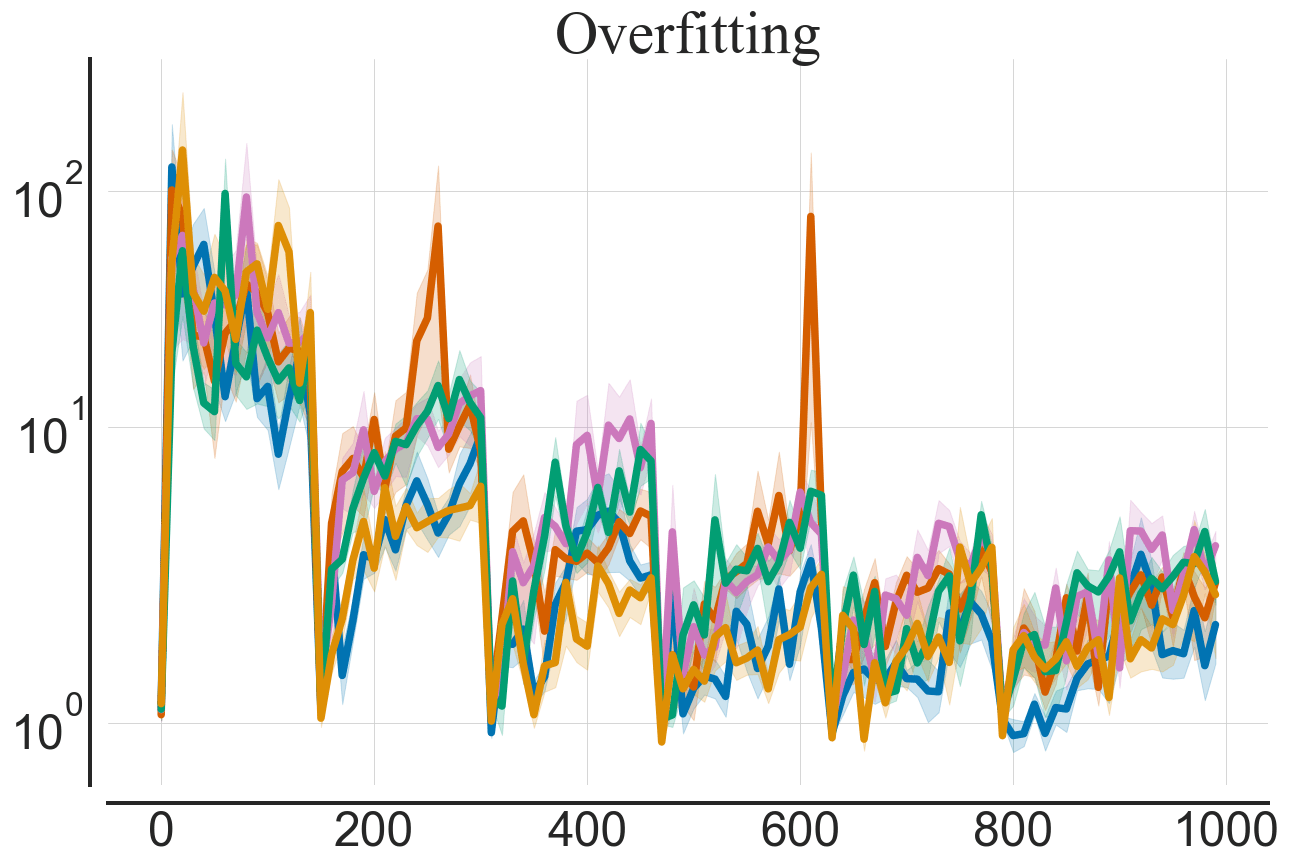}
    \hfill
    \end{subfigure}
    \subcaption{Hammer}
\end{minipage}
\bigskip
\begin{minipage}[h]{1.0\linewidth}
    \begin{subfigure}{1.0\linewidth}
    \hfill
    \includegraphics[width=0.23\linewidth]{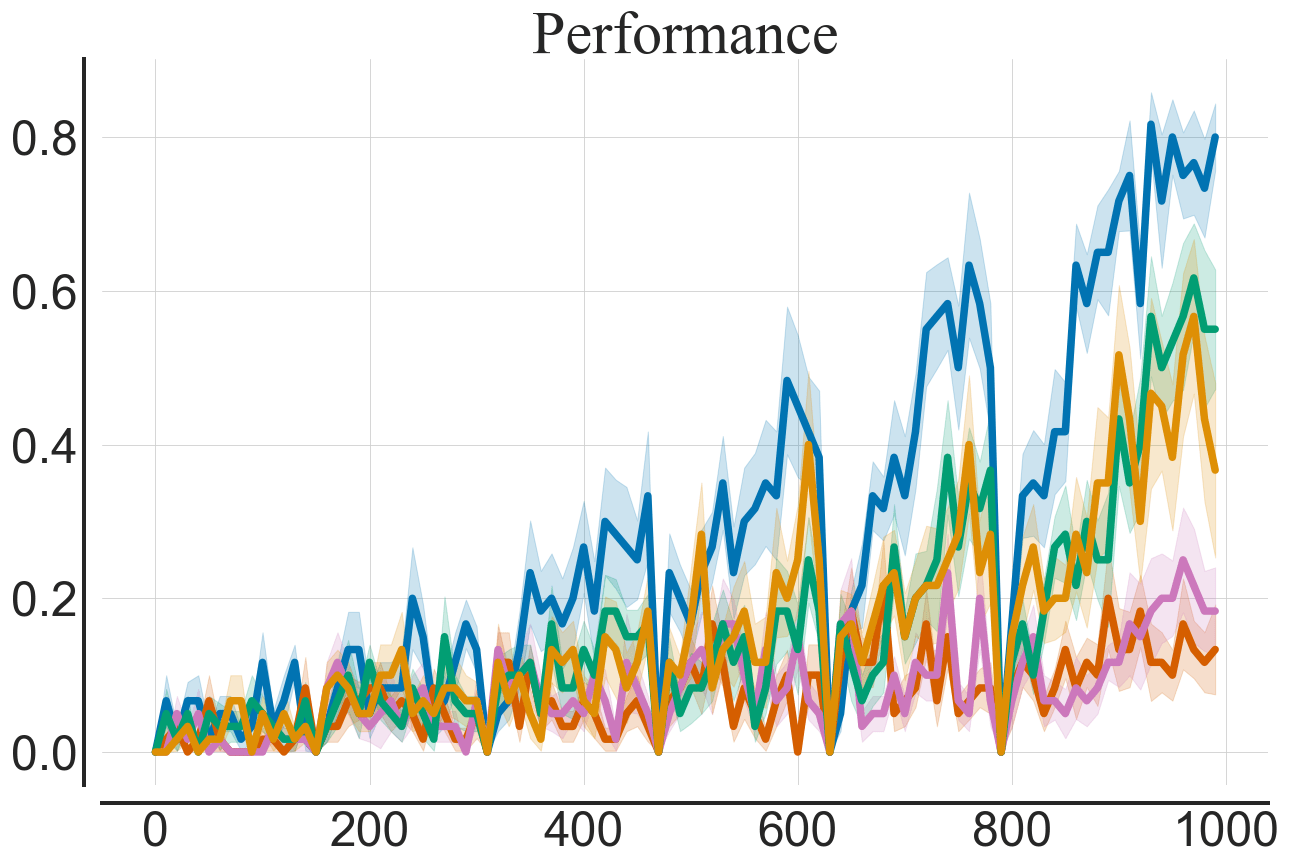}
    \hfill
    \includegraphics[width=0.23\linewidth]{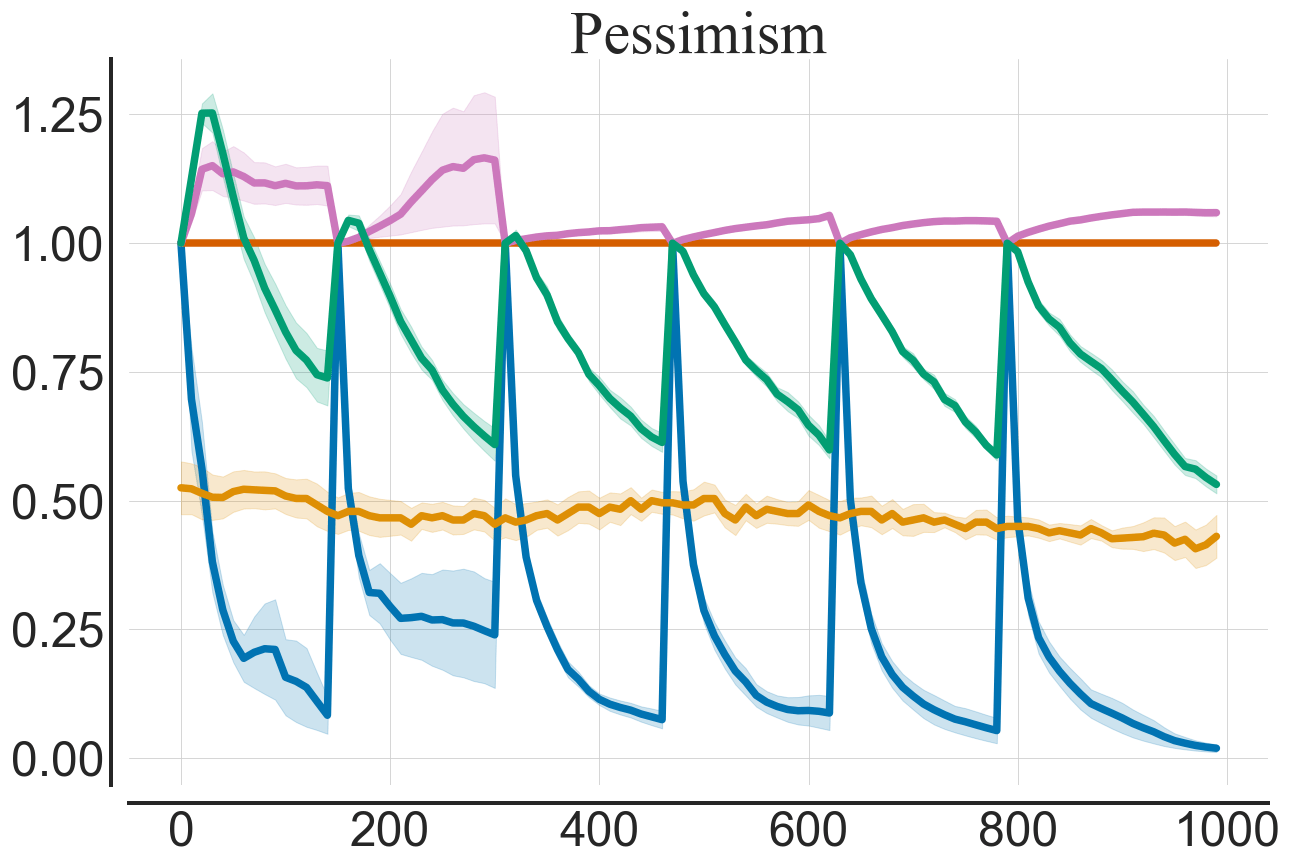}
    \hfill
    \includegraphics[width=0.23\linewidth]{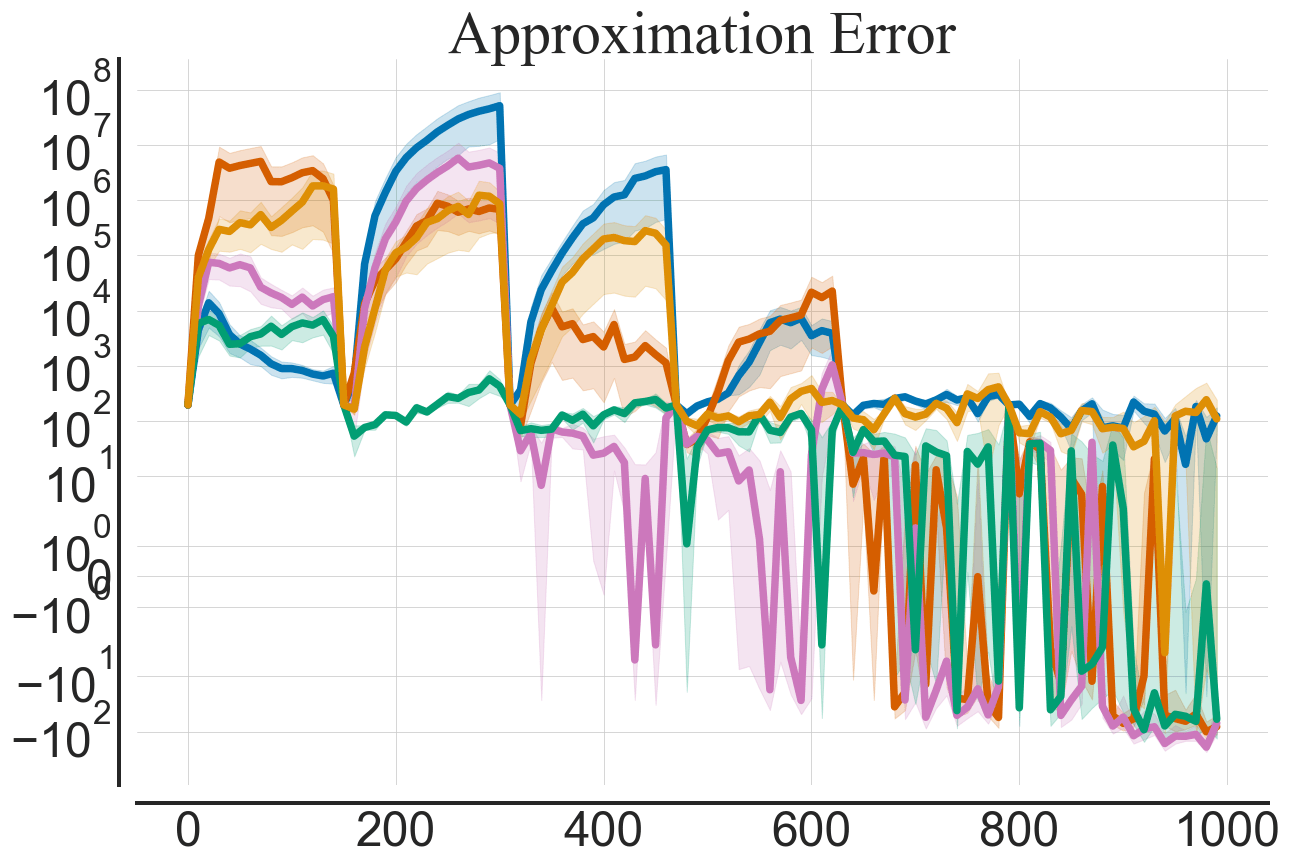}
    \hfill
    \includegraphics[width=0.23\linewidth]{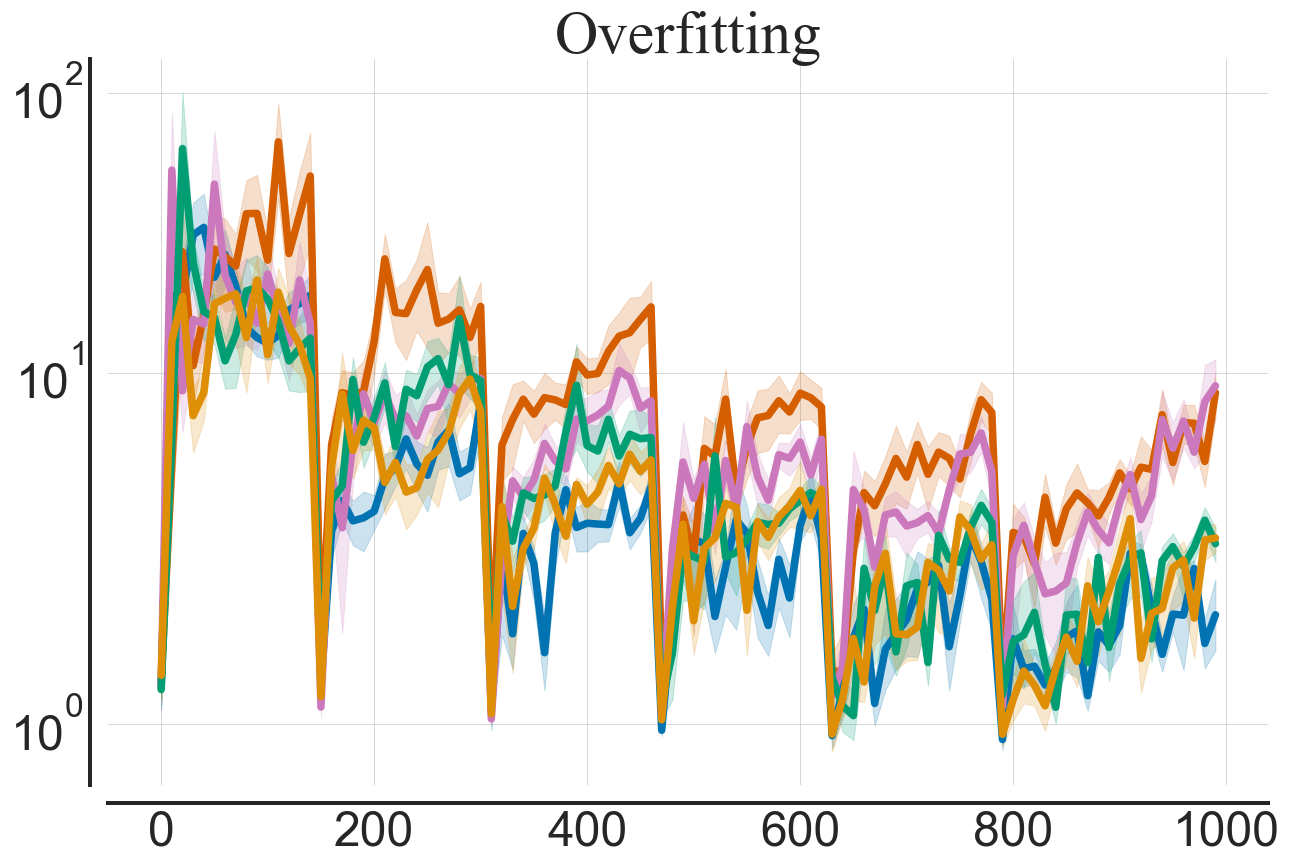}
    \hfill
    \end{subfigure}
    \subcaption{Push}
\end{minipage}
\bigskip
\begin{minipage}[h]{1.0\linewidth}
    \begin{subfigure}{1.0\linewidth}
    \hfill
    \includegraphics[width=0.23\linewidth]{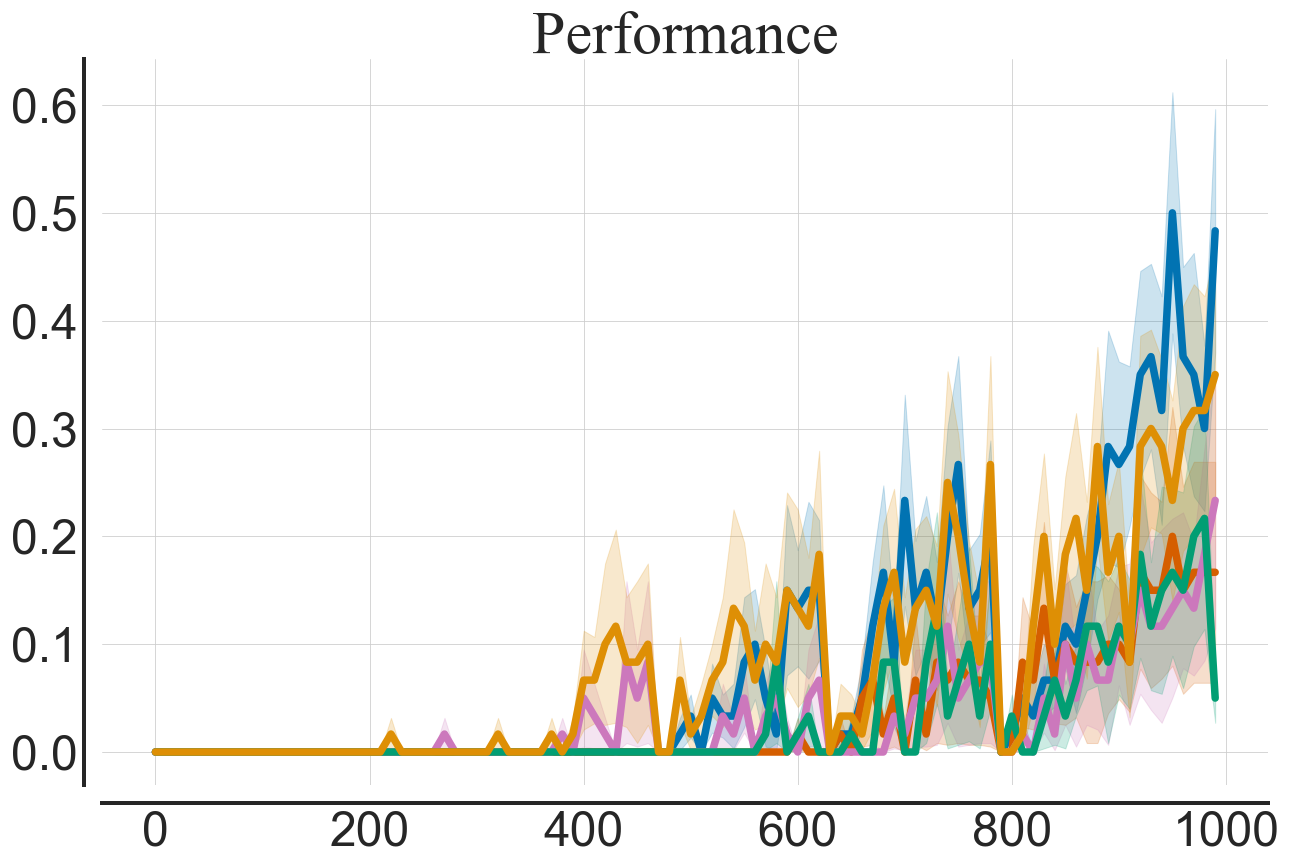}
    \hfill
    \includegraphics[width=0.23\linewidth]{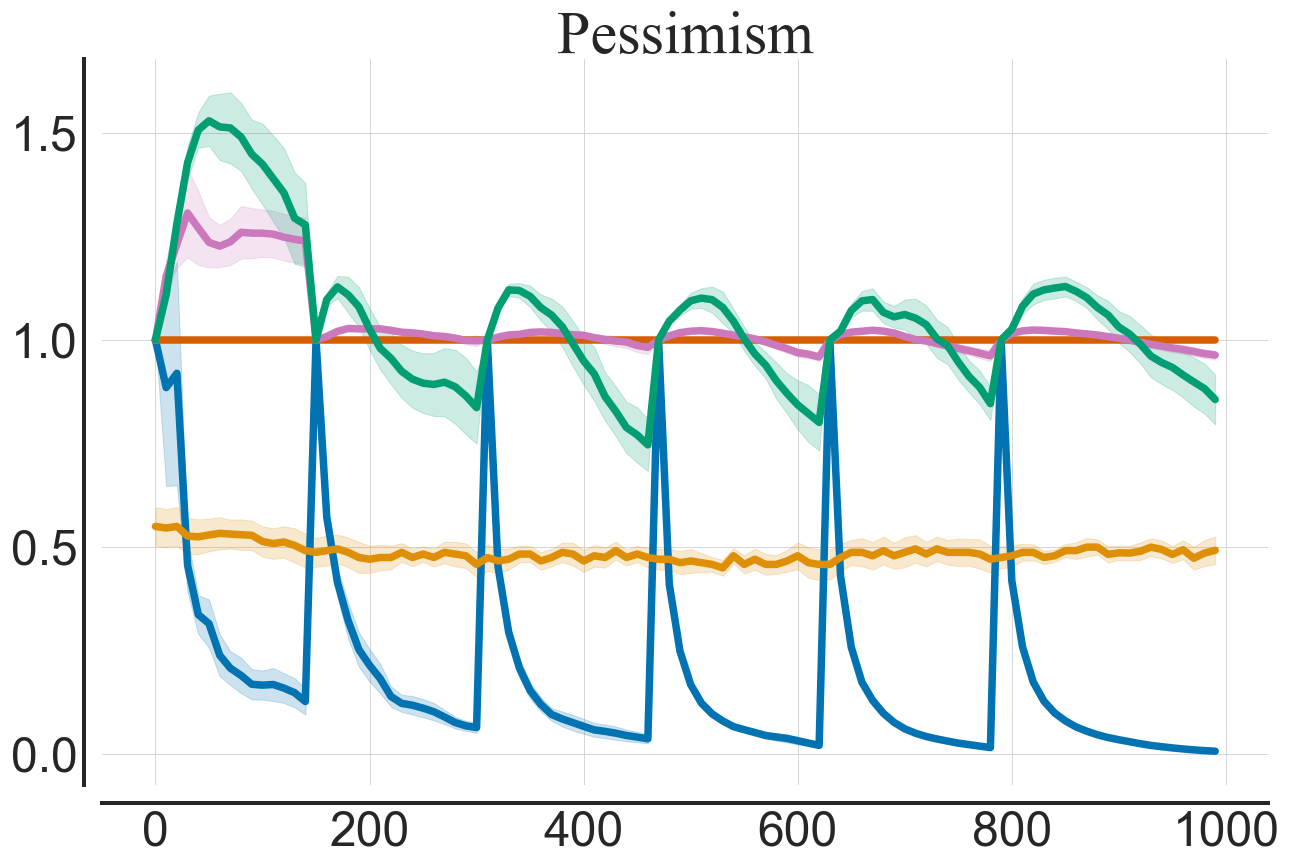}
    \hfill
    \includegraphics[width=0.23\linewidth]{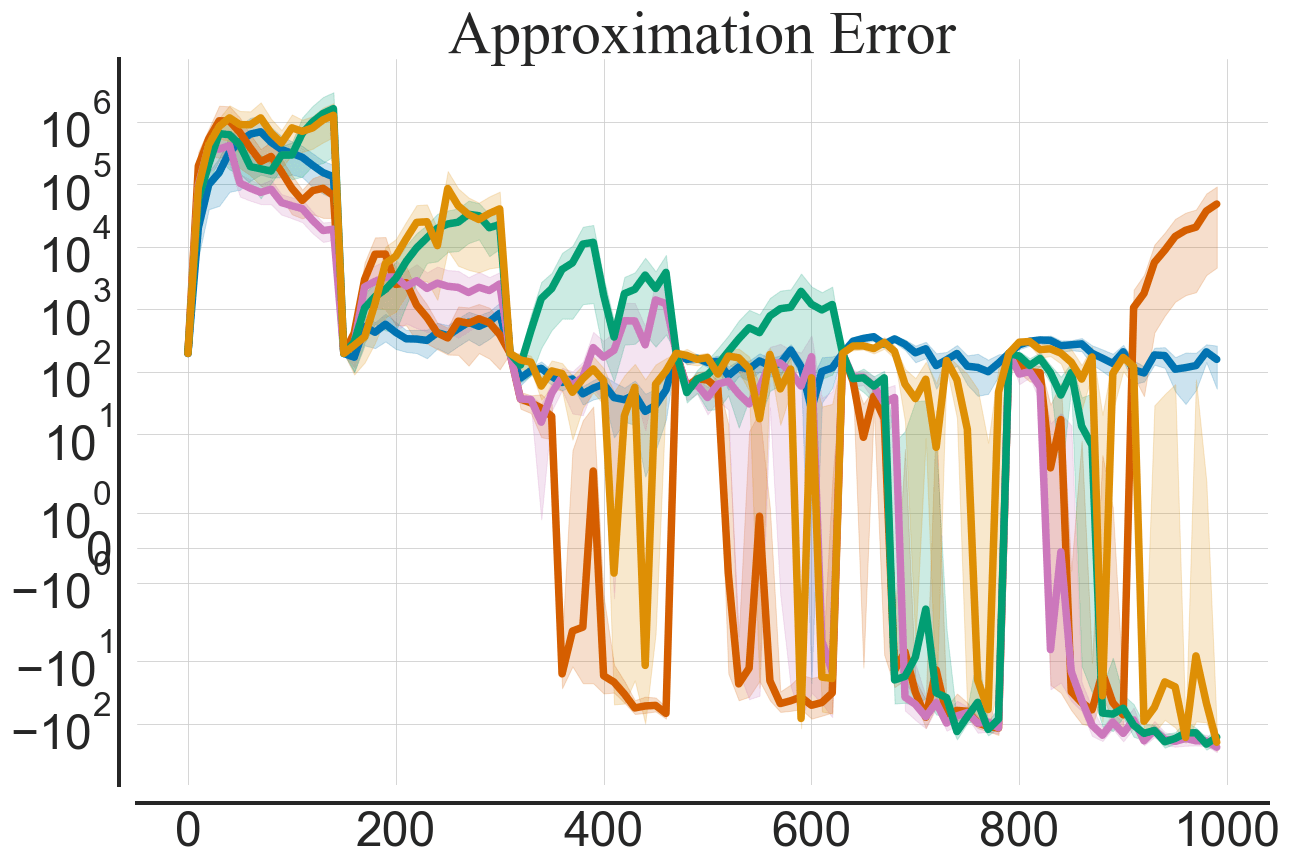}
    \hfill
    \includegraphics[width=0.23\linewidth]{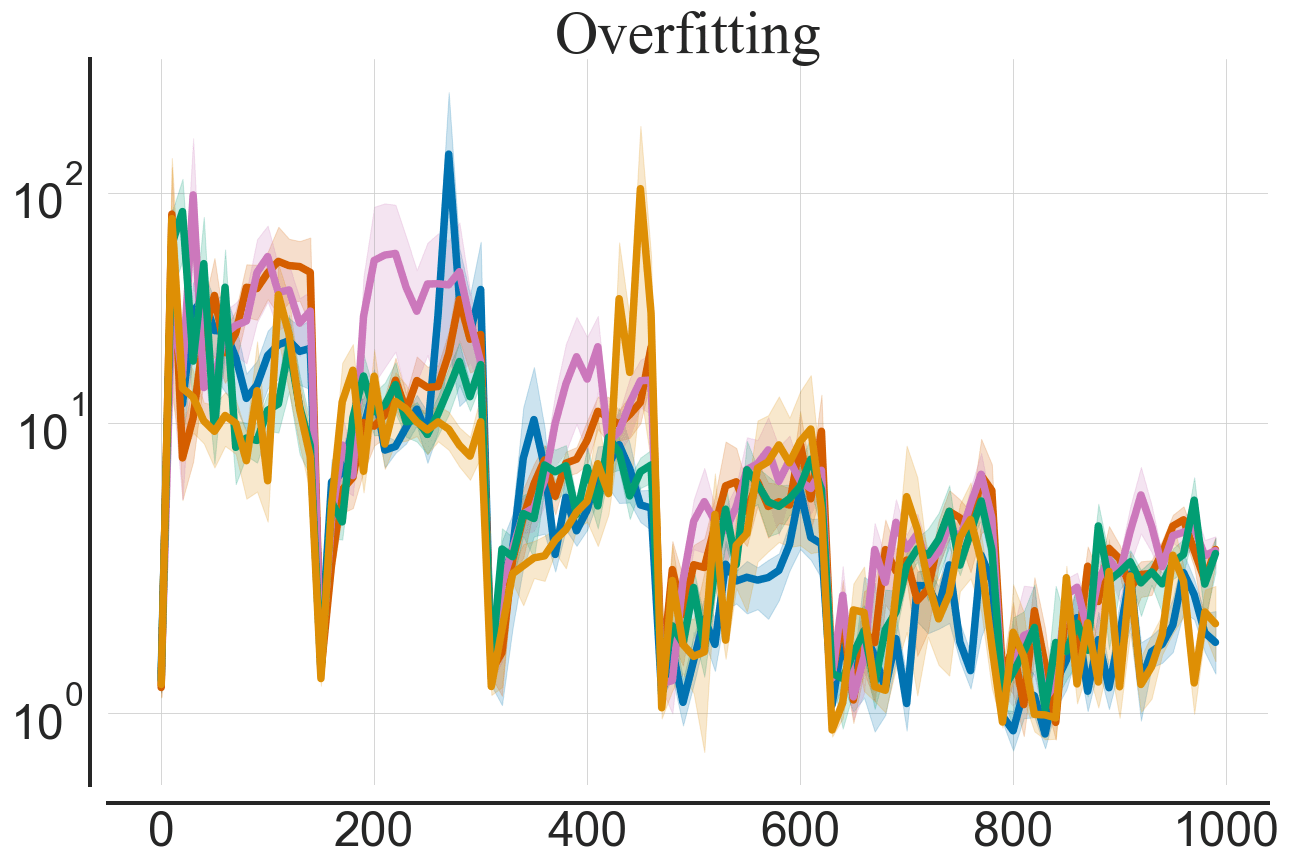}
    \hfill
    \end{subfigure}
    \subcaption{Stick Pull}
\end{minipage}
\bigskip
\begin{minipage}[h]{1.0\linewidth}
    \begin{subfigure}{1.0\linewidth}
    \hfill
    \includegraphics[width=0.23\linewidth]{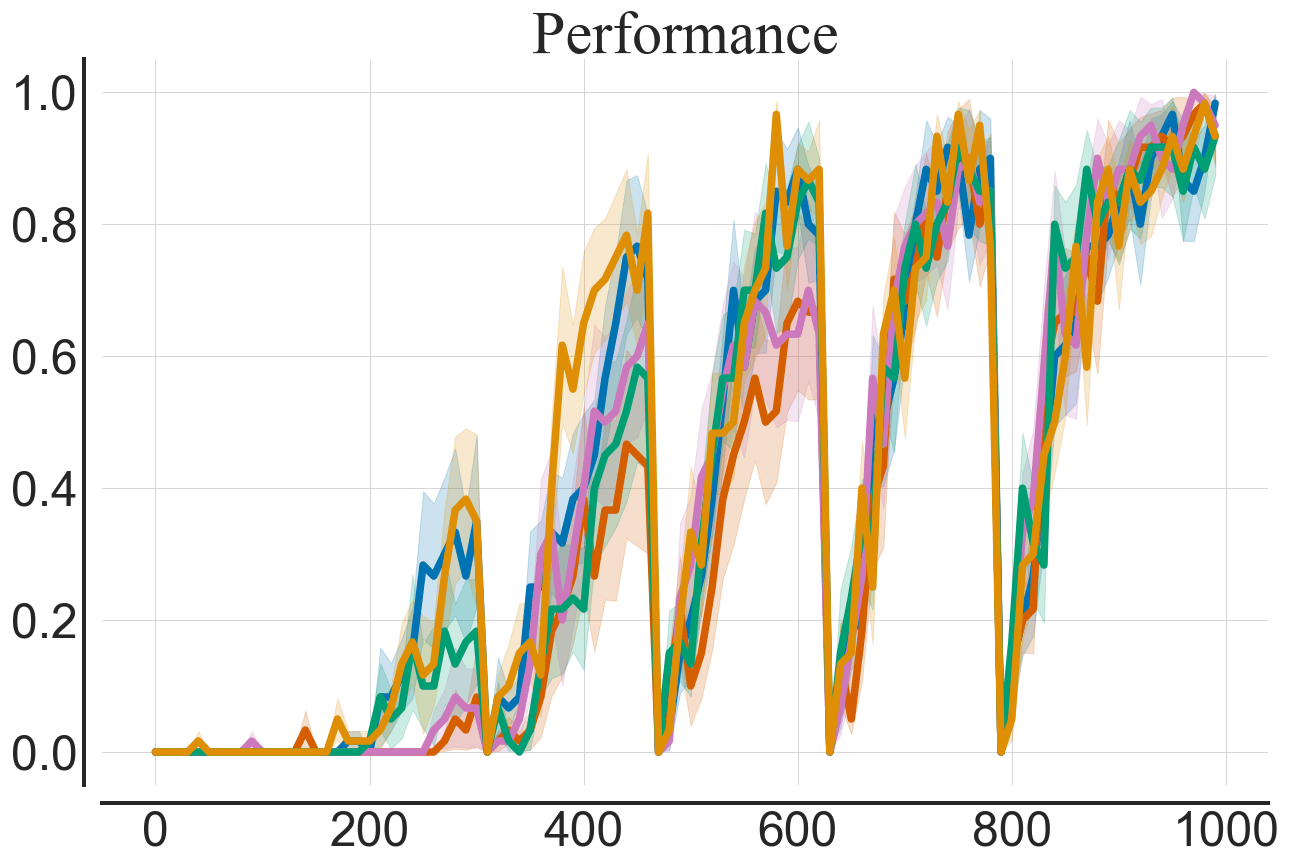}
    \hfill
    \includegraphics[width=0.23\linewidth]{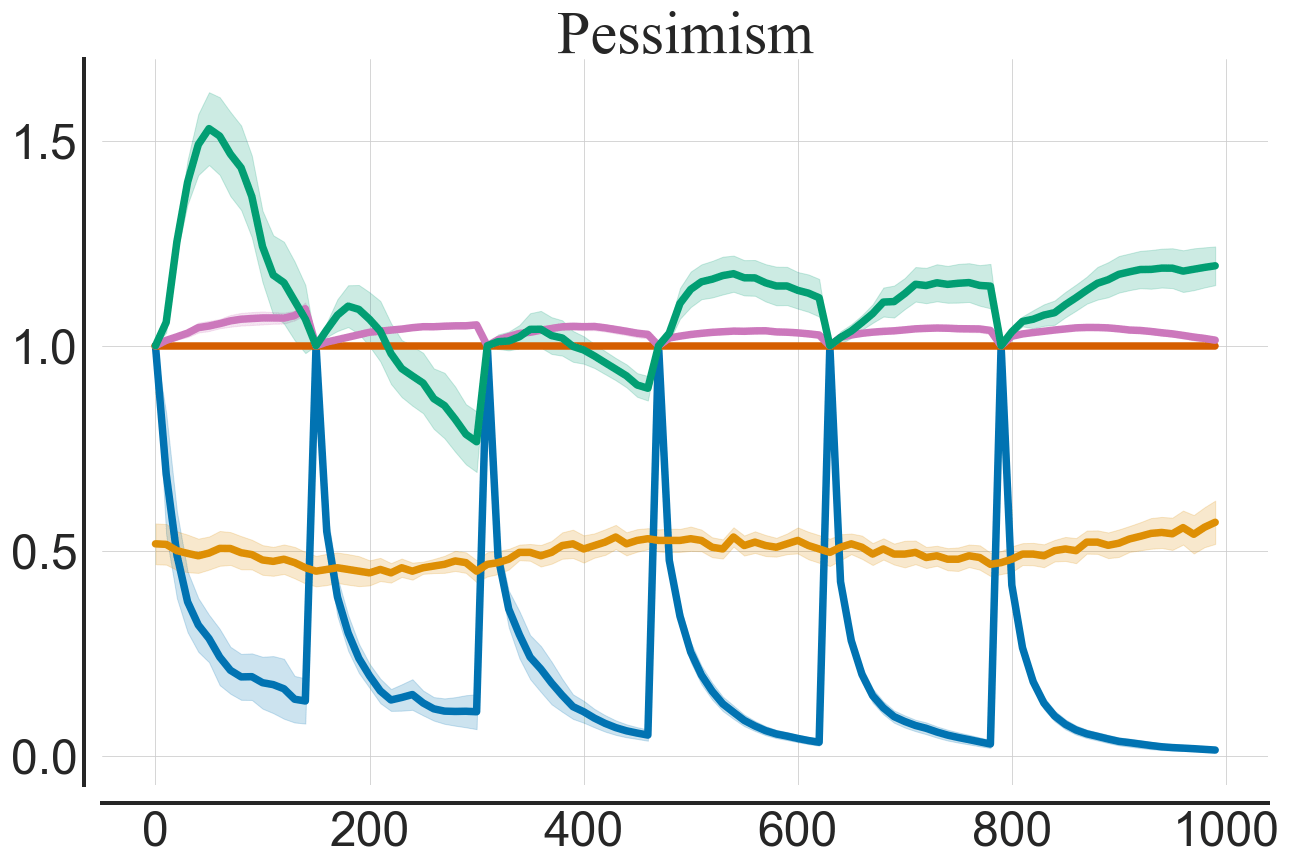}
    \hfill
    \includegraphics[width=0.23\linewidth]{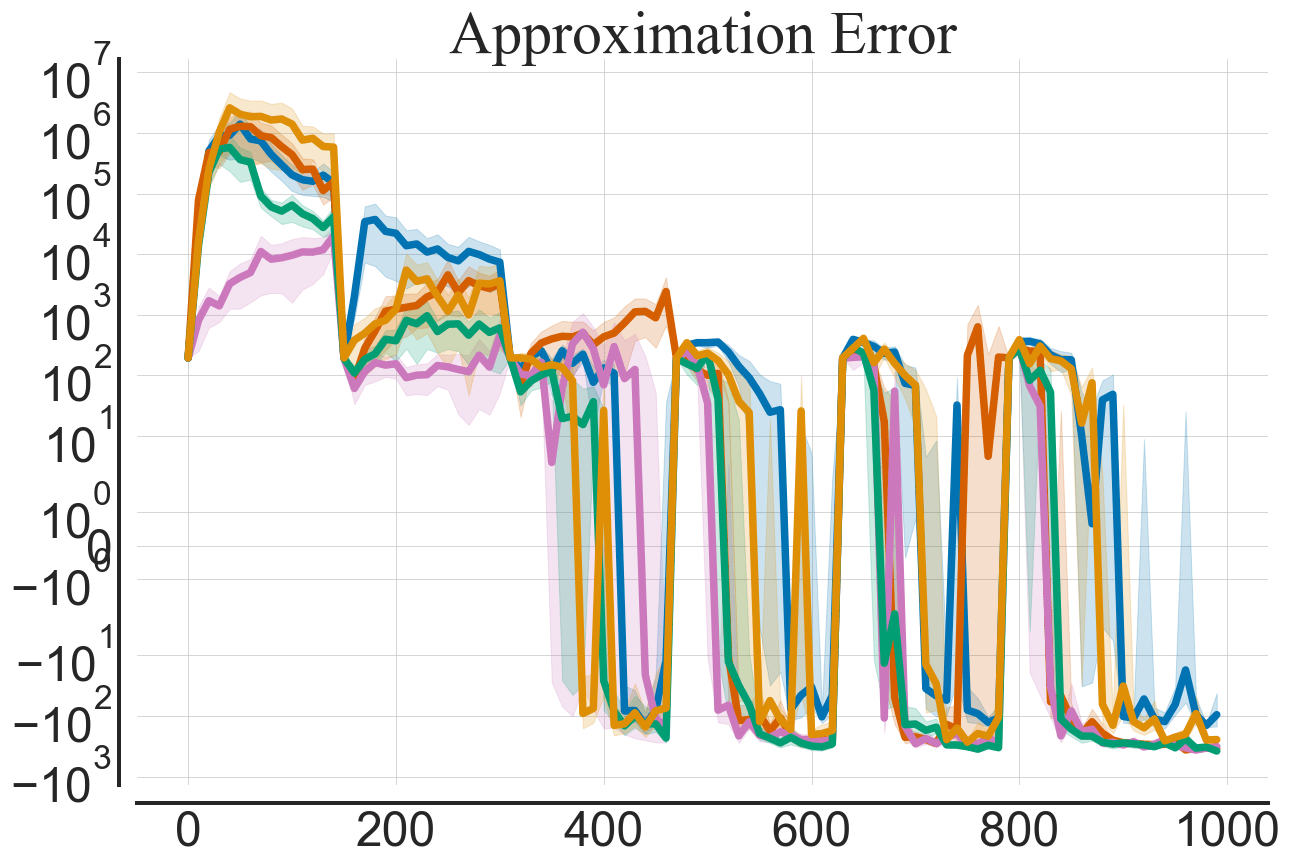}
    \hfill
    \includegraphics[width=0.23\linewidth]{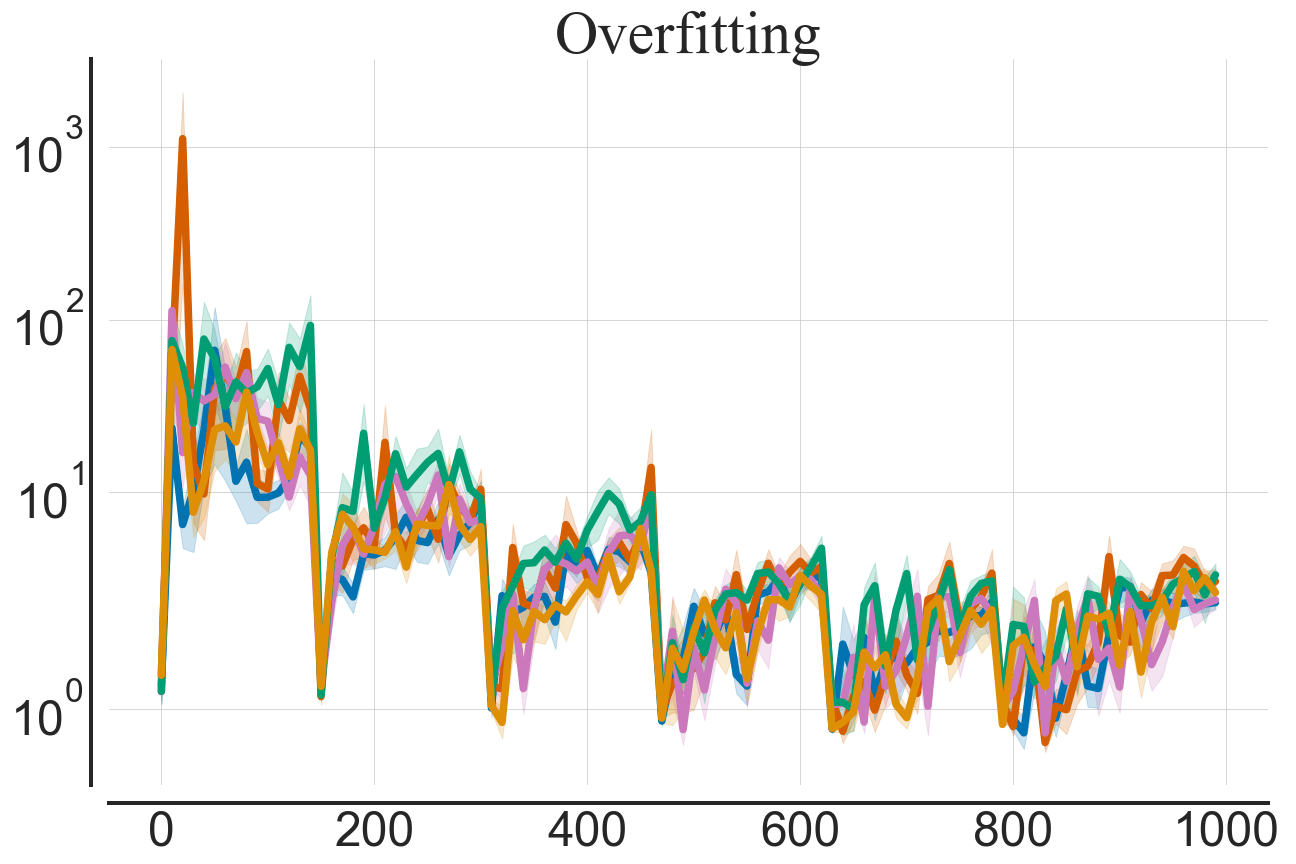}
    \hfill
    \end{subfigure}
    \subcaption{Sweep}
\end{minipage}
\caption{High replay regime results for each considered task (4/4). 10 seeds per task, mean and 3 standard deviations.}
\label{fig:learning_curves8}
\end{center}
\end{figure*}

Finally, we present the detailed training curves for performance, pessimism, approximation error and overfitting. The low replay regime results are presented in Figures \ref{fig:learning_curves1}, \ref{fig:learning_curves2}, \ref{fig:learning_curves3} and \ref{fig:learning_curves4}. the high replay regime results are presented in Figures \ref{fig:learning_curves5}, \ref{fig:learning_curves6}, \ref{fig:learning_curves7} and \ref{fig:learning_curves8}.

\end{document}